%% file: main.tex
\documentclass[a4paper,10pt,oneside,fleqn,review]{cas-dc}

\usepackage[utf8]{inputenc}
\usepackage[numbers]{natbib}
\usepackage{graphicx}
\usepackage{caption}

\usepackage{float}
\usepackage{placeins} 
\usepackage{microtype} 
\usepackage{amssymb}
\usepackage{amsmath}
\usepackage{mathtools}

\usepackage{booktabs}       
\usepackage{threeparttable}
\usepackage{multicol}
\usepackage{multirow}
\usepackage{tabularx}
\usepackage{supertabular}

\usepackage{tikz}
\usetikzlibrary{arrows,shapes,positioning,shadows,trees}
\usepackage{pgfplots}   

\definecolor{visio_R}{RGB}{192, 80, 70}
\definecolor{visio_G}{RGB}{157, 187, 97}
\definecolor{visio_B}{RGB}{75, 172, 198}

\usepackage{subfigure}

\usepackage{hyperref}

\begin{document}
\let\WriteBookmarks\relax
\def\floatpagepagefraction{1}
\def\textpagefraction{.001}
\shorttitle{Survey on pruning and quantization}
\shortauthors{T Liang et~al.}

\title [mode = title]{Pruning and Quantization for Deep Neural Network Acceleration: A Survey}

\author[1,2]{Tailin Liang}[
orcid=0000-0002-7643-912X
]
\ead{tailin.liang@xs.ustb.edu.cn}

\author[1,2,3]{John Glossner}[
]
\ead{jglossner@ustb.edu.cn}

\author[1]{Lei Wang}[]
\ead{wanglei@ustb.edu.cn}

\author[1,2]{Shaobo Shi}[
]
\ead{sbshi@hxgpt.com}

\author[1]{Xiaotong Zhang}
\ead{zxt@ies.ustb.edu.cn}
\corref{cor1}
\cortext[cor1]{Corresponding author}

\address[1]{School of Computer and Communication Engineering, University of Science and Technology Beijing, Beijing 100083, China}

\address[2]{Hua Xia General Processor Technologies, Beijing 100080, China}

\address[3]{General Processor Technologies, Tarrytown, NY 10591, United States}

\begin{abstract}
Deep neural networks have been applied in many applications exhibiting extraordinary abilities in the field of computer vision. However, complex network architectures challenge efficient real-time deployment and require significant computation resources and energy costs. These challenges can be overcome through optimizations such as network compression. Network compression can often be realized with little loss of accuracy. In some cases accuracy may even improve. This paper provides a survey on two types of network compression: pruning and quantization. Pruning can be categorized as static if it is performed offline or dynamic if it is performed at run-time. We compare pruning techniques and describe criteria used to remove redundant computations. We discuss trade-offs in element-wise, channel-wise, shape-wise, filter-wise, layer-wise and even network-wise pruning. Quantization reduces computations by reducing the precision of the datatype. Weights, biases, and activations may be quantized typically to 8-bit integers although lower bit width implementations are also discussed including binary neural networks. Both pruning and quantization can be used independently or combined. We compare current techniques, analyze their strengths and weaknesses, present compressed network accuracy results on a number of frameworks, and provide practical guidance for compressing networks.
\end{abstract}




\maketitle

\input{0-intro.tex}
\input{1-cnn.tex}

\input{2-prune.tex}
\input{3-quantize.tex}
\input{4-tail.tex}

\clearpage
\section{Quantization Performance Results} \label{sec:quant-perf-clct}
\input{ilsvrc-acc-all}
\clearpage
\bibliographystyle{cas-model2-names}
\bibliography{references}

\clearpage
\input{5-author-biography}
\end{document}

%% file: 0-intro.tex
\section{Introduction}\label{sec:intro}
Deep Neural Networks (DNNs) have shown extraordinary abilities in complicated applications such as image classification, object detection, voice synthesis, and semantic segmentation \cite{Lecun2015}.
Recent neural network designs with billions of parameters have demonstrated human-level capabilities but at the cost of significant computational complexity.
DNNs with many parameters are also time-consuming to train \cite{Brown2020}. These large networks are also difficult to deploy in embedded environments. Bandwidth becomes a limiting factor when moving weights and data between Compute Units (CUs) and memory. Over-parameterization is the property of a neural network where redundant neurons do not improve the accuracy of results. This redundancy can often be removed with little or no accuracy loss \cite{Sze2017}.

\input{cnn-acc}

\autoref{fig:cnn_acc} shows three design considerations that may contribute to over-parameterization: 1) network structure, 2) network optimization, and 3) hardware accelerator design. These design considerations are specific to Convolutional Neural Networks (CNNs) but also generally relevant to DNNs. 

Network structure encompasses three parts: 1) novel components, 2) network architecture search, and 3) knowledge distillation. Novel components is the design of efficient blocks such as separable convolution, inception blocks, and residual blocks. They are discussed in \autoref{sec:cnn-novel}. Network components also encompasses the types of connections within layers. Fully connected deep neural networks require $N^2$ connections between neurons. Feed forward layers reduce connections by considering only connections in the forward path. This reduces the number of connections to $N$. Other types of components such as dropout layers can reduce the number of connections even further. 
\par
Network Architecture Search (NAS) \cite{Wistuba2019}, also known as network auto search, programmatically searches for a highly efficient network structure from a large predefined search space. An estimator is applied to each produced architecture. While time-consuming to compute, the final architecture often outperforms manually designed networks.
\par
Knowledge Distillation (KD) \cite{Gou2020,Ruffy2019} evolved from knowledge transfer \cite{Bucilua2006}. The goal is to generate a simpler compressed model that functions as well as a larger model. KD trains a student network that tries to imitate a teacher network. The student network is usually but not always smaller and shallower than the teacher. The trained student model should be less computationally complex than the teacher. 


Network optimization \cite{Lebedev2018} includes: 1) computational convolution optimization, 2) parameter factorization, 3) network pruning, and 4) network quantization.
Convolution operations are more efficient than fully connected computations because they keep high dimensional information as a 3D tensor rather than flattening the tensors into vectors that removes the original spatial information. This feature helps CNNs to fit the underlying structure of image data in particular. Convolution layers also require significantly less coefficients compared to Fully Connected Layers (FCLs). Computational convolution optimizations include Fast Fourier Transform (FFT) based convolution \cite{Mathieu2013}, Winograd convolution \cite{Lavin2016}, and the popular image to column (im2col) \cite{Chellapilla2006} approach. 
We discuss im2col in detail in \autoref{sec:cnn-ops} since it is directly related to general pruning techniques. 
\par
Parameter factorization is a technique that decomposes higher-rank tensors into lower-rank tensors simplifying memory access and compressing model size. 
It works by breaking large layers into many smaller ones, thereby reducing the number of computations.
It can be applied to both convolutional and fully connected layers. 
This technique can also be applied with pruning and quantization.
\par
Network pruning \cite{Reed1993,Blalock2020,Augasta2013,Xu2020a} involves removing parameters that don't impact network accuracy. Pruning can be performed in many ways and is described extensively in \autoref{sec:pruning}. 
\par
Network quantization \cite{Krishnamoorthi2018, Guo2018a} involves replacing datatypes with reduced width datatypes. For example, replacing 32-bit Floating Point (FP32) with 8-bit Integers (INT8). The values can often be encoded to preserve more information than simple conversion. Quantization is described extensively in \autoref{sec:quantization}. 

Hardware accelerators \cite{Li2017a, Reuther2019} are designed primarily for network acceleration. At a high level they encompass entire processor platforms and often include hardware optimized for neural networks. Processor platforms include specialized Central Processing Unit (CPU) instructions, Graphics Processing Units (GPUs), Application Specific Integrated Circuits (ASICs), and Field Programmable Gate Arrays (FPGAs). 
\par
CPUs have been optimized with specialized Artificial Intelligence
(AI) instructions usually within specialized Single Instruction Multiple Data (SIMD) units \cite{Cornea2015, ARM2008}. While CPUs can be used for training, they have primarily been used for inference in systems that do not have specialized inference accelerators.
\par
GPUs have been used for both training and inference. nVidia has specialized tensor units incorporated into their GPUs that are optimized for neural network acceleration \cite{NVIDIACorporation2017a}. AMD \cite{AMD}, ARM \cite{Arm2020}, and Imagination \cite{Imagination} also have GPUs with instructions for neural network acceleration. 
\par
Specialized ASICs have also been designed for neural network acceleration. They typically target inference at the edge, in security cameras, or on mobile devices. Examples include: General Processor Technologies (GPT) \cite{Moudgill2020}, ARM, nVidia, and 60+ others \cite{Reuther2019} all have processors targeting this space. ASICs may also target both training and inference in datacenters. Tensor processing units (TPU) from Google \cite{Jouppi2017}, Habana from Intel \cite{Medina2019}, Kunlun from Baidu \cite{Ouyang2020}, Hanguang from Alibaba \cite{Jiao2020}, and Intelligence Processing Unit (IPU) from Graphcore \cite{Jia2019}. 
\par
Programmable reconfigurable FPGAs have been used for neural network acceleration \cite{Guo2017, Abdelouahab2018, Venieris2018, Li2020a}. FPGAs are widely used by researchers due to long ASIC design cycles. Neural network libraries are available from Xilinx \cite{Kathail2020} and Intel \cite{FPGA}.
Specific neural network accelerators are also being integrated into FPGA fabrics \cite{Xilinx2018, Achronix2020, RichardChuang2020}. Because FPGAs operate at the gate level, they are often used in low-bit width and binary neural networks \cite{Moss2017,Zhao2017,Prost-Boucle2017}. 

Neural network specific optimizations are typically incorporated into custom ASIC hardware. Lookup tables can be used to accelerate trigonometric activation functions \cite{Choi2017a} or directly generate results for low bit-width arithmetic \cite{Esser2016}, partial products can be stored in special registers and reused \cite{Chen2016b}, and memory access ordering with specialized addressing hardware can all reduce the number of cycles to compute a neural network output \cite{Judd2017}. Hardware accelerators are not the primary focus of this paper. However, we do note hardware implementations that incorporate specific acceleration techniques.
Further background information on efficient processing and hardware implementations of DNNs can be found in \cite{Sze2017}. 

We summarize our main contributions as follows:
\begin{itemize}
    \item We provide a review of two network compression techniques: pruning and quantization. We discuss methods of compression, mathematical formulations, and compare current State-Of-The-Art (SOTA) compression methods.
    \item We classify pruning techniques into static and dynamic methods, depending if they are done offline or at runtime, respectively. 
    \item We analyze and quantitatively compare quantization techniques and frameworks. 
    \item We provide practical guidance on quantization and pruning. 
\end{itemize}

This paper focuses primarily on network optimization for convolutional neural networks. It is organized as follows: In \autoref{sec:background} we give an introduction to neural networks and specifically convolutional neural networks. We also describe some of the network optimizations of convolutions. In \autoref{sec:pruning} we describe both static and dynamic pruning techniques. In \autoref{sec:quantization} we discuss quantization and its effect on accuracy. We also compare quantization libraries and frameworks. We then present quantized accuracy results for a number of common networks. We present conclusions and provide guidance on appropriate application use in \autoref{sec:summary}. Finally, we present concluding comments in \autoref{sec:discussion}.

%% file: cnn-acc.tex
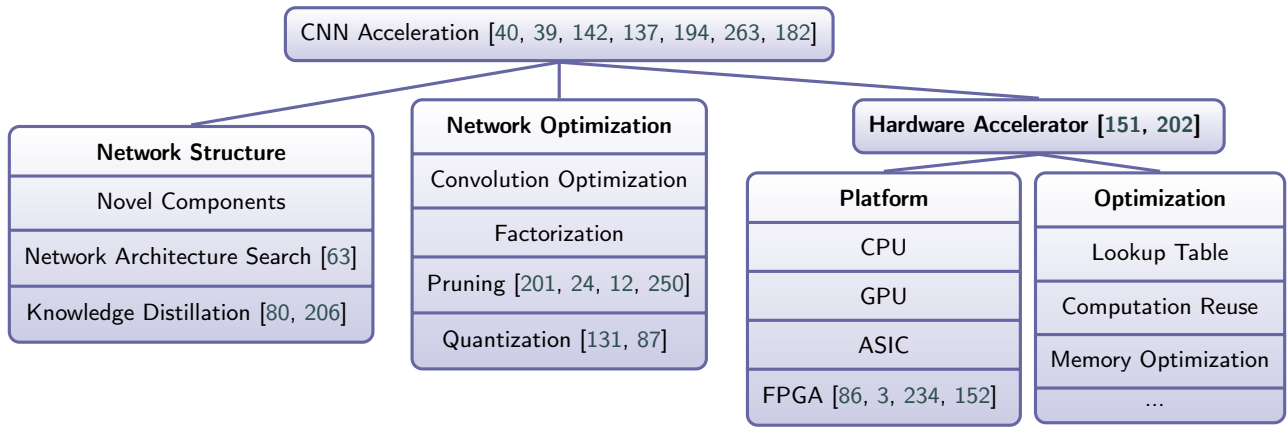
\begin{figure*}
    \centering
    \resizebox{0.98\textwidth}{!}{
        \begin{tikzpicture} [align = center] 
        \tikzset{
        grow=down,
        level 1/.style={
            sibling distance=15em,
            level distance=15ex
        },
        level 2/.style={
            sibling distance=10em,
            level distance=15ex
        },
        every node/.style={
            text ragged, inner sep=2mm
        },
        root/.style = {
            top color=white, bottom color=blue!50!black!20, draw=blue!40!black!60, 
            rectangle, rounded corners,
            minimum height = 4ex,
            very thick
        },
        punkt/.style={
            rectangle split, rounded corners, shade,
            top color=white, bottom color=blue!50!black!20, draw=blue!40!black!60, 
            very thick 
        },
        edge from parent/.style={very thick,draw=blue!40!black!60},
        edge from parent path={(\tikzparentnode.south) -- (\tikzchildnode.north)},
        }
        \node[root, anchor=south] (cnn_acc) {CNN Acceleration \cite{Cheng2017a, Cheng2018, Lei2018, Lebedev2018, Pilipovic2018, Zhang2019a, Neill2020}}
            child {
                node [punkt, rectangle split parts=5, below=1.5em of cnn_acc] (net_opt) {  
                    \textbf{Network Optimization}
                    \nodepart{two}{Convolution Optimization}
                    \nodepart{three}{Factorization}
                    \nodepart{four}{Pruning \cite{Reed1993,Blalock2020,Augasta2013,Xu2020a}}
                    \nodepart{five}{Quantization \cite{Krishnamoorthi2018, Guo2018a}}
                }
            }
            child {
                node [punkt, rectangle split parts=4, below=1.5em of cnn_acc, left=1.5em of net_opt] (net_stct) {
                    \textbf{Network Structure}
                    \nodepart{second}{Novel Components}
                    \nodepart{third}{Network Architecture Search \cite{Wistuba2019}}
                    \nodepart{fourth}{Knowledge Distillation \cite{Gou2020,Ruffy2019}}
                }
            }
            child {
                node [root, below=1.5em of cnn_acc, right=6em of net_opt.north east, anchor=north west] (hws) {
                    \textbf{Hardware Accelerator \cite{Li2017a, Reuther2019}}
                }
                    child {
                        node [punkt, rectangle split parts=5, 
                        below left=1em of hws.south
                        ](platform){
                            \textbf{Platform}
                            \nodepart{two}{CPU}
                            \nodepart{three}{GPU}
                            \nodepart{four}{ASIC}
                            \nodepart{five}{FPGA \cite{Guo2017, Abdelouahab2018, Venieris2018, Li2020a}}
                        }
                    }
                    child {
                        node [punkt, rectangle split parts=5, right=.5em of platform.north east, anchor=north west
                        ]{
                            \textbf{Optimization}
                            \nodepart{two}{Lookup Table}
                            \nodepart{three}{Computation Reuse}
                            \nodepart{four}{Memory Optimization}
                            \nodepart{five}{...}
                        }
                    }
            }
        ;
        \end{tikzpicture}
    }
    \caption{CNN Acceleration Approaches: Follow the sense from designing to implementing, CNN acceleration could fall into three categories, structure design (or generation), further optimization, and specialized hardware.}
    \label{fig:cnn_acc}
\end{figure*}

%% file: 1-cnn.tex
\section{Convolutional Neural Network} \label{sec:background}
Convolutional neural networks are a class of feed-forward DNNs that use convolution operations to extract features from a data source. CNNs have been most successfully applied to visual-related tasks however they have found use in natural language processing \cite{Hannun2014}, speech recognition \cite{Abdel-hamid2014}, recommendation systems \cite{Shen2016}, malware detection \cite{Sun2020}, and industrial sensors time series prediction \cite{Yuan2020}.
To provide a better understanding of optimization techniques, in this section, we introduce the two phases of CNN deployment - training and inference, discuss types of convolution operations, describe Batch Normalization (BN) as an acceleration technique for training, describe pooling as a technique to reduce complexity, and describe the exponential growth in parameters deployed in modern network structures.  

\subsection{Definitions} \label{sec:cnn-pre}
This section summarizes terms and definitions used to describe neural networks as well as acronyms collected in 
\autoref{tab:acronym}.
\begin{itemize}
    \item Coefficient - A constant by which an algebraic term is multiplied. Typically, a coefficient is multiplied by the data in a CNN filter.
    \item Parameter - All the factors of a layer, including coefficients and biases.
    \item Hyperparameter - A predefined parameter before network training, or fine-tunning (re-training).
    \item Activation ($\mathbf{A}\in\mathbb{R}^{h \times w \times c}$) - The activated (e.g., ReLu, Leaky, Tanh, etc.) output of one layer in a multi-layer network architecture, typically in height $h$, width $w$, and channel $c$. The $h \times w$ matrix is sometimes called an activation map. We also denote activation as output ($\mathbf{O}$) when the activation function does not matter.
    \item Feature ($\mathbf{F}\in\mathbb{R}^{h \times w \times c}$) - The input data of one layer, to distinguish the output $\mathbf{A}$. Generally the feature for the current layer is the activation of the previous layer.
    \item Kernel ($\mathbf{k} \in \mathbb{R}^{k_1 \times k_2}$) - Convolutional coefficients for a channel, excluding biases. Typically they are square (e.g. $k_1=k_2$) and sized 1, 3, 7.
    \item Filter ($\mathbf{w}\in\mathbb{R}^{k_1 \times k_2 \times c \times n}$) - Comprises all of the kernels corresponding to the $c$ channels of input features. The filter's number, $n$, results in different output channels.
    \item Weights - Two common uses: 1) kernel coefficients when describing part of a network, and 2) all the trained parameters in a neural network model when discussing the entire network.
\end{itemize}
\input{acronym}
\subsection{Training and Inference} \label{sec:cnn-tran-infer}
CNNs are deployed as a two step process: 1) training and 2) inference. Training is performed first with the result being either a continuous numerical value (regression) or a discrete class label (classification). Classification training involves applying a given annotated dataset as an input to the CNN, propagating it through the network, and comparing the output classification to the ground-truth label. The network weights are then updated typically using a backpropagation strategy such as Stochastic Gradient Descent (SGD) to reduce classification errors. This performs a search for the best weight values. Backpropogation is performed iteratively until a minimum acceptable error is reached or no further reduction in error is achieved. Backpropagation is compute intensive and traditionally performed in data centers that take advantage of dedicated GPUs or specialized training accelerators such as TPUs. 

Fine-tuning is defined as retraining a previously trained
model. It is easier to recover the accuracy of a quantized or pruned model with fine-tuning versus training from scratch.

CNN inference classification takes a previously trained classification model and predicts the class from input data not in the training dataset.
Inference is not as computationally intensive as training and can be executed on edge, mobile, and embedded devices. The size of the inference network executing on mobile devices may be limited due to memory, bandwidth, or processing constraints \cite{Gordon2018}. Pruning discussed in \autoref{sec:pruning} and quantization discussed in \autoref{sec:quantization} are two techniques that can alleviate these constraints. 
%
%

In this paper, we focus on the acceleration of CNN inference classification. We compare techniques using standard benchmarks such as ImageNet \cite{JiaDeng2009}, CIFAR \cite{Krizhevsky2009}, and MNIST \cite{LeCun1998}. The compression techniques are general and the choice of application domain doesn't restrict its use in object detection, natural language processing, etc.


\subsection{Convolution Operations} \label{sec:cnn-ops}

\begin{figure}[htb]
    \centering
    \includegraphics[width=0.95\linewidth,page=15]{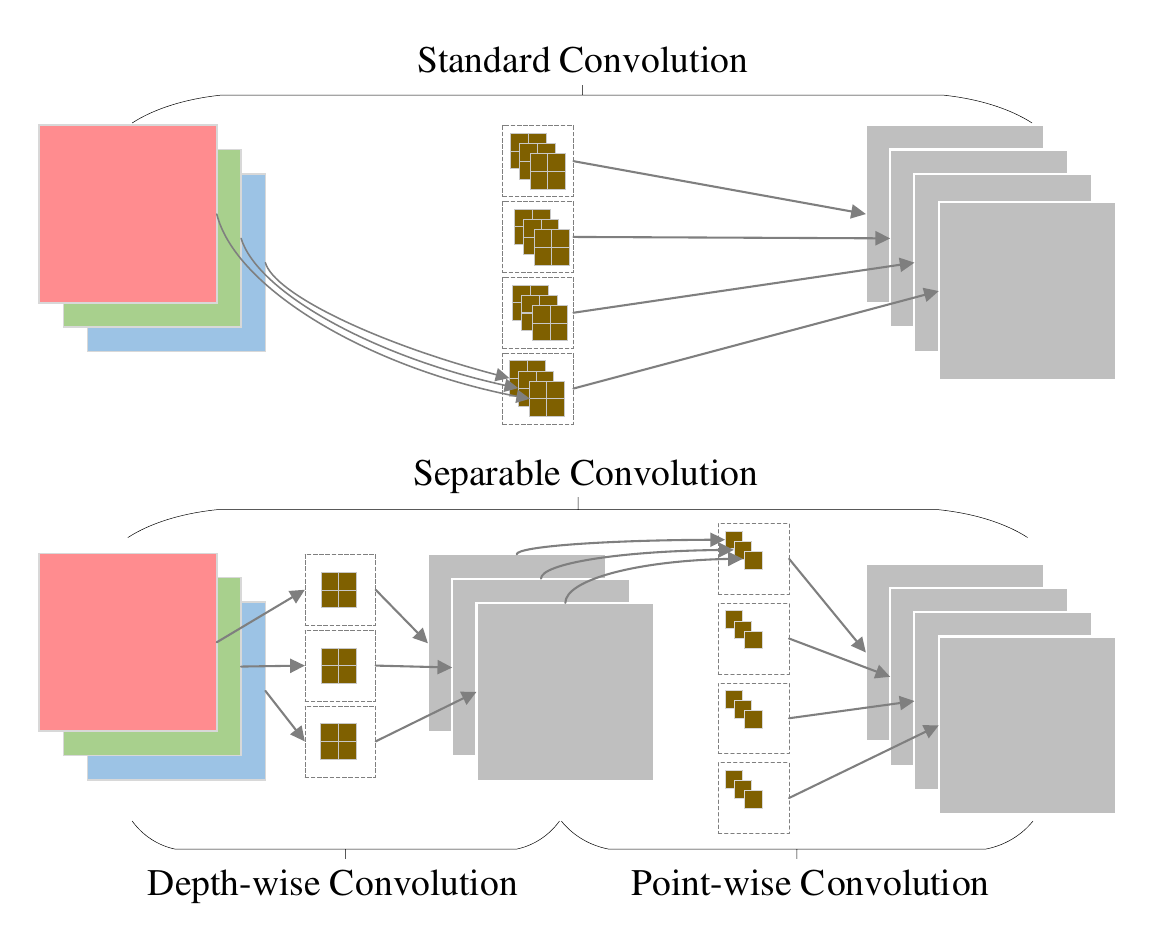}
    \caption{Separable Convolution: A standard convolution is decomposed into depth-wise convolution and point-wise convolution to reduce both the model size and computations.}
    \label{fig:separable-conv}
\end{figure}

The top of \autoref{fig:separable-conv} shows a 3-channel image (e.g., RGB) as input to a convolutional layer. Because the input image has 3 channels, the convolution kernel must also have 3 channels. 
In this figure four $2\times2\times3$ convolution filters are shown, each consisting of three $2\times2$ kernels. Data is received from all 3 channels simultaneously. 12 image values are multiplied with the kernel weights producing a single output. The kernel is moved across the 3-channel image sharing the 12 weights. If the input image is $12\times12\times3$ the resulting output will be $11\times11\times1$ (using a stride of 1 and no padding). The filters work by extracting multiple smaller bit maps known as feature maps. If more filters are desired to learn different features they can be easily added. In this case 4 filters are shown resulting in 4 feature maps.

\begin{figure}[htb]
    \centering
    \includegraphics[width=0.95\linewidth, page=7]{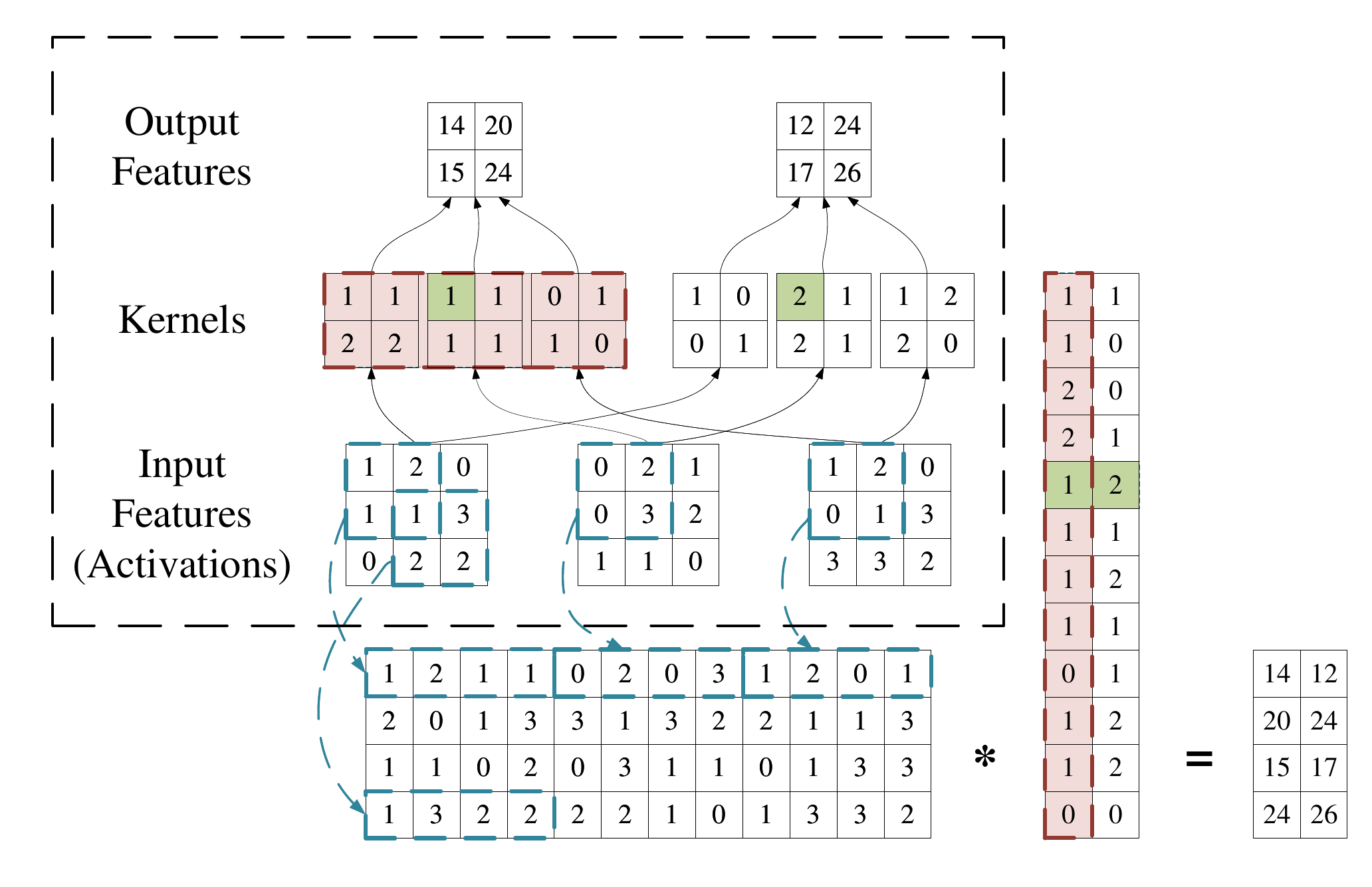}
    \caption{Convolution Performance Optimization: From traditional convolution (dot squared) to image to column (im2col) - GEMM approach, adopted from \cite{Chellapilla2006}. The red and green boxes indicate filter-wise and shape-wise elements, respectively.}
    \label{fig:im2col_and_gemm}
\end{figure}

The standard convolution operation can be computed in parallel using a GEneral Matrix Multiply (GEMM) library \cite{Dongarra1990}. \autoref{fig:im2col_and_gemm} shows a parallel column approach. The 3D tensors are first flattened into 2D matrices. The resulting matrices are multiplied by the convolutional kernel which takes each input neuron (features), multiplies it, and generates output neurons (activations) for the next layer \cite{Lecun2015}. 

\begin{equation}
    \label{eq:convolution}
    \begin{split}
        \mathbf{F}^{l+1}_{n} = \mathbf{A}^{l}_{n} = \operatorname{activate}\left\{\sum_{m=1}^{M}\left(\mathbf{W}^{l}_{mn} \ast \mathbf{F}^{l}_{m}\right) + \mathbf{b}^{l}_{n}\right\}
    \end{split}
\end{equation}

\autoref{eq:convolution} shows the layer-wise mathematical representation of the convolution layer where $\mathbf{W}$ represents the weights (filters) of the tensor with $m$ input channels and $n$ output channels, $\mathbf{b}$ represents the bias vector, and $\mathbf{F}^{l}$ represents the input feature tensor (typically from the activation of previous layer $\mathbf{A}^{l-1}$). $\mathbf{A}^l$ is the activated convolutional output. The goal of compression is to reduce the size of the $\mathbf{W}$ and $\mathbf{F}$ (or $\mathbf{A}$) without affecting accuracy. 

\begin{figure}[htb]
    \centering
    \includegraphics[width=0.95\linewidth,page=18]{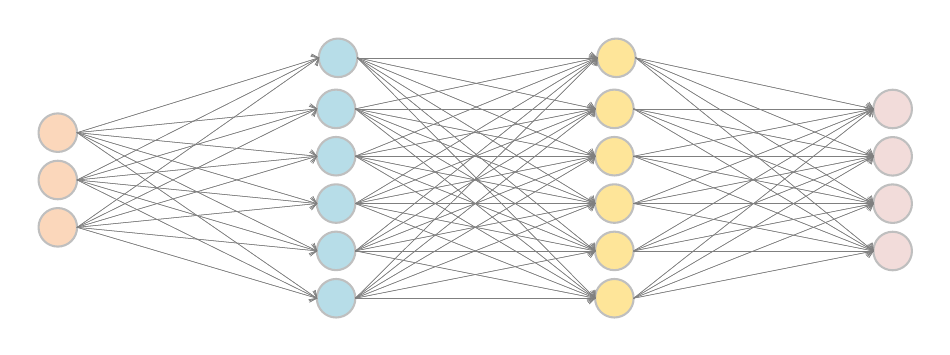}
    \caption{Fully Connected Layer: Each node in a layer connects to all the nodes in the next layer, and every line corresponds to a weight value}
    \label{fig:conn}
\end{figure}

\autoref{fig:conn} shows a FCL - also called dense layer or dense connect. Every neuron is connected to each other neuron in a crossbar configuration requiring many weights. 
As an example, if the input and output channel are 1024 and 1000, respectively, the number of parameters in the filter will be a million by $1024 \times 1000$.  
As the image size grows or the number of features increase, the number of weights grows rapidly. 

\subsection{Efficient Structure} \label{sec:cnn-novel}
The bottom of \autoref{fig:separable-conv} shows separable convolution
implemented in MobileNet \cite{Howard2017}. Separable convolution assembles a depth-wise convolution followed by a point-wise convolution. A depth-wise convolution groups the input feature by channel, and treats each channel as a single input tensor generating activations with the same number of channels. Point-wise convolution is a standard convolution with $1\times1$ kernels. It extracts mutual information across the channels with minimum computation overhead. 
For the $12\times12\times3$ image previously discussed, a standard convolution needs $2\times2\times3\times4$ multiplies to generate $1\times1\time4$ outputs. 
Separable convolution needs only $2\times2\times3$ for depth-wise convolution and $1\times1\times3\times4$ for point-wise convolution. This reduces computations by half from 48 to 24. The number of weights is also reduced from 48 to 24.

\begin{figure}[htb]
    \centering
    \includegraphics[width=0.95\linewidth,page=16]{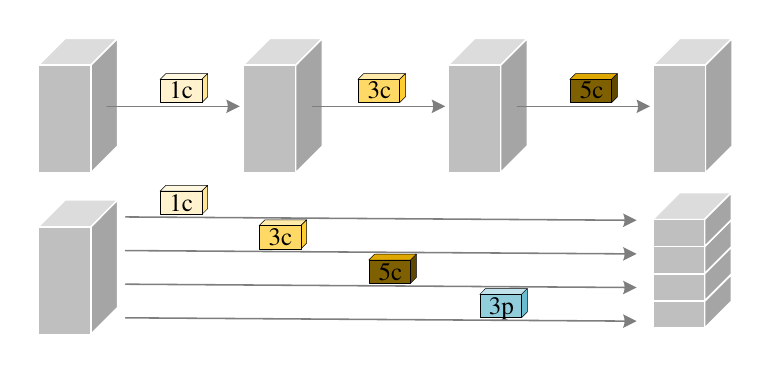}
    \caption{Inception Block: The inception block computes multiple convolutions with one input tensor in parallel, which extends the receptive field by mixing the size of kernels. The yellow - brown coloured cubes are convolutional kernels sized 1, 3, and 5. The blue cube corresponds to a $3\times3$ pooling operation.}
    \label{fig:inception-block}
\end{figure}
The receptive field is the size of a feature map used in a convolutional kernel. To extract data with a large receptive filed and high precision, cascaded layers should be applied as in the top of \autoref{fig:inception-block}. However, the number of computations can be reduced by expanding the network width with four types of filters as shown in \autoref{fig:inception-block}. The concatenated result performs better than one convolutional layer with same computation workloads \cite{Szegedy2015}. 

\begin{figure}[htb]
    \centering
    \includegraphics[width=0.95\linewidth,page=17]{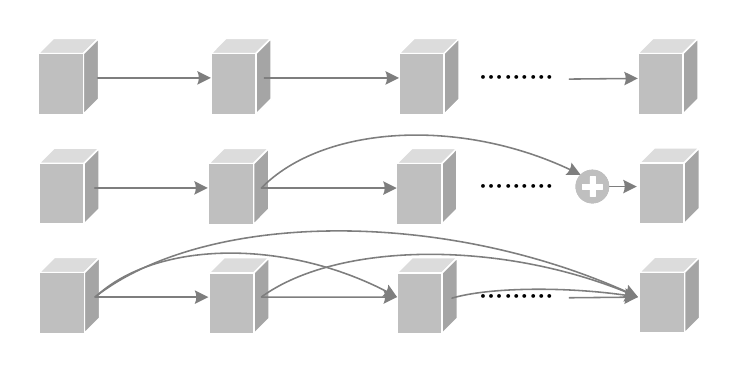}
    \caption{Conventional Network Block (top), Residual Network Block (middle), and Densely Connected Network Block (bottom)}
    \label{fig:residual-and-densenet}
\end{figure}
A residual network architecture block \cite{He2015} is a feed forward layer with a short circuit between layers as shown in the middle of \autoref{fig:residual-and-densenet}. The short circuit keeps information from the previous block to increase accuracy and avoid vanishing gradients during training. Residual networks help deep networks grow in depth by directly transferring information between deeper and shallower layers.

The bottom of \autoref{fig:residual-and-densenet} shows the densely connected convolutional block from DenseNets \cite{Huang2017a}, this block extends both the network depth and the receptive field by delivering the feature of former layers to all the later layers in a dense block using concatenation. ResNets transfer outputs from a single previous layer. DenseNets build connections across layers to fully utilize previous features. This provides weight efficiencies.

\subsection{Batch Normalization} \label{sec:cnn-bn}
BN was introduced in 2015 to speed up the training phase, and to improve the neural network performance \cite{Ioffe2015}. Most SOTA neural networks apply BN after a convolutional layer. BN addresses internal covariate shift (an altering of the network activation distribution caused by modifications to parameters during training) by normalizing layer inputs. This has been shown to reduce training time up to $14\times$.  Santurkar \cite{Santurkar2018} argues that the efficiency of BN is from its ability to smooth values during optimization.

\begin{equation}
    \begin{split}
        \mathbf{y}=\gamma \cdot \frac{\mathbf{x}-\mu}{\sqrt{\sigma^{2}+\epsilon}}+\beta
    \end{split}
\label{eq:batch-normalization}
\end{equation}

\autoref{eq:batch-normalization} gives the formula for computing inference BN, where $\mathbf{x} \text{ and } \mathbf{y}$ are the input feature and the output of BN, $\gamma$ and $\beta$ are learned parameters, $\mu$ and $\sigma$ are the mean value and standard deviation calculated from the training set, and $\epsilon$ is the additional small value (e.g., 1e-6) to prevent the denominator from being 0.
The variables of \autoref{eq:batch-normalization} are determined in the training pass and integrated into the trained weights. If the features in one channel share the same parameters, then it turns to a linear transform on each output channel. 
Channel-wise BN parameters potentially helps channel-wise pruning. BN could also raise the performance of the cluster-based quantize technique by reducing parameter dependency \cite{Choudhary2020}.


Since the parameters of the BN operation are not modified in the inference phase, they may be combined with the trained weights and biases. This is called BN folding or BN merging. \autoref{eq:bn-folding} show an example of BN folding. The new weight $\mathbf{W'}$ and bias $\mathbf{b'}$ are calculated using the pretrained weights $\mathbf{W}$ and BN parameters from \autoref{eq:batch-normalization}. Since the new weight is computed after training and prior to inference, the number of multiplies are reduced and therefore BN folding decreases inference latency and computational complexity.
\begin{equation}
        \mathbf{W'} =\gamma \cdot \frac{\mathbf{W}}{\sqrt{\sigma^{2}+\epsilon}}
        ,\quad
        \mathbf{b'} =\gamma \cdot \frac{\mathbf{b}-\mu}{\sqrt{\sigma^{2}+\epsilon}}+\beta
\label{eq:bn-folding}
\end{equation}


\subsection{Pooling} \label{sec:cnn-pool}
Pooling was first published in the 1980s with neocognitron \cite{Fukushima1988}. The technique takes a group of values and reduces them to a single value. The selection of the single replacement value can be computed as an average of the values (average pooling) or simply selecting the maximum value (max pooling). 

Pooling destroys spatial information as it is a form of down-sampling. The window size defines the area of values to be pooled. For image processing it is usually a square window with typical sizes being $2\times2$, $3\times3$ or $4\times4$. Small windows allow enough information to be propagated to successive layers while reducing the total number of computations \cite{Sun2017}. 

Global pooling is a technique where, instead of reducing a neighborhood of values, an entire feature map is reduced to a single value \cite{Lin2014}. Global Average Pooling (GAP) extracts information from multi-channel features and can be used with dynamic pruning \cite{Lin2017, Zhang2019d}.

Capsule structures have been proposed as an alternative to pooling. Capsule networks replace the scalar neuron with vectors. The vectors represent a specific entity with more detailed information, such as position and size of an object. Capsule networks void loss of spatial information by capturing it in the vector representation. Rather than reducing a neighborhood of values to a single value, capsule networks perform a dynamic routing algorithm to remove connections \cite{Sabour2017}.

\subsection{Parameters} \label{sec:cnn-param}



\begin{figure}
    \centering
    \includegraphics[width=0.98\linewidth,page=2]{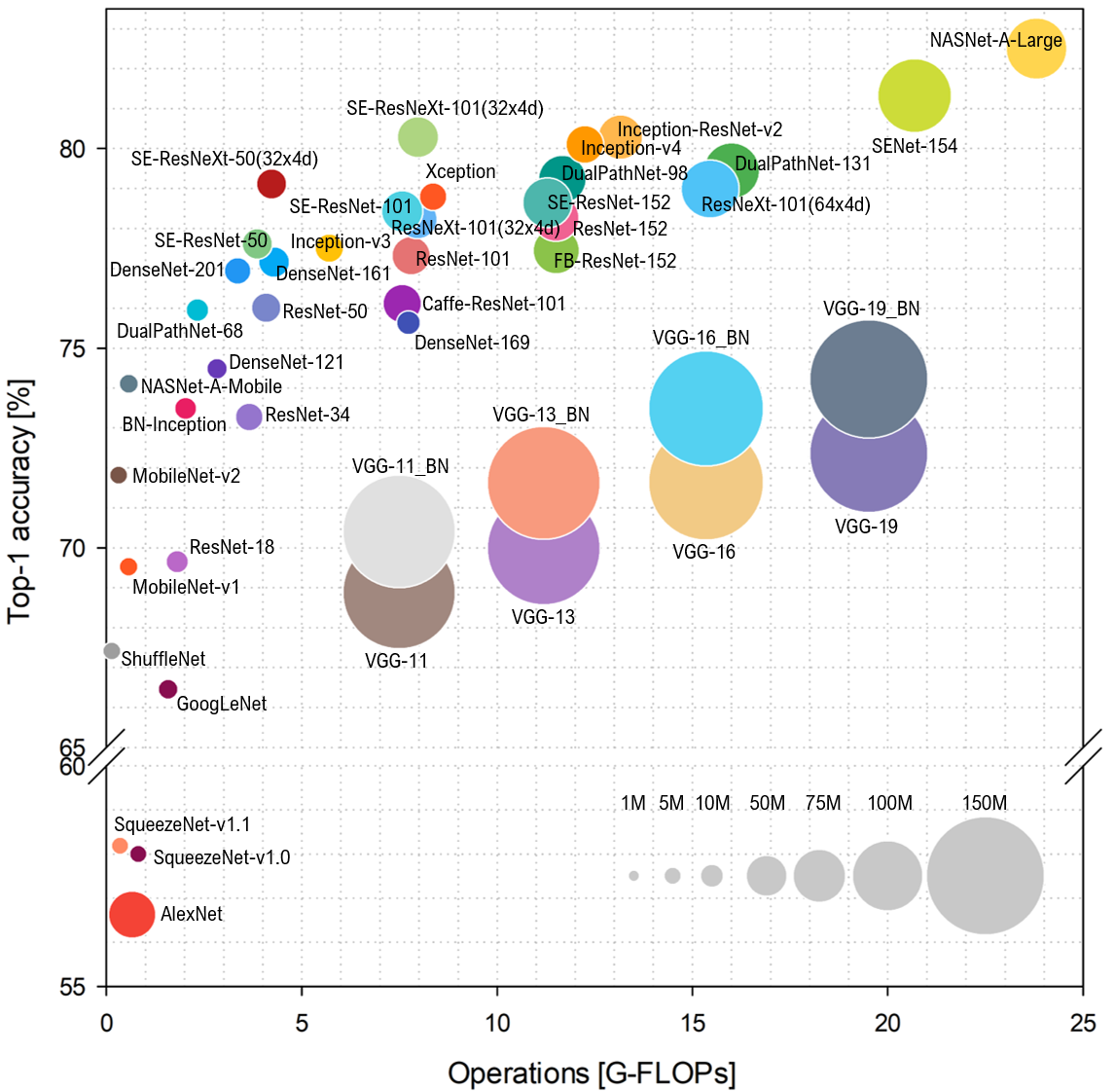}
    \caption{Popular CNN Models: Top-1 accuracy vs GFLOPs and model size, adopted from \cite{Bianco2018}}
    \label{fig:model-size-acc}
\end{figure}

\autoref{fig:model-size-acc} show top-1 accuracy percent verses the number of operations needed for a number of popular neural networks \cite{Bianco2018}. The number of parameters in each network is represented by the size of the circle. A trend (not shown in the figure) is a yearly increase in parameter complexity. In 2012, AlexNet \cite{Krizhevsky2012} was published with 60 million parameters. In 2013, VGG \cite{Simonyan2014} was introduced with 133 million parameters and achieved 71.1\% top-1 accuracy. These were part of the ImageNet large scale visual recognition challenge (ILSVRC) \cite{Russakovsky2015}. The competition's metric was top-1 absolute accuracy. Execution time was not a factor. This incentivized neural network designs with significant redundancy. As of 2020, models with more than 175 billion parameters have been published \cite{Brown2020}. 

Networks that execute in data centers can accommodate models with a large number of parameters. In resource constrained environments such as edge and mobile deployments, reduced parameter models have been designed. For example, GoogLeNet \cite{Szegedy2015} achieves similar top-1 accuracy of 69.78\% as VGG-16 but with only 7 million parameters. MobileNet \cite{Howard2017} has 70\% top-1 accuracy with only 4.2 million parameters and only 1.14 Giga FLoating-point OPerations  (GFLOPs).
A more detailed network comparison can be found in \cite{Albanie2020}.


%% file: acronym.tex
\begin{table}[h]
  \centering
  \caption{Acronyms and Abbreviations}
    \resizebox{0.98\linewidth}{!}{\begin{tabular}{ll}
    \toprule
    Acronym & Explanation \\
    \midrule
    2D & Two Dimensional \\
    3D & Three Dimensional  \\
    FP16 & 16-Bit Floating-Point \\
    FP32 & 32-Bit Floating-Point  \\
    INT16 & 16-Bit Integer  \\
    INT8 & 8-Bit Integer  \\
    IR & Intermediate Representation  \\
    OFA & One-For-All  \\
    RGB & Red, Green, And Blue \\
    SOTA & State of The Art  \\
    \midrule
    AI & Artificial Inteligence \\
    BN & Batch Normalization \\
    CBN & Conditional Batch Normalization \\
    CNN & Convolutional Neural Network \\
    DNN & Deep Neural Network \\
    EBP & Expectation Back Propagation  \\
    FCL & Fully Connected Layer \\
    FCN & Fully Connected Networks  \\
    FLOP & Floating-Point Operation  \\
    GAP & Global Average Pooling \\
    GEMM & General Matrix Multiply \\
    GFLOP & Giga Floating-Point Operation \\
    ILSVRC & Imagenet Large Visual Recognition Challenge \\
    Im2col & Image To Column \\
    KD  & Knowledge Distillation \\
    LRN & Local Response Normalization  \\
    LSTM & Long Short Term Memory  \\
    MAC & Multiply Accumulate  \\
    NAS & Network Architecture Search \\
    NN & Neural Network \\
    PTQ & Post Training Quantization  \\
    QAT & Quantization Aware Training  \\
    ReLU & Rectified Linear Unit \\
    RL & Reinforcement Learning  \\
    RNN & Recurrent Neural Network  \\
    SGD & Stochastic Gradient Descent \\
    STE & Straight-Through Estimator  \\
    \midrule
    ASIC & Application Specific Integrated Circuit \\
    AVX-512 & Advance Vector Extension 512  \\
    CPU & Central Processing Unit \\
    CU & Computing Unit \\
    FPGA & Field Programmable Gate Array \\
    GPU & Graphic Processing Unit \\
    HSA & Heterogeneous System Architecture  \\
    ISA & Instruction Set Architectures  \\
    PE & Processing Element  \\
    SIMD & Single Instruction Multiple Data \\
    SoC & System on Chip  \\
    \midrule
    DPP & Determinantal Point Process  \\
    FFT & Fast Fourier Transfer  \\
    FMA & Fused Multiply-Add  \\
    KL-divergence & Kullback-Leibler Divergence  \\
    LASSO & Least Absolute Shrinkage And Selection Operator \\
    MDP & Markov Decision Process  \\
    OLS & Ordinary Least Squares \\
    \bottomrule
    \end{tabular}}%
  \label{tab:acronym}%
\end{table}%

%% file: 2-prune.tex
\section{Pruning} \label{sec:pruning}
Network pruning is an important technique for both memory size and bandwidth reduction. 
In the early 1990s, pruning techniques were developed to reduce a trained large network into a smaller network without requiring retraining \cite{Reed1993}. This allowed neural networks to be deployed in constrained environments such as embedded systems. Pruning removes redundant parameters or neurons that do not significantly contribute to the accuracy of results. This condition may arise when the weight coefficients are zero, close to zero, or are replicated. Pruning consequently reduces the computational complexity. If pruned networks are retrained it provides the possibility of escaping a previous local minima \cite{Choi2008} and further improve accuracy. 

Research on network pruning can roughly be categorized as sensitivity calculation and penalty-term methods \cite{Reed1993}. Significant recent research interest has continued showing improvements for both network pruning categories or a further combination of them.

\begin{figure*}
    \centering
    \includegraphics[width=0.95\textwidth, page=13]{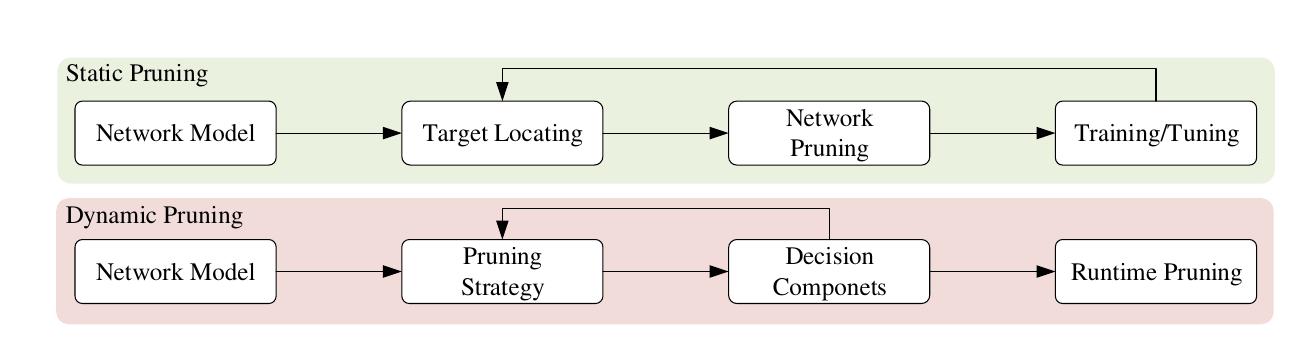}
    \caption{Pruning Categories: \textit{Static pruning} is performed offline prior to inference while \textit{Dynamic pruning} is performed at runtime.}
    \label{fig:prune-class}
\end{figure*} 

Recently, new network pruning techniques have been created. Modern pruning techniques may be classified by various aspects including: 1) \textit{structured} and \textit{unstructured pruning} depending if the pruned network is symmetric or not, 2) \textit{neuron} and \textit{connection pruning} depending on the pruned element type, or 3) \textit{static} and \textit{dynamic pruning}. \autoref{fig:prune-class} shows the processing differences between static and dynamic pruning. \textit{Static pruning} has all pruning steps performed offline prior to inference while \textit{dynamic pruning} is performed during runtime. While there is overlap between the categories, in this paper we will use \textit{static pruning} and \textit{dynamic pruning} for classification of network pruning techniques. 

\begin{figure}[htb]
    \centering
    \includegraphics[width=0.95\linewidth, page=6]{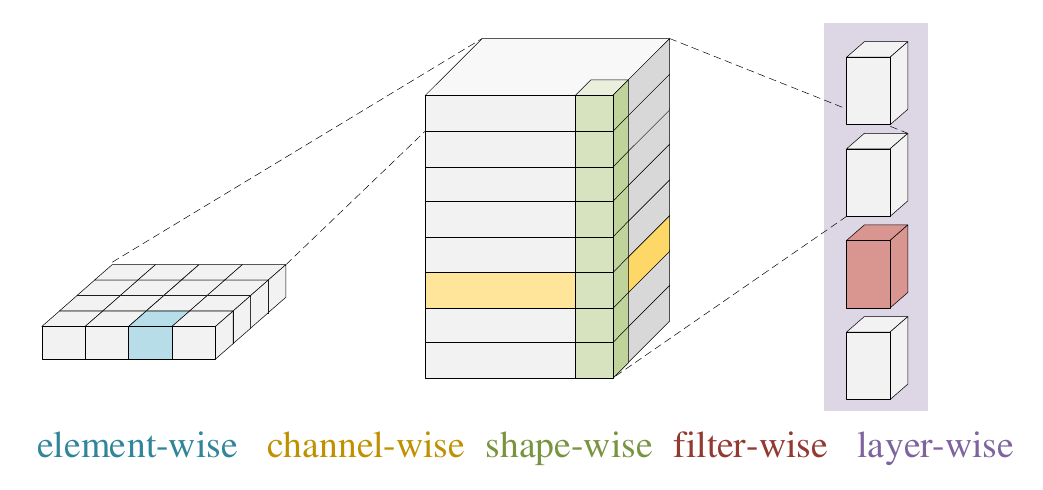}
    \caption{Pruning Opportunities: Different network sparsity results from the granularity of pruned structures. Shape-wise pruning was proposed by Wen \cite{Wen2016a}.}
    \label{fig:pruning-shapes}
\end{figure}

\autoref{fig:pruning-shapes} shows a granularity of pruning opportunities. The four rectangles on the right side correspond to the four brown filters in the top of \autoref{fig:separable-conv}. Pruning can occur on an element-by-element, row-by-row, column-by-column, filter-by-filter, or layer-by-layer basis. Typically element-by-element has the smallest sparsity impact, and results in a unstructured model. Sparsity decreases from left-to-right in \autoref{fig:pruning-shapes}.

\begin{equation}
\label{eq:pruning}
\begin{split}
\arg\min_{p} \  {L}={N(x;\mathbf{W}) - N_p(x;\mathbf{W}_p)} \\
\text{where} \  N_p(x;\mathbf{W}_p) = P\left(N\left(x;\mathbf{W}\right)\right) \\
\end{split}
\end{equation}

Independent of categorization, pruning can be described mathematically as \autoref{eq:pruning}. ${N}$ represents the entire neural network which contains a series of layers (e.g., convolutional layer, pooling layer, etc.) with $x$ as input. ${L}$ represents the pruned network with $N_p$ performance loss compared to the unpruned network. Network performance is typically defined as accuracy in classification. 
The pruning function, $P(\cdot)$, results in a different network configuration ${N}_p$ along with the pruned weights $\mathbf{W}_p$. The following sections are primarily concerned with the influence of $P(\cdot)$ on ${N}_p$. We also consider how to obtain $\mathbf{W}_p$.


\subsection{Static Pruning} \label{sec:pruning-static}
Static pruning is a network optimization technique that removes neurons offline from the network after training and before inference. During inference, no additional pruning of the network is performed. 
Static pruning commonly has three parts: 1) selection of parameters to prune, 2) the method of pruning the neurons, and 3) optionally fine-tuning or re-training \cite{Han2015}.
Retraining may improve the performance of the pruned network to achieve comparable accuracy to the unpruned network but may require significant offline computation time and energy.
\subsubsection{Pruning Criteria} \label{sec:pruning-static-criteria}
As a result of network redundancy, neurons or connections can often be removed without significant loss of accuracy.
As shown in \autoref{eq:convolution}, the core operation of a network is a convolution operation. It involves three parts: 1) input features as produced by the previous layer, 2) weights produced from the training phase, and 3) bias values produced from the training phase. The output of the convolution operation may result in either zero valued weights or features that lead to a zero output. Another possibility is that similar weights or features may be produced. These may be merged for distributive convolutions.

An early method to prune networks is brute-force pruning. In this method the entire network is traversed element-wise and weights that do not affect accuracy are removed. A disadvantage of this approach is the large solution space to traverse. 
A typical metric to determine which values to prune is given by the $l_p\text{-norm, s.t. } p\in\{N, \infty\}$, where $N$ is natural number. The $l_p$-norm of a vector $\mathbf{x}$ which consists of $n$ elements is mathematically described by \autoref{eq:lp-norm}.
\begin{equation}
\label{eq:lp-norm}
\|\mathbf{x}\|_{p}=\left(\sum_{i=1}^{n}\left|x_{i}\right|^{p}\right)^\frac{1}{p}
\end{equation}

Among the widely applied measurements, the $l_1$-norm is also known as the \textit{Manhattan norm} and the $l_2$-norm is also known as the \textit{Euclidean norm}. 
The corresponding $l_1$ and $l_2$ regularization have the names \textit{LASSO} (least absolute shrinkage and selection operator) and \textit{Ridge}, respectively \cite{Tishbirani1996}. 
The difference between the $l_2$-norm pruned tensor and an unpruned tensor is called the $l_2$-distance.
Sometimes researchers also use the term $l_0$-norm defined as the total number of nonzero elements in a vector.

\begin{equation}
\label{eq:lasso-ori}
\begin{split}
    &\arg\,\min_{\alpha, \beta} \left\{\sum_{i=1}^{N}\left(y_{i}-\alpha-\sum_{j=1}^{p} \beta_{j} \mathbf{x}_{i j}\right)^{2}\right\} \\
    &\text {subject to} \sum_{j}^{p}\left|\beta_{j}\right| \leqslant t
\end{split}
\end{equation}

Equation \autoref{eq:lasso-ori} mathematically describes $l_2$ LASSO regularization. Consider a sample consisting of $N$ cases, each of which consists of $p$ covariates and a single outcome $y_i$. Let $x^i=(x_{i1},...,x_{ip})^T$ be the standardized covariate vector for the $i$-th case (input feature in DNNs), so we have $\sum_ix_{ij}/N=0,\ \sum_ix_{ij}^2/N=1$. $\beta$ represents the coefficients ${\beta}=({\beta_1},...,{\beta_p})^T$ (weights) and $t$ is a predefined tunning parameter that determines the sparsity. The LASSO estimate $\alpha$ is 0 when the average of $y_i$ is 0 because for all $t$, the solution for $\alpha$ is $\alpha = \overline{y}$. If the constraint is $\sum_{j}^{p} \beta_j^2 \leqslant t$ then the \autoref{eq:lasso-ori} becomes Ridge regression. Removing the constraint will results in the Ordinary Least Squares (OLS) solution.

\begin{equation}
\label{eq:lasso-simp}
\begin{split}
    arg\,\min_{\beta \in \mathbb{R}}\left\{\frac{1}{N}\left\|y - \mathbf{X}\beta\right\|_2^2+\lambda\left\|\beta\right\|_1\right\}
\end{split}
\end{equation}

\autoref{eq:lasso-ori} can be simplified into the so-called Lagrangian form shown in \autoref{eq:lasso-simp}. The Lagrangian multiplier translates the objective function $f(x)$ and constraint $g(x)=0$ into the format of $\mathcal{L}(x, \lambda)=f(x)-\lambda g(x)$,  Where the $\|\cdot\|_p$ is the standard $l_p\text{-norm}$, the $\mathbf{X}$ is the covariate matrix that contains $x_{ij}$, and $\lambda$ is the data dependent parameter related to $t$ from \autoref{eq:lasso-ori}.

Both magnitude-based pruning and penalty based pruning may generate zero values or near-zero values for the weights. In this section we discuss both methods and their impact.
\paragraph{Magnitude-based pruning:}
It has been proposed and is widely accepted that trained weights with large values are more important than trained weights with smaller values  \cite{Lei2017}. This observation is the key to magnitude-based methods.
Magnitude-based pruning methods seek to identify unneeded weights or features to remove them from runtime evaluation. Unneeded values may be pruned either in the kernel or at the activation map. The most intuitive magnitude-based pruning methods is to prune all zero-valued weights or all weights within an absolute value threshold.

LeCun as far back as 1990 proposed Optimal Brain Damage (OBD) to prune single non-essential weights \cite{LeCun1990}. By using the second derivative (Hessian matrix) of the loss function, this static pruning technique reduced network parameters by a quarter. For a simplified derivative computation, OBD functions under three assumptions: 1) \textit{quadratic} - the cost function is near-quadratic, 2) \textit{extremal} - the pruning is done after the network converged, and 3) \textit{diagonal} - sums up the error of individual weights by pruning the result of the error caused by their co-consequence.
This research also suggested that the sparsity of DNNs could provide opportunities to accelerate network performance.
Later Optimal Brain Surgeon (OBS) \cite{Hassibi1993} extended OBD with a similar second-order method but removed the \textit{diagonal} assumption in OBD. OBS considers the Hessian matrix is usually non-diagonal for most applications. OBS improved the neuron removal precision with up to a 90\% reduction in weights for XOR networks.

These early methods reduced the number of connections based on the second derivative of the loss function. The training procedure did not consider future pruning but still resulted in networks that were amenable to pruning. They also suggested that methods based on Hessian pruning would exhibit higher accuracy than those pruned with only magnitude-based algorithms \cite{Hassibi1993}.
More recent DNNs exhibit larger weight values when compared to early DNNs. Early DNNs were also much shallower with orders of magnitude less neurons. GPT-3 \cite{Brown2020}, for example, contains 175-billion parameters while VGG-16 \cite{Simonyan2014} contains just 133-million parameters. Calculating the Hessian matrix during training for networks with the complexity of GPT-3 is not currently feasible as it has the complexity of $O(W^2)$. Because of this simpler magnitude-based algorithms have been developed \cite{Molchanov2017, Lee2019a}.

Filter-wise pruning \cite{Li2017} uses the $l_1$-norm to remove filters that do not affect the accuracy of the classification. Pruning entire filters and their related feature maps resulted in a reduced inference cost of 34\% for VGG-16 and 38\% for ResNet-110 on the CIFAR-10 dataset with improved accuracy 0.75\% and 0.02\%, respectively.

Most network pruning methods choose to measure weights rather than activations when rating the effectiveness of pruning \cite{Guo2016}.
However, activations may also be an indicator to prune corresponding weights. Average Percentage Of Zeros (APoZ) \cite{Hu2016} was introduced to judge if one output activation map is contributing to the result. Certain activation functions, particularly rectification such as Rectified Linear Unit (ReLU), may result in a high percentage of zeros in activations and thus be amenable to pruning. \autoref{eq:apoz} shows the definition of $\operatorname{APoZ}_{c}^{(i)}$ of the $c$-th neuron in the $i$-th layer, where $\mathbf{O}_{c}^{(i)}$ denotes the activation, $N$ is the number of calibration (validation) images, and $M$ is the dimension of activation map. $f(\text{true})=1$ and $f(\text{false})=0$.
\begin{equation}
\label{eq:apoz}
\operatorname{APoZ}_{c}^{(i)}=
\operatorname{APoZ}\left(\mathbf{O}_{c}^{(i)}\right)=\frac{\sum\limits_{k=0}^{N} \sum\limits_{j=0}^{M} f\left(\mathbf{O}_{c, j}^{(i)}(k)=0\right)}{N \times M}
\end{equation}
Similarly, inbound pruning \cite{Polyak2015}, also an activation technique, considers channels that do not contribute to the result. If the top activation channel in the standard convolution of \autoref{fig:separable-conv} are determined to be less contributing, the corresponding channel of the filter in the bottom of the figure will be removed. After pruning this technique achieved about $1.5\times$ compression.

Filter-wise pruning using a threshold from the sum of filters' absolute values can directly take advantage of the structure in the network. In this way, the ratio of pruned to unpruned neurons (i.e. the pruning ratio) is positively correlated to the percentage of kernel weights with zero values, which can be further improved by penalty-based methods. 

\paragraph{Penalty-based pruning:}
In penalty-based pruning, the goal is to modify an error function or add other constraints, known as bias terms, in the training process. A penalty value is used to update some weights to zero or near zero values. These values are then pruned. 

Hanson \cite{HANSON1989} explored hyperbolic and exponential bias terms for pruning in the late 80s. This method uses weight decay in backpropagation to determine if a neuron should be pruned. Low-valued weights are replaced by zeros. Residual zero valued weights after training are then used to prune unneeded neurons. 

Feature selection \cite{Dash1997} is a technique that selects a subset of relevant features that contribute to the result. It is also known as attribute selection or variable selection. Feature selection helps algorithms avoiding over-fitting and accelerates both training and inference by removing features and/or connections that don't contribute to the results. Feature selection also aids model understanding by simplifying them to the most important features. Pruning in DNNs can be considered to be a kind of feature selection \cite{JianchangMao1994}.

LASSO was previously introduced as a penalty term. LASSO shrinks the least absolute valued feature's corresponding weights. This increases weight sparsity. This operation is also referred to as LASSO feature selection and has been shown to perform better than traditional procedures such as OLS by selecting the most significantly contributed variables instead of using all the variables. This lead to approximately 60\% more sparsity than OLS \cite{Muthukrishnan2016}.

Element-wise pruning may result in an unstructured network organizations. This leads to sparse weight matrices that are not efficiently executed on instruction set processors. In addition they are usually hard to compress or accelerate without specialized hardware support \cite{Han2016}. Group LASSO \cite{Yuan2006} mitigates these inefficiencies by using a structured pruning method that removes entire groups of neurons while maintaining structure in the network organization \cite{Foroosh2015}. 

Group LASSO is designed to ensure that all the variables sorted into one group could be either included or excluded as a whole. \autoref{eq:group-lasso} gives the pruning constraint where $\mathbf{X}$ and $\beta$ in \autoref{eq:lasso-simp} are replaced by the higher dimensional $\mathbf{X_j}$ and $\beta_j$ for the $j$ groups.

\begin{equation}
\label{eq:group-lasso}
\begin{split}
    &arg\,\min_{\beta \in \mathbb{R}^{p}}\left\{\left\|y-\sum_{j=1}^{J} \mathbf{X}_{j} \beta_{j}\right\|_{2}^{2}+\lambda \sum_{j=1}^{J}\left\|\beta_{j}\right\|_{K_{j}}\right\} \\
\end{split}
\end{equation}

\begin{figure}
    \centering
    \includegraphics[width=.95\linewidth]{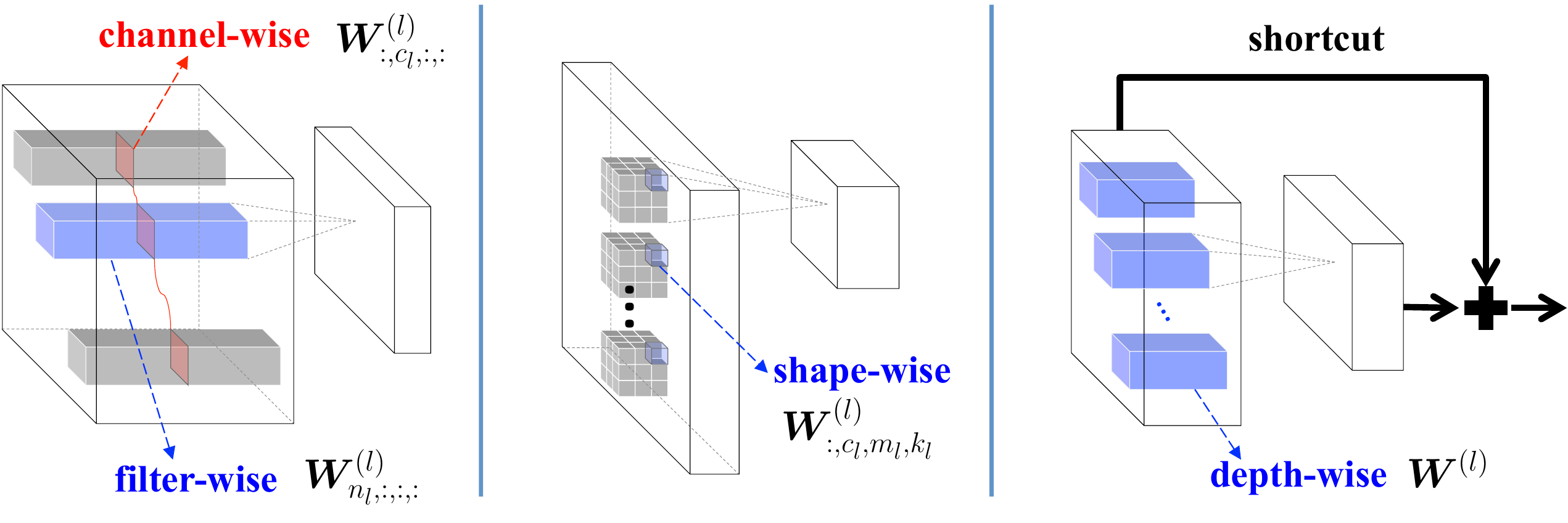}
    \caption{Types of Sparsity Geometry, adopted from \cite{Wen2016a}}
    \label{fig:ssl}
\end{figure}
\autoref{fig:ssl} shows Group LASSO with group shapes used in Structured Sparsity Learning (SSL) \cite{Wen2016a}. Weights are split into multiple groups. Unneeded groups of weights are removed using LASSO feature selection. Groups may be determined based on geometry, computational complexity, group sparsity, etc. SSL describes an example where group sparsity in row and column directions may be used to reduce the execution time of GEMM. SSL has shown improved inference times on AlexNet with both CPUs and GPUs by $5.1\times$ and $3.1\times$, respectively.

Group-wise brain damage \cite{Lebedev2016} also introduced the group LASSO constraint but applied it to filters. This simulates brain damage and introduces sparsity. It achieved $2\times$ speedup with 0.7\% ILSVRC-2012 accuracy loss on the VGG Network.

Sparse Convolutional Neural Networks (SCNN) \cite{Foroosh2015} take advantage of two-stage tensor decomposition. By decomposing the input feature map and convolutional kernels, the tensors are transformed into two tensor multiplications. Group LASSO is then applied. SCNN also proposed a hardware friendly algorithm to further accelerate sparse matrix computations. They achieved $2.47\times$ to $6.88\times$ speed-up on various types of convolution.

Network slimming \cite{Liu2017a} applies LASSO on the scaling factors of BN. BN normalizes the activation by statistical parameters which are obtained during the training phase. Network slimming has the effect of introducing forward invisible additional parameters without additional overhead. 
Specifically, by setting the BN scaler parameter to zero, channel-wise pruning is enabled. They achieved 82.5\% size reduction with VGG and 30.4\% computation compression without loss of accuracy on ILSVRC-2012.  

Sparse structure selection \cite{Huang2018a} is a generalized network slimming method. It prunes by applying LASSO to sparse scaling factors in neurons, groups, or residual blocks. Using an improved gradient method, Accelerated Proximal Gradient (APG), the proposed method shows better performance without fine-tunning achieving $4\times$ speed-up on VGG-16 with 3.93\% ILSVRC-2012 top-1 accuracy loss.

\paragraph{Dropout:}
While not specifically a technique to prune networks, dropout does reduce the number of parameters \cite{Srivastava2014}. It was originally designed as a stochastic regularizer to avoid over-fitting of data \cite{Hinton2012a}. The technique randomly omits a percentage of neurons typically up to 50\%, This \textit{dropout} operation breaks off part of the connections between neurons to avoid co-adaptations. Dropout could also be regarded as an operation that separately trains many sub-networks and takes the average of them during the inference phase. Dropout increases training overhead but it does not affect the inference time. 

Sparse variational dropout \cite{Molchanov2017a} added a dropout hyperparameter called the dropout rate to reduce the weights of VGG-like networks by $68\times$. During training the dropout rate can be used to identify single weights to prune. This can also be applied with other compression approaches for further reduction in weights.  

\paragraph{Redundancies:}
The goal of norm-based pruning algorithms is to remove zeros. This implies that the distribution of values should wide enough to retain some values but contain enough values close to zero such that a smaller network organization is still accurate. This does not hold in some circumstances. For example, filters that have small norm deviations or a large minimum norm have small search spaces making it difficult to prune based on a threshold \cite{He2018}. 
Even when parameter values are wide enough, in some networks smaller values may still play an important role in producing results. One example of this is when large valued parameters saturate \cite{Engelbrecht2001}. In these cases magnitude-based pruning of zero values may decrease result accuracy. 

Similarly, penalty-based pruning may cause network accuracy loss. In this case, the filters identified as unneeded due to similar coefficient values in other filters may actually be required. Removing them may significantly decrease network accuracy \cite{Guo2016}. \autoref{sec:prunning-static-tunning} describes techniques to undo pruning by tuning the weights to minimize network loss while this section describes redundancy based pruning. 

Using BN parameters, feature map channel distances can be computed by layer \cite{Zhang2019}. Using a clustering approach for distance, nearby features can be tuned. An advantage of clustering is that redundancy is not measured with an absolute distance but a relative value. With about 60 epochs of training they were able to prune the network resulting in a 50\% reduction in FLOPs (including non-convolutional operations) with a reduction in accuracy of only 1\% for both top-1 and top-5 on the ImageNet dataset. 

Filter pruning via geometric median (FPGM) \cite{He2018} identifies filters to prune by measuring the $l_2$-distance using the geometric median. FPGM found 42\% FLOPs reduction with 0.05\% top-1 accuracy drop on ILSVRC-2012 with ResNet-101.

The reduce and reused (also described as outbound) \linebreak[4] method \cite{Polyak2015} prunes entire filters by computing the statistical variance of each filter's output using a calibration set. Filters with low variance are pruned. The outbound method obtained $2.37\times$ acceleration with 1.52\% accuracy loss on Labeled Faces in the Wild (LFW) dataset \cite{Huang2014} in the filed of face recognition.

A method that iteratively removes redundant neurons for FCLs without requiring special validation data is proposed in \cite{Srinivas2015}. This approach measures the similarity of weight groups after a normalization. It removes redundant weights and merges the weights into a single value. This lead to a 34.89\% reduction of FCL weights on AlexNet with 2.24\% top-1 accuracy loss on ILSVRC-2012.

Comparing with the similarity based approach above, DIVersity NETworks (DIVNET) \cite{Mariet2016} considers the calculation redundancy based on the activations. DIVNET introduces Determinantal Point Process (DPP) \cite{Macchi1975} as a pruning tool. DPP sorts neurons into categories including dropped and retained. Instead of forcing the removal of elements with low contribution factors, they fuse the neurons by a process named re-weighting. Re-weighting works by minimizing the impact of neuron removal. This minimizes pruning influence and mitigates network information loss. They found 3\% loss on CIFAR-10 dataset when compressing the network into half weight. 

ThiNet \cite{Luo2017} adopts statistics information from the next layer to determine the importance of filters. It uses a greedy search to prune the channel that has the smallest reconstruction cost in the next layer. ThiNet prunes layer-by-layer instead of globally to minimize large errors in classification accuracy. It also prunes less during each training epoch to allow for coefficient stability. The pruning ratio is a predefined hyper-parameter and the runtime complexity is directly related to the pruning ratio. ThiNet compressed ResNet-50 FLOPs to 44.17\% with a top-1 accuracy reduction of 1.87\%.  

He \cite{He2017} adopts LASSO regression instead of a greedy algorithm to estimate the channels. Specifically, in one iteration, the first step is to evaluate the most important channel using the $l_1$-norm. The next step is to prune the corresponding channel that has the smallest Mean Square Error (MSE). Compared to an unpruned network, this approach obtained $2\times$ acceleration of ResNet-50 on ILSVRC-2012 with about 1.4\% accuracy loss on top-5, and a $4\times$ reduction in execution time with top-5 accuracy loss of 1.0\% for VGG-16. The authors categorize their approach as dynamic inference-time channel pruning. However it requires 5000 images for calibration with 10 samples per image and more importantly results in a statically pruned network. Thus we have placed it under static pruning.


\subsubsection{Pruning combined with Tuning or Retraining} \label{sec:prunning-static-tunning}
Pruning removes network redundancies and has the benefit of reducing the number of computations without significant impact on accuracy for some network architectures. However, as the estimation criterion is not always accurate, some important elements may be eliminated resulting in a decrease in accuracy. Because of the loss of accuracy, time-consuming fine-tuning or re-training may be employed to increase accuracy \cite{Yu2017a}.

Deep compression \cite{Han2015}, for example, describes a static method to prune connections that don't contribute to classification accuracy. In addition to feature map pruning they also remove weights with small values. After pruning they re-train the network to improve accuracy. This process is performed iteratively three times resulting in a $9\times$ to $13\times$ reduction in total parameters with no loss of accuracy. Most of the removed parameters were from FCLs.

\paragraph{Recoverable Pruning:}
Pruned elements usually cannot be recovered. This may result in reduced network capability. Recovering lost network capability requires significant re-training. Deep compression required millions of iterations to retrain the network \cite{Han2015}. To avoid this shortcoming, many approaches adopt recoverable pruning algorithms. The pruned elements may also be involved in the subsequent training process and adjust themselves to fit the pruned network.

Guo \cite{Guo2016} describes a recoverable pruning method using binary mask matrices to indicate whether a single weight value is pruned or not. The $l_1$-norm pruned weights can be stochastically spliced back into the network. Using this approach AlexNet was able to be reduced by a factor of $17.7\times$ with no accuracy loss. Re-training iterations were significantly reduced to 14.58\% of Deep compression \cite{Han2015}. 
However this type of pruning still results in an asymmetric network complicating hardware implementation. 

Soft Filter Pruning (SFP) \cite{He2017a} further extended recoverable pruning using a dimension of filter. SFP obtained structured compression results with an additional benefit or reduced inference time. Furthermore, SFP can be used on difficult to compress networks achieving a 29.8\% speed-up on ResNet-50 with 1.54\% ILSVRC-2012 top-1 accuracy loss. Comparing with Guo's recoverable weight \cite{Guo2016} technique, SFP achieves inference speed-ups closer to theoretical results on general purpose hardware by taking advantage of the structure of the filter.

\paragraph{Increasing Sparsity:}
Another motivation to apply fine-tuning is to increase network sparsity. Sparse constraints \cite{Zhou2016} applied low rank tensor constraints \cite{Liu2013} and group sparsity \cite{Deng2013} achieving a 70\% reduction of neurons with a 0.57\% drop of AlexNet in ILSVRC-2012 top-1 accuracy.

\paragraph{Adaptive Sparsity:}
No matter what kind of pruning criteria is applied, a layer-wise pruning ratio usually requires a human decision. Too high a ratio resulting in very high sparsity may cause the network to diverge requiring heavy re-tuning.

Network slimming \cite{Liu2017a}, previously discussed, addresses this problem by automatically computing layer-wise sparsity. This achieved a $20\times$ model size compression, $5\times$ computing reduction, and less than 0.1\% accuracy loss on the VGG network. 

Pruning can also be performed using a min-max optimization module \cite{Singh2019} that maintains network accuracy during tuning by keeping a pruning ratio. This technique compressed the VGG network by a factor of $17.5\times$ and resulted in a theoretical execution time (FLOPs) of $15.56\%$ of the unpruned network. 
A similar approach was proposed with an estimation of weights sets \cite{Carreira-Perpinan2018}. By avoiding the use of a greedy search to keep the best pruning ratio, they achieved the same ResNet classification accuracy with only 5\% to 10\% size of original weights.

AutoPruner \cite{Luo2020} integrated the pruning and fine-tuning of a three-stage pipeline as an independent training-friendly layer. The layer helped gradually prune during training eventually resulting in a less complex network. AutoPruner pruned 73.59\% of compute operations on VGG-16 with 2.39\% ILSVRC-2012 top-1 loss. ResNet-50 resulted in a 65.80\% of compute operations with 3.10\% loss of accuracy.  

\paragraph{Training from Scratch:}
Observation shows that network training efficiency and accuracy is inversely proportional to structure sparsity. The more dense the network, the less training time \cite{Han2015a, Li2017, Frankle2019}. This is one reason that current pruning techniques tend to follow a train-prune-tune pipeline rather than training a pruned structure from scratch.

However, the lottery ticket hypothesis \cite{Frankle2019} shows that it is not of primary importance to preserve the original weights but the initialization. Experiments show that dense, randomly-initialized pruned sub-networks can be trained effectively and reach comparable accuracy to the original network with the same number of training iterations.
Furthermore, standard pruning techniques can uncover the aforementioned sub-networks from a large oversized network - the \textit{Winning Tickets}.
In contrast with current static pruning techniques, the lottery ticket hypothesis after a period of time drops all well-trained weights and resets them to an initial random state. This technique found that ResNet-18 could maintain comparable performance with a pruning ratio up to 88.2\% on the CIFAR-10 dataset.  

\paragraph{Towards Better Accuracy:}
By reducing the number of network parameters, pruning techniques can also help to reduce over-fitting. Dense-Sparse-Dense (DSD) training \cite{Han2016d} helps various network improve classification accuracy by 1.1\% to 4.3\%. DSD uses a three stage pipeline: 1) dense training to identify important connections, 2) prune insignificant weights and sparse training with a sparsity constraint to take reduce the number of parameters, and 3) re-dense the structure to recover the original symmetric structure, this also increase the model capacity. The DSD approach has also shown impressive performance on the other type of deep networks such as Recurrent Neural Networks (RNNs) and Long Short Term Memory networks (LSTMs).

\subsection{Dynamic Pruning} \label{sec:pruning-dynamic}

Except for recoverable techniques, static pruning permanently destroys the original network structure which may lead to a decrease in model capability. Techniques have been researched to recover lost network capabilities but once pruned and re-trained, the static pruning approach can't recover destroyed information. Additionally, observations shows that the importance of neuron binding is input-independent \cite{Gao2019}. 

\begin{figure*}
    \centering
    \includegraphics[page=14, width=\textwidth]{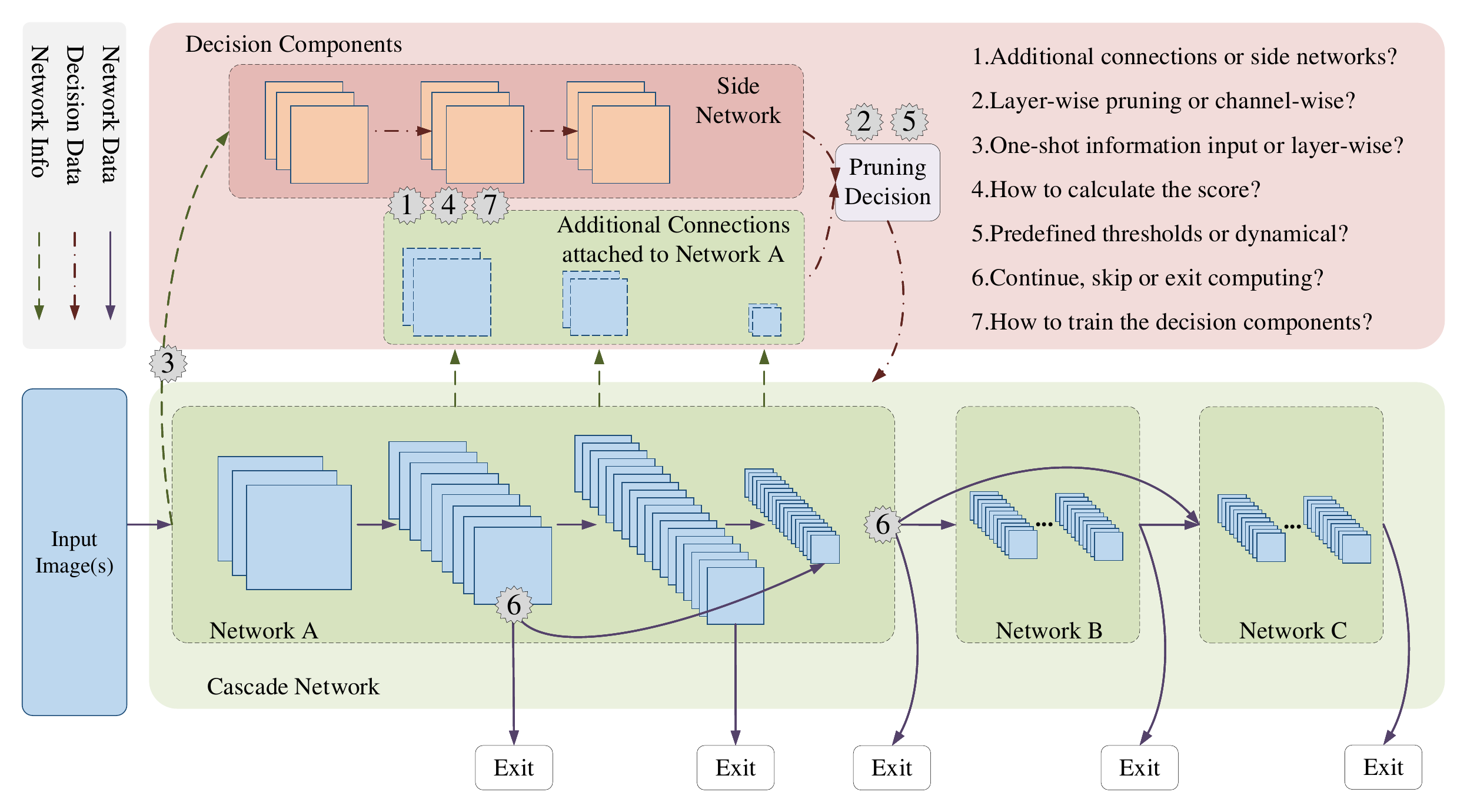}
    \caption{Dynamic Pruning System Considerations}
    \label{fig:dynamic-pruning-tips}
\end{figure*}

Dynamic pruning determines at runtime which layers, channels, or neurons will not participate in further activity. Dynamic pruning can overcome limitations of static pruning by taking advantage of changing input data potentially reducing computation, bandwidth, and power dissipation. Dynamic pruning typically doesn't perform runtime fine-tuning or re-training. 
In \autoref{fig:dynamic-pruning-tips}, we show an overview of dynamic pruning systems. The most important consideration is the decision system that decides what to prune. The related issues are:
\begin{enumerate}
    \item The type of the decision components: a) additional connections attached to the original network used during the inference phase and/or the training phase, b) characteristics of the connections that can be learned by standard backpropagation algorithms \cite{Gao2019}, and c) a side decision network which tends to perform well but is often difficult to train \cite{Lin2017}.
    \item The pruning level (shape): a) channel-wise \cite{Lin2017, Gao2019, Zhang2019d}, b) layer-wise \cite{Leroux2017}, c) block-wise \cite{Wu2018a}, or d) network-wise \cite{Bolukbasi2017}. The pruning level chosen influences hardware design. 
    \item Input data: a) one-shot information feeding \cite{Wu2018a} feeds the entire input to the decision system, and b) layer-wise information feeding \cite{Bolukbasi2017, Figurnov2017} where a window of data is iteratively fed to the decision system along with the forwarding. 
    \item Computing a decision score: $l_p$-norm \cite{Gao2019}, or b) other approaches \cite{Huang2018b}. 
    \item Score comparison: a) human experience/experiment results \cite{Leroux2017} or b) automatic threshold or dynamic mechanisms \cite{Huang2018b}. 
    \item Stopping criteria: a) in the case of layer-wise and network-wise pruning, some pruning algorithms skip the pruned layer/network \cite{Bengio2015, Wu2018a}, b) some algorithms dynamically choose the data path \cite{Odena2017, Yu2018a}, and c) ending the computation and outputing the predicting results \cite{Figurnov2017, Leroux2017, Li2019}. In this case the remaining layers are considered to be pruned. 
    \item Training the decision component: a) attached connections can be trained along with the original network \cite{Leroux2017, Li2019, Gao2019}, b) side networks are typically trained using reinforcement learning (RL) algorithms \cite{Bengio2015, Lin2017, Odena2017, Wu2018a}. 
\end{enumerate}
For instruction set processors, feature maps or the number of filters used to identify objects is a large portion of bandwidth usage \cite{Sze2017} - especially for depth-wise or point-wise convolutions where features consume a larger portion of the bandwidth \cite{Chollet2017}. Dynamic tuning may also be applied to statically pruned networks potentially further reducing compute and bandwidth requirements.

A drawback of dynamic pruning is that the criteria to determine which elements to prune must be computed at runtime. This adds overhead to the system requiring additional compute, bandwidth, and power. A trade-off between dynamic pruning overhead, reduced network computation, and accuracy loss, should be considered.
One method to mitigate power consumption inhibits computations from 0-valued parameters within a Processing Element (PE) \cite{Lin2017}. 

\subsubsection{Conditional Computing} \label{sec:pruning-dynamic-conditional}

Conditional computing involves activating an optimal part of a network without activating the entire network. Non-activated neurons are considered to be pruned. They do not participate in the result thereby reducing the number of computations required. Conditional computing applies to training and inference \cite{Bengio2013, Davis2013}.


Conditional computing has a similarity with RL in that they both learn a pattern to achieve a reward. Bengio \cite{Bengio2015} split the network into several blocks and formulates the block chosen policies as an RL problem. This approach consists of only fully connected neural networks and achieved a $5.3\times$ speed-up on CIFAR-10 dataset without loss of accuracy.

\subsubsection{Reinforcement Learning Adaptive Networks} \label{sec:pruning-dynamic-adaptive}

Adaptive networks aim to accelerating network inference by conditionally determining early exits. A trade-off between network accuracy and computation can be applied using thresholds. Adaptive networks have multiple intermediate classifiers to provide the ability of an early exit.
A cascade network is a type of adaptive network. Cascade networks are the combinations of serial networks which all have output layers rather than per-layer outputs. 
Cascade networks have a natural advantage of an early exit by not requiring all output layers to be computed. If the early accuracy of a cascade network is not sufficient, inference could potentially be dispatched to a cloud device \cite{Leroux2017, Bolukbasi2017}.
A disadvantage of adaptive networks is that they usually need hyper-parameters optimized manually (e.g., confidence score \cite{Leroux2017}). This introduces automation challenges as well as classification accuracy loss.  They found 28.75\% test error on CIFAR-10 when setting the threshold to 0.5. A threshold of 0.99 lowered the error to 15.74\% at a cost of 3x to inference time. 

A cascading network \cite{Odena2017} is an adaptive network with an RL trained \textit{Composer} that can determine a reasonable computation graph for each input. An adaptive controller \textit{Policy Preferences} is used to intelligently enhance the \textit{Composer} allowing an adjustment of the network computation graph from sub-graphs. The \textit{Composer} performs much better in terms of accuracy than the baseline network with the same number of computation-involved parameters on a modified dataset, namely Wide-MNIST. For example, when invoking 1k parameters, the baseline achieves 72\% accuracy while the \textit{Composer} obtained 85\%.

BlockDrop \cite{Wu2018a} introduced a \textit{policy network} that \linebreak[3] trained using RL to make an image-specific determination whether a residual network block should participate in the following computation. While the other approaches compute an exit confidence score per layer, the \textit{policy network} runs only once when an image is loaded. It generates a boolean vector that indicates which residual blocks are activate or inactive. BlockDrop adds more flexibility to the early exit mechanism by allowing a decision to be made on any block and not just early blocks in Spatially Adaptive Computation Time (SACT) \cite{Figurnov2017}. This is discussed further in \autoref{sec:pruning-dynamic-wo-RL}. BlockDrop achieves an average speed-up of 20\% on ResNet-101 for ILSVRC-2012 without accuracy loss. Experiments using the CIFAR dataset showed better performance than other SOTA counterparts at that time \cite{Figurnov2017, Graves2016, Li2017}.

Runtime Neural Pruning (RNP) \cite{Lin2017} is a framework that prunes neural networks dynamically. RNP formulates the feature selection problem as a Markov Decision Process (MDP) and then trains an RNN-based decision network by RL. The MDP reward function in the state-action-reward sequence is computation efficiency. Rather than removing layers, a side network of RNP predicts which feature maps are not needed. They found $2.3\times$ to $5.9\times$ reduction in execution time with top-5 accuracy loss from 2.32\% to 4.89\% for VGG-16. 
 
\subsubsection{Differentiable Adaptive Networks} \label{sec:pruning-dynamic-wo-RL}
Most of the aforementioned decision components are non-differential, thus computationally expensive RL is adopted for training. A number of techniques have been developed to reduce training complexity by using differentiable methods.

Dynamic channel pruning \cite{Gao2019} proposes a method to dynamically select which channel to skip or to process using Feature Boosting and Suppression (FBS). FBS is a side network that guides channel amplification and omission. FBS is trained along with convolutional networks using SGD with LASSO constraints. The selecting indicator can be merged into BN parameters.
FBS achieved $5\times$ acceleration on VGG-16 with 0.59\% ILSVRC-2012 top-5 accuracy loss, and $2\times$ acceleration on ResNet-18 with 2.54\% top-1, 1.46\% top-5 accuracy loss.

Another approach, Dynamic Channel Pruning (DCP) \cite{Zhang2019d} dynamically prunes channels using a \textit{channel threshold weighting (T-Weighting)} decision. Specifically, this module prunes the channels whose score is lower than a given threshold. The score is calculated by a T-sigmoid activation function, which is mathematically described in \autoref{eq:t-sigmoid}, where $\sigma(x)={1}/(1+e^{-x})$ is the sigmoid function. The input to the T-sigmoid activation function is down sampled by a FCL from the feature maps. The threshold is found using iterative training which can be a computationally expensive process.
DCP increased VGG-16 top-5 error by 4.77\% on ILSVRC-2012 for $5\times$ computation speed-up. By comparison, RNP increased VGG-16 top-5 error by 4.89\% \cite{Lin2017}.
\begin{equation}
\label{eq:t-sigmoid}
    h(x)=\begin{cases}
    \sigma(x), &\text{if } x > T \\
    0, &\text{otherwise}
    \end{cases}
\end{equation}

The cascading neural network by Leroux \cite{Leroux2017} reduced the average inference time of overfeat network \cite{Sermanet2013} by 40\% with a 2\% ILSVRC-2012 top-1 accuracy loss. Their criteria for early exit is based on the confidence score generated by an output layer. The auxiliary layers were trained with general backpropagation. The adjustable score threshold provides a trade-off between accuracy and efficiency.

Bolukbasi \cite{Bolukbasi2017} reports a system that contains a combination of other SOTA networks (e.g., AlexNet, ResNet, GoogLeNet, etc.). A policy adaptively chooses a point to exit early. This policy can be trained by minimizing its cost function. They format the system as a directed acyclic graph with various pre-trained networks as basic components. They evaluate this graph to determine leaf nodes for early exit.  
The cascade of acyclic graphs with a combination of various networks reduces computations while maintaining prediction accuracy. ILSVRC-2012 experiments show ResNet-50 acceleration of $2.8\times$ with 1\% top-5 accuracy loss and $1.9\times$ speed-up with no accuracy loss.

Considering the similarity of RNNs and residual networks \cite{Greff2016}, Spatially Adaptive Computation Time (SACT) \cite{Figurnov2017} explored an early stop mechanism of residual networks in the spatial domain. SACT can be applied to various tasks including image classification, object detection, and image segmentation. SACT achieved about 20\% acceleration with no accuracy loss for ResNet-101 on ILSVRC-2012. 
 

To meet the computation constraints, Multi-Scale Dense Networks (MSDNets) \cite{Huang2018b} designed an adaptive network using two techniques: 1) an \textit{anytime-prediction} to generate prediction results at many nodes to facilitate the network's early exit and 2) \textit{batch computational budget} to enforce a simpler exit criteria such as a computation limit. MSDNets combine multi-scale feature maps \cite{Zhang2016b} and dense connectivity \cite{Huang2017a} to enable  accurate early exit while maintaining higher accuracy. The classifiers are differentiable so that MSDNets can be trained using stochastic gradient descent. MSDNets achieve $2.2\times$ speed-up at the same accuracy for ResNet-50 on ILSVRC-2012 dataset.

To address the training complexity of adaptive networks, Li \cite{Li2019} proposed two methods. The first method is gradient equilibrium (GE). This technique helps backbone networks converge by using multiple intermediate classifiers across multiple different network layers. This improves the gradient imbalance issue found in MSDNets \cite{Huang2018b}. The second method is an Inline Subnetwork Collaboration (ISC) and a One-For-All knowledge distillation (OFA). Instead of independently training different exits, ISC takes early predictions into later predictors to enhance their input information. OFA supervises all the intermediate exits using a final classifier.  At a same ILSVRC-2012 top-1 accuracy of 73.1\%, their network takes only one-third the computational budget of ResNet.

Slimmable Neural Networks (SNN) \cite{Yu2018a} are a type of networks that can be executed at different widths. Also known as switchable networks, the network enables dynamically selecting network architectures (width) without much computation overhead. Switchable networks are designed to adaptively and efficiently make trade-offs between accuracy and on-device inference latency across different hardware platforms. SNN found that the difference of feature mean and variance may lead to training faults. SNN solves this issue with a novel switchable BN technique and then trains a wide enough network. Unlike cascade networks which primarily benefit from specific blocks, SNN can be applied with many more types of operations. As BN already has two parameters as mentioned in \autoref{sec:background}, the network switch that controls the network width comes with little additional cost. SNN increased top-1 error by 1.4\% on ILSVRC-2012 while achieving about $2\times$ speed-up.


\subsection{Comparisons} \label{sec:pruning-comparison}

Pruning techniques are diverse and difficult to compare. Shrinkbench \cite{Blalock2020} is a unified benchmark framework aiming to provide pruning performance comparisons.

There exist ambiguities about the value of the pre-trained weights. Liu \cite{Liu2019Rethingking} argues that the pruned model could be trained from scratch using a random weight initialization. This implies the pruned architecture itself is crucial to success. By this observation, the pruning algorithms could be seen as a type of NAS. Liu concluded that because the weight values can be re-trained, by themselves they are not efficacious. However, the lottery ticket hypothesis \cite{Frankle2019} achieved comparable accuracy only when the weight initialization was exactly the same as the unpruned model.
Glae \cite{Gale2019} resolved the discrepancy by showing that what really matters is the pruning form. Specifically, unstructured pruning can only be fine-tuned to restore accuracy but structured pruning can be trained from scratch. In addition, they explored the performance of dropout and $l_0$ regularization. The results showed that simple magnitude based pruning can perform better. They developed a magnitude based pruning algorithm and showed the pruned ResNet-50 obtained higher accuracy than SOTA at the same computational complexity. 

%% file: 3-quantize.tex
\section{Quantization} \label{sec:quantization}
Quantization is known as \textit{the process of approximating a continuous signal by a set of discrete symbols or integer values}. Clustering and parameter sharing also fall within this definition \cite{Han2015}. 
Partial quantization uses clustering algorithms such as k-means to quantize weight states and then store the parameters in a compressed file. The weights can be decompressed using either a lookup table or a linear transformation. This is typically performed during runtime inference. This scheme only reduces the storage cost of a model. This is discussed in \autoref{sec:quant-method-other}. In this section we focus on numerical low-bit quantization.

Compressing CNNs by reducing precision values has been previously proposed. Converting floating-point parameters into low numerical precision datatypes for quantizing neural networks was proposed as far back as the 1990s \cite{Fiesler1990, balzer1991weight}. Renewed interest in quantization began in the 2010s when 8-bit weight values were shown to accelerate inference without a significant drop in accuracy \cite{Vanhoucke2011}.
 
\begin{figure*}
    \centering
    \includegraphics[width=0.95\textwidth, page=12]{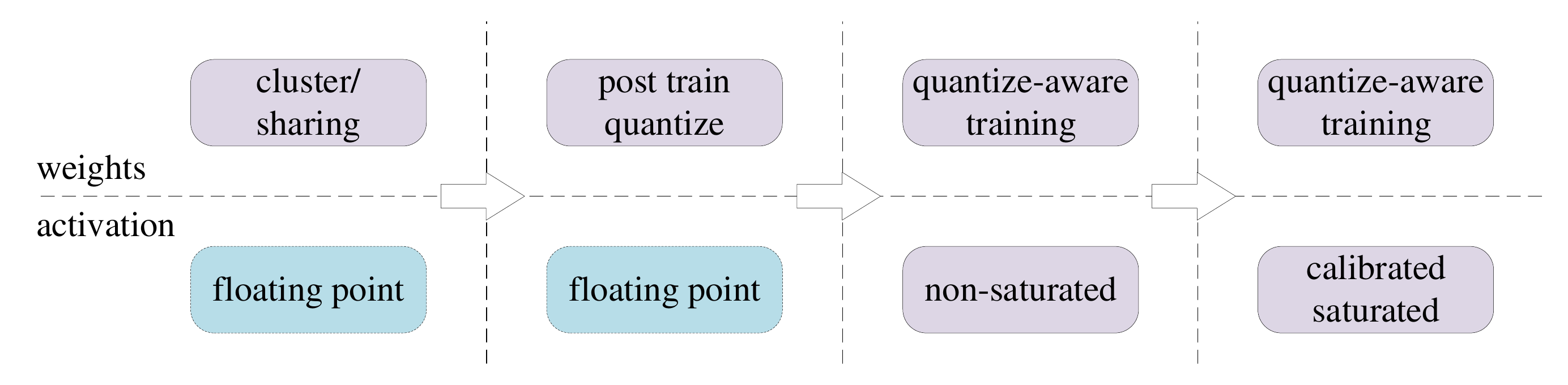}
    \caption{Quantization Evolution: The development of quantization techniques, from left to right. Purple rectangles indicated quantized data while blue rectangles represent full precision 32-bit floating point format.}
    \label{fig:quantize-roadmap}
\end{figure*} 

Historically most networks are trained using FP32 numbers \cite{Sze2017}. For many networks an FP32 representation has greater precision than needed. Converting FP32 parameters to lower bit representations can significantly reduce bandwidth, energy, and on-chip area. 

\autoref{fig:quantize-roadmap} shows the evolution of quantization techniques. Initially, only weights were quantized. By quantizing, clustering, and sharing, weight storage requirements can be reduced by nearly $4\times$. Han \cite{Han2015} combined these techniques to reduce weight storage requirements from 27MB to 6.9MB. 
Post training quantization involves taking a trained model, quantizing the weights, and then re-optimizing the model to generate a quantized model with scales \cite{Banner2019}.
Quantization-aware training involves fine-tuning a stable full precision model or re-training the quantized model. During this process real-valued weights are often down-scaled to integer values - typically 8-bits  \cite{Jacob2017}. 
Saturated quantization can be used to generate feature scales using a calibratation algorithm with a calibration set. Quantized activations show similar distributions with previous real-valued data \cite{Migacz2017}. Kullback-Leibler divergence (KL-divergence, also known as relative entropy or information divergence) calibrated quantization is typically applied and can accelerate the network without accuracy loss for many well known models \cite{Migacz2017}. Fine-tuning can also be applied with this approach.

KL-divergence is a measure to show the relative entropy of probability distributions between two sets. \autoref{eq:kl-divergence} gives the equation for KL-divergence. $P$ and $Q$ are defined as discrete probability distributions on the same probability space. Specifically, $P$ is the original data (floating-point) distribution that falls in several bins. $Q$ is the quantized data histogram. 
\begin{equation}
\label{eq:kl-divergence}
    D_{ \mathrm{KL} }( P \| Q ) = \sum_{i = 0} ^{N} P ( x_i ) \log \left( \frac { P ( x_i ) } { Q ( x_i ) } \right)
\end{equation}

Depending upon the processor and execution environment, quantized parameters can often accelerate neural network inference. 

Quantization research can be categorized into two focus areas: 1) quantization aware training (QAT) and 2) post training quantization (PTQ). The difference depends on whether training progress is is taken into account during training. Alternatively, we could also categorize quantization by where data is grouped for quantization: 1) layer-wise and 2) channel-wise. Further, while evaluating parameter widths, we could further classify by length: N-bit quantization.

Reduced precision techniques do not always achieve the expected speedup. For example, INT8 inference doesn't \linebreak[4] achieve exactly $4\times$ speedup over 32-bit floating point due to the additional operations of quantization and dequantization. 
For instance, Google's TensorFlow-Lite \cite{tflite} and nVidia's Tensor RT \cite{Migacz2017} INT8 inference speedup is about $2\text{-}3 \times$. Batch size is the capability to process more than one image in the forward pass. Using larger batch sizes, Tensor RT does achieve $3\text{-}4\times$ acceleration with INT8 \cite{Migacz2017}.

\autoref{sec:quant-perf-clct} summarizes current quantization techniques used on the ILSVRC-2012 dataset along with their bit-widths for weights and activation.

\subsection{Quantization Algebra} \label{quant-algo}
\begin{equation}
    \mathbf{X}_q = f(s \times g(\mathbf{X}_r) + z)
\label{eq:quantization}
\end{equation}
There are many methods to quantize a given network. Generally, they are formulated as \autoref{eq:quantization} where $s$ is a scalar that can be calculated using various methods. $g(\cdot)$ is the clamp function applied to floating-point values $\mathbf{X}_r$ performing the quantization. $z$ is the zero-point to adjust the true zero in some asymmetrical quantization approaches. $f(\cdot)$ is the rounding function. This section introduces quantization using the mathematical framework of \autoref{eq:quantization}.

\begin{equation}
\label{eq:quantization-clamp}
    clamp(x, \alpha, \beta) = max(min(x, \beta), \alpha)
\end{equation}

\autoref{eq:quantization-clamp} defines a clamp function. 
The \textit{min-max} \linebreak[4] method is given by \autoref{eq:quantization-min-max} where $[m, M]$ are the bounds for the minimum and maximum values of the parameters, respectively. $n$ is the maximum representable number derived from the bit-width (e.g., $256 = 2 ^ 8$ in case of 8-bit), and $z, s$ are the same as in \autoref{eq:quantization}. $z$ is typically non-zero in the \textit{min-max} method \cite{Jacob2017}. 
\begin{equation}
\label{eq:quantization-min-max}
    \begin{split}
         &g(x) = clamp(x, m, M) \\
         &s = \frac {n - 1}{M - m}, \ z = \frac{m \times (1 - n)}{M - m} \\
         &\text{where} \ m = \min\{\mathbf{X}_i\}, \ M = \max\{\mathbf{X}_i\} \\
    \end{split}
\end{equation}

The \textit{max-abs} method uses a symmetry bound shown in \autoref{eq:quantization-max-abs}. The quantization scale $s$ is calculated from the largest one $R$ among the data to be quantized. Since the bound is symmetrical, the zero point $z$ will be zero. In such a situation, the overhead of computing an offset-involved convolution will be reduced but the dynamic range is reduced since the valid range is narrower. This is especially noticeable for ReLU activated data where all of which values fall on the positive axis.
\begin{equation}
\label{eq:quantization-max-abs}
    \begin{split}
        &g(x) = clamp(x, -M, M)      \\
        & s = \frac{n - 1}{R}, \ z = 0 \\
        &\text{where} \ R = \max\{abs\{\mathbf{X}_i\}\} \\
    \end{split}
\end{equation}

Quantization can be applied on input features $\mathbf{F}$, weights $\mathbf{W}$, and biases $\mathbf{b}$. Taking feature $\mathbf{F}$ and weights $\mathbf{W}$ as an example (ignoring the biases) and using the \textit{min-max} method gives \autoref{eq:quantize-quantize-weights-and-feature}. The subscripts $r \text{ and } q$ denote the real-valued and quantized data, respectively. The $max$ suffix is from $R$ in \autoref{eq:quantization-max-abs}, while $s_f=(n-1)/F_{max},\ s_w=(n-1)/W_{max}$.
\begin{equation}
\label{eq:quantize-quantize-weights-and-feature}
        \mathbf{F}_q = \frac{n - 1}{F_{max}} \times \mathbf{F}_{r}
        \ ,\quad
        \mathbf{W}_q = \frac{n - 1}{W_{max}} \times \mathbf{W}_{r} 
\end{equation}

Integer quantized convolution is shown in \autoref{eq:quantize-quantized-convolution} and follows the same form as convolution with real values. In \autoref{eq:quantize-quantized-convolution}, the $*$ denotes the convolution operation, $\mathbf{F}$ the feature, $\mathbf{W}$ the weights, and $\mathbf{O}_q$, the quantized convolution result. Numerous third party libraries support this type of integer quantized convolution acceleration. They are discussed in \autoref{sec:quant-deploy-framework}.
\begin{equation}
    \label{eq:quantize-quantized-convolution}
    \mathbf{O}_q = \mathbf{F}_q \ast \mathbf{W}_q \quad \text{s.t.} \ \mathbf{F}, \mathbf{W} \in \mathbb{Z}
\end{equation}

De-quantizing converts the quantized value $\mathbf{O}_{q}$ back to floating-point $\mathbf{O}_{r}$ using the feature scales $s_f$ and weights scales $s_w$. A symmetric example with $z=0$ is shown in \autoref{eq:quantize-dequantize}. This is useful for layers that process floating-point tensors.
Quantization libraries are discussed in \autoref{sec:quant-deploy-framework}. 
\begin{equation}
\label{eq:quantize-dequantize}
    \mathbf{O}_{r} = \frac{\mathbf{O}_q}{s_f \times s_w} = \mathbf{O}_q \times \frac{{F}_{max}}{(n - 1)} \times \frac{{W}_{max}}{(n - 1)}
\end{equation}

In most circumstances, consecutive layers can compute with quantized parameters. This allows dequantization to be merged in one operation as in \autoref{eq:quantize-merged}. $\mathbf{F}_{q}^{l+1}$ is the quantized feature for next layer and $s_{f}^{l+1}$ is the feature scale for next layer. 
\begin{equation}
\label{eq:quantize-merged}
    \mathbf{F}_{q}^{l+1} = \frac{\mathbf{O}_q \times s_{f}^{l+1}}{s_f \times s_w}
\end{equation}

The activation function can be placed following either the quantized output $\mathbf{O}_q$, the de-quantized output $\mathbf{O}_r$, or after a re-quantized output $\mathbf{F}_{q}^{l+1}$. The different locations may lead to different numerical outcomes since they typically have different precision. 

Similar to convolutional layers, FCLs can also be quantized. 
K-means clustering can be used to aid in the compression of weights. In 2014 Gong \cite{Gong2014} used k-means clustering on FCLs and achieved a compression ratio of more than $20 \times$ with 1\% top-5 accuracy loss.

Bias terms in neural networks introduce intercepts in linear equations. They are typically regarded as constants that help the network to train and best fit given data.
Bias quantization is not widely mentioned in the literature. \cite{Jacob2017} maintained 32-bit biases while quantizing weights to 8-bit. Since biases account for minimal memory usage (e.g. 12 values for a 10-in/12-out FCL vs 120 weight values) it is recommended to leave biases in full precision.
If bias quantization is performed it can be a multiplication by both the feature scale and weight scale \cite{Jacob2017}, as shown in \autoref{eq:quant-bias-scale}. However, in some circumstances they may have their own scale factor. For example, when the bit-lengths are limited to be shorter than the multiplication results.
\begin{equation}
    s_b = s_w \times s_f
    ,\quad
    \mathbf{b}_q = \mathbf{b}_r \times s_b
\label{eq:quant-bias-scale}
\end{equation}

\subsection{Quantization Methodology} \label{quant-method}
We describe PTQ and QAT quantization approaches based on back-propagation use. We can also categorize them based on bit-width. In the following subsections, we introduce common quantization methods. In \autoref{sec:quant-method-low-num-precision} low bit-width quantization is discussed. In \autoref{sec:quant-method-log} and \autoref{sec:quant-method-plus} special cases of low bit-width quantization is discussed. In \autoref{sec:quant-method-training} difficulties with training quantized networks are discussed. Finally, in \autoref{sec:quant-method-other}, alternate approached to quantization are discussed. 

\subsubsection{Lower Numerical Precision} \label{sec:quant-method-low-num-precision}
Half precision floating point (16-bit floating-point, FP16) has been widely used in nVidia GPUs and ASIC accelerators with minimal accuracy loss \cite{Das2018}. Mixed precision training with weights, activations, and gradients using FP16 while the accumulated error  for updating weights remains in FP32 has shown SOTA performance - sometimes even improved performance \cite{Micikevicius2017}. 

Researchers \cite{Ma2016, He2015, Vanhoucke2011} have shown that FP32 parameters produced during training can be reduced to 8-bit integers for inference without significant loss of accuracy. Jacob \cite{Jacob2017} applied 8-bit integers for both training and inference, with an accuracy loss of 1.5\% on ResNet-50. Xilinx \cite{Settle2018} showed that 8-bit numerical precision could also achieve lossless performance with only one batch inference to adjust quantization parameters and without retraining.

Quantization can be considered an exhaustive search optimizing the scale found to reduce an error term. Given a floating-point network, the quantizer will take an initial scale, typically calculated by minimizing the $l_2$-error, and use it to quantize the first layer weights. Then the quantizer will adjust the scale to find the lowest output error. It performans this operation on every layer.

\begin{figure}[!tbh]
    \centering
    \includegraphics[width=0.45\textwidth, page=11]{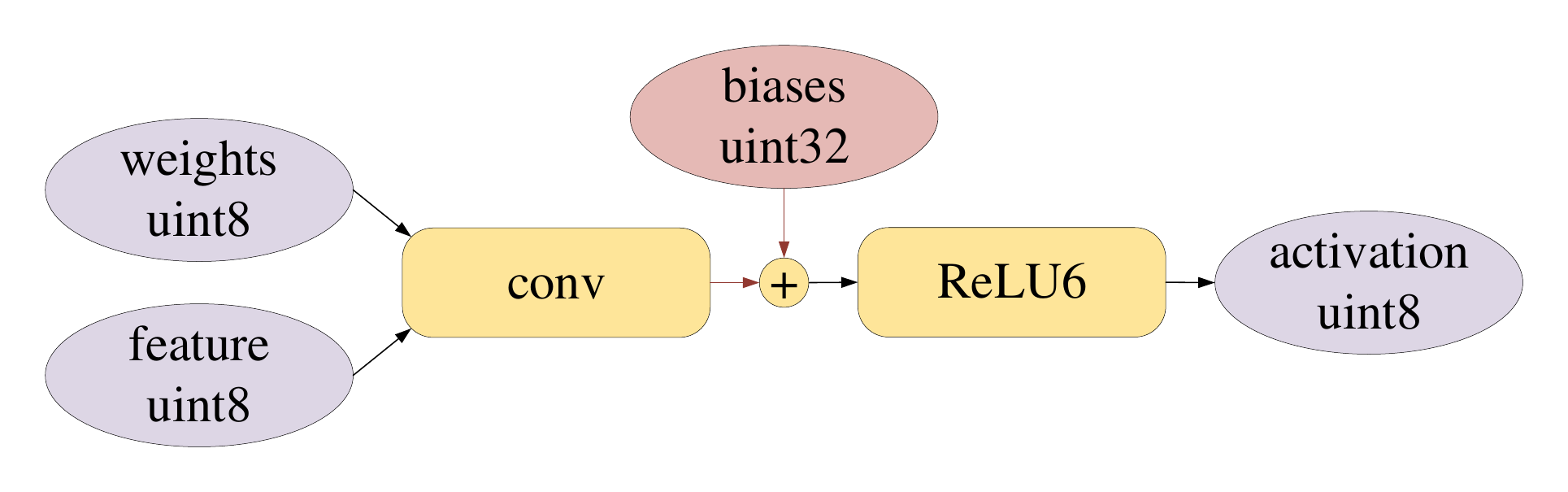}
    \caption{Integer Arithmetic-only Inference: The convolution operation takes unsigned int8 weights and inputs, accumulates them to unsigned int32, and then performs a 32-bit addition with biases. The ReLU6 operation outputs 8-bit integers. Adopted from \cite{Jacob2017}}
    \label{fig:quantize-interger-arithmetic-only-inference}
\end{figure}

Integer Arithmetic-only Inference (IAI) \cite{Jacob2017} proposed a practical quantization scheme able to be adopted by industry using standard datatypes. IAI trades off accuracy and inference latency by compressing compact networks into integers. Previous techniques only compressed the weights of redundant networks resulting in better storage efficiency. 
IAI quantizes $z \not= 0$ in \autoref{eq:quantization} requiring additional zero-point handling but resulting in higher efficiency by making use of unsigned 8-bit integers. The data-flow is described in \autoref{fig:quantize-interger-arithmetic-only-inference}.
TensorFlow-Lite \cite{Jacob2017, Krishnamoorthi2018} deployed IAI with an accuracy loss of 2.1\% using ResNet-150 on the ImageNet dataset. This is described in more detail in \autoref{sec:quant-deploy-framework}.

Datatypes other than INT8 have been used to quantize parameters. Fixed point, where the radix point is not at the right-most binary digit, is one format that has been found to be useful. It provides little loss or even higher accuracy but with a lower computation budget. Dynamic scaled fixed-point representation \cite{Vanhoucke2011} obtained a $4\times$ acceleration on CPUs. However, it requires specialized hardware including 16-bit fixed-point \cite{Gupta2015}, 16-bit flex point \cite{Koster2017}, and 12-bit operations using dynamic fixed-point format (DFXP) \cite{Courbariaux2014}.
The specialized hardware is mentioned in \autoref{sec:quant-deploy-hw}.

\subsubsection{Logarithmic Quantization} \label{sec:quant-method-log}
Bit-shift operations are inexpensive to implement in hardware compared to multiplication operations. FPGA implementations \cite{Alemdar2017} specifically benefit by converting floating-point multiplication into bit shifts. Network inference can be further optimized if weights are also constrained to be power-of-two with variable-length encoding. Logarithmic quantization takes advantage of this by being able to express a larger dynamic range compared to linear quantization.

Inspired by binarized networks \cite{Courbariaux2015}, introduced in \autoref{sec:quant-method-plus}, Lin \cite{Lin2015} forced the neuron output into a power-of-two value. This converts multiplications into bit-shift operations by quantizing the representations at each layer of the binarized network. Both training and inference time are thus reduced by eliminating multiplications. 

Incremental Network Quantization (INQ) \cite{Zhou2017a} replaces weights with power-of-two values. This reduces computation time by converting multiplies into shifts. 
INQ weight quantization is performed iteratively. In one iteration, weight pruning-inspired weight partitioning is performed using group-wise quantization. These weights are then fine-tuned by using a pruning-like measurement \cite{Han2015, Guo2016}. Group-wise retraining fine-tunes a subset of weights in full precision to preserve ensemble accuracy. The other weights are converted into power-of-two format. After multiple iterations most of the full precision weights are converted to power-of-two. The final networks have weights from 2 (ternary) to 5 bits with values near zero set to zero. 
Results of group-wise iterative quantization show lower error rates than a random power-of-two strategy. Specifically, INQ obtained $71\times$ compression with 0.52\% top-1 accuracy loss on the ILSVRC-2012 with AlexNet. 

Logarithmic Neural Networks (LogNN) \cite{Miyashita2016} quantize weights and features into a log-based representation. Logarithmic backpropagation during training is performed using shift operations. Bases other than $log_{2}$ can be used. $log_{\sqrt{2}}$ based arithmetic is described as a trade-off between dynamic range and representation precision. $log_{2}$ showed $7\times$ compression with 6.2\% top-5 accuracy loss on AlexNet, while $log_{\sqrt{2}}$ showed 1.7\% top-5 accuracy loss.

Shift convolutional neural networks (ShiftCNN) \cite{Gudovskiy2017} improve efficiency by quantizing and decomposing the real-valued weights matrix into an $N$ times $B$ ranged bit-shift, and encoding them with code-books $\mathbf{C}$ as shown in \autoref{eq:shiftcnn}. $idx_i(n)$ is the index for the $i^{th}$ weights in the $n^{th}$ code-book. Each coded weight ${w}_{i}$ can be indexed by the NB-bit expression.
\begin{equation}
    \begin{split}
        &{w}_{i}=\sum_{n=1}^{N} \mathbf{C}_{n}\left[\operatorname{idx}_{i}(n)\right] \\
        &\mathbf{C}_{n}=\left\{0, \pm 2^{-n+1}, \pm 2^{-n}, \pm 2^{-n-1}, \ldots, \pm 2^{-n-\lfloor M / 2\rfloor+2}\right\} \\
        &\text{where}\ M = 2^B - 1 
    \end{split}
\label{eq:shiftcnn}
\end{equation}
Note that the number of code-books $C_n$ can be greater than one. This means the encoded weight might be a combination of multiple shift operations. This property allows ShiftCNN to expand to a relatively large-scale quantization or to shrink to binarized or ternary weights. We discuss ternary weights in \autoref{sec:quant-method-plus}. 
ShiftCNN was deployed on an FPGA platform and achieved comparable accuracy on the ImageNet dataset with 75\% power saving and up to $1090\times$ clock cycle speed-up. ShiftCNN achieves this impressive result without requiring retraining.  With $N=2$ and $B=4$ encoding, SqueezeNet \cite{Iandola2016} has only 1.01\% top-1 accuracy loss. The loss for GoogLeNet, ResNet-18, and ResNet-50 is 0.39\%, 0.54\%, and 0.67\%, respectively, While compressing the weights into 7/32 of the original size. This implies that the weights have significant redundancy.

Based on LogNN, Cai \cite{Cai2018} proposed improvements by disabling activation quantization to reduce overhead during inference. This also reduced the clamp bound hyperparameter tuning during training.
These changes resulted in many low-valued weights that are rounded to the nearest value during encoding. As $2^n\  \text{s.t.}\ n\in N$ increases quantized weights sparsity as $n$ increases. In this research, $n$ is allowed to be real-valued numbers as $n \in R$ to quantize the weights. This makes weight quantization more complex. However, a code-book helps to reduce the complexity. 

In 2019, Huawei proposed DeepShift, a method of saving computing power by shift convolution \cite{Elhoushi2019}. DeepShift removed all floating-point multiply operations and replaced them with bit reverse and bit shift. The quantized weight $W_q$ transformation is shown mathematically in \autoref{eq:deepshift}, where $S$ is a sign matrix, $P$ is a shift matrix, and $Z$ is the set of integers.
\begin{equation}
    W_q = S \times 2^P,\ \text{s.t.} \ P \in \mathbb{Z}, S \in \{-1, 0, +1\}
    \label{eq:deepshift}
\end{equation}
Results indicate that DeepShift networks cannot be easily trained from scratch. They also show that shift-format networks do not directly learn for lager datasets such as Imagenet. Similar to INQ, they show that fine-tuning a pre-trained network can improve performance. For example, with the same configuration of 32-bit activations and 6-bit shift-format weights, the top-1 ILSVRC-2012 accuracy loss on ResNet-18 for trained from scratch and tuned from a pre-trained model are 4.48\% and 1.09\%, respectively.
 
DeepShift proposes models with differential backpropagation for generating shift coefficients during the retraining process.  DeepShift-Q \cite{Elhoushi2019} is trained with floating-point parameters in backpropagation with values rounded to a suitable format during inference. DeepShift-PS directly adopts the shift $P$ and sign $S$ parameters as trainable parameters. 

Since logarithmic encoding has larger dynamic range, redundant networks particularly benefit. However, less redundant networks show significant accuracy loss. For example, VGG-16 which is a redundant network shows 1.31\% accuracy loss on top-1 while DenseNet-121 shows 4.02\% loss. 
\subsubsection{Plus-minus Quantization} \label{sec:quant-method-plus}
Plus-minus quantization was in 1990 \cite{Saad1990}. This technique reduces all weights to 1-bit representations. Similar to logarithmic quantization, expensive multiplications are removed. In this section, we provide an overview of significant binarized network results. Simons \cite{Simons2019} and Qin \cite{Qin2020} provide an in-depth review of BNNs.

Binarized neural networks (BNN) have only 1-bit weights and often 1-bit activations. 0 and 1 are encoded to represent -1 and +1, respectively. Convolutions can be separated into multiplies and additions. In binary arithmetic, single bit operations can be performed using \textit{and}, \textit{xnor}, and \textit{bit-count}. We follow the introduction from \cite{Zhou2017c} to explain bit-wise operation.
Single bit fixed point dot products are calculated as in \autoref{eq:bitcount-bit_prod}, where \textit{and} is a bit-wise AND operation and \textit{bitcount} counts the number of 1's in the bit string.
\begin{equation}
    \label{eq:bitcount-bit_prod}
    \boldsymbol{x} \cdot \boldsymbol{y} = \operatorname{bitcount}(\operatorname{and}(\boldsymbol{x}, \boldsymbol{y})), \text{ s.t. }\forall i, x_{i}, y_{i} \in\{0,1\}
\end{equation}
This can be extended into multi-bit computations as in \autoref{eq:bitcount-multi_bit_prod} \cite{Courbariaux2016}. $\boldsymbol{x} \text{ and } \boldsymbol{y}$ are M-bit and K-bit fixed point integers, subject to $\boldsymbol{x}=\sum_{m=0}^{M-1}c_m(\boldsymbol{x})2^m$ and $\boldsymbol{y}=\sum_{k=0}^{K-1}c_k(\boldsymbol{y})2^k$ , where $(c_m({\boldsymbol{x}}))_{m=0}^{M-1}$ and $(c_k({\boldsymbol{y}}))_{k=0}^{K-1}$ are bit vectors.
\begin{equation}
    \label{eq:bitcount-multi_bit_prod}
    \begin{split}
        \mathrm{x} \cdot \mathrm{y}=\sum_{m=0}^{M-1} \sum_{k=0}^{K-1} 2^{m+k} \operatorname{bitcount}\left[\operatorname{and}\left(c_{m}(\mathrm{x}), c_{k}(\mathrm{y})\right)\right],\\
        \text{ s.t. } c_{m}(\mathrm{x})_{i}, c_{k}(\mathrm{y})_{i} \in\{0,1\} \forall i, m, k.
    \end{split}
\end{equation}

By removing complicated floating-point multiplications, networks are dramatically simplified with simple accumulation hardware. Binarization not only reduces the network size by up-to 32$\times$, but also drastically reduces memory usage resulting in significantly lower energy consumption \cite{Mishra2018, Hubara2016b}. However, reducing 32-bit parameters into a single bit results in a significant loss of information, which decreases  prediction accuracy. Most quantized binary networks significantly under-perform compared to 32-bit competitors. 

There are two primary methods to reduce floating-point values into a single bit: 1) stochastic and 2) deterministic \cite{Courbariaux2015}. Stochastic methods consider global statistics or the value of input data to determine the probability of some parameter to be -1 or +1. Deterministic binarization directly computes the bit value based on a threshold, usually 0, resulting in a sign function. Deterministic binarization is much simpler to implement in hardware. 


Binary Connect (BC), proposed by Courbariaux \cite{Courbariaux2015}, is an early stochastic approach to binarize neural networks. They binarized the weights both in forward and backward propagation.  \autoref{eq:bnn-stochastic-binarization} shows the stochastic binarization $x^b$ with a \textit{hard sigmoid} probability $\sigma(x)$. Both the activations and the gradients use 32-bit single precision floating point. The trained BC network shows 1.18\% classification error on the small MNIST dataset but 8.27\% classification on the larger CIFAR-10 dataset.
\begin{equation}
\label{eq:bnn-stochastic-binarization}
    \begin{split}
        x^{b}=
        \begin{cases}
            +1, & \text{with probability } p=\sigma(x) \\
            -1, & \text{with probability } 1-p 
        \end{cases} \\
        \text{where } \sigma(x)=\operatorname{clamp}\left(\frac{x+1}{2}, 0,1\right)
    \end{split}
\end{equation}

 Courbariaux extended BC networks by binarizing the activations. He named them BinaryNets \cite{Courbariaux2016}, which is recognized as the first BNN. They also report a customized binary matrix multiplication GPU kernel that accelerates the calculation by $7\times$. BNN is considered the first binarized neural network where both weights and activations are quantized to binary values \cite{Simons2019}. Considering the hardware cost of stochastic binarization, they made a trade-off to apply deterministic binarization in most circumstances. BNN reported 0.86\% error on MNIST, 2.53\% error on SVHN, and 10.15\% error on CIFAR-10. The ILSVRC-2012 dataset accuracy results for binarized AlexNet and GoogleNet are 36.1\% top-1 and 47.1\%, respectively while the FP32 original networks achieve 57\% and 68\%, respectively \cite{Hubara2016b}.



Rastegari \cite{Rastegari2016} explored binary weight networks (BWN) on the ILSVRC dataset with AlexNet and achieved the same classification accuracy as the single precision version. The key is a scaling factor $\alpha \in \mathbb{R}^{+}$ applied to an entire layer of binarized weights $\mathbf{B}$. This results in similar weights values as if they were computed using FP32 $\mathbf{W} \approx \alpha \mathbf{B}$. They also applied weight binarization on ResNet-18 and GoogLeNet, resulting in 9.5\% and 5.8\% top-1 accuracy loss compared to the FP32 version, respectively.
They also extended binarization to activations called XNOR-Net and evaluated it on the large ILSVRC-2012 dataset. Compared to BNN, XNOR-Net also applied a scaling factor on the input feature and a rearrangement of the network structure (swapping the convolution, activation, and BN). Finally, XNOR-Net achieved 44.2\% top-1 classification accuracy on ILSVRC-2012 with AlexNet, while accelerating execution time $58\times$ on CPUs. 
The attached scaling factor extended the binarized value expression, which reduced the network distortion and lead to better ImageNet accuracy. 

DoReFa-Net \cite{Zhou2016a} also adopts plus-minus arithmetic for quantized network. DoReFa additionally quantizes gradients to low-bit widths within 8-bit expressions during the backward pass. The gradients are quantized stochastically in back propagation. For example, it takes 1 bit to represent weights layer-wise, 2-bit activations, and 6-bits for gradients. We describe training details in \autoref{sec:quant-method-training}. They found 9.8\% top-1 accuracy loss on AlexNet with ILSVRC-2012 using the 1-2-6 combination. The result for the 1-4-32 combination is 2.9\%.

Li \cite{Li2016} and Leng \cite{Leng2017} showed that for ternary weights ($-1, 0,\text{ and }+1$), in Ternary Weight Networks (TWN), only a slight accuracy loss was realized. 
Compared to BNN, TWN has an additional value to reduce information loss while still keeping computational complexity similar to BNN's.
Ternary logic may be implemented very efficiently in hardware, as the additional value (zero) do not actually participate in computations \cite{Cotofana1997}. 
TWN adopts the $l_2$-distance to find the scale and formats the weights into $-1, 0,\text{ and }+1$ with a threshold generated by an assumption that the weighs are uniformly distributed such as in $[-a, a]$. This resulted in up to $16 \times$ model compression with 3.6\% ResNet-18 top-1 accuracy loss on ILSVRC-2012.

Trained Ternary Quantization (TTQ) \cite{Zhu2016} extended TWN by introducing two dynamic constraints to adjust the quantization threshold. TTQ outperformed the full precision AlexNet on the ILSVRC-2012 top-1 classification accuracy by 0.3\%. It also outperformed TWN by 3\%. 

Ternary Neural Networks (TNN) \cite{Alemdar2017} extend TWN by quantizing the activations into ternary values. A teacher network is trained with full precision and then using transfer learning the same structure is used but replacing the full precision values with a ternarized student in a layer-wise greedy method. A small difference between the real-valued teacher network and the ternarized student network is that they activate the output with a ternary output activation function to simulate the real TNN output. TNN achieves 1.67\% MNIST classification error and 12.11\% classification error on CIFAR10. TNN has slightly lower accuracy compared to TWN (an additional 1.02\% MNIST error). 

Intel proposed Fine-Grained Quantization (FGQ) \cite{Mellempudi2017} to generalize ternary weights
by splitting them into several groups and with independent ternary values. 
The FGQ quantized ResNet-101 network achieved 73.85\% top-1 accuracy on the ImageNet dataset (compared with 77.5\% for the baseline) using four groups weights and without re-training. FGQ also showed improvements in (re)training demonstrating a top-1 accuracy improvement from 48\% on non-trained to 71.1\% top-1 on ResNet-50. ResNet-50's baseline accuracy is 75\%. Four groups FGQ with ternary weights and low bit-width activations achieves about $9\times$ acceleration.  

MeliusNet \cite{Bethge2020} is a binary neural network that consist of two types of binary blocks. To mitigate drawbacks of low bit width networks, reduced information quality, and reduced network capacity, MeliusNet used a combination of \textit{dense block} \cite{Bethge2019} which increases network channels by concatenating derived channels from the input to improve capacity and \textit{improvement block} \cite{Liu2018c} which improves the quality of features by adding additional convolutional activations onto existing extra channels from \textit{dense block}. 
They achieved accuracy results comparable to MobileNet on the ImageNet dataset with MeliusNet-59 reporting 70.7\% top-1 accuracy while requiring only 0.532 BFLOPs. A similar sized 17MB MobileNet required 0.569 BFLOPs achieving 70.6\% accuracy.

AdderNet \cite{Chen2019b} is another technique that replaces multiply arithmetic but allows larger than 1-bit parameters. It replaces all convolutions with addition.
\autoref{eq:addernet-conv} shows that for a standard convolution, AdderNet formulates it as a similarity measure problem
\begin{equation}
\label{eq:addernet-conv}
    Y(m, n, t)=\sum_{i=0}^{d} \sum_{j=0}^{d} \sum_{k=0}^{c_{i n}} S(\mathbf{X}(m+i, n+j, k), \mathbf{F}(i, j, k, t))
\end{equation}
where $\mathbf{F} \in \mathbb{R}^{d \times d \times c_{i n} \times c_{\text {out}}}$ is a filter, $d$ is the kernel size, $c_{in}$ is an input channel and $c_{\text {out}}$ is an output channel. $\mathbf{X} \in \mathbb{R}^{h \times w \times c_{i n}}$ stands for the input feature height $h$ and width $w$. With this formulation, the output $Y$ is calculated with the similarity $S(\cdot,\cdot)$, i.e., $S(x, y) =x \times y$ for conventional convolution where the similarity measure is calculated by cross correlation.
\autoref{eq:addernet-add} mathematically describes AdderNet, which replaces the multiply with subtraction. The $l_1$-distance is applied to calculate the distance between the filter and the input feature. By replacing multiplications with subtractions, AdderNet speeds up inference by transforming 3.9 billion multiplications into subtractions with a loss in ResNet-50 accuracy of 1.3\%. 
\begin{equation}
\label{eq:addernet-add}
Y(m, n, t)=-\sum_{i=0}^{d} \sum_{j=0}^{d} \sum_{k=0}^{c_{i n}}|\mathbf{X}(m+i, n+j, k)-\mathbf{F}(i, j, k, t)|
\end{equation}

NAS can be applied to BNN construction. Shen \cite{Shen2019} adopted evolutionary algorithms to find compact but accurate models achieving 69.65\% top-1 accuracy on ResNet-18 with ImageNet at $2.8\times$ speed-up. This is better performance than the 32-bit single precision baseline ResNet-18 accuracy of 69.6\%. 
However, the search approach is time consuming taking 1440 hours on an nVidia V100 GPU to search 50k ImageNet images to process an initial network.

\subsubsection{Other Approaches to Quantization} \label{sec:quant-method-other}
Weight sharing by vector quantization can also be considered a type of quantization. In order to compress parameters to reduce memory space usage, parameters can be clustered and shared. K-means is a widely used clustering algorithm and has been successfully applied to DNNs with minimal loss of accuracy \cite{Gong2014, Wu2016, Lei2017} achieving 16-24 times compression with 1\% accuracy loss on the ILSVRC-2012 dataset \cite{Gong2014, Wu2016}.

HashNet \cite{Chen2015} uses a hash to cluster weights. Each hash group is replaced with a single floating-point weight value. This was applied to FCLs and shallow CNN models. They found a compression factor of $64\times$ outperforms equivalent-sized networks on MNIST and seven other datasets they evaluated.

In 2016 Han applied Huffman coding with Deep Compression \cite{Han2015}. The combination of weight sharing, pruning, and huffman coding achieved $49\times$ compression on VGG-16 with no loss of accuracy on ILSVRC-2012, which was SOTA at the time.

The Hessian method was applied to measure the importance of network parameters and therefore improve weight quantization \cite{Choi2017}. They minimized the average Hessian weighted quantization errors to cluster parameters. They found compression ratios of 40.65 on AlexNet with 0.94\% accuracy loss on ILSVRC-2012.
Weight regularization can slightly improve the accuracy of quantized networks by penalizing weights with large magnitudes \cite{Sheng2018}. Experiments showed that $l_2$ regularization improved 8-bit quantized MobileNet top-1 accuracy by 0.23\% on ILSVRC-2012.

BN has proved to have many advantages including addressing the internal covariate shift issue \cite{Ioffe2015}. It can also be considered a type of quantization. However, quantization performed with BN may have numerical instabilities. The BN layer has nonlinear square and square root operations. Low bit representations may be problematic when using non-linear operations. To solve this, $l_1$-norm BN \cite{Wu2019} has only linear operations in both forward and backward training. It provided $1.5\times$ speedup at half the power on FPGA platforms and can be used with both training and inference. 

\subsubsection{Quantization-aware Training} \label{sec:quant-method-training}
Most quantization methods use a global (layer-wise) quantization to reduce the full precision model into a reduced bit model. Thus can result in non-negligible accuracy loss.
A significant drawback of quantization is information loss caused by the irreversible precision reducing transform. Accuracy loss is particularly visible in binary networks and shallow networks. Applying binary weights and activations to ResNet-34 or GoogLeNet resulted in 29.10\% and 24.20\% accuracy loss, respectively \cite{Courbariaux2016}.
It has been shown that backward propagation fine-tunes (retrains) a quantized network and can recover losses in accuracy caused by the quantization process \cite{Merolla2016}. The retraining is even resilient to binarization information distortions. Thus training algorithms play a crucial role when using quantization. In this section, we introduce (re)training of quantized networks.

\paragraph{BNN Training:}
For a binarized network that has binary valued weights it is not effective to update the weights using gradient decent methods due to typically small derivatives. 
Early quantized networks were trained with a variation of Bayesian inference named Expectation Back Propagation (EBP) \cite{Soudry2014, Cheng2015}. This method assigns limited parameter precision (e.g., binarized) weights and activations. EBP infers networks with quantized weights by updating the posterior distributions over the weights. The posterior distributions are updated by differentiating the parameters of the backpropagation. 

BinaryConnect \cite{Courbariaux2015} adopted the probabilistic idea of EBP but instead of optimizing the weights posterior distribution, BC preserved floating-point weights for updates and then quantized them into binary values. The real-valued \linebreak[4] weights update using the back propagated error by simply ignoring the binarization in the update.

A binarized Network has only 1-bit parameters - $\pm 1$ quantized from a sign function. Single bit parameters are non-differentiable and therefore it is not possible to calculate gradients needed for parameter updating \cite{Saad1990}. SGD algorithms have been shown to need 6 to 8 bits to be effective \cite{Muller2015}. To work around these limitations the Straight-Through Estimator (STE), previously introduced by Hinton \cite{Hinton2012}, was applied for propagating gradients by using discretization \cite{Hubara2016b}. \autoref{eq:STE} show the STE for sign binarization, where $c$ denotes the 
cost function, $w_r$ is the real-valued weights, and $w_b$ is the binarized weight produced by the \textit{sign} function. STE bypasses the binarization function to directly calculate real-valued gradients. The floating-point weights are then updated using methods like SGD. To avoid real-valued weights approaching infinity, BNNs typically clamp floating-point weights to the desired range of $\pm 1$ \cite{Hubara2016b}.
\begin{equation}
\label{eq:STE}
    \begin{split}
        \text { Forward : } &w_{b}=\operatorname{sign}\left(w_{r}\right) \\
        \text { Backward : } &\frac{\partial c}{\partial w_{r}}=\frac{\partial c}{\partial w_{b}} \mathbf{1}_{\left|w_{r}\right| \leq 1}
    \end{split}
\end{equation}


Unlike the forward phase where weights and activations are produced with deterministic quantization, in the gradient phase, the low bit gradients should be generated by stochastic quantization \cite{Gupta2015, Zhou2018}. 
DoReFa \cite{Zhou2016a} first successfully trained a network with gradient bit-widths less than eight and achieved a comparable result with $k$-bit quantization arithmetic. 
This low bit-width gradient scheme could accelerate training in edge devices with little impact to network accuracy but minimal inference acceleration compared to BNNs. 
DoReFa quantizes the weights, features, and gradients into many levels obtaining a larger dynamic range than BNNs. They trained AlexNet on ImageNet from scratch with 1-bit weights, 2-bit activations, and 6-bit gradients. They obtained 46.1\% top-1 accuracy (9.8\% loss comparing with the full precision counterpart).
\autoref{eq:quant-dorefa} shows the weight quantizing approach. $w$ is the weights (the same as in \autoref{eq:STE}), $\operatorname{limit}$ is a limit function applied to the weights keeping them in the range of [0, 1], and $\operatorname{quantize}_k$ quantizes the weights into $k$-levels. Feature quantization is performed using the $f_{\alpha}^{k}=\operatorname{quantize}_k$ function.
\begin{equation} 
\label{eq:quant-dorefa}
\begin{split}
    &f_{w}^{k} = 2 \operatorname{quantize}_k \left( \operatorname{limit}(w_r) \right) - 1 \\
    \text{where } &\operatorname{quantize}_k(w_r) = \frac{1}{2^k - 1} \operatorname{round}\left( \left( 2^k - 1\right) w_r \right), \\
    \text{and } &\operatorname{limit}(x)=\frac{\operatorname{tanh}(x)}{2\operatorname{max}(|\operatorname{tanh}(x)|)} + \frac{1}{2}
\end{split}
\end{equation}
In DoReFa, gradient quantization is shown in \autoref{eq:quant-dorefa-gradients}, where $\mathrm{d}r={\partial c}/{\partial r}$ is the backprogagated gradient of the cost function $c$ to output $r$.
\begin{equation} \label{eq:quant-dorefa-gradients}
\tilde{f}_{\gamma}^{k} = 2 \operatorname{max}_{0}(|\mathrm{d} r|)\left[\operatorname{quantize}_{k}\left(\frac{\mathrm{d} r}{2 \max_{0}(|\mathrm{d} r|)}+\frac{1}{2}\right)-\frac{1}{2}\right]
\end{equation}

As in deep feed forward networks, the exploding gradient problem can cause BNN's not to train. To address this issue, Hou \cite{Hou2017} formulated the binarization effect on the network loss as an optimization problem which was solved by a proximal Newton's algorithm with diagonal Hessian approximation that directly minimizes the loss with respect to the binary weights. This optimization found 0.09\% improvement on MNIST dataset compared with BNN.

\begin{figure}[h]
    \centering
    \begin{subfigure}[Straight-through Estimator]{
        \includegraphics[width=0.45\linewidth,page=8]{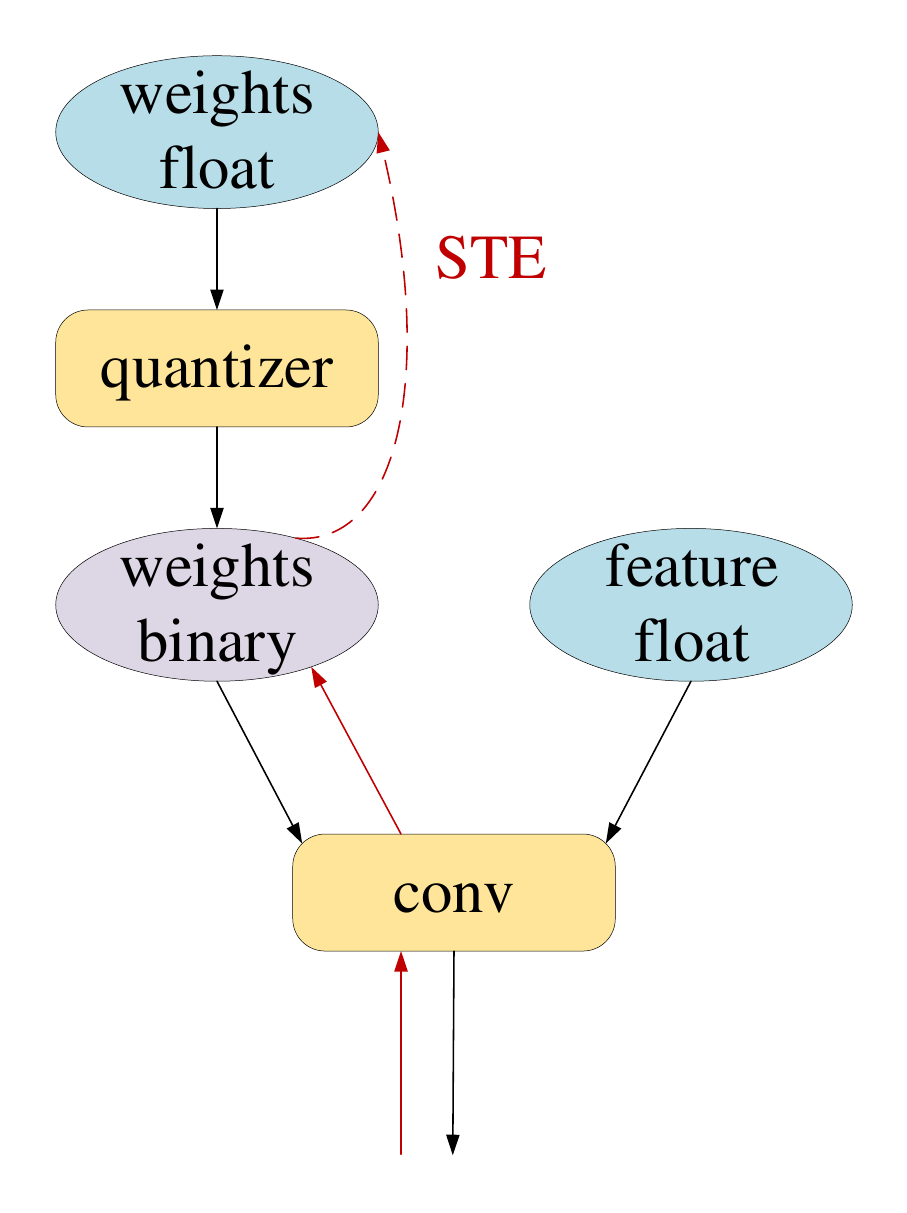}
        \label{fig:bnn-training-ste}}
    \end{subfigure}
    \vspace{1em}
    \begin{subfigure}[Alpha-Blending Approach]{
        \includegraphics[width=0.45\linewidth,page=9]{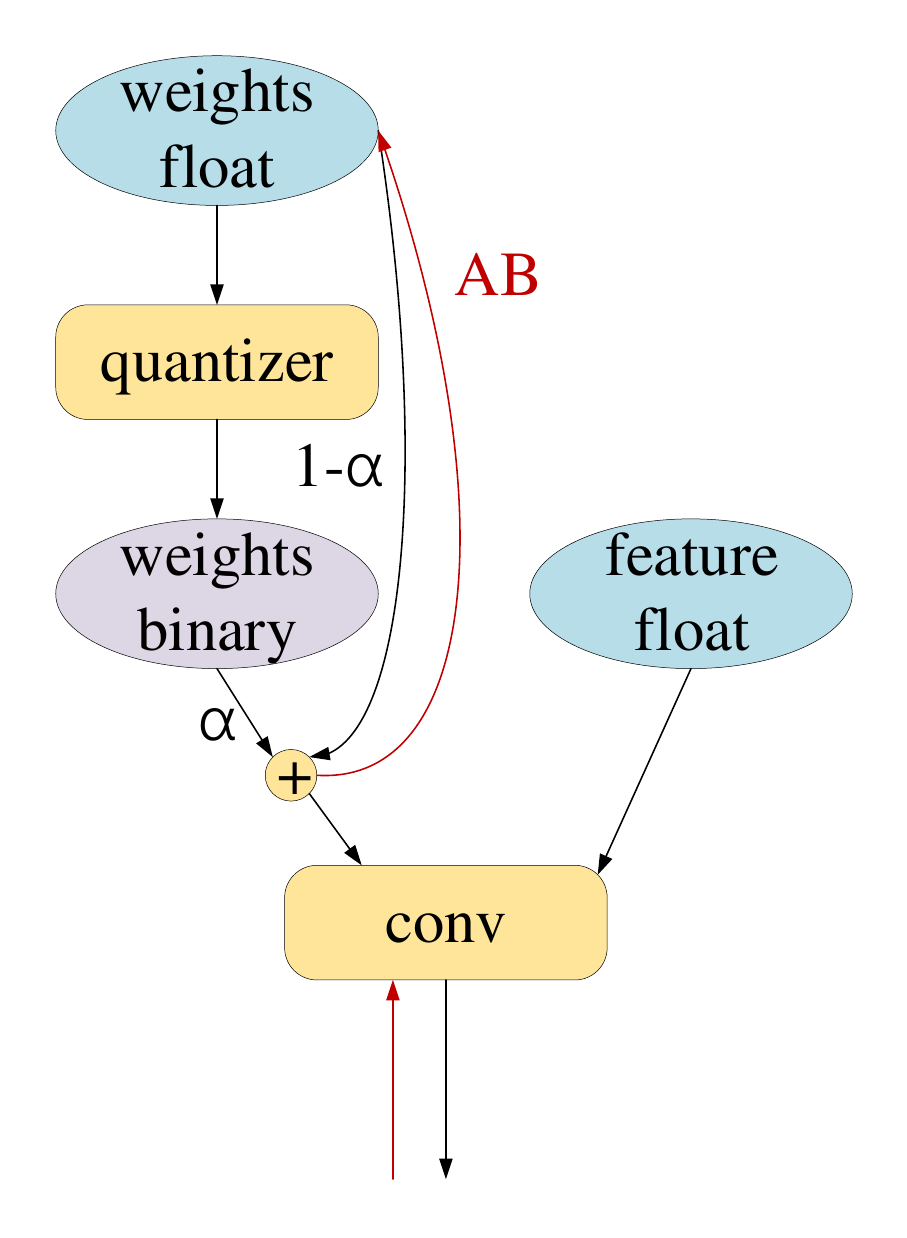}
        \label{fig:bnn-training-ab}}
    \end{subfigure}
    \caption{STE and AB: STE directly bypasses the quantizer while AB calculates gradients for real-valued weights by introducing additional coefficients $\alpha$ \cite{Liu2019c}}
    \label{fig:bnn-training}
\end{figure}

Alpha-Blending (AB) \cite{Liu2019c} was proposed as a replacement for STE. Since STE directly sets the quantization function gradients to 1, a hypothesis was made that STE tuned networks could suffer accuracy losses. \autoref{fig:bnn-training} shows that AB introduces an additional scale coefficient $\alpha$. Real-valued weights and quantized weights are both kept. During training $\alpha$ is gradually raised to 1 until a fully quantized network is realized.

\begin{figure}[!tbh]
    \centering
    \includegraphics[width=0.45\textwidth, page=10]{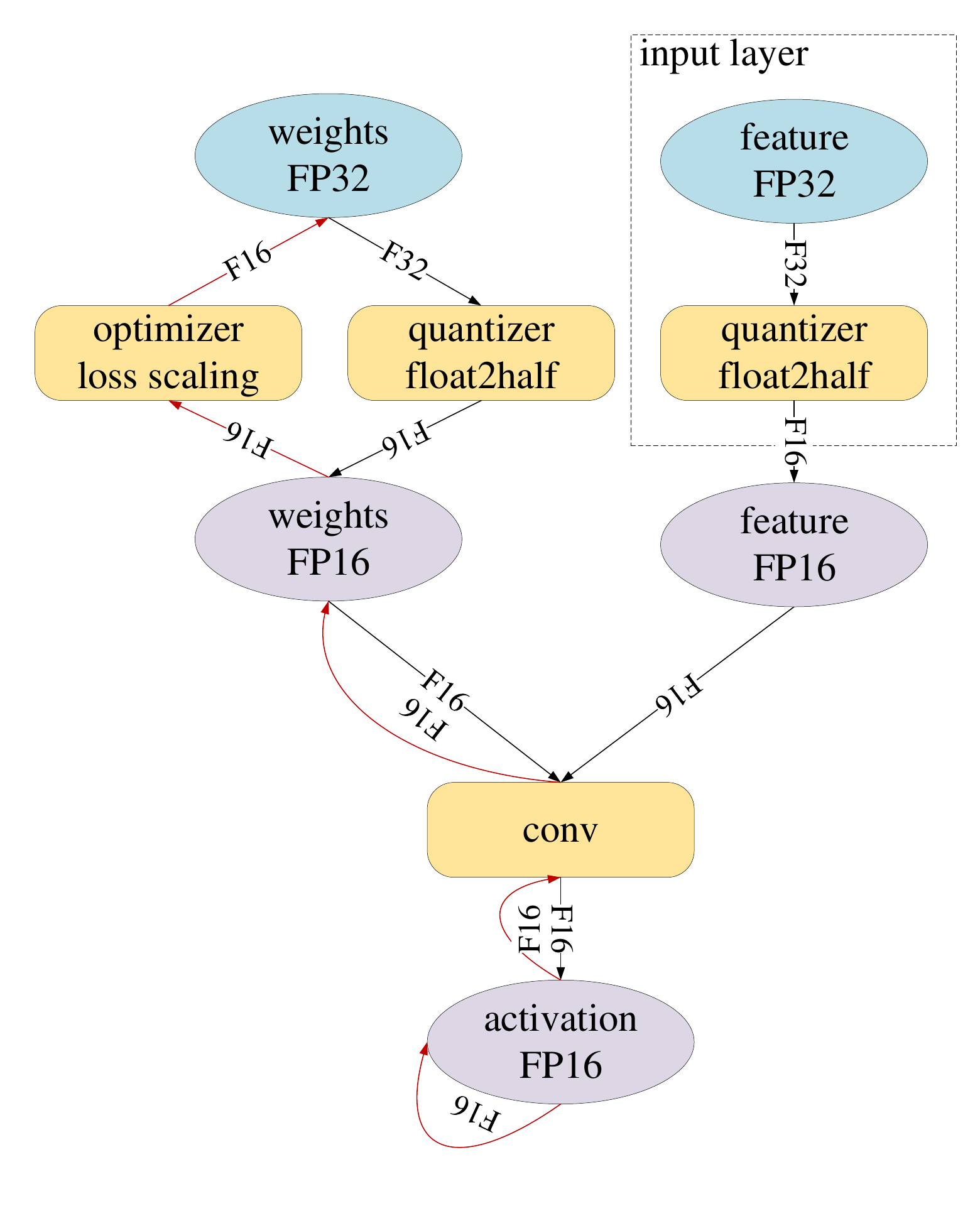}
    \caption{Mixed Precision Training \cite{Micikevicius2017}: FP16 is applied in the forward and backward pass, while FP32 weights are maintained for the update.}
    \label{fig:quantize-mixed-precision-training}
\end{figure}

\paragraph{Low Numerical Precision Training:} 
Training with low numerical precision involves taking the low precision values into both forward and backward propagation while maintaining the full precision accumulated results. Mixed Precision \cite{Micikevicius2017, Das2018} training uses FP16 or 16-bit integer (INT16) for weight precision. This has been shown to be inaccurate for gradient values. As shown in \autoref{fig:quantize-mixed-precision-training}, full precision weights are maintained for gradient updating, while other operands use half-float. A \textit{loss scaling} technique is applied to keep very small magnitude gradients from affecting the computation since any value less than $2^{-24}$ becomes zero in half-precision \cite{Micikevicius2017}. Specifically, a scaler is introduced to the loss value before backpropagation. Typically, the scaler is a bit-shift optimal value $2^n$ obtained empirically or by statistical information.

In TensorFlow-Lite \cite{Jacob2017}, training proceeds with real values while quantization effects are simulated in the forward pass. Real-valued parameters are quantized to lower precision before convolutional layers. BN layers are folded into convolution layers. More details are described in \autoref{sec:quant-deploy-framework}.

As in binarized networks, STE can also be applied to reduced precision training such as 8-bit integers \cite{Krishnamoorthi2018}.

\subsection{Quantization Deployment} \label{sec:quant-deploy}
In this section, we describe implementations of quantization deployed in popular frameworks and hardware. In \autoref{sec:quant-deploy-intro} we give an introduction to deployment issues. In \autoref{sec:quant-deploy-framework}, we discuss deep learning libraries and frameworks. We introduce their specification in \autoref{tab:quantize-libraries} and then compare their performance in \autoref{tab:quant-framework-acc}. We also discuss hardware implementations of DNNs in \autoref{sec:quant-deploy-hw}. Dedicated hardware is designed or programmed to support efficient processing of quantized networks. Specialized CPU and GPU operations are discussed. Finally, in \autoref{sec:quant-deploy-compiler} we discuss DNN compilers. 

\subsubsection{Deployment Introduction} \label{sec:quant-deploy-intro}

With significant resource capability, large organizations and institutions usually have their own proprietary solutions for applications and heterogeneous platforms. Their support to the quantization is either inference only or as well as training. The frameworks don't always follow the same idea of quantization. Therefore there are differences between them, so performs.

With DNNs being applied in many application areas, the issue of efficient use of hardware has received considerable attention.
Multicore processors and accelerators have been developed to accelerate DNN processing. Many types of accelerators have been deployed, including CPUs with instruction enhancements, GPUs, FPGAs, and specialized AI accelerators. Often accelerators are incorporated as part of a heterogeneous system. 
A Heterogeneous System Architecture (HSA) allows the different processors to integrate into a system to simultaneously access shared memory. For example, CPUs and GPUs using cache coherent shared virtual memory on the same System of Chip (SoC) or connected by PCIe with platform atomics can share the same address space \cite{Glossner2016}.
Floating-point arithmetic units consume more energy and take longer to compute compared to integer arithmetic units. Consequently, low-bitwidth architectures are designed to accelerate computation \cite{Moudgill2020}.
Specialized algorithms and efficient hardware can accelerate neural network processing during both training and inference \cite{Reuther2019}.

\subsubsection{Efficient Kernels} \label{sec:quant-deploy-framework}
Typically low precision inference in only executed on convolutional layers. Intermediate values passed between layers use 32-bit floating-point. This makes many of the frameworks amenable to modifications. 

\autoref{tab:quantize-libraries} gives a list of major low precision acceleration frameworks and libraries. Most of them use INT8 precision.  We will next describe some popular and open-source libraries in more detail. 
\input{quant-lib-general}

\paragraph{Tensor RT} \cite{Vanholder2016, Wu2020} is an nVidia developed C++ library that facilitates high-performance inference on NVIDIA GPUs. 
It is a low precision inference library that eliminates the bias term in convolutional layers. 
It requires a calibration set to adjust the quantization thresholds for each layer or channel. Afterwards the quantized parameters are represented by 32-bit floating-point scalar and INT8 weights.

Tensor RT takes a pre-trained floating-point model and generates a reusable optimized 8-bit integer or 16-bit half float model. The optimizer performs network profiling, layer fusion, memory management, and operation concurrency. \autoref{eq:tensorRT} shows the convolution-dequantization dataflow in Tensor RT for 8-bit integers. The intermediate result of convolution by INT8 input feature $\mathbf{F}_{i8}$ and weights $\mathbf{W}_{i8}$ are accumulated into INT32 tensor $\mathbf{O}_{i32}$. They are dequantized by dividing by the feature and weight scales $s_f, s_w$.

\begin{equation}
\label{eq:tensorRT}
        \mathbf{O}_{i32} = \mathbf{F}_{i8} \ast \mathbf{W}_{i8}
        ,\quad
        \mathbf{O}_{f32} = \frac{\mathbf{O}_{i32}}{ s_f \times s_w} 
\end{equation}

Tensor RT applies a variant of \textit{max-abs} quantization to reduce storage requirements and calculation time of the zero point term $z$ in \autoref{eq:quantization-max-abs} by finding the proper threshold instead of the absolute value in the floating-point tensor. 
KL-divergence is introduced to make a trade-off between numerical dynamic range and precision of the INT8 representation \cite{Migacz2017}. KL calibration can significantly help to avoid accuracy loss. 

The method traverses a predefined possible range of scales and calculates the KL-divergences for all the points. It then selects the scale which minimizes the KL-divergence.
KL-divergence is widely used in many post training acceleration frameworks. nVidia found a model calibrated with 125 images showed only 0.36\% top-1 accuracy loss using GoogLeNet on the Imagenet dataset.

\paragraph{Intel MKL-DNN} \cite{Rodriguez2018} is an optimized computing library for Intel processors with Intel AVX-512, AVX-2, and SSE4.2 Instruction Set Architectures (ISA). The library uses FP32 for training and inference. Inference can also be performed using 8-bits in convolutional layers, ReLU activations, and pooling layers. It also uses Winograd convolutions.
MKL-DNN uses \textit{max-abs} quantization shown in \autoref{eq:quantization-max-abs}, where the feature adopts unsigned 8-bit integer $n_f = 256$ and signed 8-bit integer weights  $n_w = 128$. The rounding function $f(\cdot)$ in \autoref{eq:quantization} uses nearest integer rounding. \autoref{eq:quantization-mkl-dnn-quantize} shows the quantization applied on a given tensor or each channel in a tensor. The maximum of weights $R_w$ and features $R_f$ is calculated from the maximum of the absolute value (nearest integer rounding) of the tensor $\mathbb{T}_f$ and $\mathbb{T}_w$. The feature scale $s_f$ and weights scale $s_w$ are generated using $R_w$ and $R_f$. Then quantized 8-bit signed integer weights $\mathbf{W}_{s8}$, 8-bit unsigned integer feature $\mathbf{F}_{u8}$ and 32-bit unsigned integer biases $\mathbf{B}_{u32}$ are generated using the scales and a nearest rounding function $\|\cdot\|$.

\begin{equation}
\label{eq:quantization-mkl-dnn-quantize}
    \begin{split}
        R_{\{f,w\}} &= max((abs(\mathbb{T}_{\{f,w\}})) \\
        s_f &= \frac{255}{R_f},\ s_w = \frac{127}{R_w} \\
        \mathbf{W}_{s8}&=\|s_{w}\times\mathbf{W}_{f32}\| \in[-127,127] \\ 
        \mathbf{F}_{u8}&=\|s_{f}\times\mathbf{F}_{f32}\| \in[0,255] \\ 
        \mathbf{B}_{s32}&=\|s_{f}\times s_{w}\times \mathbf{B}_{f32}\| \in[-2^{31}, 2^{31}-1]
    \end{split}
\end{equation}

An affine transformation using 8-bit multipliers and 32-bit accumulates results in \autoref{eq:quantization-mkl-dnn-f2i} with the same scale factors as defined in \autoref{eq:quantization-mkl-dnn-quantize} and $\ast$ denoting convolution. It is an approximation since rounding is ignored.

\begin{equation}
\label{eq:quantization-mkl-dnn-f2i}
\begin{split}
    \mathbf{O}_{s 32}&=\mathbf{W}_{s 8} \ast \mathbf{F}_{u 8}+\mathbf{b}_{s 32}  \\
    &\approx s_{f} s_{w}\left(\mathbf{W}_{f 32} \ast \mathbf{F}_{f 32}+\mathbf{b}_{f 32}\right) \\
    &=s_{f}\times s_{w}\times \mathbf{O}_{f 32}
\end{split}
\end{equation}

\autoref{eq:quantization-mkl-dnn-appr} is the affine transformation with FP32 format. $D$ is the dequantization factor.

\begin{equation}
\label{eq:quantization-mkl-dnn-appr}
\begin{split}
    \mathbf{O}_{f32}&=\mathbf{W}_{f32} \ast \mathbf{F}_{f32}+\mathbf{b}_{f32} \\
    &\approx \frac{1}{s_{f} s_{w}} \mathbf{O}_{s32}=D \times \mathbf{O}_{s32} \\
    &\text{where}\ D = \frac{1}{s_{f} s_{w}}
\end{split}
\end{equation}

Weight quantization is done prior to inference. Activation quantization factors are prepared by sampling the validation dataset to find a suitable range (similar to Tensor RT). The quantization factors can be either FP32 in the supported devices, or rounded to the nearest power-of-two format to enable bit-shifts. Rounding reduces accuracy by about 1\%.

MKL-DNN assumes activations are non-negative (ReLU activated). Local Response Normalization (LRN), a function to pick the local maximum in a local distribution, is used to avoid over-fitting. BN, FCL, and soft-max using 8-bit inference are not currently supported.

\paragraph{TensorFlow-Lite (TF-Lite)} \cite{Abadi2016} is an open source framework by Google for performing inference on mobile or embedded devices. It consists of two sets of tools for converting and interpreting quantized networks. Both PTQ and QAT are available in TF-Lite.

\textit{GEMM low-precision (Gemmlowp)} \cite{Google2018} is a Google open source gemm library for low precision calculations on mobile and embedded devices. It is used in TF-Lite. Gemmlowp uses asymmetric quantzation as shown in \autoref{eq:quantization-gemmlowp-quantize} where $\mathbf{F}, \mathbf{W}, \mathbf{O}$ denotes feature, weights and output, respectively. $s_f, s_w$ are the scales for feature and weights, respectively. $\mathbf{F}_{f32}$ is Feature value in 32-bit floating. Similarly, $\mathbf{W}_{f32}$ is the Weight value in 32-bit floating point. $\mathbf{F}_q, \mathbf{W}_q$ are the quantized Features and Weights, respectively. Asymmetric quantization introduces the zero points ($\mathbf{z}_f$ and $\mathbf{z}_w$). This produces a more accurate numerical encoding.

\begin{equation}
\label{eq:quantization-gemmlowp-quantize}
\begin{split}
\mathbf{O}_{f32} &= \mathbf{F}_{f32} \ast \mathbf{W}_{f32} \\
&= s_f \times (\mathbf{F}_q + \mathbf{z}_f) \ast s_w \times (\mathbf{W}_q + \mathbf{z}_w) \\
&= s_f \times s_w \times \underline{(\mathbf{F}_q + \mathbf{z}_f) \ast (\mathbf{W}_q + \mathbf{z}_k)}
\end{split}
\end{equation}

The underlined part in \autoref{eq:quantization-gemmlowp-quantize} is the most computationally intensive. In addition to the convolution, the zero point also requires calculation. Gemmlowp reduces many multi-add operations by multiplying an all-ones matrix as the \textit{bias} matrix $P$ and $Q$ in \autoref{eq:quantization-gemmlowp-trick}. This allows four multiplies to be dispatched in a three stage pipeline \cite{Krishnamoorthi2018},  to produce the quantized output $\mathbf{O}_q$. $\mathbf{F}, \mathbf{W}, \mathbf{z}$ are the same as in \autoref{eq:quantization-gemmlowp-quantize}.

\begin{equation} 
\label{eq:quantization-gemmlowp-trick}
\begin{split}
\mathbf{O}_q &= (\mathbf{F}_q + \mathbf{z}_f \times P) \ast (\mathbf{W}_q + \mathbf{z}_w \times Q) \\
&= \mathbf{F}_q \ast \mathbf{W}_q \\
&+ \mathbf{z}_f \times \mathbf{P} \times \mathbf{W}_q \\
&+ \mathbf{z}_w \times \mathbf{Q} \times \mathbf{F}_q \\
&+ \mathbf{z}_f \times \mathbf{z}_w \times \mathbf{P} \times \mathbf{Q}
\end{split}
\end{equation}

\paragraph{Ristretto} \cite{Gysel2018} is a tool for Caffe quantization. It uses retraining to adjust the quantized parameters. 
Ristretto uses a three-part quantization strategy: 1) a modified fixed-point format Dynamic Fixed Point (DFP) which permits the limited bit-width precision to dynamically carry data, 2) bit-width reduced floating-point numbers called mini float which follows the IEEE-754 standard \cite{Society2008}, and 3) integer power of 2 weights that force parameters into power of 2 values to replace multiplies with bit shift operations.

DPF is shown in \autoref{eq:quantization-ristretto-dynamic-fixed-point} where $s$ takes one sign bit, $\text{FL}$ denotes the fractional length, and $x$ is the mantissa. The total bit-width is $B$. This quantization can encode data from various ranges to a proper format by adjusting the fractional length.

\begin{equation}
\label{eq:quantization-ristretto-dynamic-fixed-point}
    (-1)^{s} \cdot 2^\text{-FL} \sum_{i=0}^{B-2} 2^{i} \cdot x_{i}
\end{equation}

A bit shift convolution conversion is shown in \autoref{eq:quantization-ristretto-pow-of-2-weiths}. The convolution by input $\mathbf{F}_j$ and weights $\mathbf{W}_j$ and bias $\mathbf{b}_i$ are transformed into shift arithmetic by rounding the weights to the nearest power of 2 values. Power of 2 weights provides inference acceleration while dynamic fixed point provides better accuracy.

\begin{equation}
\label{eq:quantization-ristretto-pow-of-2-weiths}
    \begin{split}
        \mathbf{O}_{i}&=\sum_{j}\left[\mathbf{F}_{j} \cdot \mathbf{W}_{j}\right]+\mathbf{b}_{i} \\
        &\approx \sum_{j}\left[\mathbf{F}_{j} \ll \operatorname{ round }\left(\log _{2}\left(\mathbf{W}_{j}\right)\right)\right]+\mathbf{b}_{i}
    \end{split}
\end{equation}

\paragraph{NCNN} \cite{Tencent2019} is a standalone framework from Tencent for efficient inference on mobile devices. Inspired by Ristretto and Tensor-RT, it works with multiple operating systems and supports low precision inference \cite{BUG19892019}. It performs channel-wise quantization with KL calibration. The quantization results in 0.04\% top-1 accuracy loss on ILSVRC-2012.
NCNN has implementations optimized for ARM NEON. NCNN also replaces $3\times3$ convolutions with simpler Winograd convolutions \cite{Lavin2016}.

\paragraph{Mobile AI Compute Engine (MACE)} \cite{Xiaomi2019} from Xiaomi supports both post-training quantization and quantization-aware training. Quantization-aware training is recommended as it exhibits lower accuracy loss
. Post-training quantization requires statistical information from activations collected while performing inference. This is typically performed with batch calibration of input data.
MACE also supports processor implementations optimized for ARM NEON and Qualcomm's Hexagon digital signal processor. OpenCL acceleration is also supported. Winograd convolutions can be applied for further acceleration as discussed in \autoref{sec:quant-method-log}.

\paragraph{Quantized Neural Network PACKage (QNNPACK)} \cite{Dukhan2018}
is a Facebook produced open-source library optimized for edge computing especially for mobile low precision neural network inference. 
It has the same method of quantization as TF-Lite including using a zero-point.
The library has been integrated into PyTorch \cite{Paszke2019} to provide users a high-level interface.
In addition to Winograd and FFT convolution operations, the library has optimized gemm for cache indexing and feature packing.
QNNPACK has a full compiled solution for many mobile devices and has been deployed on millions of devices with Facebook applications. 

\begin{figure}[!tbh]
    \centering
    \includegraphics[width=0.45\textwidth, page=1]{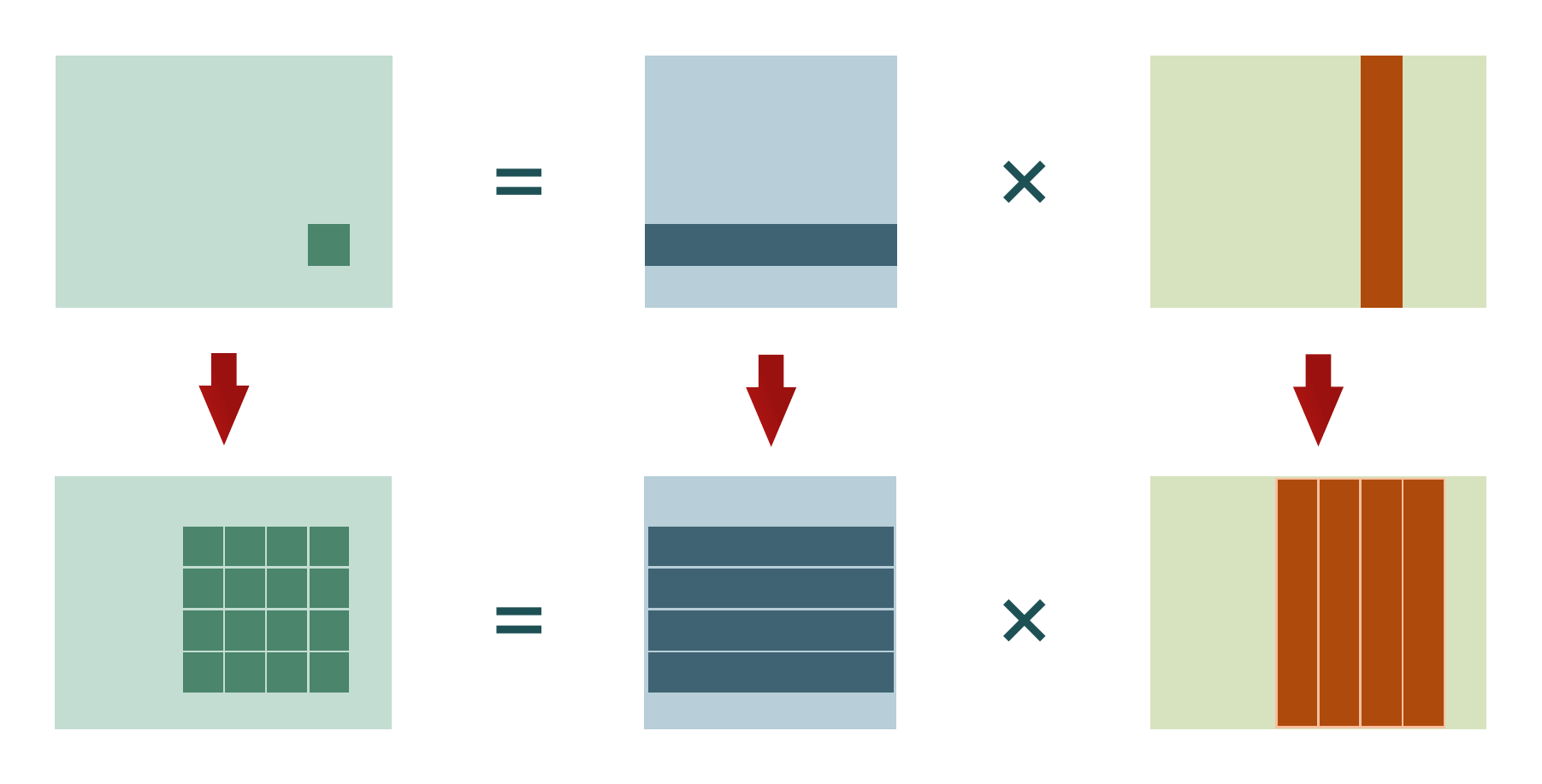}
    \caption{PDOT: computing dot product for several points in parallel.}
    \label{fig:quantize-qnnpack-optimize}
\end{figure}

Panel Dot product (PDOT) is a key feature of QNNPACK's highly efficient gemm library. It assumes computing efficiency is limited with memory, cache, and bandwidth instead of Multiply and Accumulate (MAC) performance. PDOT computes multiple dot products in parallel as shown in \autoref{fig:quantize-qnnpack-optimize}. Rather than loading just two operands per MAC operation, PDOT loads multiple columns and rows. This improves convolution performance about $1.41\times$~$2.23\times$ speedup for MobileNet on mobile devices \cite{Dukhan2018}.

\paragraph{Paddle} \cite{Baidu2019}
applies both QAT and PTQ quantization with using zero-points. The dequantization operation can be performed prior to convolution as shown in \autoref{eq:quantize-dequantize-before-gemm}. Paddle uses this feature to do floating-point gemm-based convolutions with quantize-dequantized weights and features within the framework data-path. It introduces quantization error while maintaining the data in format of floating-point. This quantize-dequantize-convolution pipeline is called simu-quantize and its results are approximately equal to a FP32->INT8->Convolutional->FP32 (quantize - convolutional - dequantize) three stage model.

Simu-quantize maintains the data at each phase in 32-bit floating-point facilitating backward propagation. In the Paddle framework, during backpropagation, gradients are added to the original 32-bit floating-point weights rather than the quantized or the quantize-dequantized weights.
\begin{equation}
    \mathbf{O}_{f32} = (\frac{\mathbf{F}_q}{(n - 1)} \times \mathbf{F}_{max}) \ast (\frac{\mathbf{W}_q}{(n - 1)} \times \mathbf{W}_{max})
\label{eq:quantize-dequantize-before-gemm}
\end{equation}

Paddle uses \textit{max-abs} in three ways to quantize parameters: 
1) the average of the max absolute value in a calculation window, 2) the max absolute value during a calculation window, and 3) a sliding average of the max absolute value of the window. 
The third method is described in \autoref{eq:quantize-paddle-sliding-avg-max}, where $V$ is the max absolute value in the current batch, $V_t$ is the average value of the sliding window, and $k$ is a coefficient chosen by default as 0.9.

The Paddle framework uses a specialized toolset, PaddleSlim, which supports Quantization, Pruning, Network Architecture Search, and Knowledge Distilling. They found 86.47\% size reduction of ResNet-50, with 1.71\% ILSVRC-2012 top-1 accuracy loss.
\begin{equation}
V_t = (1 - k) \times V + k \times V_{t-1}
\label{eq:quantize-paddle-sliding-avg-max}
\end{equation}
\input{quant-lib-acc}

\subsubsection{Hardware Platforms} \label{sec:quant-deploy-hw}
\begin{figure*}
    \centering
    \includegraphics[width=0.95\textwidth]{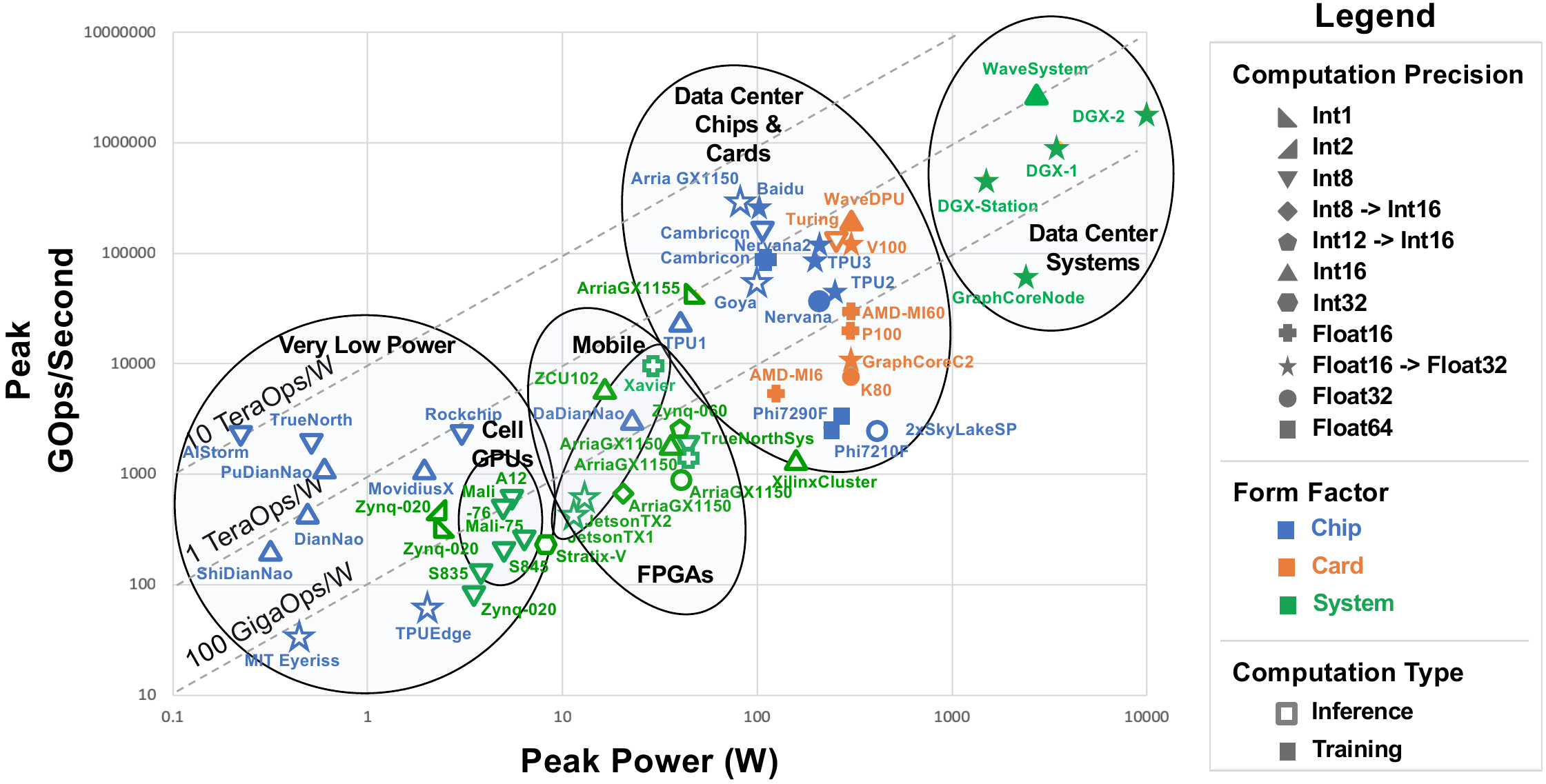}
    \caption{Hardware platforms for neural networks efficiency deploy, adopted from \cite{Reuther2019}.}
    \label{fig:quant-hw-all}
\end{figure*}

\autoref{fig:quant-hw-all} shows AI chips, cards, and systems plotted by peak operations verses power in log scale originally published in \cite{Reuther2019}. Three normalizing lines are shown at 100 GOPS/Watt, 1 TOP/Watt, and 10 TOPs/Watt. Hardware platforms are classified along several dimensions including: 1) training or inference, 2) chip, card, or system form factors, 3) datacenter or mobile, and 4) numerical precision. We focus on low precision general and specialized hardware in this section.
\paragraph{Programmable Hardware:}

Quantized networks with less than 8-bits of precision are typically implemented in FPGAs but may also be executed on general purpose processors.

BNN's have been implemented on a Xilinx Zynq heterogeneous FPGA platform \cite{Zhao2017}. They have also been implemented on Intel Xeon CPUs and Intel Arria 10 FPGA heterogeneous platforms by dispatching bit operation to FPGAs and other operations to CPUs \cite{Moss2017}. The heterogeneous system shares the same memory address space. Training is typically mapped to CPUs.
FINN \cite{Umuroglu2016a} is a specialized framework for BNN inference on FPGAs. It contains binarized fully connected, convolutional, and pooling layers. When deployed on a Zynq-7000 SoC, FINN has achieved 12.36 million images per second on the MNIST dataset with 4.17\% accuracy loss. 

Binarized weights with 3-bit features have been implemented on Xilinx Zynq FPGAs and Arm NEON processors \cite{Preuser2018}. The first and last layer of the network use 8-bit quantities but all other layers use binary weights and 3-bit activation values. On an embedded platform, Zynq XCZU3EG, they performed 16 images per second for inference.
To accelerate Tiny-YOLO inference, significant efforts were taken including: 1) replacing max-pool with stride 2 convolution, 2) replacing leaky ReLU with ReLU, and 3) revising the hidden layer output channel. The improved efficiency on the FPGA from 2.5 to 5 frames per second with 1.3\% accuracy loss.

TNN \cite{Alemdar2017} is deployed on an FPGA with specialized computation units optimized for ternary value multiplication. A specific FPGA structure (dimensions) is determined during synthesis to improve hardware efficiency. On the Sakura-X FPGA board they achieved 255k MNIST image classifications per second with an accuracy of 98.14\%. A scalable design implemented on a Xilinx Virtex-7 VC709 board dramatically reduced hardware resources and power consumption but at a significantly reduced throughput of 27k CIFAR-10 images per second \cite{Prost-Boucle2017}. Power consumption for CIFAR-10 was 6.8 Watts.

Reducing hardware costs is a key objective of logarithmic hardware. Xu \cite{Xu2019} adopted $\sqrt{2}$ based logarithmic quantization with 5-bits of resolution. This showed 50.8\% top-1 accuracy and dissipated a quarter of the power while using half the chip area. Half precision inference has a top-1 accuracy of 53.8\%.

\paragraph{General Hardware:}
In addition to specialized hardware, INT8 quantization has been widely adopted in many general purpose processor architectures. In this section we provide a high-level overview. A detailed survey on hardware efficiency for processing DNNs can be found in \cite{Reuther2019}.

CNN acceleration on ARM CPUs was originally implemented by ARM advanced SIMD extensions known as NEON. The ARM 8.2 ISA extension added NEON support for 8-bit integer matrix operations \cite{Arm2015}. These were implemented in the CPU IP cores Cortex-A75 and A55 \cite{Arm2020a} as well as the Mali-G76 GPU IP core \cite{Arm2020}. These cores have been integrated into the Kirin SoC by Huawei, Qualcomm Snapdragon SoC, MediaTek Helio SoC, and Samsung Exynos \cite{Ignatov2019}. 
For example on Exynos 9825 Octa, 8-bit integer quantized MobileNet v2 can process an image in 19ms (52 images per second) using the Mali-G76 \cite{Ignatov2019}.


Intel improved the integer performance about 33\% with Intel Advanced Vector Extension 512 (AVX-512) ISA \cite{Rodriguez2018}. This 512-bit SIMD ISA extension included a Fused Multiply-Add (FMA) instruction. 


Low precision computation on nVidia GPUs was enabled since the Pascal series of GPUs \cite{NVIDIACorporation2015}. The Turing GPU architecture \cite{NVIDIACorporation2018} introduced specialized units to processes INT4 and INT8. This provides real-time integer performance on AI algorithms used in games.
For embedded platforms, nVidia developed Jetson platforms \cite{NVIDIACorporation2018a}. They use CUDA Maxwell cores \cite{NVIDIACorporation2014} that can process half-precision types.
For the data center, nVidia developed the extremely high performance DGX system \cite{NVIDIACorporation2017}. It contains multiple high-end GPUs interconnected using nVidia's proprietary bus nVLINK. A DGX system can perform 4-bit integer to 32-bit floating point operations.

\subsubsection{DNN Compilers}  \label{sec:quant-deploy-compiler}

Heterogeneous neural networks hardware accelerators are accelerating deep learning algorithm deployment \cite{Reuther2019}. Often exchange formats can be used to import/export models. Further, compilers have been developed to optimize models and generate code for specific processors. However several challenges remain: 

\begin{itemize}
    \item Network Parsing: Developers design neural network models on different platforms using various frameworks and programming languages. However, they have common parts, such as convolution, activation, pooling, etc. Parsing tools analyze the model compositions and transfer them into the unified representation.
    \item Structure Optimization: The model may contain operations used in training that aren't required for inference. Tool-kits and compilers should optimize these structures (e.g. BN folding as discussed in \autoref{sec:cnn-bn}). 
    \item Intermediate Representation (IR): An optimized model should be properly stored for further deployment. Since the inference engine is uncertain, the stored IR should include the model architecture and the trained weights. A compiler can then read the model and optimize it for a specific inference engine.
    \item Compression: Compilers and optimizers should optionally be able to automatically compress arbitrary network structures using pruning and quantization.
    \item Deployment: The final optimized model should be mapped \linebreak[4] to the target engine(s) which may be heterogeneous.
\end{itemize}

Open Neural Network Exchange (ONNX) \cite{ONNX} is an open-source tool to parse AI models written for a variety diverse frameworks. It imports and exports models using an open-source format facilitating the translation of neural network models between frameworks. It is thus capable of network parsing provided low-level operations are defined in all target frameworks.

TVM \cite{Chen2018}, Glow \cite{Rotem2018}, OpenVINO \cite{Intel}, and MLIR \cite{Lattner2020} are deep learning compilers. They differ from frameworks such as Caffe in that they store intermediate representations and optimize those to map models onto specific hardware engines. They typically integrate both quantization-aware training and calibration-based post-training quantization. We summarize key features below. They perform all the operations noted in our list. A detailed survey can be found in \cite{Li2020}.

TVM \cite{Chen2018} leverages the efficiency of quantization by enabling deployment of quantized models from PyTorch and TF-Lite. As a compiler, TVM has the ability to map the model on general hardware such as Intel's AVX and nVidia's CUDA.

Glow \cite{Rotem2018} enables quantization with zero points and converts the data into 8-bit signed integers using a calibration-based method. Neither Glow or TVM currently support quantization-aware training although they both announced future support for it \cite{Rotem2018}.

MLIR \cite{Lattner2020} and OpenVINO \cite{Intel} have sophisticated quantization support including quantization-aware training. OpenVINO integrates it in TensorFlow and PyTorch while MLIR natively supports quantization-aware training. This allows users to fine-tune an optimized model when it doesn't satisfy accuracy criteria. 

\subsection{Quantization Reduces Over-fitting} \label{sec:quant-reduce-OF}



In addition to accelerating neural networks, quantization has also been found in some cases to result in higher accuracy. As examples: 
1) 3-bit weights VGG-16 outperforms its full precision counterpart by 1.1\% top-1 \cite{Leng2017}, 
2) AlexNet reduces 1.0\% top-1 error of the reference with 2-bit weights and 8-bit activations \cite{Faraone2018}, 
3) ResNet-34 with 4-bit weights and activation obtained 74.52\% top-1 accuracy while the 32-bit version is 73.59\% \cite{Mishra2018}, 
4) Zhou showed a quantized model reduced the classification error by 0.15\%, 2.28\%, 0.13\%, 0.71\%, and 1.59\% on AlexNet, VGG-16, GoogLeNet, ResNet-18 and ResNet-50, respectively \cite{Zhou2017a}, and
5) Xu showed reduced bit quantized networks help to reduce over-fitting on Fully Connected Networks (FCNs). By taking advantage of strict constraints in biomedical image segmentation they improved segmentation accuracy by 1\% combined with a $6.4\times$ memory usage reduction \cite{Xu2018a}.


%% file: quant-lib-general.tex
\begin{table*}
    \centering
    \caption{Low Precision Libraries Using Quantization: QAT is quantization-aware training, PTQ is  post-training quantization, and offset indicates the zero point $z$ in \autoref{eq:quantization}.}
    \resizebox{1\textwidth}{!}{
        \begin{tabular}{lllllll}
        \toprule
        Name & \multicolumn{1}{p{7.72em}}{Institution} & Core Lib & \multicolumn{1}{p{4.5em}}{Precision} & \multicolumn{1}{p{12.89em}}{Method} & Platform & Open-sourced \\
        \midrule
        ARM CMSIS NN \cite{Keil2018} & Arm & CMSIS & 8-bit & deploy only & Arm Cortex-M Processor & No \\
        MACE \cite{Xiaomi2019} & XiaoMi & -  & 8-bit & QAT and PTQ & Mobile - CPU, Hexagon Chips, MTK APU & Yes \\
        MKL-DNN \cite{Rodriguez2018} & Intel               & -  & 8-bit & PTQ, mixed offset, and QAT & Intel AVX Core & Yes \\
        NCNN \cite{Tencent2019} & Tencent             & -  & 8-bit & PTQ w/o offset & Mobile Platform & Yes \\
        Paddle \cite{Baidu2019} & Baidu              & -  & 8-bit & QAT and PTQ w/o offset & Mobile Platform & Yes \\
        QNNPACK \cite{Dukhan2018} & Fackbook            & -  & 8-bit & PTQ w/ offset & Mobile Platform & Yes \\
        Ristretto \cite{Gysel2018} & LEPS & gemm & 3 method & QAT & Desktop Platform & Yes \\
        SNPE \cite{SNPE} & Qualcomm & -  & 16/8-bit & PTQ w/ offset, max-min & Snapdragon CPU, GPU, DSP & No \\
        Tensor-RT \cite{Migacz2017} & nVidia              & -  & 8-bit & PTQ w/o offset & nVidia GPU & Yes \\
        TF-Lite \cite{Abadi2016} & Google              & gemmlowp & 8-bit & PTQ w/ offset & Mobile Platform & Yes \\
        \bottomrule
        \end{tabular}%
  }
    \label{tab:quantize-libraries}
\end{table*}%

%% file: quant-lib-acc.tex
\begin{table*}[htbp]
  \centering 
  \caption{Low Precision Libraries versus Accuracy for Common Networks in Multiple Frameworks.}
  \resizebox{0.95\textwidth}{!}{
    \begin{tabular}{lllcccccc}
    \toprule
    \multicolumn{1}{c}{\multirow{2}[4]{*}{Name}} & \multicolumn{1}{c}{\multirow{2}[4]{*}{Framework}} & \multicolumn{1}{c}{\multirow{2}[4]{*}{Method}} & \multicolumn{2}{c}{Accuracy Float} & \multicolumn{2}{c}{Accuracy Quant} & \multicolumn{2}{c}{Accuracy Diff} \\
\cmidrule{4-9}       &    &    & Top-1 & Top-5 & Top-1 & Top-5 & Top-1 & Top-5 \\
    \midrule
    AlexNet & TensorRT \cite{Migacz2017} & PTQ, w/o offset & 57.08\% & 80.06\% & 57.05\% & 80.06\% & -0.03\% & 0.00\% \\
       & Ristretto \cite{Gysel2018} & Dynamic FP & 56.90\% & 80.09\% & 56.14\% & 79.50\% & -0.76\% & -0.59\% \\
       & Ristretto \cite{Gysel2018} & Minifloat & 56.90\% & 80.09\% & 52.26\% & 78.23\% & -4.64\% & -1.86\% \\
       & Ristretto \cite{Gysel2018} & Pow-of-two & 56.90\% & 80.09\% & 53.57\% & 78.25\% & -3.33\% & -1.84\% \\
    \midrule
    \multicolumn{1}{p{6em}}{GoogleNet} & NCNN \cite{BUG19892019} & PTQ, w/o offset & 68.50\% & 88.84\% & 68.62\% & 88.68\% & 0.12\% & -0.16\% \\
       & TensorRT \cite{Migacz2017} & PTQ, w/o offset & 68.57\% & 88.83\% & 68.12\% & 88.64\% & -0.45\% & -0.19\% \\
       & Ristretto \cite{Gysel2018} & Dynamic FP & 68.93\% & 89.16\% & 68.37\% & 88.63\% & -0.56\% & -0.53\% \\
       & Ristretto \cite{Gysel2018} & Minifloat & 68.93\% & 89.16\% & 64.02\% & 87.69\% & -4.91\% & -1.47\% \\
       & Ristretto \cite{Gysel2018} & Pow-of-two & 68.93\% & 89.16\% & 57.63\% & 81.38\% & -11.30\% & -7.78\% \\
    \midrule
    Inception v3 & TF-Lite \cite{Google-tf-lite} & PTQ & 78.00\% & 93.80\% & 77.20\% & \multicolumn{1}{p{3.39em}}{-} & -0.80\% & - \\
       & TF-Lite \cite{Google-tf-lite} & QAT & 78.00\% & 93.80\% & 77.50\% & 93.70\% & -0.50\% & -0.10\% \\
    \midrule
    \multicolumn{1}{p{6em}}{MobileNet v1} & NCNN \cite{BUG19892019} & PTQ, w/o offset & 67.26\% & 87.92\% & 66.74\% & 87.43\% & -0.52\% & -0.49\% \\
       & Paddle \cite{Baidu2019} & QAT+Pruning & 70.91\% & \multicolumn{1}{p{3.39em}}{-} & 69.20\% & \multicolumn{1}{p{3.39em}}{-} & -1.71\% & - \\
       & TF-Lite \cite{Google-tf-lite} & PTQ & 70.90\% & \multicolumn{1}{p{3.39em}}{-} & 65.70\% & \multicolumn{1}{p{3.39em}}{-} & -5.20\% & - \\
       & TF-Lite \cite{Google-tf-lite} & QAT & 70.90\% & \multicolumn{1}{p{3.39em}}{-} & 70.00\% & \multicolumn{1}{p{3.39em}}{-} & -0.90\% & - \\
    \midrule
    \multicolumn{1}{p{6em}}{MobileNet v2} & QNNPACK \cite{Dukhan2018} & PTQ, w/ offset & 71.90\% & \multicolumn{1}{p{3.39em}}{-} & 72.14\% & \multicolumn{1}{p{3.39em}}{-} & 0.24\% & - \\
       & TF-Lite \cite{Google-tf-lite} & PTQ & 71.90\% & \multicolumn{1}{p{3.39em}}{-} & 63.70\% & \multicolumn{1}{p{3.39em}}{-} & -8.20\% & - \\
       & TF-Lite \cite{Google-tf-lite} & QAT & 71.90\% & \multicolumn{1}{p{3.39em}}{-} & 70.90\% & \multicolumn{1}{p{3.39em}}{-} & -1.00\% & - \\
    \midrule
    ResNet-101 & TensorRT \cite{Migacz2017} & PTQ, w/o offset & 74.39\% & 91.78\% & 74.40\% & 91.73\% & 0.01\% & -0.05\% \\
       & TF-Lite \cite{Google-tf-lite} & PTQ & 77.00\% & \multicolumn{1}{p{3.39em}}{-} & 76.80\% & \multicolumn{1}{p{3.39em}}{-} & -0.20\% & - \\
    \midrule
    ResNet-152 & TensorRT \cite{Migacz2017} & PTQ, w/o offset & 74.78\% & 91.82\% & 74.70\% & 91.78\% & -0.08\% & -0.04\% \\
    \midrule
    \multicolumn{1}{p{6em}}{ResNet-18} & NCNN \cite{BUG19892019} & PTQ, w/o offset & 65.49\% & 86.56\% & 65.30\% & 86.52\% & -0.19\% & -0.04\% \\
    \midrule
    \multicolumn{1}{p{6em}}{ResNet-50} & NCNN \cite{BUG19892019} & PTQ, w/o offset & 71.80\% & 89.90\% & 71.76\% & 90.06\% & -0.04\% & 0.16\% \\
       & TensorRT \cite{Migacz2017} & PTQ, w/o offset & 73.23\% & 91.18\% & 73.10\% & 91.06\% & -0.13\% & -0.12\% \\
    \midrule
    \multicolumn{1}{p{6em}}{SqueezeNet} & NCNN \cite{BUG19892019} & PTQ, w/o offset & 57.78\% & 79.88\% & 57.82\% & 79.84\% & 0.04\% & -0.04\% \\
       & Ristretto \cite{Gysel2018} & Dynamic FP & 57.68\% & 80.37\% & 57.21\% & 79.99\% & -0.47\% & -0.38\% \\
       & Ristretto \cite{Gysel2018} & Minifloat & 57.68\% & 80.37\% & 54.80\% & 78.28\% & -2.88\% & -2.09\% \\
       & Ristretto \cite{Gysel2018} & Pow-of-two & 57.68\% & 80.37\% & 41.60\% & 67.37\% & -16.08\% & -13.00\% \\
    \midrule
    VGG-19 & TensorRT \cite{Migacz2017} & PTQ, w/o offset & 68.41\% & 88.78\% & 68.38\% & 88.70\% & -0.03\% & -0.08\% \\
    \bottomrule
    \end{tabular}%
    }
  \label{tab:quant-framework-acc}%
\end{table*}%

%% file: 4-tail.tex
\section{Summary} \label{sec:summary}
In this section we summarize the results of Pruning and Quantization.

\subsection{Pruning} \label{sec:summary-pruning}

\autoref{sec:pruning} shows pruning is an important technique for compressing neural networks. In this paper, we discussed pruning techniques categorized as 1) static pruning and 2) dynamic pruning. Previously, static pruning was the dominant area of research. Recently, dynamic pruning has become a focus because it can further improve performance even if static pruning has first been performed. 


Pruning can be performed in multiple ways. Element-wise pruning improves weight compression and storage. \linebreak[4] Channel-wise and shape-wise pruning can be accelerated with specialized hardware and computation libraries. Filter-wise and layer-wise pruning can dramatically reduce computational complexity. 

Though pruning sometimes introduces incremental improvement in accuracy by escaping a local minima \cite{Augasta2013}, accuracy improvements are better realized by switching to a better network architecture \cite{Blalock2020}. For example, a separable block may provide better accuracy with reduced computational complexity \cite{Howard2017}. 
Considering the evolution of network structures, performance may also be bottlenecked by the structure itself. From this point of view, Network Architecture Search and Knowledge Distillation can be options for further compression.
Network pruning can be viewed as a subset of NAS but with a smaller searching space. This is especially true when the pruned architecture no longer needs to use weights from the unpruned network (see \autoref{sec:pruning-comparison}). In addition, some NAS techniques can also be applied to the pruning approach including borrowing trained coefficients and reinforcement learning search. 

Typically, compression is evaluated on large data-sets such as the ILSVRC-2012 dataset with one thousand object categories. In practice, resource constraints in embedded devices don't allow a large capacity of optimized networks. Compressing a model to best fit a constrained environment should consider but not be limited to the deployment environment, target device, speed/compression trade-offs, and accuracy requirements \cite{Cai2019a}.

Based on the reviewed pruning techniques, we recommend the following for effective pruning:
\begin{itemize}
    \item Uniform pruning introduces accuracy loss therefore setting the pruning ratio to vary by layers is better \cite{Liu2019}. 
    \item Dynamic pruning may result in higher accuracy and maintain higher network capacity \cite{Wu2018a}.
    \item Structurally pruning a network may benefit from maturing libraries especially when pruning at a high level \cite{Wen2016a}.
    \item Training a pruned model from scratch sometimes, but not always (see \autoref{sec:pruning-comparison}), is more efficient than tunning from the unpruned weights \cite{Liu2019Rethingking}. 
    \item Penalty-based pruning typically reduces accuracy loss compared with magnitude-based pruning \cite{Ye2018}. However, recent efforts are narrowing the gap \cite{Gale2019}.
\end{itemize}

\subsection{Quantization} \label{sec:summary-quantization}
\autoref{sec:quantization} discusses quantization techniques. It describes binarized quantized neural networks, and reduced precision networks, along with their training methods. We described low-bit dataset validation techniques and results. We also list the accuracy of popular quantization frameworks and described hardware implementations in \autoref{sec:quant-deploy}.

Quantization usually results in a loss of accuracy due to information lost during the quantization process. This is particularly evident on compact networks. Most of the early low bit quantization approaches only compare performance on small datasets (e.g., MNIST, and CIFAR-10) \cite{Dettmers2015,Han2015a,Lin2015,Rastegari2016,Venkatesh2017b,Zhou2017a}. However, observations showed that some quantized networks could outperform the original network (see: \autoref{sec:quant-reduce-OF}).
Additionally,  non-uniform distribution data may lead to further deterioration in quantization performance \cite{Zhu2019}. Sometimes this can be ameliorated by normalization in fine-tuning \cite{Micikevicius2017} or by non-linear quantization (e.g., log representation) \cite{Miyashita2016}.

Advanced quantization techniques have improved accuracy. Asymmetric quantization \cite{Jacob2017} maintains higher dynamic range by using a zero point in addition to a regular scale parameter. Overheads introduced by the zero point were minimized by pipelining the processing unit. Calibration based quantization \cite{Migacz2017} removed zero points and replaced them with precise scales obtained from a calibrating dataset. Quantization-aware training was shown to further improve quantization accuracy.

8-bit quantization is widely applied in practice as a good trade-off between accuracy and compression. It can easily be deployed on current processors and custom hardware. Minimal accuracy loss is experienced especially when quantization-aware training is enabled. 
Binarized networks have also achieved reasonable accuracy with specialized hardware designs. 

Though BN has advantages to help training and pruning, an issue with BN is that it may require a large dynamic range across a single layer kernel or between different channels. This may make layer-wise quantization more difficult. Because of this per channel quantization is recommended \cite{Krishnamoorthi2018}.


To achieve better accuracy following quantization, we recommend:  
\begin{itemize}
    \item Use asymmetrical quantization. It preserves flexibility over the quantization range even though it has computational overheads \cite{Jacob2017}.
    \item Quantize the weights rather than the activations. Activation is more sensitive to numerical precision \cite{Gong2019}. 
    \item Do not quantize biases. They do not require significant storage. High precision biases in all layers \cite{Hwang2014}, and first/last layers \cite{Rastegari2016, Zhou2016a}, maintain higher network accuracy. 
    \item Quantize kernels channel-wise instead of layer-wise to significantly improve accuracy \cite{Krishnamoorthi2018}.
    \item Fine-tune the quantized model. It reduces the accuracy gap between the quantized model and the real-valued model \cite{Wu2018}.
    \item Initially train using a 32-bit floating point model. Low-bit quantized model can be difficult to train from scratch - especially compact models on large-scaled data-sets \cite{Zhou2016a}. 
    \item The sensitivity of quantization is ordered as gradients, activations, and then weights \cite{Zhou2016a}.
    \item Stochastic quantization of gradients is necessary when training quantized models \cite{Gupta2015, Zhou2016a}.
\end{itemize}

\section{Future Work} \label{sec:future-work}

Although punning and quantization algorithms help reduce the computation cost and bandwidth burden, there are still areas for improvement. In this section we highlight future work to further improvement quantization and prunning. 

\textbf{Automatic Compression.} Low bit width quantization can cause significant accuracy loss, especially when the quantized bit-width is very narrow and the dataset is large \cite{Zhou2016a, Lin2017a}. Automatic quantization is a technique to automatically search quantization encoding to evaluate accuracy loss verses compression ratio. Similarly, automatic prunning is a technique to automatically search different prunning approaches to evaluate the sparsity ratio versus accuracy. Similar to hyperparameter tuning \cite{Yogatama2014}, this can be performed without human intervention using any number of search techniques (e.g. random search, genetic search, etc.). 

\textbf{Compression on Other Types of Neural Networks.} Current compression research is primarily focused on CNNs. More specifically, research is primarily directed towards CNN classification tasks. Future work should also consider other types of applications such as object detection, speech recognition, language translation, etc. Network compression verses accuracy for different applications is an interesting area of research.

\textbf{Hardware Adaptation.} Hardware implementations may limit the effectiveness of pruning algorithms. For example, element-wise pruning only slightly reduces computations or bandwidth when using im2col-gemm on GPU \cite{Zhang2016}. Similarly, shape-wise pruning is not typically able to be implemented on dedicated CNN accelerators. Hardware-software co-design of compression techniques for hardware accelerators should be considered to achieve the best system efficiency.

\textbf{Global Methods.} Network optimizations are typically applied separately without information from one optimization informing any other optimization. Recently, approaches that consider optimization effectiveness at multiple layers have been proposed. \cite{Li2020b} discusses pruning combined with tensor factorization that results in better overall compression. Similar techniques can be considered using different types and levels of compression and factorization.


\section{Conclusions} \label{sec:discussion}

Deep neural networks have been applied in many applications exhibiting extraordinary abilities in the field of computer vision. However, complex network architectures challenge efficient real-time deployment and require significant computation resources and energy costs. These challenges can be overcome through optimizations such as network compression. Network compression can often be realized with little loss of accuracy. In some cases accuracy may even improve. 

Pruning can be categorized as static (\autoref{sec:pruning-static}) if it is performed offline or dynamic (\autoref{sec:pruning-dynamic}) if it is performed at run-time. The criteria applied to removing redundant computations if often just a simple magnitude of weights with values near zero being pruned. More complicated methods include checking the $l_p$-norm. Techniques such as LASSO and Ridge are built around $l_1$ and $l_2$ norms. Pruning can be performed element-wise, channel-wise, shape-wise, filter-wise, layer-wise and even network-wise. Each has trade-offs in compression, accuracy, and speedup. 

Quantization reduces computations by reducing the precision of the datatype. Most networks are trained using 32-bit floating point. Weights, biases, and activations may then be quantized typically to 8-bit integers. Lower bit width quantizations have been performed with single bit being termed a binary neural network. It is difficult to (re)train very low bit width neural networks. A single bit is not differentiable thereby prohibiting back propagation. Lower bit widths cause difficulties for computing gradients. The advantage of quantization is significantly improved performance (usually 2-3x) and dramatically reduced storage requirements. In addition to describing how quantization is performed we also included an overview of popular libraries and frameworks that support quantization. We further provided a comparison of accuracy for a number of networks using different frameworks \autoref{tab:quantize-libraries}.

In this paper, we summarized pruning and quantization techniques. Pruning removes redundant computations that have limited contribution to a result. Quantization reduces computations by reducing the precision of the datatype. Both can be used independently or in combination to reduce storage requirements and accelerate inference.

%% file: ilsvrc-acc-all.tex
\begin{center}
    
\scriptsize

\tablefirsthead{
    \toprule
    \multirow{2}{*}{Model} & \multirow{2}{*}{Deployment} & \multicolumn{2}{c}{Bit-width} & \multicolumn{2}{c}{Acc. Drop} & \multirow{2}{*}{Ref.} \\
    \cline{3-6} & & W & A & Top-1 & Top-5 & \\
    \midrule
}

\tablehead{
    \toprule
    Model & Deployment & W & A & Top-1 & Top-5 & Ref.\\
    \midrule
}
\tabletail{\hline}


\topcaption{Quantization Network Performance on ILSVRC2012 for various bit-widths of the weights W and activation A (aka. feature)}
\begin{supertabular}{p{0.15\linewidth}p{0.20\linewidth}p{0.04\linewidth}p{0.04\linewidth}p{0.09\linewidth}p{0.09\linewidth}p{0.09\linewidth}}


AlexNet & QuantNet & 1  & 32  & -1.70\% & -1.50\% & \cite{Yang2019} \\
  & BWNH & 1  & 32  & -1.40\% & -0.70\% & \cite{Hu2018} \\
  & SYQ & 2  & 8  & -1.00\% & -0.60\% & \cite{Faraone2018} \\
  & TSQ & 2  & 2  & -0.90\% & -0.30\% & \cite{Wang2018a} \\
  & INQ & 5  & 32  & -0.87\% & -1.39\% & \cite{Zhou2017a} \\
  & PACT & 4  & 3  & -0.60\% & -1.00\% & \cite{Choi2018} \\
  & QIL & 4  & 4  & -0.20\% & - & \cite{Jung2018} \\
  & Mixed-Precision & 16  & 16  & -0.16\% & - & \cite{Micikevicius2017} \\
  & PACT & 32  & 5  & -0.10\% & -0.20\% & \cite{Choi2018} \\
  & QIL & 5  & 5  & -0.10\% & - & \cite{Jung2018} \\
  & QuantNet & 3(±4) & 32  & -0.10\% & -0.10\% & \cite{Yang2019} \\
  & ELNN & 3(±4) & 32  & 0.00\% & 0.20\% & \cite{Leng2017} \\
  & DoReFa-Net & 32  & 3  & 0.00\% & -0.90\% & \cite{Zhou2016a} \\
  & TensorRT & 8  & 8  & 0.03\% & 0.00\% & \cite{Migacz2017} \\
  & PACT & 2  & 2  & 0.10\% & -0.70\% & \cite{Choi2018} \\
  & PACT & 32  & 2  & 0.20\% & -0.20\% & \cite{Choi2018} \\
  & DoReFa-Net & 32  & 5  & 0.20\% & -0.50\% & \cite{Zhou2016a} \\
  & QuantNet & 3(±2) & 32  & 0.30\% & 0.00\% & \cite{Yang2019} \\
  & DoReFa-Net & 32  & 4  & 0.30\% & -0.50\% & \cite{Zhou2016a} \\
  & WRPN & 2  & 32  & 0.40\% & - & \cite{Mishra2018} \\
  & DFP16 & 16  & 16  & 0.49\% & 0.59\% & \cite{Das2018} \\
  & PACT & 3  & 2  & 0.50\% & -0.10\% & \cite{Choi2018} \\
  & PACT & 4  & 2  & 0.50\% & -0.10\% & \cite{Choi2018} \\
  & SYQ & 1  & 8  & 0.50\% & 0.80\% & \cite{Faraone2018} \\
  & QIL & 3  & 3  & 0.50\% & - & \cite{Jung2018} \\
  & FP8 & 8  & 8  & 0.50\% & - & \cite{Wang2018b} \\
  & BalancedQ & 32  & 2  & 0.60\% & -2.00\% & \cite{Zhou2017c} \\
  & ELNN & 3(±2) & 32  & 0.80\% & 0.60\% & \cite{Leng2017} \\
  & SYQ & 1  & 4  & 0.90\% & 0.80\% & \cite{Faraone2018} \\
  & QuantNet & 2  & 32  & 0.90\% & 0.30\% & \cite{Yang2019} \\
  & FFN & 2  & 32  & 1.00\% & 0.30\% & \cite{Wang2017} \\
  & DoReFa-Net & 32  & 2  & 1.00\% & 0.10\% & \cite{Zhou2016a} \\
  & Unified INT8 & 8  & 8  & 1.00\% & - & \cite{Zhu2019} \\
  & DeepShift-PS & 6  & 32  & 1.19\% & 0.67\% & \cite{Elhoushi2019} \\
  & WEQ & 4  & 4  & 1.20\% & 1.00\% & \cite{Park2017} \\
  & LQ-NETs & 2  & 32  & 1.30\% & 0.80\% & \cite{Zhang2018d} \\
  & SYQ & 2  & 2  & 1.30\% & 1.00\% & \cite{Faraone2018} \\
  & LQ-NETs & 1  & 2  & 1.40\% & 1.40\% & \cite{Zhang2018d} \\
  & BalancedQ & 2  & 2  & 1.40\% & -1.00\% & \cite{Zhou2017c} \\
  & WRPN-2x & 8  & 8  & 1.50\% & - & \cite{Mishra2018} \\
  & DoReFa-Net & 1  & 4  & 1.50\% & - & \cite{Zhou2016a} \\
  & DeepShift-Q & 6  & 32  & 1.55\% & 0.81\% & \cite{Elhoushi2019} \\
  & WRPN-2x & 32  & 8  & 1.60\% & - & \cite{Mishra2018} \\
  & WEQ & 3  & 4  & 1.60\% & 1.10\% & \cite{Park2017} \\
  & WRPN-2x & 8  & 4  & 1.70\% & - & \cite{Mishra2018} \\
  & WRPN-2x & 4  & 8  & 1.70\% & - & \cite{Mishra2018} \\
  & SYQ & 1  & 2  & 1.70\% & 1.60\% & \cite{Faraone2018} \\
  & ELNN & 2  & 32  & 1.80\% & 1.80\% & \cite{Leng2017} \\
  & WRPN-2x & 4  & 4  & 1.90\% & - & \cite{Mishra2018} \\
  & WRPN-2x & 32  & 4  & 1.90\% & - & \cite{Mishra2018} \\
\hline \hline
GoogLeNet & Mixed-Precision & 16  & 16  & -0.10\% & - & \cite{Micikevicius2017} \\
  & DeepShift-PS & 6  & 32  & -0.09\% & -0.09\% & \cite{Elhoushi2019} \\
  & DFP16 & 16  & 16  & -0.08\% & 0.00\% & \cite{Das2018} \\
  & AngleEye & 16  & 16  & 0.05\% & 0.45\% & \cite{Guo2018} \\
  & AngleEye & 16  & 16  & 0.05\% & 0.45\% & \cite{Guo2018} \\
  & ShiftCNN & 3  & 4  & 0.05\% & 0.09\% & \cite{Gudovskiy2017} \\
  & DeepShift-Q & 6  & 32  & 0.27\% & 0.29\% & \cite{Elhoushi2019} \\
  & LogQuant & 32  & 6  & 0.36\% & 0.28\% & \cite{Cai2018} \\
  & ShiftCNN & 2  & 4  & 0.39\% & 0.29\% & \cite{Gudovskiy2017} \\
  & TensorRT & 8  & 8  & 0.45\% & 0.19\% & \cite{Migacz2017} \\
  & LogQuant & 6  & 32  & 0.64\% & 0.67\% & \cite{Cai2018} \\
  & INQ & 5  & 32  & 0.76\% & 0.25\% & \cite{Zhou2017a} \\
  & ELNN & 3(±4) & 32  & 2.40\% & 1.40\% & \cite{Leng2017} \\
  & ELNN & 3(±2) & 32  & 2.80\% & 1.60\% & \cite{Leng2017} \\
  & LogQuant & 6  & 6  & 3.43\% & 0.78\% & \cite{Cai2018} \\
  & QNN & 4  & 4  & 5.10\% & 7.80\% & \cite{Hubara2016a} \\
  & QNN & 6  & 6  & 5.20\% & 8.10\% & \cite{Hubara2016a} \\
  & ELNN & 2  & 32  & 5.60\% & 3.50\% & \cite{Leng2017} \\
  & BWN & 1  & 32  & 5.80\% & 4.80\% & \cite{Rastegari2016} \\
  & AngleEye & 8  & 8  & 6.00\% & 3.20\% & \cite{Guo2018} \\
  & TWN & 2  & 32  & 7.50\% & 4.80\% & \cite{Li2016} \\
  & ELNN & 1  & 32  & 8.40\% & 5.70\% & \cite{Leng2017} \\
  & BWN & 2  & 32  & 9.70\% & 6.50\% & \cite{Rastegari2016} \\
  & ShiftCNN & 1  & 4  & 11.26\% & 7.36\% & \cite{Gudovskiy2017} \\
  & LogQuant & 32  & 3  & 13.50\% & 8.93\% & \cite{Cai2018} \\
  & LogQuant & 3  & 3  & 18.07\% & 12.85\% & \cite{Cai2018} \\
  & LogQuant & 4  & 4  & 18.57\% & 13.21\% & \cite{Cai2018} \\
  & LogQuant & 32  & 4  & 18.57\% & 13.21\% & \cite{Cai2018} \\
  & BNN & 1  & 1  & 24.20\% & 20.90\% & \cite{Courbariaux2016} \\
  & AngleEye & 6  & 6  & 52.10\% & 57.35\% & \cite{Guo2018} \\
\hline \hline
MobileNet & HAQ-Cloud & 6  & 6  & -0.38\% & -0.23\% & \cite{Wang2018} \\
V1  & HAQ-Edge & 6  & 6  & -0.38\% & -0.34\% & \cite{Wang2018} \\
  & MelinusNet59 & 1  & 1  & -0.10\% & - & \cite{Bethge2020} \\
  & HAQ-Edge & 5  & 5  & 0.24\% & 0.08\% & \cite{Wang2018} \\
  & PACT & 6  & 6  & 0.36\% & 0.26\% & \cite{Choi2018} \\
  & PACT & 6  & 6  & 0.36\% & 0.26\% & \cite{Choi2018} \\
  & HAQ-Cloud & 5  & 5  & 0.85\% & 0.48\% & \cite{Wang2018} \\
  & HAQ-Edge & 4  & 4  & 3.42\% & 1.95\% & \cite{Wang2018} \\
  & PACT & 5  & 5  & 3.82\% & 2.20\% & \cite{Choi2018} \\
  & PACT & 5  & 5  & 3.82\% & 2.20\% & \cite{Choi2018} \\
  & HAQ-Cloud & 4  & 4  & 5.49\% & 3.25\% & \cite{Wang2018} \\
  & PACT & 4  & 4  & 8.38\% & 5.66\% & \cite{Choi2018} \\
\hline \hline
MobileNet & HAQ-Edge & 6  & 6  & -0.08\% & -0.11\% & \cite{Wang2018} \\
V2  & HAQ-Cloud & 6  & 6  & -0.04\% & 0.01\% & \cite{Wang2018} \\
  & Unified INT8 & 8  & 8  & 0.00\% & - & \cite{Zhu2019} \\
  & PACT & 6  & 6  & 0.56\% & 0.25\% & \cite{Choi2018} \\
  & HAQ-Edge & 5  & 5  & 0.91\% & 0.34\% & \cite{Wang2018} \\
  & HAQ-Cloud & 5  & 5  & 2.36\% & 1.31\% & \cite{Wang2018} \\
  & PACT & 5  & 5  & 2.97\% & 1.67\% & \cite{Choi2018} \\
  & HAQ-Cloud & 4  & 4  & 4.80\% & 2.79\% & \cite{Wang2018} \\
  & HAQ-Edge & 4  & 4  & 4.82\% & 2.92\% & \cite{Wang2018} \\
  & PACT & 4  & 4  & 10.42\% & 6.53\% & \cite{Choi2018} \\
\hline \hline
ResNet-18 & RangeBN & 8  & 8  & -0.60\% & - & \cite{Banner2018} \\
  & LBM & 8  & 8  & -0.60\% & - & \cite{Zhong2020} \\
  & QuantNet & 5  & 32  & -0.30\% & -0.10\% & \cite{Yang2019} \\
  & QIL & 5  & 5  & -0.20\% & - & \cite{Jung2018} \\
  & QuantNet & 3(±4) & 32  & -0.10\% & -0.10\% & \cite{Yang2019} \\
  & ShiftCNN & 3  & 4  & 0.03\% & 0.12\% & \cite{Gudovskiy2017} \\
  & LQ-NETs & 4  & 32  & 0.20\% & 0.50\% & \cite{Zhang2018d} \\
  & QIL & 3  & 32  & 0.30\% & 0.30\% & \cite{Jung2018} \\
  & LPBN & 32  & 5  & 0.30\% & 0.40\% & \cite{Cai2017} \\
  & QuantNet & 3(±2) & 32  & 0.40\% & 0.20\% & \cite{Yang2019} \\
  & PACT & 32  & 4  & 0.40\% & 0.30\% & \cite{Choi2018} \\
  & SeerNet & 4  & 1  & 0.42\% & 0.18\% & \cite{Cao2019} \\
  & ShiftCNN & 2  & 4  & 0.54\% & 0.34\% & \cite{Gudovskiy2017} \\
  & PACT & 5  & 5  & 0.60\% & 0.30\% & \cite{Choi2018} \\
  & INQ & 4  & 32  & 0.62\% & 0.10\% & \cite{Zhou2017a} \\
  & Unified INT8 & 8  & 8  & 0.63\% & - & \cite{Zhu2019} \\
  & QIL & 5  & 5  & 0.80\% & - & \cite{Jung2018} \\
  & LQ-NETs & 3(±4) & 32  & 0.90\% & 0.80\% & \cite{Zhang2018d} \\
  & QIL & 3  & 3  & 1.00\% & - & \cite{Jung2018} \\
  & DeepShift-Q & 6  & 32  & 1.09\% & 0.47\% & \cite{Elhoushi2019} \\
  & ELNN & 3(±2) & 32  & 1.10\% & 0.70\% & \cite{Leng2017} \\
  & PACT & 32  & 3  & 1.20\% & 0.70\% & \cite{Choi2018} \\
  & PACT & 4  & 4  & 1.20\% & 0.60\% & \cite{Choi2018} \\
  & QuantNet & 2  & 32  & 1.20\% & 0.60\% & \cite{Yang2019} \\
  & ELNN & 3(±4) & 32  & 1.30\% & 0.60\% & \cite{Leng2017} \\
  & DeepShift-PS & 6  & 32  & 1.44\% & 0.67\% & \cite{Elhoushi2019} \\
  & ABC-Net & 5  & 32  & 1.46\% & 1.18\% & \cite{Lin2017a} \\
  & ELNN & 3(±2) & 32  & 1.60\% & 1.10\% & \cite{Leng2017} \\
  & DoReFa-Net & 32  & 5  & 1.70\% & 1.00\% & \cite{Zhou2016a} \\
  & SYQ & 2  & 8  & 1.90\% & 1.40\% & \cite{Faraone2018} \\
  & DoReFa-Net & 32  & 4  & 1.90\% & 1.10\% & \cite{Zhou2016a} \\
  & LQ-NETs & 3  & 3  & 2.00\% & 1.60\% & \cite{Zhang2018d} \\
  & DoReFa-Net & 5  & 5  & 2.00\% & 1.30\% & \cite{Zhou2016a} \\
  & ELNN & 2  & 32  & 2.10\% & 1.50\% & \cite{Leng2017} \\
  & QIL & 2  & 32  & 2.10\% & 1.30\% & \cite{Jung2018} \\
  & DoReFa-Net & 32  & 3  & 2.10\% & 1.40\% & \cite{Zhou2016a} \\
  & QIL & 4  & 4  & 2.20\% & - & \cite{Jung2018} \\
  & LQ-NETs & 2  & 32  & 2.20\% & 1.60\% & \cite{Zhang2018d} \\
  & GroupNet-8 & 1  & 1  & 2.20\% & 1.40\% & \cite{Zhuang2019} \\
  & PACT & 3  & 3  & 2.30\% & 1.40\% & \cite{Choi2018} \\
  & DoReFa-Net & 4  & 4  & 2.30\% & 1.50\% & \cite{Zhou2016a} \\
  & TTN & 2  & 32  & 2.50\% & 1.80\% & \cite{Zhu2016} \\
  & TTQ & 2  & 32  & 2.70\% & 2.00\% & \cite{Zoph2017} \\
  & AddNN & 32  & 32  & 2.80\% & 1.50\% & \cite{Chen2019b} \\
  & ELNN & 2  & 32  & 2.80\% & 1.50\% & \cite{Leng2017} \\
  & LPBN & 32  & 4  & 2.90\% & 1.70\% & \cite{Cai2017} \\
  & PACT & 32  & 2  & 2.90\% & 2.00\% & \cite{Choi2018} \\
  & DoReFa-Net & 3  & 3  & 2.90\% & 2.00\% & \cite{Zhou2016a} \\
  & QuantNet & 1  & 32  & 3.10\% & 1.90\% & \cite{Yang2019} \\
  & INQ & 2  & 32  & 3.10\% & 1.90\% & \cite{Zhou2017a} \\
\hline \hline
ResNet-34 & WRPN-2x & 4  & 4  & -0.93\% & - & \cite{Mishra2018} \\
  & WRPN-2x & 4  & 8  & -0.89\% & - & \cite{Mishra2018} \\
  & QIL & 4  & 4  & 0.00\% & - & \cite{Jung2018} \\
  & QIL & 5  & 5  & 0.00\% & - & \cite{Jung2018} \\
  & WRPN-2x & 4  & 2  & 0.01\% & - & \cite{Mishra2018} \\
  & WRPN-2x & 2  & 4  & 0.09\% & - & \cite{Mishra2018} \\
  & WRPN-2x & 2  & 2  & 0.27\% & - & \cite{Mishra2018} \\
  & SeerNet & 4  & 1  & 0.35\% & 0.17\% & \cite{Cao2019} \\
  & Unified INT8 & 8  & 8  & 0.39\% & - & \cite{Zhu2019} \\
  & LCCL &  &  & 0.43\% & 0.17\% & \cite{Dong2017a} \\
  & QIL & 3  & 3  & 0.60\% & - & \cite{Jung2018} \\
  & WRPN-3x & 1  & 1  & 0.90\% & - & \cite{Mishra2018} \\
  & WRPN-3x & 1  & 1  & 1.21\% & - & \cite{Mishra2018} \\
  & GroupNet-8  & 1  & 1  & 1.40\% & 1.00\% & \cite{Zhuang2019} \\
  & dLAC & 2  & 16  & 1.67\% & 0.89\% & \cite{Venkatesh2017b} \\
  & LQ-NETs & 3  & 3  & 1.90\% & 1.20\% & \cite{Zhang2018d} \\
  & GroupNet**-5 & 1  & 1  & 2.70\% & 2.10\% & \cite{Zhuang2019} \\
  & IR-Net & 1  & 32  & 2.90\% & 1.80\% & \cite{Qin2020a} \\
  & QIL & 2  & 2  & 3.10\% & - & \cite{Jung2018} \\
  & WRPN-2x & 1  & 1  & 3.40\% & - & \cite{Mishra2018} \\
  & WRPN-2x & 1  & 1  & 3.74\% & - & \cite{Mishra2018} \\
  & LQ-NETs & 2  & 2  & 4.00\% & 2.30\% & \cite{Zhang2018d} \\
  & GroupNet-5 & 1  & 1  & 4.70\% & 3.40\% & \cite{Zhuang2019} \\
  & ABC-Net & 5  & 5  & 4.90\% & 3.10\% & \cite{Lin2017a} \\
  & HWGQ & 1  & 32  & 5.10\% & 3.40\% & \cite{Cai2017} \\
  & WAGEUBN & 8  & 8  & 5.18\% & - & \cite{Yang2020} \\
  & ABC-Net & 3  & 3  & 6.60\% & 3.90\% & \cite{Lin2017a} \\
  & LQ-NETs & 1  & 2  & 6.70\% & 4.40\% & \cite{Zhang2018d} \\
  & LQ-NETs & 4  & 4  & 6.70\% & 4.40\% & \cite{Zhang2018d} \\
  & BCGD & 1  & 4  & 7.60\% & 4.70\% & \cite{Yin2019} \\
  & HWGQ & 1  & 2  & 9.00\% & 5.60\% & \cite{Cai2017} \\
  & IR-Net & 1  & 1  & 9.50\% & 6.20\% & \cite{Qin2020a} \\
  & CI-BCNN (add) & 1  & 1  & 11.07\% & 6.39\% & \cite{Wang2019} \\
  & Bi-Real & 1  & 1  & 11.10\% & 7.40\% & \cite{Xu2018b} \\
  & WRPN-1x & 1  & 1  & 12.80\% & - & \cite{Mishra2018} \\
  & WRPN & 1  & 1  & 13.05\% & - & \cite{Mishra2018} \\
  & CI-BCNN & 1  & 1  & 13.59\% & 8.65\% & \cite{Wang2019} \\
  & DoReFa-Net & 1  & 4  & 14.60\% & - & \cite{Zhou2016a} \\
  & DoReFa-Net & 1  & 2  & 20.40\% & - & \cite{Zhou2016a} \\
  & ABC-Net & 1  & 1  & 20.90\% & 14.80\% & \cite{Lin2017a} \\
  & BNN & 1  & 1  & 29.10\% & 24.20\% & \cite{Zhou2016a} \\
\hline \hline
ResNet-50 & Mixed-Precision & 16  & 16  & -0.12\% & - & \cite{Micikevicius2017} \\
  & DFP16 & 16  & 16  & -0.07\% & -0.06\% & \cite{Das2018} \\
  & QuantNet & 5  & 32  & 0.00\% & 0.00\% & \cite{Yang2019} \\
  & LQ-NETs & 4  & 32  & 0.00\% & 0.10\% & \cite{Zhang2018d} \\
  & FGQ & 32  & 32  & 0.00\% & - & \cite{Mellempudi2017} \\
  & TensorRT & 8  & 8  & 0.13\% & 0.12\% & \cite{Migacz2017} \\
  & PACT & 5  & 5  & 0.20\% & -0.20\% & \cite{Choi2018} \\
  & QuantNet & 3(±4) & 32  & 0.20\% & 0.00\% & \cite{Yang2019} \\
  & Unified INT8 & 8  & 8  & 0.26\% & - & \cite{Zhu2019} \\
  & ShiftCNN & 3  & 4  & 0.29\% & 0.15\% & \cite{Gudovskiy2017} \\
  & ShiftCNN & 3  & 4  & 0.31\% & 0.16\% & \cite{Gudovskiy2017} \\
  & PACT & 4  & 4  & 0.40\% & -0.10\% & \cite{Choi2018} \\
  & LPBN & 32  & 5  & 0.40\% & 0.40\% & \cite{Graham2017} \\
  & ShiftCNN & 2  & 4  & 0.67\% & 0.41\% & \cite{Gudovskiy2017} \\
  & DeepShift-Q & 6  & 32  & 0.81\% & 0.21\% & \cite{Elhoushi2019} \\
  & DeepShift-PS & 6  & 32  & 0.84\% & 0.31\% & \cite{Elhoushi2019} \\
  & PACT & 5  & 32  & 0.90\% & 0.20\% & \cite{Choi2018} \\
  & QuantNet & 3(±2) & 32  & 0.90\% & 0.40\% & \cite{Yang2019} \\
  & PACT & 4  & 32  & 1.00\% & 0.20\% & \cite{Choi2018} \\
  & dLAC & 2  & 16  & 1.20\% & - & \cite{Venkatesh2017b} \\
  & QuantNet & 2  & 32  & 1.20\% & 0.60\% & \cite{Yang2019} \\
  & AddNN & 32  & 32  & 1.30\% & 1.20\% & \cite{Chen2019b} \\
  & LQ-NETs & 4  & 4  & 1.30\% & 0.80\% & \cite{Zhang2018d} \\
  & LQ-NETs & 2  & 32  & 1.30\% & 0.90\% & \cite{Zhang2018d} \\
  & INQ & 5  & 32  & 1.32\% & 0.41\% & \cite{Zhou2017a} \\
  & PACT & 3  & 32  & 1.40\% & 0.50\% & \cite{Choi2018} \\
  & IAO & 8  & 8  & 1.50\% & - & \cite{Jacob2017} \\
  & PACT & 3  & 3  & 1.60\% & 0.50\% & \cite{Choi2018} \\
  & HAQ & 2MP & 4MP & 1.91\% & - & \cite{Wang2018} \\
  & HAQ & MP & MP & 2.09\% & - & \cite{Wang2018} \\
  & LQ-NETs & 3  & 3  & 2.20\% & 1.60\% & \cite{Zhang2018d} \\
  & LPBN & 32  & 4  & 2.20\% & 1.20\% & \cite{Graham2017} \\
  & Deep Comp. & 3  & MP & 2.29\% & - & \cite{Han2015} \\
  & PACT & 4  & 2  & 2.40\% & 1.20\% & \cite{Choi2018} \\
  & ShiftCNN & 2  & 4  & 2.49\% & 1.64\% & \cite{Gudovskiy2017} \\
  & FFN & 2  & 32  & 2.50\% & 1.30\% & \cite{Wang2017} \\
  & UNIQ & 4  & 8  & 2.60\% & - & \cite{Baskin2018} \\
  & QuantNet & 1  & 32  & 3.20\% & 1.70\% & \cite{Yang2019} \\
  & SYQ & 2  & 8  & 3.70\% & 2.10\% & \cite{Faraone2018} \\
  & FGQ-TWN & 2  & 8  & 4.29\% & - & \cite{Mellempudi2017} \\
  & PACT & 2  & 2  & 4.70\% & 2.60\% & \cite{Choi2018} \\
  & LQ-NETs & 2  & 2  & 4.90\% & 2.90\% & \cite{Zhang2018d} \\
  & SYQ & 1  & 8  & 5.40\% & 3.40\% & \cite{Faraone2018} \\
  & DoReFa-Net & 4  & 4  & 5.50\% & 3.30\% & \cite{Zhou2016a} \\
  & DoReFa-Net & 5  & 5  & 5.50\% & -0.20\% & \cite{Zhou2016a} \\
  & FGQ & 2  & 8  & 5.60\% & - & \cite{Mellempudi2017} \\
  & ABC-Net & 5  & 5  & 6.30\% & 3.50\% & \cite{Lin2017a} \\
  & FGQ-TWN & 2  & 4  & 6.67\% & - & \cite{Mellempudi2017} \\
  & HWGQ & 1  & 2  & 6.90\% & 4.60\% & \cite{Cai2017} \\
\hline \hline
ResNet-100 & IAO & 8  & 8  & 1.40\% & - & \cite{Jacob2017} \\
ResNet-101 & TensorRT & 8  & 8  & -0.01\% & 0.05\% & \cite{Migacz2017} \\
  & FGQ-TWN & 2  & 8  & 3.65\% & - & \cite{Mellempudi2017} \\
  & FGQ-TWN & 2  & 4  & 6.81\% & - & \cite{Mellempudi2017} \\
ResNet-150 & IAO & 8  & 8  & 2.10\% & - & \cite{Jacob2017} \\
ResNet-152 & TensorRT & 8  & 8  & 0.08\% & 0.04\% & \cite{Migacz2017} \\
  & dLAC & 2  & 16  & 1.20\% & 0.64\% & \cite{Venkatesh2017b} \\
\hline \hline
SqueezeNet & AngleEye & 16  & 16  & 0.00\% & 0.01\% & \cite{Guo2018} \\
  & ShiftCNN & 3  & 4  & 0.01\% & 0.01\% & \cite{Gudovskiy2017} \\
  & ShiftCNN & 2  & 4  & 1.01\% & 0.71\% & \cite{Gudovskiy2017} \\
  & AngleEye & 8  & 8  & 1.42\% & 1.05\% & \cite{Guo2018} \\
  & AngleEye & 6  & 6  & 28.13\% & 27.43\% & \cite{Guo2018} \\
  & ShiftCNN & 1  & 4  & 35.39\% & 35.09\% & \cite{Gudovskiy2017} \\
\hline \hline
VGG-16 & ELNN & 3(±4) & 32  & -1.10\% & -1.00\% & \cite{Leng2017} \\
 & ELNN & 3(±2) & 32  & -0.60\% & -0.80\% & \cite{Leng2017} \\
 & AngleEye & 16  & 16  & 0.09\% & -0.05\% & \cite{Guo2018} \\
 & DFP16 & 16  & 16  & 0.11\% & 0.29\% & \cite{Das2018} \\
 & AngleEye & 8  & 8  & 0.21\% & 0.08\% & \cite{Guo2018} \\
 & SeerNet & 4  & 1  & 0.28\% & 0.10\% & \cite{Cao2019} \\
 & DeepShift-Q & 6  & 32  & 0.29\% & 0.11\% & \cite{Elhoushi2019} \\
 & FFN & 2  & 32  & 0.30\% & -0.20\% & \cite{Wang2017} \\
 & DeepShift-PS & 6  & 32  & 0.47\% & 0.30\% & \cite{Elhoushi2019} \\
 & DeepShift-Q & 6  & 32  & 0.72\% & 0.29\% & \cite{Elhoushi2019} \\
 & INQ & 5  & 32  & 0.77\% & 0.08\% & \cite{Elhoushi2019} \\
 & TWN & 2  & 32  & 1.10\% & 0.30\% & \cite{Li2016} \\
 & ELNN & 2  & 32  & 2.00\% & 0.90\% & \cite{Leng2017} \\
 & TSQ & 2  & 2  & 2.00\% & 0.70\% & \cite{Wang2018a} \\
 & AngleEye & 16  & 16  & 2.15\% & 1.49\% & \cite{Guo2018} \\
 & BWN & 2  & 32  & 2.20\% & 1.20\% & \cite{Rastegari2016} \\
 & AngleEye & 8  & 8  & 2.35\% & 1.76\% & \cite{Guo2018} \\
 & ELNN & 1  & 32  & 3.30\% & 1.80\% & \cite{Leng2017} \\
 & AngleEye & 6  & 6  & 9.07\% & 6.58\% & \cite{Guo2018} \\
 & AngleEye & 6  & 6  & 22.38\% & 17.75\% & \cite{Guo2018} \\
 & LogQuant & 3  & 3  & - & 0.99\% & \cite{Cai2018} \\
 & LogQuant & 4  & 4  & - & 0.51\% & \cite{Cai2018} \\
 & LogQuant & 6  & 6  & - & 0.83\% & \cite{Cai2018} \\
 & LogQuant & 32  & 3  & - & 0.82\% & \cite{Cai2018} \\
 & LogQuant & 32  & 4  & - & 0.36\% & \cite{Cai2018} \\
 & LogQuant & 32  & 6  & - & 0.31\% & \cite{Cai2018} \\
 & LogQuant & 6  & 32  & - & 0.76\% & \cite{Cai2018} \\
 & LDR & 5  & 4  & - & 0.90\% & \cite{Miyashita2016} \\
 & LogNN & 5  & 4  & - & 1.38\% & \cite{Miyashita2016} \\

\end{supertabular}
\end{center}

%% file: 5-author-biography.tex
\bio{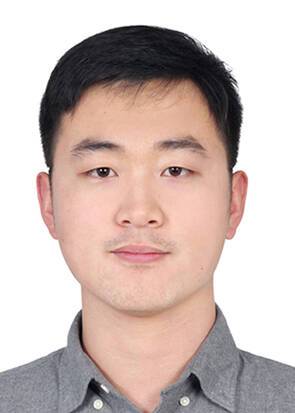}
\textbf{Tailin Liang} received the B.E. degree in Computer Science and B.B.A from the University of Science and Technology Beijing in 2017. He is currently working toward a Ph.D. degree in Computer Science at the School of Computer and Communication Engineering, University of Science and Technology Beijing. His current research interests include deep learning domain-specific processors and co-designed optimization algorithms.
\endbio

\bio{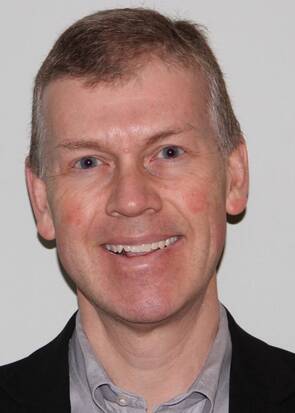}
\textbf{John Glossner} received the Ph.D. degree in Electrical Engineering from TU Delft in 2001. He is the Director of the Computer Architecture, Heterogeneous Computing, and AI Lab at the University of Science and Technology Beijing. He is also the CEO of Optimum Semiconductor Technologies and President of both the Heterogeneous System Architecture Foundation and Wireless Innovation Forum. John's research interests include the design of heterogeneous computing systems, computer architecture, embedded systems, digital signal processors, software defined radios, artificial intelligence algorithms, and machine learning systems.
\endbio

\bio{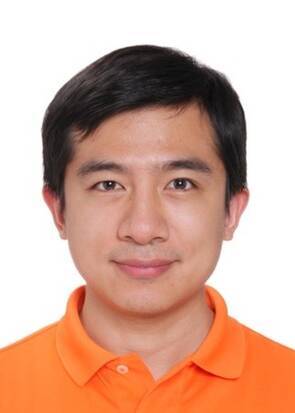}
\textbf{Lei Wang} received the B.E. and Ph.D. degrees in 2006 and 2012 from the University of Science and Technology Beijing. He then served as an assistant researcher at the Institute of Automation of the Chinese Academy of Sciences during 2012-2015. He was a joint Ph.D. of Electronic Engineering at The University of Texas at Dallas during 2009-2011. Currently, he is an adjunct professor at the School of Computer and Communication Engineering, University of Science and Technology Beijing.
\endbio

\bio{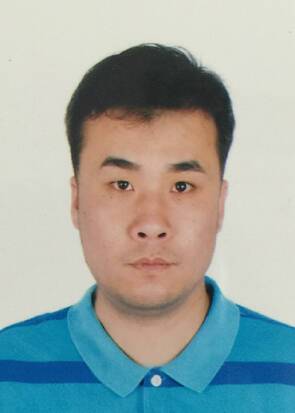}
\textbf{Shaobo Shi} received the B.E. and Ph.D. degrees in 2008 and 2014 from the University of Science and Technology Beijing. He then served as an assistant researcher at the Institute of Automation of the Chinese Academy of Sciences during 2014-2017. Currently, he is a deep learning domain-specific processor engineer at Huaxia General Processor Technology. As well serve as an adjunct professor at the School of Computer and Communication Engineering, University of Science and Technology Beijing.
\endbio

\bio{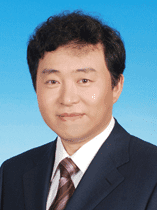}
\textbf{Xiaotong Zhang} received the M.E. and Ph.D. degrees from the University of Science and Technology Beijing in 1997 and 2000, respectively, where he was a professor of Computer Science and Technology. His research interest includes the quality of wireless channels and networks, wireless sensor networks, networks management, cross-layer design and resource allocation of broadband and wireless networks, and the signal processing of communication and computer architecture.
\endbio

%% file: main.bbl
\begin{thebibliography}{277}
\expandafter\ifx\csname natexlab\endcsname\relax\def\natexlab#1{#1}\fi
\providecommand{\url}[1]{\texttt{#1}}
\providecommand{\href}[2]{#2}
\providecommand{\path}[1]{#1}
\providecommand{\DOIprefix}{doi:}
\providecommand{\ArXivprefix}{arXiv:}
\providecommand{\URLprefix}{URL: }
\providecommand{\Pubmedprefix}{pmid:}
\providecommand{\doi}[1]{\href{http://dx.doi.org/#1}{\path{#1}}}
\providecommand{\Pubmed}[1]{\href{pmid:#1}{\path{#1}}}
\providecommand{\bibinfo}[2]{#2}
\ifx\xfnm\relax \def\xfnm[#1]{\unskip,\space#1}\fi
\bibitem[{Abadi et~al.(2016)Abadi, Agarwal, Barham, Brevdo, Chen, Citro,
  Corrado, Davis, Dean, Devin, Ghemawat, Goodfellow, Harp, Irving, Isard, Jia,
  Jozefowicz, Kaiser, Kudlur, Levenberg, Mane, Monga, Moore, Murray, Olah,
  Schuster, Shlens, Steiner, Sutskever, Talwar, Tucker, Vanhoucke, Vasudevan,
  Viegas, Vinyals, Warden, Wattenberg, Wicke, Yu and Zheng}]{Abadi2016}
\bibinfo{author}{Abadi, M.}, \bibinfo{author}{Agarwal, A.},
  \bibinfo{author}{Barham, P.}, \bibinfo{author}{Brevdo, E.},
  \bibinfo{author}{Chen, Z.}, \bibinfo{author}{Citro, C.},
  \bibinfo{author}{Corrado, G.S.}, \bibinfo{author}{Davis, A.},
  \bibinfo{author}{Dean, J.}, \bibinfo{author}{Devin, M.},
  \bibinfo{author}{Ghemawat, S.}, \bibinfo{author}{Goodfellow, I.},
  \bibinfo{author}{Harp, A.}, \bibinfo{author}{Irving, G.},
  \bibinfo{author}{Isard, M.}, \bibinfo{author}{Jia, Y.},
  \bibinfo{author}{Jozefowicz, R.}, \bibinfo{author}{Kaiser, L.},
  \bibinfo{author}{Kudlur, M.}, \bibinfo{author}{Levenberg, J.},
  \bibinfo{author}{Mane, D.}, \bibinfo{author}{Monga, R.},
  \bibinfo{author}{Moore, S.}, \bibinfo{author}{Murray, D.},
  \bibinfo{author}{Olah, C.}, \bibinfo{author}{Schuster, M.},
  \bibinfo{author}{Shlens, J.}, \bibinfo{author}{Steiner, B.},
  \bibinfo{author}{Sutskever, I.}, \bibinfo{author}{Talwar, K.},
  \bibinfo{author}{Tucker, P.}, \bibinfo{author}{Vanhoucke, V.},
  \bibinfo{author}{Vasudevan, V.}, \bibinfo{author}{Viegas, F.},
  \bibinfo{author}{Vinyals, O.}, \bibinfo{author}{Warden, P.},
  \bibinfo{author}{Wattenberg, M.}, \bibinfo{author}{Wicke, M.},
  \bibinfo{author}{Yu, Y.}, \bibinfo{author}{Zheng, X.}, \bibinfo{year}{2016}.
\newblock \bibinfo{title}{{TensorFlow: Large-Scale Machine Learning on
  Heterogeneous Distributed Systems}}.
\newblock \bibinfo{journal}{arXiv preprint arXiv: 1603.04467} \URLprefix
  \url{https://arxiv.org/abs/1603.04467}.
\bibitem[{Abdel-Hamid et~al.(2014)Abdel-Hamid, Mohamed, Jiang, Deng, Penn and
  Yu}]{Abdel-hamid2014}
\bibinfo{author}{Abdel-Hamid, O.}, \bibinfo{author}{Mohamed, A.r.},
  \bibinfo{author}{Jiang, H.}, \bibinfo{author}{Deng, L.},
  \bibinfo{author}{Penn, G.}, \bibinfo{author}{Yu, D.}, \bibinfo{year}{2014}.
\newblock \bibinfo{title}{{Convolutional Neural Networks for Speech
  Recognition}}.
\newblock \bibinfo{journal}{IEEE/ACM Transactions on Audio, Speech, and
  Language Processing} \bibinfo{volume}{22}, \bibinfo{pages}{1533--1545}.
\newblock \URLprefix \url{http://ieeexplore.ieee.org/document/6857341/},
  \DOIprefix\doi{10.1109/TASLP.2014.2339736}.
\bibitem[{Abdelouahab et~al.(2018)Abdelouahab, Pelcat, Serot and
  Berry}]{Abdelouahab2018}
\bibinfo{author}{Abdelouahab, K.}, \bibinfo{author}{Pelcat, M.},
  \bibinfo{author}{Serot, J.}, \bibinfo{author}{Berry, F.},
  \bibinfo{year}{2018}.
\newblock \bibinfo{title}{{Accelerating CNN inference on FPGAs: A Survey}}.
\newblock \bibinfo{journal}{ArXiv preprint} \URLprefix
  \url{http://arxiv.org/abs/1806.01683}.
\bibitem[{{Achronix Semiconductor Corporation}(2020)}]{Achronix2020}
\bibinfo{author}{{Achronix Semiconductor Corporation}}, \bibinfo{year}{2020}.
\newblock \bibinfo{title}{{FPGAs Enable the Next Generation of Communication
  and Networking Solutions}}.
\newblock \bibinfo{journal}{White Paper} \bibinfo{volume}{WP021},
  \bibinfo{pages}{1--15}.
\bibitem[{{Albanie}(2020)}]{Albanie2020}
\bibinfo{author}{{Albanie}}, \bibinfo{year}{2020}.
\newblock \bibinfo{title}{{convnet-burden}}.
\newblock \URLprefix \url{https://github.com/albanie/convnet-burden}.
\bibitem[{Alemdar et~al.(2017)Alemdar, Leroy, Prost-Boucle and
  Petrot}]{Alemdar2017}
\bibinfo{author}{Alemdar, H.}, \bibinfo{author}{Leroy, V.},
  \bibinfo{author}{Prost-Boucle, A.}, \bibinfo{author}{Petrot, F.},
  \bibinfo{year}{2017}.
\newblock \bibinfo{title}{{Ternary neural networks for resource-efficient AI
  applications}}, in: \bibinfo{booktitle}{2017 International Joint Conference
  on Neural Networks (IJCNN)}, \bibinfo{publisher}{IEEE}. pp.
  \bibinfo{pages}{2547--2554}.
\newblock \URLprefix
  \url{https://ieeexplore.ieee.org/abstract/document/7966166/},
  \DOIprefix\doi{10.1109/IJCNN.2017.7966166}.
\bibitem[{{AMD}()}]{AMD}
\bibinfo{author}{{AMD}}, .
\newblock \bibinfo{title}{{Radeon Instinct™ MI25 Accelerator}}.
\newblock \URLprefix
  \url{https://www.amd.com/en/products/professional-graphics/instinct-mi25}.
\bibitem[{{Arm}(2015)}]{Arm2015}
\bibinfo{author}{{Arm}}, \bibinfo{year}{2015}.
\newblock \bibinfo{title}{{ARM Architecture Reference Manual ARMv8, for ARMv8-A
  architecture profile}}.
\newblock
  \bibinfo{howpublished}{https://developer.arm.com/documentation/ddi0487/latest}.
\newblock \URLprefix
  \url{https://developer.arm.com/documentation/ddi0487/latest}.
\bibitem[{{Arm}(2020)}]{Arm2020a}
\bibinfo{author}{{Arm}}, \bibinfo{year}{2020}.
\newblock \bibinfo{title}{{Arm Cortex-M Processor Comparison Table}}.
\newblock \URLprefix
  \url{https://developer.arm.com/ip-products/processors/cortex-a}.
\bibitem[{{Arm} and Graphics(2020)}]{Arm2020}
\bibinfo{author}{{Arm}}, \bibinfo{author}{Graphics, C.}, \bibinfo{year}{2020}.
\newblock \bibinfo{title}{{MALI-G76 High-Performance GPU for Complex Graphics
  Features and Bene ts High Performance for Mixed Realities}}.
\newblock \URLprefix
  \url{https://www.arm.com/products/silicon-ip-multimedia/gpu/mali-g76}.
\bibitem[{{ARM} and Reddy(2008)}]{ARM2008}
\bibinfo{author}{{ARM}}, \bibinfo{author}{Reddy, V.G.}, \bibinfo{year}{2008}.
\newblock \bibinfo{title}{{Neon technology introduction}}.
\newblock \bibinfo{journal}{ARM Corporation} , \bibinfo{pages}{1--34}\URLprefix
  \url{http://caxapa.ru/thumbs/301908/AT_-_NEON_for_Multimedia_Applications.pdf}.
\bibitem[{Augasta and Kathirvalavakumar(2013)}]{Augasta2013}
\bibinfo{author}{Augasta, M.G.}, \bibinfo{author}{Kathirvalavakumar, T.},
  \bibinfo{year}{2013}.
\newblock \bibinfo{title}{{Pruning algorithms of neural networks - A
  comparative study}}.
\newblock \bibinfo{journal}{Open Computer Science} \bibinfo{volume}{3},
  \bibinfo{pages}{105--115}.
\newblock \DOIprefix\doi{10.2478/s13537-013-0109-x}.
\bibitem[{{Baidu}(2019)}]{Baidu2019}
\bibinfo{author}{{Baidu}}, \bibinfo{year}{2019}.
\newblock \bibinfo{title}{{PArallel Distributed Deep LEarning: Machine Learning
  Framework from Industrial Practice}}.
\newblock \URLprefix \url{https://github.com/PaddlePaddle/Paddle}.
\bibitem[{Balzer et~al.(1991)Balzer, Takahashi, Ohta and
  Kyuma}]{balzer1991weight}
\bibinfo{author}{Balzer, W.}, \bibinfo{author}{Takahashi, M.},
  \bibinfo{author}{Ohta, J.}, \bibinfo{author}{Kyuma, K.},
  \bibinfo{year}{1991}.
\newblock \bibinfo{title}{{Weight quantization in Boltzmann machines}}.
\newblock \bibinfo{journal}{Neural Networks} \bibinfo{volume}{4},
  \bibinfo{pages}{405--409}.
\newblock \DOIprefix\doi{10.1016/0893-6080(91)90077-I}.
\bibitem[{Banner et~al.(2018)Banner, Hubara, Hoffer and Soudry}]{Banner2018}
\bibinfo{author}{Banner, R.}, \bibinfo{author}{Hubara, I.},
  \bibinfo{author}{Hoffer, E.}, \bibinfo{author}{Soudry, D.},
  \bibinfo{year}{2018}.
\newblock \bibinfo{title}{{Scalable methods for 8-bit training of neural
  networks}}, in: \bibinfo{booktitle}{Advances in Neural Information Processing
  Systems (NIPS)}, pp. \bibinfo{pages}{5145--5153}.
\newblock \URLprefix
  \url{http://papers.nips.cc/paper/7761-scalable-methods-for-8-bit-training-of-neural-networks}.
\bibitem[{Banner et~al.(2019)Banner, Nahshan and Soudry}]{Banner2019}
\bibinfo{author}{Banner, R.}, \bibinfo{author}{Nahshan, Y.},
  \bibinfo{author}{Soudry, D.}, \bibinfo{year}{2019}.
\newblock \bibinfo{title}{{Post training 4-bit quantization of convolutional
  networks for rapid-deployment}}, in: \bibinfo{booktitle}{Advances in Neural
  Information Processing Systems (NIPS)}, pp. \bibinfo{pages}{7950--7958}.
\bibitem[{{Baoyuan Liu} et~al.(2015){Baoyuan Liu}, {Min Wang}, Foroosh, Tappen
  and Penksy}]{Foroosh2015}
\bibinfo{author}{{Baoyuan Liu}}, \bibinfo{author}{{Min Wang}},
  \bibinfo{author}{Foroosh, H.}, \bibinfo{author}{Tappen, M.},
  \bibinfo{author}{Penksy, M.}, \bibinfo{year}{2015}.
\newblock \bibinfo{title}{{Sparse Convolutional Neural Networks}}, in:
  \bibinfo{booktitle}{2015 IEEE Conference on Computer Vision and Pattern
  Recognition (CVPR)}, \bibinfo{publisher}{IEEE}. pp.
  \bibinfo{pages}{806--814}.
\newblock \URLprefix \url{http://ieeexplore.ieee.org/document/7298681/},
  \DOIprefix\doi{10.1109/CVPR.2015.7298681}.
\bibitem[{Baskin et~al.(2018)Baskin, Schwartz, Zheltonozhskii, Liss, Giryes,
  Bronstein and Mendelson}]{Baskin2018}
\bibinfo{author}{Baskin, C.}, \bibinfo{author}{Schwartz, E.},
  \bibinfo{author}{Zheltonozhskii, E.}, \bibinfo{author}{Liss, N.},
  \bibinfo{author}{Giryes, R.}, \bibinfo{author}{Bronstein, A.M.},
  \bibinfo{author}{Mendelson, A.}, \bibinfo{year}{2018}.
\newblock \bibinfo{title}{{UNIQ: Uniform Noise Injection for Non-Uniform
  Quantization of Neural Networks}}.
\newblock \bibinfo{journal}{arXiv preprint arXiv:1804.10969} \URLprefix
  \url{http://arxiv.org/abs/1804.10969}.
\bibitem[{Bengio et~al.(2015)Bengio, Bacon, Pineau and Precup}]{Bengio2015}
\bibinfo{author}{Bengio, E.}, \bibinfo{author}{Bacon, P.L.},
  \bibinfo{author}{Pineau, J.}, \bibinfo{author}{Precup, D.},
  \bibinfo{year}{2015}.
\newblock \bibinfo{title}{{Conditional Computation in Neural Networks for
  faster models}}.
\newblock \bibinfo{journal}{ArXiv preprint} \URLprefix
  \url{http://arxiv.org/abs/1511.06297}.
\bibitem[{Bengio(2013)}]{Bengio2013}
\bibinfo{author}{Bengio, Y.}, \bibinfo{year}{2013}.
\newblock \bibinfo{title}{{Estimating or Propagating Gradients Through
  Stochastic Neurons}}.
\newblock \bibinfo{journal}{ArXiv preprint} \URLprefix
  \url{http://arxiv.org/abs/1305.2982}.
\bibitem[{Bethge et~al.(2020)Bethge, Bartz, Yang, Chen and Meinel}]{Bethge2020}
\bibinfo{author}{Bethge, J.}, \bibinfo{author}{Bartz, C.},
  \bibinfo{author}{Yang, H.}, \bibinfo{author}{Chen, Y.},
  \bibinfo{author}{Meinel, C.}, \bibinfo{year}{2020}.
\newblock \bibinfo{title}{{MeliusNet: Can Binary Neural Networks Achieve
  MobileNet-level Accuracy?}}
\newblock \bibinfo{journal}{ArXiv preprint} \URLprefix
  \url{http://arxiv.org/abs/2001.05936}.
\bibitem[{Bethge et~al.(2019)Bethge, Yang, Bornstein and Meinel}]{Bethge2019}
\bibinfo{author}{Bethge, J.}, \bibinfo{author}{Yang, H.},
  \bibinfo{author}{Bornstein, M.}, \bibinfo{author}{Meinel, C.},
  \bibinfo{year}{2019}.
\newblock \bibinfo{title}{{BinaryDenseNet: Developing an architecture for
  binary neural networks}}.
\newblock \bibinfo{journal}{Proceedings - 2019 International Conference on
  Computer Vision Workshop, ICCVW 2019} ,
  \bibinfo{pages}{1951--1960}\DOIprefix\doi{10.1109/ICCVW.2019.00244}.
\bibitem[{Bianco et~al.(2018)Bianco, Cadene, Celona and
  Napoletano}]{Bianco2018}
\bibinfo{author}{Bianco, S.}, \bibinfo{author}{Cadene, R.},
  \bibinfo{author}{Celona, L.}, \bibinfo{author}{Napoletano, P.},
  \bibinfo{year}{2018}.
\newblock \bibinfo{title}{{Benchmark analysis of representative deep neural
  network architectures}}.
\newblock \bibinfo{journal}{IEEE Access} \bibinfo{volume}{6},
  \bibinfo{pages}{64270--64277}.
\newblock \DOIprefix\doi{10.1109/ACCESS.2018.2877890}.
\bibitem[{Blalock et~al.(2020)Blalock, Ortiz, Frankle and Guttag}]{Blalock2020}
\bibinfo{author}{Blalock, D.}, \bibinfo{author}{Ortiz, J.J.G.},
  \bibinfo{author}{Frankle, J.}, \bibinfo{author}{Guttag, J.},
  \bibinfo{year}{2020}.
\newblock \bibinfo{title}{{What is the State of Neural Network Pruning?}}
\newblock \bibinfo{journal}{ArXiv preprint} \URLprefix
  \url{http://arxiv.org/abs/2003.03033}.
\bibitem[{Bolukbasi et~al.(2017)Bolukbasi, Wang, Dekel and
  Saligrama}]{Bolukbasi2017}
\bibinfo{author}{Bolukbasi, T.}, \bibinfo{author}{Wang, J.},
  \bibinfo{author}{Dekel, O.}, \bibinfo{author}{Saligrama, V.},
  \bibinfo{year}{2017}.
\newblock \bibinfo{title}{{Adaptive Neural Networks for Efficient Inference}}.
\newblock \bibinfo{journal}{Thirty-fourth International Conference on Machine
  Learning} \URLprefix \url{https://arxiv.org/abs/1702.07811
  http://arxiv.org/abs/1702.07811}.
\bibitem[{Brown et~al.(2020)Brown, Mann, Ryder, Subbiah, Kaplan, Dhariwal,
  Neelakantan, Shyam, Sastry, Askell, Agarwal, Herbert-Voss, Krueger, Henighan,
  Child, Ramesh, Ziegler, Wu, Winter, Hesse, Chen, Sigler, Litwin, Gray, Chess,
  Clark, Berner, McCandlish, Radford, Sutskever and Amodei}]{Brown2020}
\bibinfo{author}{Brown, T.B.}, \bibinfo{author}{Mann, B.},
  \bibinfo{author}{Ryder, N.}, \bibinfo{author}{Subbiah, M.},
  \bibinfo{author}{Kaplan, J.}, \bibinfo{author}{Dhariwal, P.},
  \bibinfo{author}{Neelakantan, A.}, \bibinfo{author}{Shyam, P.},
  \bibinfo{author}{Sastry, G.}, \bibinfo{author}{Askell, A.},
  \bibinfo{author}{Agarwal, S.}, \bibinfo{author}{Herbert-Voss, A.},
  \bibinfo{author}{Krueger, G.}, \bibinfo{author}{Henighan, T.},
  \bibinfo{author}{Child, R.}, \bibinfo{author}{Ramesh, A.},
  \bibinfo{author}{Ziegler, D.M.}, \bibinfo{author}{Wu, J.},
  \bibinfo{author}{Winter, C.}, \bibinfo{author}{Hesse, C.},
  \bibinfo{author}{Chen, M.}, \bibinfo{author}{Sigler, E.},
  \bibinfo{author}{Litwin, M.}, \bibinfo{author}{Gray, S.},
  \bibinfo{author}{Chess, B.}, \bibinfo{author}{Clark, J.},
  \bibinfo{author}{Berner, C.}, \bibinfo{author}{McCandlish, S.},
  \bibinfo{author}{Radford, A.}, \bibinfo{author}{Sutskever, I.},
  \bibinfo{author}{Amodei, D.}, \bibinfo{year}{2020}.
\newblock \bibinfo{title}{{Language Models are Few-Shot Learners}}.
\newblock \bibinfo{journal}{ArXiv preprint} \URLprefix
  \url{http://arxiv.org/abs/2005.14165}.
\bibitem[{Buciluǎ et~al.(2006)Buciluǎ, Caruana and
  Niculescu-Mizil}]{Bucilua2006}
\bibinfo{author}{Buciluǎ, C.}, \bibinfo{author}{Caruana, R.},
  \bibinfo{author}{Niculescu-Mizil, A.}, \bibinfo{year}{2006}.
\newblock \bibinfo{title}{{Model compression}}, in:
  \bibinfo{booktitle}{Proceedings of the 12th ACM SIGKDD international
  conference on Knowledge discovery and data mining - KDD '06},
  \bibinfo{publisher}{ACM Press}, \bibinfo{address}{New York, New York, USA}.
  p. \bibinfo{pages}{535}.
\newblock \URLprefix \url{https://dl.acm.org/doi/abs/10.1145/1150402.1150464},
  \DOIprefix\doi{10.1145/1150402.1150464}.
\bibitem[{{BUG1989}(2019)}]{BUG19892019}
\bibinfo{author}{{BUG1989}}, \bibinfo{year}{2019}.
\newblock \bibinfo{title}{{BUG1989/caffe-int8-convert-tools: Generate a
  quantization parameter file for ncnn framework int8 inference}}.
\newblock \URLprefix \url{https://github.com/BUG1989/caffe-INT8-convert-tools}.
\bibitem[{Cai et~al.(2019)Cai, Gan, Wang, Zhang and Han}]{Cai2019a}
\bibinfo{author}{Cai, H.}, \bibinfo{author}{Gan, C.}, \bibinfo{author}{Wang,
  T.}, \bibinfo{author}{Zhang, Z.}, \bibinfo{author}{Han, S.},
  \bibinfo{year}{2019}.
\newblock \bibinfo{title}{{Once-for-All: Train One Network and Specialize it
  for Efficient Deployment}}.
\newblock \bibinfo{journal}{ArXiv preprint} , \bibinfo{pages}{1--15}\URLprefix
  \url{http://arxiv.org/abs/1908.09791}.
\bibitem[{Cai et~al.(2018)Cai, Takemoto and Nakajo}]{Cai2018}
\bibinfo{author}{Cai, J.}, \bibinfo{author}{Takemoto, M.},
  \bibinfo{author}{Nakajo, H.}, \bibinfo{year}{2018}.
\newblock \bibinfo{title}{{A Deep Look into Logarithmic Quantization of Model
  Parameters in Neural Networks}}, in: \bibinfo{booktitle}{Proceedings of the
  10th International Conference on Advances in Information Technology - IAIT
  2018}, \bibinfo{publisher}{ACM Press}, \bibinfo{address}{New York, New York,
  USA}. pp. \bibinfo{pages}{1--8}.
\newblock \URLprefix \url{http://dl.acm.org/citation.cfm?doid=3291280.3291800},
  \DOIprefix\doi{10.1145/3291280.3291800}.
\bibitem[{Cai et~al.(2017)Cai, He, Sun and Vasconcelos}]{Cai2017}
\bibinfo{author}{Cai, Z.}, \bibinfo{author}{He, X.}, \bibinfo{author}{Sun, J.},
  \bibinfo{author}{Vasconcelos, N.}, \bibinfo{year}{2017}.
\newblock \bibinfo{title}{{Deep Learning with Low Precision by Half-Wave
  Gaussian Quantization}}, in: \bibinfo{booktitle}{2017 IEEE Conference on
  Computer Vision and Pattern Recognition (CVPR)}, \bibinfo{publisher}{IEEE}.
  pp. \bibinfo{pages}{5406--5414}.
\newblock \URLprefix \url{http://ieeexplore.ieee.org/document/8100057/},
  \DOIprefix\doi{10.1109/CVPR.2017.574}.
\bibitem[{Cao et~al.(2019)Cao, Ma, Xiao, Zhang, Liu, Zhang, Nie and
  Yang}]{Cao2019}
\bibinfo{author}{Cao, S.}, \bibinfo{author}{Ma, L.}, \bibinfo{author}{Xiao,
  W.}, \bibinfo{author}{Zhang, C.}, \bibinfo{author}{Liu, Y.},
  \bibinfo{author}{Zhang, L.}, \bibinfo{author}{Nie, L.},
  \bibinfo{author}{Yang, Z.}, \bibinfo{year}{2019}.
\newblock \bibinfo{title}{{SeerNet : Predicting Convolutional Neural Network
  Feature-Map Sparsity through Low-Bit Quantization}}.
\newblock \bibinfo{journal}{Proceedings of the IEEE/CVF Conference on Computer
  Vision and Pattern Recognition (CVPR)} \URLprefix
  \url{http://openaccess.thecvf.com/content_CVPR_2019/papers/Cao_SeerNet_Predicting_Convolutional_Neural_Network_Feature-Map_Sparsity_Through_Low-Bit_Quantization_CVPR_2019_paper.pdf}.
\bibitem[{Carreira-Perpinan and Idelbayev(2018)}]{Carreira-Perpinan2018}
\bibinfo{author}{Carreira-Perpinan, M.A.}, \bibinfo{author}{Idelbayev, Y.},
  \bibinfo{year}{2018}.
\newblock \bibinfo{title}{{"Learning-Compression" Algorithms for Neural Net
  Pruning}}, in: \bibinfo{booktitle}{IEEE/CVF Conference on Computer Vision and
  Pattern Recognition (CVPR)}, \bibinfo{publisher}{IEEE}. pp.
  \bibinfo{pages}{8532--8541}.
\newblock \URLprefix \url{https://ieeexplore.ieee.org/document/8578988/},
  \DOIprefix\doi{10.1109/CVPR.2018.00890}.
\bibitem[{Chellapilla et~al.(2006)Chellapilla, Puri and
  Simard}]{Chellapilla2006}
\bibinfo{author}{Chellapilla, K.}, \bibinfo{author}{Puri, S.},
  \bibinfo{author}{Simard, P.}, \bibinfo{year}{2006}.
\newblock \bibinfo{title}{{High Performance Convolutional Neural Networks for
  Document Processing}}, in: \bibinfo{booktitle}{Tenth International Workshop
  on Frontiers in Handwriting Recognition}.
\newblock \URLprefix \url{https://hal.inria.fr/inria-00112631/},
  \DOIprefix\doi{10.1.1.137.482}.
\bibitem[{Chen et~al.(2020)Chen, Wang, Xu, Shi, Xu, Tian and Xu}]{Chen2019b}
\bibinfo{author}{Chen, H.}, \bibinfo{author}{Wang, Y.}, \bibinfo{author}{Xu,
  C.}, \bibinfo{author}{Shi, B.}, \bibinfo{author}{Xu, C.},
  \bibinfo{author}{Tian, Q.}, \bibinfo{author}{Xu, C.}, \bibinfo{year}{2020}.
\newblock \bibinfo{title}{{AdderNet: Do We Really Need Multiplications in Deep
  Learning?}}, in: \bibinfo{booktitle}{Proceedings of the IEEE/CVF Conference
  on Computer Vision and Pattern Recognition (CVPR)}, pp.
  \bibinfo{pages}{1468--1477}.
\newblock \URLprefix \url{http://arxiv.org/abs/1912.13200}.
\bibitem[{Chen et~al.(2018)Chen, Moreau, Jiang, Zheng, Yan, Cowan, Shen, Wang,
  Hu, Ceze, Guestrin and Krishnamurthy}]{Chen2018}
\bibinfo{author}{Chen, T.}, \bibinfo{author}{Moreau, T.},
  \bibinfo{author}{Jiang, Z.}, \bibinfo{author}{Zheng, L.},
  \bibinfo{author}{Yan, E.}, \bibinfo{author}{Cowan, M.},
  \bibinfo{author}{Shen, H.}, \bibinfo{author}{Wang, L.}, \bibinfo{author}{Hu,
  Y.}, \bibinfo{author}{Ceze, L.}, \bibinfo{author}{Guestrin, C.},
  \bibinfo{author}{Krishnamurthy, A.}, \bibinfo{year}{2018}.
\newblock \bibinfo{title}{{TVM: An automated end-to-end optimizing compiler for
  deep learning}}, in: \bibinfo{booktitle}{Proceedings of the 13th USENIX
  Symposium on Operating Systems Design and Implementation, OSDI 2018}, pp.
  \bibinfo{pages}{579--594}.
\newblock \URLprefix \url{http://arxiv.org/abs/1802.04799}.
\bibitem[{Chen et~al.(2015)Chen, Wilson, Tyree, Weinberger and Chen}]{Chen2015}
\bibinfo{author}{Chen, W.}, \bibinfo{author}{Wilson, J.},
  \bibinfo{author}{Tyree, S.}, \bibinfo{author}{Weinberger, K.},
  \bibinfo{author}{Chen, Y.}, \bibinfo{year}{2015}.
\newblock \bibinfo{title}{{Compressing neural networks with the hashing
  trick.}}, in: \bibinfo{booktitle}{In International Conference on Machine
  Learning}, pp. \bibinfo{pages}{2285--2294}.
\newblock \URLprefix \url{http://arxiv.org/abs/1504.04788}.
\bibitem[{Chen et~al.(2016)Chen, Chen, Xu, Sun and Temam}]{Chen2016b}
\bibinfo{author}{Chen, Y.}, \bibinfo{author}{Chen, T.}, \bibinfo{author}{Xu,
  Z.}, \bibinfo{author}{Sun, N.}, \bibinfo{author}{Temam, O.},
  \bibinfo{year}{2016}.
\newblock \bibinfo{title}{{DianNao family: Energy-Efficient Hardware
  Accelerators for Machine Learning}}.
\newblock \bibinfo{journal}{Communications of the ACM} \bibinfo{volume}{59},
  \bibinfo{pages}{105--112}.
\newblock \URLprefix
  \url{10.1145/2594446%5Cnhttps://ejwl.idm.oclc.org/login?url=http://search.ebscohost.com/login.aspx?direct=true&db=bth&AN=95797996&site=ehost-live
  http://dl.acm.org/citation.cfm?doid=3013530.2996864},
  \DOIprefix\doi{10.1145/2996864}.
\bibitem[{Cheng et~al.(2018)Cheng, Wang, Li, Hu and Lu}]{Cheng2018}
\bibinfo{author}{Cheng, J.}, \bibinfo{author}{Wang, P.s.}, \bibinfo{author}{Li,
  G.}, \bibinfo{author}{Hu, Q.h.}, \bibinfo{author}{Lu, H.q.},
  \bibinfo{year}{2018}.
\newblock \bibinfo{title}{{Recent advances in efficient computation of deep
  convolutional neural networks}}.
\newblock \bibinfo{journal}{Frontiers of Information Technology {\&} Electronic
  Engineering} \bibinfo{volume}{19}, \bibinfo{pages}{64--77}.
\newblock \URLprefix \url{http://link.springer.com/10.1631/FITEE.1700789},
  \DOIprefix\doi{10.1631/FITEE.1700789}.
\bibitem[{Cheng et~al.(2017)Cheng, Wang, Zhou and Zhang}]{Cheng2017a}
\bibinfo{author}{Cheng, Y.}, \bibinfo{author}{Wang, D.}, \bibinfo{author}{Zhou,
  P.}, \bibinfo{author}{Zhang, T.}, \bibinfo{year}{2017}.
\newblock \bibinfo{title}{{A Survey of Model Compression and Acceleration for
  Deep Neural Networks}}.
\newblock \bibinfo{journal}{ArXiv preprint} \URLprefix
  \url{http://arxiv.org/abs/1710.09282}.
\bibitem[{Cheng et~al.(2015)Cheng, Soudry, Mao and Lan}]{Cheng2015}
\bibinfo{author}{Cheng, Z.}, \bibinfo{author}{Soudry, D.},
  \bibinfo{author}{Mao, Z.}, \bibinfo{author}{Lan, Z.}, \bibinfo{year}{2015}.
\newblock \bibinfo{title}{{Training Binary Multilayer Neural Networks for Image
  Classification using Expectation Backpropagation}}.
\newblock \bibinfo{journal}{ArXiv preprint} \URLprefix
  \url{http://cn.arxiv.org/pdf/1503.03562.pdf http://arxiv.org/abs/1503.03562}.
\bibitem[{Chiliang et~al.(2019)Chiliang, Tao, Yingda and Zuochang}]{Zhang2019d}
\bibinfo{author}{Chiliang, Z.}, \bibinfo{author}{Tao, H.},
  \bibinfo{author}{Yingda, G.}, \bibinfo{author}{Zuochang, Y.},
  \bibinfo{year}{2019}.
\newblock \bibinfo{title}{{Accelerating Convolutional Neural Networks with
  Dynamic Channel Pruning}}, in: \bibinfo{booktitle}{2019 Data Compression
  Conference (DCC)}, \bibinfo{publisher}{IEEE}. pp. \bibinfo{pages}{563--563}.
\newblock \URLprefix \url{https://ieeexplore.ieee.org/document/8712710/},
  \DOIprefix\doi{10.1109/DCC.2019.00075}.
\bibitem[{Choi et~al.(2008)Choi, Lee and Kim}]{Choi2008}
\bibinfo{author}{Choi, B.}, \bibinfo{author}{Lee, J.H.}, \bibinfo{author}{Kim,
  D.H.}, \bibinfo{year}{2008}.
\newblock \bibinfo{title}{{Solving local minima problem with large number of
  hidden nodes on two-layered feed-forward artificial neural networks}}.
\newblock \bibinfo{journal}{Neurocomputing} \bibinfo{volume}{71},
  \bibinfo{pages}{3640--3643}.
\newblock \DOIprefix\doi{10.1016/j.neucom.2008.04.004}.
\bibitem[{Choi et~al.(2018)Choi, Wang, Venkataramani, Chuang, Srinivasan and
  Gopalakrishnan}]{Choi2018}
\bibinfo{author}{Choi, J.}, \bibinfo{author}{Wang, Z.},
  \bibinfo{author}{Venkataramani, S.}, \bibinfo{author}{Chuang, P.I.j.},
  \bibinfo{author}{Srinivasan, V.}, \bibinfo{author}{Gopalakrishnan, K.},
  \bibinfo{year}{2018}.
\newblock \bibinfo{title}{{PACT: Parameterized Clipping Activation for
  Quantized Neural Networks}}.
\newblock \bibinfo{journal}{ArXiv preprint} , \bibinfo{pages}{1--15}\URLprefix
  \url{http://arxiv.org/abs/1805.06085}.
\bibitem[{Choi et~al.(2017a)Choi, El-Khamy and Lee}]{Choi2017}
\bibinfo{author}{Choi, Y.}, \bibinfo{author}{El-Khamy, M.},
  \bibinfo{author}{Lee, J.}, \bibinfo{year}{2017}a.
\newblock \bibinfo{title}{{Towards the Limit of Network Quantization}}, in:
  \bibinfo{booktitle}{International Conference on Learning
  Representations(ICLR)}, \bibinfo{publisher}{IEEE}.
\newblock \URLprefix \url{https://arxiv.org/abs/1612.01543
  http://arxiv.org/abs/1612.01543}.
\bibitem[{Choi et~al.(2017b)Choi, Member, Bae, Sim, Member, Choi, Kim, Member,
  Kim and Member}]{Choi2017a}
\bibinfo{author}{Choi, Y.}, \bibinfo{author}{Member, S.S.},
  \bibinfo{author}{Bae, D.}, \bibinfo{author}{Sim, J.},
  \bibinfo{author}{Member, S.S.}, \bibinfo{author}{Choi, S.},
  \bibinfo{author}{Kim, M.}, \bibinfo{author}{Member, S.S.},
  \bibinfo{author}{Kim, L.s.S.}, \bibinfo{author}{Member, S.S.},
  \bibinfo{year}{2017}b.
\newblock \bibinfo{title}{{Energy-Efficient Design of Processing Element for
  Convolutional Neural Network}}.
\newblock \bibinfo{journal}{IEEE Transactions on Circuits and Systems II:
  Express Briefs} \bibinfo{volume}{64}, \bibinfo{pages}{1332--1336}.
\newblock \URLprefix \url{http://ieeexplore.ieee.org/document/7893765/},
  \DOIprefix\doi{10.1109/TCSII.2017.2691771}.
\bibitem[{Chollet and Google(2017)}]{Chollet2017}
\bibinfo{author}{Chollet, F.}, \bibinfo{author}{Google, C.},
  \bibinfo{year}{2017}.
\newblock \bibinfo{title}{{Xception : Deep Learning with Depthwise Separable
  Convolutions}}, in: \bibinfo{booktitle}{The IEEE Conference on Computer
  Vision and Pattern Recognition (CVPR)}, \bibinfo{publisher}{IEEE}. pp.
  \bibinfo{pages}{1251--1258}.
\newblock \URLprefix \url{http://ieeexplore.ieee.org/document/8099678/},
  \DOIprefix\doi{10.1109/CVPR.2017.195}.
\bibitem[{Choudhary et~al.(2020)Choudhary, Mishra, Goswami and
  Sarangapani}]{Choudhary2020}
\bibinfo{author}{Choudhary, T.}, \bibinfo{author}{Mishra, V.},
  \bibinfo{author}{Goswami, A.}, \bibinfo{author}{Sarangapani, J.},
  \bibinfo{year}{2020}.
\newblock \bibinfo{title}{{A comprehensive survey on model compression and
  acceleration}}.
\newblock \bibinfo{journal}{Artificial Intelligence Review}
  \bibinfo{volume}{53}, \bibinfo{pages}{5113--5155}.
\newblock \URLprefix \url{https://doi.org/10.1007/s10462-020-09816-7},
  \DOIprefix\doi{10.1007/s10462-020-09816-7}.
\bibitem[{Cornea(2015)}]{Cornea2015}
\bibinfo{author}{Cornea, M.}, \bibinfo{year}{2015}.
\newblock \bibinfo{title}{{Intel {\textregistered} AVX-512 Instructions and
  Their Use in the Implementation of Math Functions}}.
\newblock \bibinfo{journal}{Intel Corporation} .
\bibitem[{Cotofana et~al.(1997)Cotofana, Vassiliadis, Logic, Addition and
  Addition}]{Cotofana1997}
\bibinfo{author}{Cotofana, S.}, \bibinfo{author}{Vassiliadis, S.},
  \bibinfo{author}{Logic, T.}, \bibinfo{author}{Addition, B.},
  \bibinfo{author}{Addition, S.}, \bibinfo{year}{1997}.
\newblock \bibinfo{title}{{Low Weight and Fan-In Neural Networks for Basic
  Arithmetic Operations}}, in: \bibinfo{booktitle}{15th IMACS World Congress},
  pp. \bibinfo{pages}{227--232}.
\newblock \DOIprefix\doi{10.1.1.50.4450}.
\bibitem[{Courbariaux et~al.(2014)Courbariaux, Bengio and
  David}]{Courbariaux2014}
\bibinfo{author}{Courbariaux, M.}, \bibinfo{author}{Bengio, Y.},
  \bibinfo{author}{David, J.P.}, \bibinfo{year}{2014}.
\newblock \bibinfo{title}{{Training deep neural networks with low precision
  multiplications}}, in: \bibinfo{booktitle}{International Conference on
  Learning Representations(ICLR)}, pp. \bibinfo{pages}{1--10}.
\newblock \URLprefix \url{http://arxiv.org/abs/1412.7024},
  \DOIprefix\doi{arXiv: 1412.7024}.
\bibitem[{Courbariaux et~al.(2015)Courbariaux, Bengio and
  David}]{Courbariaux2015}
\bibinfo{author}{Courbariaux, M.}, \bibinfo{author}{Bengio, Y.},
  \bibinfo{author}{David, J.P.}, \bibinfo{year}{2015}.
\newblock \bibinfo{title}{{BinaryConnect: Training Deep Neural Networks with
  binary weights during propagations}}, in: \bibinfo{booktitle}{Advances in
  Neural Information Processing Systems (NIPS)}, pp. \bibinfo{pages}{1--9}.
\newblock \URLprefix \url{http://arxiv.org/abs/1511.00363},
  \DOIprefix\doi{10.5555/2969442.2969588}.
\bibitem[{Courbariaux et~al.(2016)Courbariaux, Hubara, Soudry, El-Yaniv and
  Bengio}]{Courbariaux2016}
\bibinfo{author}{Courbariaux, M.}, \bibinfo{author}{Hubara, I.},
  \bibinfo{author}{Soudry, D.}, \bibinfo{author}{El-Yaniv, R.},
  \bibinfo{author}{Bengio, Y.}, \bibinfo{year}{2016}.
\newblock \bibinfo{title}{{Binarized Neural Networks: Training Deep Neural
  Networks with Weights and Activations Constrained to +1 or -1}}.
\newblock \bibinfo{journal}{ArXiv preprint} \URLprefix
  \url{https://github.com/MatthieuCourbariaux/
  http://arxiv.org/abs/1602.02830}.
\bibitem[{Das et~al.(2018)Das, Mellempudi, Mudigere, Kalamkar, Avancha,
  Banerjee, Sridharan, Vaidyanathan, Kaul, Georganas, Heinecke, Dubey, Corbal,
  Shustrov, Dubtsov, Fomenko and Pirogov}]{Das2018}
\bibinfo{author}{Das, D.}, \bibinfo{author}{Mellempudi, N.},
  \bibinfo{author}{Mudigere, D.}, \bibinfo{author}{Kalamkar, D.},
  \bibinfo{author}{Avancha, S.}, \bibinfo{author}{Banerjee, K.},
  \bibinfo{author}{Sridharan, S.}, \bibinfo{author}{Vaidyanathan, K.},
  \bibinfo{author}{Kaul, B.}, \bibinfo{author}{Georganas, E.},
  \bibinfo{author}{Heinecke, A.}, \bibinfo{author}{Dubey, P.},
  \bibinfo{author}{Corbal, J.}, \bibinfo{author}{Shustrov, N.},
  \bibinfo{author}{Dubtsov, R.}, \bibinfo{author}{Fomenko, E.},
  \bibinfo{author}{Pirogov, V.}, \bibinfo{year}{2018}.
\newblock \bibinfo{title}{{Mixed Precision Training of Convolutional Neural
  Networks using Integer Operations}}, in: \bibinfo{booktitle}{International
  Conference on Learning Representations(ICLR)}, pp. \bibinfo{pages}{1--11}.
\newblock \URLprefix
  \url{https://www.anandtech.com/show/11741/hot-chips-intel-knights-mill-live-blog-445pm-pt-1145pm-utc
  http://arxiv.org/abs/1802.00930}.
\bibitem[{Dash and Liu(1997)}]{Dash1997}
\bibinfo{author}{Dash, M.}, \bibinfo{author}{Liu, H.}, \bibinfo{year}{1997}.
\newblock \bibinfo{title}{{Feature selection for classification}}.
\newblock \bibinfo{journal}{Intelligent Data Analysis} \bibinfo{volume}{1},
  \bibinfo{pages}{131--156}.
\newblock \DOIprefix\doi{10.3233/IDA-1997-1302}.
\bibitem[{Davis and Arel(2013)}]{Davis2013}
\bibinfo{author}{Davis, A.}, \bibinfo{author}{Arel, I.}, \bibinfo{year}{2013}.
\newblock \bibinfo{title}{{Low-Rank Approximations for Conditional Feedforward
  Computation in Deep Neural Networks}}, in: \bibinfo{booktitle}{International
  Conference on Learning Representations Workshops (ICLRW)}, pp.
  \bibinfo{pages}{1--10}.
\newblock \URLprefix \url{http://arxiv.org/abs/1312.4461}.
\bibitem[{Deng et~al.(2013)Deng, Yin and Zhang}]{Deng2013}
\bibinfo{author}{Deng, W.}, \bibinfo{author}{Yin, W.}, \bibinfo{author}{Zhang,
  Y.}, \bibinfo{year}{2013}.
\newblock \bibinfo{title}{{Group sparse optimization by alternating direction
  method}}, in: \bibinfo{editor}{Van De~Ville, D.}, \bibinfo{editor}{Goyal,
  V.K.}, \bibinfo{editor}{Papadakis, M.} (Eds.), \bibinfo{booktitle}{Wavelets
  and Sparsity XV}, p. \bibinfo{pages}{88580R}.
\newblock \URLprefix
  \url{http://proceedings.spiedigitallibrary.org/proceeding.aspx?doi=10.1117/12.2024410},
  \DOIprefix\doi{10.1117/12.2024410}.
\bibitem[{Dettmers(2015)}]{Dettmers2015}
\bibinfo{author}{Dettmers, T.}, \bibinfo{year}{2015}.
\newblock \bibinfo{title}{{8-Bit Approximations for Parallelism in Deep
  Learning}}, in: \bibinfo{booktitle}{International Conference on Learning
  Representations(ICLR)}.
\newblock \URLprefix \url{https://github.com/soumith/convnet-benchmarks
  http://arxiv.org/abs/1511.04561}.
\bibitem[{Dong et~al.(2017)Dong, Huang, Yang and Yan}]{Dong2017a}
\bibinfo{author}{Dong, X.}, \bibinfo{author}{Huang, J.}, \bibinfo{author}{Yang,
  Y.}, \bibinfo{author}{Yan, S.}, \bibinfo{year}{2017}.
\newblock \bibinfo{title}{{More is less: A more complicated network with less
  inference complexity}}.
\newblock \bibinfo{journal}{Proceedings - 30th IEEE Conference on Computer
  Vision and Pattern Recognition, CVPR 2017} \bibinfo{volume}{2017-Janua},
  \bibinfo{pages}{1895--1903}.
\newblock \URLprefix \url{http://arxiv.org/abs/1703.08651},
  \DOIprefix\doi{10.1109/CVPR.2017.205}.
\bibitem[{Dongarra et~al.(1990)Dongarra, Du~Croz, Hammarling and
  Duff}]{Dongarra1990}
\bibinfo{author}{Dongarra, J.J.}, \bibinfo{author}{Du~Croz, J.},
  \bibinfo{author}{Hammarling, S.}, \bibinfo{author}{Duff, I.S.},
  \bibinfo{year}{1990}.
\newblock \bibinfo{title}{{A set of level 3 basic linear algebra subprograms}}.
\newblock \bibinfo{journal}{ACM Transactions on Mathematical Software (TOMS)}
  \bibinfo{volume}{16}, \bibinfo{pages}{1--17}.
\newblock \DOIprefix\doi{10.1145/77626.79170}.
\bibitem[{Dukhan et~al.(2019)Dukhan, Yiming, Hao and Lu}]{Dukhan2018}
\bibinfo{author}{Dukhan, M.}, \bibinfo{author}{Yiming, W.},
  \bibinfo{author}{Hao, L.}, \bibinfo{author}{Lu, H.}, \bibinfo{year}{2019}.
\newblock \bibinfo{title}{{QNNPACK: Open source library for optimized mobile
  deep learning - Facebook Engineering}}.
\newblock \URLprefix \url{https://engineering.fb.com/ml-applications/qnnpack/}.
\bibitem[{Elhoushi et~al.(2019)Elhoushi, Chen, Shafiq, Tian and
  Li}]{Elhoushi2019}
\bibinfo{author}{Elhoushi, M.}, \bibinfo{author}{Chen, Z.},
  \bibinfo{author}{Shafiq, F.}, \bibinfo{author}{Tian, Y.H.},
  \bibinfo{author}{Li, J.Y.}, \bibinfo{year}{2019}.
\newblock \bibinfo{title}{{DeepShift: Towards Multiplication-Less Neural
  Networks}}.
\newblock \bibinfo{journal}{ArXiv preprint} \URLprefix
  \url{http://arxiv.org/abs/1905.13298}.
\bibitem[{Elsken et~al.(2019)Elsken, Metzen and Hutter}]{Wistuba2019}
\bibinfo{author}{Elsken, T.}, \bibinfo{author}{Metzen, J.H.},
  \bibinfo{author}{Hutter, F.}, \bibinfo{year}{2019}.
\newblock \bibinfo{title}{{Neural Architecture Search}}.
\newblock \bibinfo{journal}{Journal of Machine Learning Research}
  \bibinfo{volume}{20}, \bibinfo{pages}{63--77}.
\newblock \URLprefix
  \url{http://link.springer.com/10.1007/978-3-030-05318-5_3},
  \DOIprefix\doi{10.1007/978-3-030-05318-5{\_}3}.
\bibitem[{Engelbrecht(2001)}]{Engelbrecht2001}
\bibinfo{author}{Engelbrecht, A.P.}, \bibinfo{year}{2001}.
\newblock \bibinfo{title}{{A new pruning heuristic based on variance analysis
  of sensitivity information}}.
\newblock \bibinfo{journal}{IEEE Transactions on Neural Networks}
  \bibinfo{volume}{12}, \bibinfo{pages}{1386--1389}.
\newblock \DOIprefix\doi{10.1109/72.963775}.
\bibitem[{Esser et~al.(2016)Esser, Merolla, Arthur, Cassidy, Appuswamy,
  Andreopoulos, Berg, McKinstry, Melano, Barch, di~Nolfo, Datta, Amir, Taba,
  Flickner and Modha}]{Esser2016}
\bibinfo{author}{Esser, S.K.}, \bibinfo{author}{Merolla, P.A.},
  \bibinfo{author}{Arthur, J.V.}, \bibinfo{author}{Cassidy, A.S.},
  \bibinfo{author}{Appuswamy, R.}, \bibinfo{author}{Andreopoulos, A.},
  \bibinfo{author}{Berg, D.J.}, \bibinfo{author}{McKinstry, J.L.},
  \bibinfo{author}{Melano, T.}, \bibinfo{author}{Barch, D.R.},
  \bibinfo{author}{di~Nolfo, C.}, \bibinfo{author}{Datta, P.},
  \bibinfo{author}{Amir, A.}, \bibinfo{author}{Taba, B.},
  \bibinfo{author}{Flickner, M.D.}, \bibinfo{author}{Modha, D.S.},
  \bibinfo{year}{2016}.
\newblock \bibinfo{title}{{Convolutional networks for fast, energy-efficient
  neuromorphic computing}}.
\newblock \bibinfo{journal}{Proceedings of the National Academy of Sciences}
  \bibinfo{volume}{113}, \bibinfo{pages}{11441--11446}.
\newblock \URLprefix
  \url{http://www.pnas.org/lookup/doi/10.1073/pnas.1604850113},
  \DOIprefix\doi{10.1073/pnas.1604850113}.
\bibitem[{Faraone et~al.(2018)Faraone, Fraser, Blott and Leong}]{Faraone2018}
\bibinfo{author}{Faraone, J.}, \bibinfo{author}{Fraser, N.},
  \bibinfo{author}{Blott, M.}, \bibinfo{author}{Leong, P.H.},
  \bibinfo{year}{2018}.
\newblock \bibinfo{title}{{SYQ: Learning Symmetric Quantization for Efficient
  Deep Neural Networks}}, in: \bibinfo{booktitle}{Proceedings of the IEEE/CVF
  Conference on Computer Vision and Pattern Recognition (CVPR)}.
\bibitem[{Fiesler et~al.(1990)Fiesler, Choudry and Caulfield}]{Fiesler1990}
\bibinfo{author}{Fiesler, E.}, \bibinfo{author}{Choudry, A.},
  \bibinfo{author}{Caulfield, H.J.}, \bibinfo{year}{1990}.
\newblock \bibinfo{title}{{Weight discretization paradigm for optical neural
  networks}}.
\newblock \bibinfo{journal}{Optical Interconnections and Networks}
  \bibinfo{volume}{1281}, \bibinfo{pages}{164}.
\newblock \DOIprefix\doi{10.1117/12.20700}.
\bibitem[{Figurnov et~al.(2017)Figurnov, Collins, Zhu, Zhang, Huang, Vetrov and
  Salakhutdinov}]{Figurnov2017}
\bibinfo{author}{Figurnov, M.}, \bibinfo{author}{Collins, M.D.},
  \bibinfo{author}{Zhu, Y.}, \bibinfo{author}{Zhang, L.},
  \bibinfo{author}{Huang, J.}, \bibinfo{author}{Vetrov, D.},
  \bibinfo{author}{Salakhutdinov, R.}, \bibinfo{year}{2017}.
\newblock \bibinfo{title}{{Spatially Adaptive Computation Time for Residual
  Networks}}, in: \bibinfo{booktitle}{IEEE/CVF Conference on Computer Vision
  and Pattern Recognition (CVPR)}, \bibinfo{publisher}{IEEE}. pp.
  \bibinfo{pages}{1790--1799}.
\newblock \URLprefix \url{http://ieeexplore.ieee.org/document/8099677/},
  \DOIprefix\doi{10.1109/CVPR.2017.194}.
\bibitem[{FPGA()}]{FPGA}
\bibinfo{author}{FPGA, I.}, .
\newblock \bibinfo{title}{{Intel{\textregistered} FPGA Development Tools -
  Intel FPGA}}.
\newblock \URLprefix
  \url{https://www.intel.com/content/www/us/en/software/programmable/overview.html}.
\bibitem[{Frankle and Carbin(2019)}]{Frankle2019}
\bibinfo{author}{Frankle, J.}, \bibinfo{author}{Carbin, M.},
  \bibinfo{year}{2019}.
\newblock \bibinfo{title}{{The lottery ticket hypothesis: Finding sparse,
  trainable neural networks}}, in: \bibinfo{booktitle}{International Conference
  on Learning Representations(ICLR)}.
\newblock \URLprefix \url{http://arxiv.org/abs/1803.03635}.
\bibitem[{Fukushima(1988)}]{Fukushima1988}
\bibinfo{author}{Fukushima, K.}, \bibinfo{year}{1988}.
\newblock \bibinfo{title}{{Neocognitron: A hierarchical neural network capable
  of visual pattern recognition}}.
\newblock \bibinfo{journal}{Neural Networks} \bibinfo{volume}{1},
  \bibinfo{pages}{119--130}.
\newblock \DOIprefix\doi{10.1016/0893-6080(88)90014-7}.
\bibitem[{Gale et~al.(2019)Gale, Elsen and Hooker}]{Gale2019}
\bibinfo{author}{Gale, T.}, \bibinfo{author}{Elsen, E.},
  \bibinfo{author}{Hooker, S.}, \bibinfo{year}{2019}.
\newblock \bibinfo{title}{{The State of Sparsity in Deep Neural Networks}}.
\newblock \bibinfo{journal}{ArXiv preprint} \URLprefix
  \url{http://arxiv.org/abs/1902.09574}.
\bibitem[{Gao et~al.(2019)Gao, Zhao, Dudziak, Mullins, Xu, Dudziak, Mullins and
  Xu}]{Gao2019}
\bibinfo{author}{Gao, X.}, \bibinfo{author}{Zhao, Y.},
  \bibinfo{author}{Dudziak, L.}, \bibinfo{author}{Mullins, R.},
  \bibinfo{author}{Xu, C.Z.}, \bibinfo{author}{Dudziak, L.},
  \bibinfo{author}{Mullins, R.}, \bibinfo{author}{Xu, C.Z.},
  \bibinfo{year}{2019}.
\newblock \bibinfo{title}{{Dynamic Channel Pruning: Feature Boosting and
  Suppression}}, in: \bibinfo{booktitle}{International Conference on Learning
  Representations (ICLR)}, pp. \bibinfo{pages}{1--14}.
\newblock \URLprefix \url{http://arxiv.org/abs/1810.05331}.
\bibitem[{Glossner et~al.(2016)Glossner, Blinzer and Takala}]{Glossner2016}
\bibinfo{author}{Glossner, J.}, \bibinfo{author}{Blinzer, P.},
  \bibinfo{author}{Takala, J.}, \bibinfo{year}{2016}.
\newblock \bibinfo{title}{{HSA-enabled DSPs and accelerators}}.
\newblock \bibinfo{journal}{2015 IEEE Global Conference on Signal and
  Information Processing, GlobalSIP 2015} ,
  \bibinfo{pages}{1407--1411}\DOIprefix\doi{10.1109/GlobalSIP.2015.7418430}.
\bibitem[{Gong et~al.(2019)Gong, Liu, Jiang, Li, Hu, Lin, Yu and
  Yan}]{Gong2019}
\bibinfo{author}{Gong, R.}, \bibinfo{author}{Liu, X.}, \bibinfo{author}{Jiang,
  S.}, \bibinfo{author}{Li, T.}, \bibinfo{author}{Hu, P.},
  \bibinfo{author}{Lin, J.}, \bibinfo{author}{Yu, F.}, \bibinfo{author}{Yan,
  J.}, \bibinfo{year}{2019}.
\newblock \bibinfo{title}{{Differentiable soft quantization: Bridging
  full-precision and low-bit neural networks}}, in:
  \bibinfo{booktitle}{Proceedings of the IEEE International Conference on
  Computer Vision (ICCV)}, pp. \bibinfo{pages}{4851--4860}.
\newblock \DOIprefix\doi{10.1109/ICCV.2019.00495}.
\bibitem[{Gong et~al.(2014)Gong, Liu, Yang and Bourdev}]{Gong2014}
\bibinfo{author}{Gong, Y.}, \bibinfo{author}{Liu, L.}, \bibinfo{author}{Yang,
  M.}, \bibinfo{author}{Bourdev, L.}, \bibinfo{year}{2014}.
\newblock \bibinfo{title}{{Compressing Deep Convolutional Networks using Vector
  Quantization}}, in: \bibinfo{booktitle}{International Conference on Learning
  Representations(ICLR)}.
\newblock \URLprefix \url{http://arxiv.org/abs/1412.6115}.
\bibitem[{{Google}()}]{Google-tf-lite}
\bibinfo{author}{{Google}}, .
\newblock \bibinfo{title}{{Hosted models | TensorFlow Lite}}.
\newblock \URLprefix \url{https://www.tensorflow.org/lite/guide/hosted_models}.
\bibitem[{{Google}(2018)}]{Google2018}
\bibinfo{author}{{Google}}, \bibinfo{year}{2018}.
\newblock \bibinfo{title}{{google/gemmlowp: Low-precision matrix
  multiplication}}.
\newblock \bibinfo{howpublished}{https://github.com/google/gemmlowp}.
\newblock \URLprefix \url{https://github.com/google/gemmlowp}.
\bibitem[{Gordon et~al.(2018)Gordon, Eban, Nachum, Chen, Wu, Yang and
  Choi}]{Gordon2018}
\bibinfo{author}{Gordon, A.}, \bibinfo{author}{Eban, E.},
  \bibinfo{author}{Nachum, O.}, \bibinfo{author}{Chen, B.},
  \bibinfo{author}{Wu, H.}, \bibinfo{author}{Yang, T.J.},
  \bibinfo{author}{Choi, E.}, \bibinfo{year}{2018}.
\newblock \bibinfo{title}{{MorphNet: Fast {\&} Simple Resource-Constrained
  Structure Learning of Deep Networks}}, in: \bibinfo{booktitle}{Proceedings of
  the IEEE/CVF Conference on Computer Vision and Pattern Recognition (CVPR)},
  \bibinfo{publisher}{IEEE}. pp. \bibinfo{pages}{1586--1595}.
\newblock \URLprefix \url{https://ieeexplore.ieee.org/document/8578269/},
  \DOIprefix\doi{10.1109/CVPR.2018.00171}.
\bibitem[{Gou et~al.(2020)Gou, Yu, Maybank and Tao}]{Gou2020}
\bibinfo{author}{Gou, J.}, \bibinfo{author}{Yu, B.}, \bibinfo{author}{Maybank,
  S.J.}, \bibinfo{author}{Tao, D.}, \bibinfo{year}{2020}.
\newblock \bibinfo{title}{{Knowledge Distillation: A Survey}}.
\newblock \bibinfo{journal}{ArXiv preprint} \URLprefix
  \url{http://arxiv.org/abs/2006.05525}.
\bibitem[{Graham(2017)}]{Graham2017}
\bibinfo{author}{Graham, B.}, \bibinfo{year}{2017}.
\newblock \bibinfo{title}{{Low-Precision Batch-Normalized Activations}}.
\newblock \bibinfo{journal}{ArXiv preprint} , \bibinfo{pages}{1--16}\URLprefix
  \url{http://arxiv.org/abs/1702.08231}.
\bibitem[{Graves(2016)}]{Graves2016}
\bibinfo{author}{Graves, A.}, \bibinfo{year}{2016}.
\newblock \bibinfo{title}{{Adaptive Computation Time for Recurrent Neural
  Networks}}.
\newblock \bibinfo{journal}{ArXiv preprint} , \bibinfo{pages}{1--19}\URLprefix
  \url{http://arxiv.org/abs/1603.08983}.
\bibitem[{Greff et~al.(2016)Greff, Srivastava and Schmidhuber}]{Greff2016}
\bibinfo{author}{Greff, K.}, \bibinfo{author}{Srivastava, R.K.},
  \bibinfo{author}{Schmidhuber, J.}, \bibinfo{year}{2016}.
\newblock \bibinfo{title}{{Highway and Residual Networks learn Unrolled
  Iterative Estimation}}, in: \bibinfo{booktitle}{International Conference on
  Learning Representations(ICLR)}, pp. \bibinfo{pages}{1--14}.
\newblock \URLprefix \url{http://arxiv.org/abs/1612.07771}.
\bibitem[{Gudovskiy and Rigazio(2017)}]{Gudovskiy2017}
\bibinfo{author}{Gudovskiy, D.A.}, \bibinfo{author}{Rigazio, L.},
  \bibinfo{year}{2017}.
\newblock \bibinfo{title}{{ShiftCNN: Generalized Low-Precision Architecture for
  Inference of Convolutional Neural Networks}}.
\newblock \bibinfo{journal}{ArXiv preprint} \URLprefix
  \url{http://arxiv.org/abs/1706.02393}.
\bibitem[{Guo et~al.(2018)Guo, Sui, Qiu, Yu, Wang, Yao, Han, Wang and
  Yang}]{Guo2018}
\bibinfo{author}{Guo, K.}, \bibinfo{author}{Sui, L.}, \bibinfo{author}{Qiu,
  J.}, \bibinfo{author}{Yu, J.}, \bibinfo{author}{Wang, J.},
  \bibinfo{author}{Yao, S.}, \bibinfo{author}{Han, S.}, \bibinfo{author}{Wang,
  Y.}, \bibinfo{author}{Yang, H.}, \bibinfo{year}{2018}.
\newblock \bibinfo{title}{{Angel-Eye: A complete design flow for mapping CNN
  onto embedded FPGA}}.
\newblock \bibinfo{journal}{IEEE Transactions on Computer-Aided Design of
  Integrated Circuits and Systems} \bibinfo{volume}{37},
  \bibinfo{pages}{35--47}.
\newblock \URLprefix
  \url{https://ieeexplore.ieee.org/abstract/document/7930521/},
  \DOIprefix\doi{10.1109/TCAD.2017.2705069}.
\bibitem[{Guo et~al.(2017)Guo, Zeng, Yu, Wang and Yang}]{Guo2017}
\bibinfo{author}{Guo, K.}, \bibinfo{author}{Zeng, S.}, \bibinfo{author}{Yu,
  J.}, \bibinfo{author}{Wang, Y.}, \bibinfo{author}{Yang, H.},
  \bibinfo{year}{2017}.
\newblock \bibinfo{title}{{A Survey of FPGA-Based Neural Network Accelerator}}.
\newblock \bibinfo{journal}{ACM Transactions on Reconfigurable Technology and
  Systems} \bibinfo{volume}{9}.
\newblock \URLprefix \url{http://arxiv.org/abs/1712.08934
  https://arxiv.org/abs/1712.08934}.
\bibitem[{Guo(2018)}]{Guo2018a}
\bibinfo{author}{Guo, Y.}, \bibinfo{year}{2018}.
\newblock \bibinfo{title}{{A Survey on Methods and Theories of Quantized Neural
  Networks}}.
\newblock \bibinfo{journal}{ArXiv preprint} \URLprefix
  \url{http://arxiv.org/abs/1808.04752}.
\bibitem[{Guo et~al.(2016)Guo, Yao and Chen}]{Guo2016}
\bibinfo{author}{Guo, Y.}, \bibinfo{author}{Yao, A.}, \bibinfo{author}{Chen,
  Y.}, \bibinfo{year}{2016}.
\newblock \bibinfo{title}{{Dynamic Network Surgery for Efficient DNNs}}, in:
  \bibinfo{booktitle}{Advances in Neural Information Processing Systems
  (NIPS)}, pp. \bibinfo{pages}{1379--1387}.
\newblock \URLprefix
  \url{http://papers.nips.cc/paper/6165-dynamic-network-surgery-for-efficient-dnns}.
\bibitem[{Gupta et~al.(2015)Gupta, Agrawal, Gopalakrishnan and
  Narayanan}]{Gupta2015}
\bibinfo{author}{Gupta, S.}, \bibinfo{author}{Agrawal, A.},
  \bibinfo{author}{Gopalakrishnan, K.}, \bibinfo{author}{Narayanan, P.},
  \bibinfo{year}{2015}.
\newblock \bibinfo{title}{{Deep learning with limited numerical precision}},
  in: \bibinfo{booktitle}{International Conference on Machine Learning (ICML)},
  pp. \bibinfo{pages}{1737--1746}.
\bibitem[{Gysel et~al.(2018)Gysel, Pimentel, Motamedi and Ghiasi}]{Gysel2018}
\bibinfo{author}{Gysel, P.}, \bibinfo{author}{Pimentel, J.},
  \bibinfo{author}{Motamedi, M.}, \bibinfo{author}{Ghiasi, S.},
  \bibinfo{year}{2018}.
\newblock \bibinfo{title}{{Ristretto: A Framework for Empirical Study of
  Resource-Efficient Inference in Convolutional Neural Networks}}.
\newblock \bibinfo{journal}{IEEE Transactions on Neural Networks and Learning
  Systems} \bibinfo{volume}{29}, \bibinfo{pages}{1--6}.
\newblock \URLprefix
  \url{https://ieeexplore.ieee.org/abstract/document/8318896/},
  \DOIprefix\doi{10.1109/TNNLS.2018.2808319}.
\bibitem[{Han et~al.(2016a)Han, Liu, Mao, Pu, Pedram, Horowitz and
  Dally}]{Han2016}
\bibinfo{author}{Han, S.}, \bibinfo{author}{Liu, X.}, \bibinfo{author}{Mao,
  H.}, \bibinfo{author}{Pu, J.}, \bibinfo{author}{Pedram, A.},
  \bibinfo{author}{Horowitz, M.A.}, \bibinfo{author}{Dally, W.J.},
  \bibinfo{year}{2016}a.
\newblock \bibinfo{title}{{EIE: Efficient Inference Engine on Compressed Deep
  Neural Network}}, in: \bibinfo{booktitle}{2016 ACM/IEEE 43rd Annual
  International Symposium on Computer Architecture (ISCA)},
  \bibinfo{publisher}{IEEE}. pp. \bibinfo{pages}{243--254}.
\newblock \URLprefix \url{http://ieeexplore.ieee.org/document/7551397/
  http://arxiv.org/abs/1602.01528}, \DOIprefix\doi{10.1109/ISCA.2016.30}.
\bibitem[{Han et~al.(2016b)Han, Mao and Dally}]{Han2015}
\bibinfo{author}{Han, S.}, \bibinfo{author}{Mao, H.}, \bibinfo{author}{Dally,
  W.J.}, \bibinfo{year}{2016}b.
\newblock \bibinfo{title}{{Deep compression: Compressing deep neural networks
  with pruning, trained quantization and Huffman coding}}, in:
  \bibinfo{booktitle}{International Conference on Learning
  Representations(ICLR)}, pp. \bibinfo{pages}{199--203}.
\newblock \URLprefix \url{http://arxiv.org/abs/1510.00149}.
\bibitem[{Han et~al.(2016c)Han, Pool, Narang, Mao, Gong, Tang, Elsen, Vajda,
  Paluri, Tran, Catanzaro and Dally}]{Han2016d}
\bibinfo{author}{Han, S.}, \bibinfo{author}{Pool, J.}, \bibinfo{author}{Narang,
  S.}, \bibinfo{author}{Mao, H.}, \bibinfo{author}{Gong, E.},
  \bibinfo{author}{Tang, S.}, \bibinfo{author}{Elsen, E.},
  \bibinfo{author}{Vajda, P.}, \bibinfo{author}{Paluri, M.},
  \bibinfo{author}{Tran, J.}, \bibinfo{author}{Catanzaro, B.},
  \bibinfo{author}{Dally, W.J.}, \bibinfo{year}{2016}c.
\newblock \bibinfo{title}{{DSD: Dense-Sparse-Dense Training for Deep Neural
  Networks}}, in: \bibinfo{booktitle}{International Conference on Learning
  Representations(ICLR)}.
\newblock \URLprefix \url{http://arxiv.org/abs/1607.04381}.
\bibitem[{Han et~al.(2015)Han, Pool, Tran and Dally}]{Han2015a}
\bibinfo{author}{Han, S.}, \bibinfo{author}{Pool, J.}, \bibinfo{author}{Tran,
  J.}, \bibinfo{author}{Dally, W.J.}, \bibinfo{year}{2015}.
\newblock \bibinfo{title}{{Learning both Weights and Connections for Efficient
  Neural Networks}}, in: \bibinfo{booktitle}{Advances in Neural Information
  Processing Systems (NIPS)}, pp. \bibinfo{pages}{1135--1143}.
\newblock \URLprefix \url{http://arxiv.org/abs/1506.02626},
  \DOIprefix\doi{10.1016/S0140-6736(95)92525-2}.
\bibitem[{Hannun et~al.(2014)Hannun, Case, Casper, Catanzaro, Diamos, Elsen,
  Prenger, Satheesh, Sengupta, Coates and Ng}]{Hannun2014}
\bibinfo{author}{Hannun, A.}, \bibinfo{author}{Case, C.},
  \bibinfo{author}{Casper, J.}, \bibinfo{author}{Catanzaro, B.},
  \bibinfo{author}{Diamos, G.}, \bibinfo{author}{Elsen, E.},
  \bibinfo{author}{Prenger, R.}, \bibinfo{author}{Satheesh, S.},
  \bibinfo{author}{Sengupta, S.}, \bibinfo{author}{Coates, A.},
  \bibinfo{author}{Ng, A.Y.}, \bibinfo{year}{2014}.
\newblock \bibinfo{title}{{Deep Speech: Scaling up end-to-end speech
  recognition}}.
\newblock \bibinfo{journal}{ArXiv preprint} , \bibinfo{pages}{1--12}\URLprefix
  \url{http://arxiv.org/abs/1412.5567}.
\bibitem[{HANSON(1989)}]{HANSON1989}
\bibinfo{author}{HANSON, S.}, \bibinfo{year}{1989}.
\newblock \bibinfo{title}{{Comparing biases for minimal network construction
  with back-propagation}}, in: \bibinfo{booktitle}{Advances in Neural
  Information Processing Systems (NIPS)}, pp. \bibinfo{pages}{177--185}.
\bibitem[{Hassibi et~al.(1993)Hassibi, Stork and Wolff}]{Hassibi1993}
\bibinfo{author}{Hassibi, B.}, \bibinfo{author}{Stork, D.G.},
  \bibinfo{author}{Wolff, G.J.}, \bibinfo{year}{1993}.
\newblock \bibinfo{title}{{Optimal brain surgeon and general network pruning}}.
\newblock \DOIprefix\doi{10.1109/icnn.1993.298572}.
\bibitem[{He et~al.(2015)He, Zhang, Ren and Sun}]{He2015}
\bibinfo{author}{He, K.}, \bibinfo{author}{Zhang, X.}, \bibinfo{author}{Ren,
  S.}, \bibinfo{author}{Sun, J.}, \bibinfo{year}{2015}.
\newblock \bibinfo{title}{{Deep Residual Learning for Image Recognition}}, in:
  \bibinfo{booktitle}{IEEE/CVF Conference on Computer Vision and Pattern
  Recognition (CVPR)}, \bibinfo{publisher}{IEEE}. pp.
  \bibinfo{pages}{171--180}.
\newblock \URLprefix \url{http://arxiv.org/abs/1512.03385
  http://ieeexplore.ieee.org/document/7780459/},
  \DOIprefix\doi{10.3389/fpsyg.2013.00124}.
\bibitem[{He et~al.(2018)He, Kang, Dong, Fu and Yang}]{He2017a}
\bibinfo{author}{He, Y.}, \bibinfo{author}{Kang, G.}, \bibinfo{author}{Dong,
  X.}, \bibinfo{author}{Fu, Y.}, \bibinfo{author}{Yang, Y.},
  \bibinfo{year}{2018}.
\newblock \bibinfo{title}{{Soft Filter Pruning for Accelerating Deep
  Convolutional Neural Networks}}, in: \bibinfo{booktitle}{Proceedings of the
  Twenty-Seventh International Joint Conference on Artificial Intelligence
  (IJCAI-18)}, \bibinfo{publisher}{International Joint Conferences on
  Artificial Intelligence Organization}, \bibinfo{address}{California}. pp.
  \bibinfo{pages}{2234--2240}.
\newblock \URLprefix \url{http://arxiv.org/abs/1808.06866},
  \DOIprefix\doi{10.24963/ijcai.2018/309}.
\bibitem[{He et~al.(2019)He, Liu, Wang, Hu and Yang}]{He2018}
\bibinfo{author}{He, Y.}, \bibinfo{author}{Liu, P.}, \bibinfo{author}{Wang,
  Z.}, \bibinfo{author}{Hu, Z.}, \bibinfo{author}{Yang, Y.},
  \bibinfo{year}{2019}.
\newblock \bibinfo{title}{{Filter Pruning via Geometric Median for Deep
  Convolutional Neural Networks Acceleration}}.
\newblock \bibinfo{journal}{IEEE/CVF Conference on Computer Vision and Pattern
  Recognition (CVPR)} \URLprefix \url{http://arxiv.org/abs/1811.00250}.
\bibitem[{He et~al.(2017)He, Zhang and Sun}]{He2017}
\bibinfo{author}{He, Y.}, \bibinfo{author}{Zhang, X.}, \bibinfo{author}{Sun,
  J.}, \bibinfo{year}{2017}.
\newblock \bibinfo{title}{{Channel Pruning for Accelerating Very Deep Neural
  Networks}}, in: \bibinfo{booktitle}{IEEE International Conference on Computer
  Vision (ICCV)}, \bibinfo{publisher}{IEEE}. pp. \bibinfo{pages}{1398--1406}.
\newblock \URLprefix
  \url{http://openaccess.thecvf.com/content_ICCV_2017/papers/He_Channel_Pruning_for_ICCV_2017_paper.pdf
  http://ieeexplore.ieee.org/document/8237417/},
  \DOIprefix\doi{10.1109/ICCV.2017.155}.
\bibitem[{Hinton(2012)}]{Hinton2012}
\bibinfo{author}{Hinton, G.}, \bibinfo{year}{2012}.
\newblock \bibinfo{title}{{Neural networks for machine learning.}}
\newblock \bibinfo{type}{Technical Report}. Coursera.
\bibitem[{Hinton et~al.(2012)Hinton, Srivastava, Krizhevsky, Sutskever and
  Salakhutdinov}]{Hinton2012a}
\bibinfo{author}{Hinton, G.E.}, \bibinfo{author}{Srivastava, N.},
  \bibinfo{author}{Krizhevsky, A.}, \bibinfo{author}{Sutskever, I.},
  \bibinfo{author}{Salakhutdinov, R.R.}, \bibinfo{year}{2012}.
\newblock \bibinfo{title}{{Improving neural networks by preventing
  co-adaptation of feature detectors}}.
\newblock \bibinfo{journal}{ArXiv preprint} , \bibinfo{pages}{1--18}\URLprefix
  \url{http://arxiv.org/abs/1207.0580}.
\bibitem[{Hou et~al.(2017)Hou, Yao and Kwok}]{Hou2017}
\bibinfo{author}{Hou, L.}, \bibinfo{author}{Yao, Q.}, \bibinfo{author}{Kwok,
  J.T.}, \bibinfo{year}{2017}.
\newblock \bibinfo{title}{{Loss-aware Binarization of Deep Networks}}, in:
  \bibinfo{booktitle}{International Conference on Learning
  Representations(ICLR)}.
\newblock \URLprefix \url{http://arxiv.org/abs/1611.01600}.
\bibitem[{Howard et~al.(2017)Howard, Zhu, Chen, Kalenichenko, Wang, Weyand,
  Andreetto and Adam}]{Howard2017}
\bibinfo{author}{Howard, A.G.}, \bibinfo{author}{Zhu, M.},
  \bibinfo{author}{Chen, B.}, \bibinfo{author}{Kalenichenko, D.},
  \bibinfo{author}{Wang, W.}, \bibinfo{author}{Weyand, T.},
  \bibinfo{author}{Andreetto, M.}, \bibinfo{author}{Adam, H.},
  \bibinfo{year}{2017}.
\newblock \bibinfo{title}{{MobileNets: Efficient Convolutional Neural Networks
  for Mobile Vision Applications}}.
\newblock \bibinfo{journal}{ArXiv preprint} \URLprefix
  \url{http://arxiv.org/abs/1704.04861}.
\bibitem[{Hu et~al.(2016)Hu, Peng, Tai and Tang}]{Hu2016}
\bibinfo{author}{Hu, H.}, \bibinfo{author}{Peng, R.}, \bibinfo{author}{Tai,
  Y.W.}, \bibinfo{author}{Tang, C.K.}, \bibinfo{year}{2016}.
\newblock \bibinfo{title}{{Network Trimming: A Data-Driven Neuron Pruning
  Approach towards Efficient Deep Architectures}}.
\newblock \bibinfo{journal}{ArXiv preprint} \URLprefix
  \url{http://arxiv.org/abs/1607.03250}.
\bibitem[{Hu et~al.(2018)Hu, Wang and Cheng}]{Hu2018}
\bibinfo{author}{Hu, Q.}, \bibinfo{author}{Wang, P.}, \bibinfo{author}{Cheng,
  J.}, \bibinfo{year}{2018}.
\newblock \bibinfo{title}{{From hashing to CNNs: Training binary weight
  networks via hashing}}, in: \bibinfo{booktitle}{AAAI Conference on Artificial
  Intelligence}, pp. \bibinfo{pages}{3247--3254}.
\bibitem[{Huang et~al.(2018)Huang, Chen, Li, Wu, Van Der~Maaten and
  Weinberger}]{Huang2018b}
\bibinfo{author}{Huang, G.}, \bibinfo{author}{Chen, D.}, \bibinfo{author}{Li,
  T.}, \bibinfo{author}{Wu, F.}, \bibinfo{author}{Van Der~Maaten, L.},
  \bibinfo{author}{Weinberger, K.}, \bibinfo{year}{2018}.
\newblock \bibinfo{title}{{Multi-scale dense networks for resource efficient
  image classification}}, in: \bibinfo{booktitle}{International Conference on
  Learning Representations(ICLR)}.
\newblock \URLprefix \url{http://image-net.org/challenges/talks/}.
\bibitem[{Huang et~al.(2017)Huang, Liu, Van Der~Maaten and
  Weinberger}]{Huang2017a}
\bibinfo{author}{Huang, G.}, \bibinfo{author}{Liu, Z.}, \bibinfo{author}{Van
  Der~Maaten, L.}, \bibinfo{author}{Weinberger, K.Q.}, \bibinfo{year}{2017}.
\newblock \bibinfo{title}{{Densely Connected Convolutional Networks}}, in:
  \bibinfo{booktitle}{IEEE/CVF Conference on Computer Vision and Pattern
  Recognition (CVPR)}, \bibinfo{publisher}{IEEE}. pp.
  \bibinfo{pages}{2261--2269}.
\newblock \URLprefix \url{https://ieeexplore.ieee.org/document/8099726/},
  \DOIprefix\doi{10.1109/CVPR.2017.243}.
\bibitem[{Huang and Learned-miller(2014)}]{Huang2014}
\bibinfo{author}{Huang, G.B.}, \bibinfo{author}{Learned-miller, E.},
  \bibinfo{year}{2014}.
\newblock \bibinfo{title}{{Labeled faces in the wild: Updates and new reporting
  procedures}}.
\newblock \bibinfo{journal}{Dept. Comput. Sci., Univ. Massachusetts Amherst,
  Amherst, MA, USA, Tech. Rep} \bibinfo{volume}{14}, \bibinfo{pages}{1--5}.
\bibitem[{Huang and Wang(2018)}]{Huang2018a}
\bibinfo{author}{Huang, Z.}, \bibinfo{author}{Wang, N.}, \bibinfo{year}{2018}.
\newblock \bibinfo{title}{{Data-Driven Sparse Structure Selection for Deep
  Neural Networks}}, in: \bibinfo{booktitle}{Lecture Notes in Computer Science
  (including subseries Lecture Notes in Artificial Intelligence and Lecture
  Notes in Bioinformatics)}. volume \bibinfo{volume}{11220 LNCS}, pp.
  \bibinfo{pages}{317--334}.
\newblock \URLprefix
  \url{http://link.springer.com/10.1007/978-3-030-01270-0_19},
  \DOIprefix\doi{10.1007/978-3-030-01270-0{\_}19}.
\bibitem[{Hubara et~al.(2016a)Hubara, Courbariaux, Soudry, El-Yaniv and
  Bengio}]{Hubara2016b}
\bibinfo{author}{Hubara, I.}, \bibinfo{author}{Courbariaux, M.},
  \bibinfo{author}{Soudry, D.}, \bibinfo{author}{El-Yaniv, R.},
  \bibinfo{author}{Bengio, Y.}, \bibinfo{year}{2016}a.
\newblock \bibinfo{title}{{Binarized Neural Networks}}, in:
  \bibinfo{booktitle}{Advances in Neural Information Processing Systems
  (NIPS)}, pp. \bibinfo{pages}{4114--4122}.
\newblock \URLprefix
  \url{http://papers.nips.cc/paper/6573-binarized-neural-networks}.
\bibitem[{Hubara et~al.(2016b)Hubara, Courbariaux, Soudry, El-Yaniv and
  Bengio}]{Hubara2016a}
\bibinfo{author}{Hubara, I.}, \bibinfo{author}{Courbariaux, M.},
  \bibinfo{author}{Soudry, D.}, \bibinfo{author}{El-Yaniv, R.},
  \bibinfo{author}{Bengio, Y.}, \bibinfo{year}{2016}b.
\newblock \bibinfo{title}{{Quantized Neural Networks: Training Neural Networks
  with Low Precision Weights and Activations}}.
\newblock \bibinfo{journal}{Journal of Machine Learning Research 18}
  \bibinfo{volume}{18}, \bibinfo{pages}{187--1}.
\newblock \URLprefix \url{http://arxiv.org/abs/1609.07061}.
\bibitem[{Hwang and Sung(2014)}]{Hwang2014}
\bibinfo{author}{Hwang, K.}, \bibinfo{author}{Sung, W.}, \bibinfo{year}{2014}.
\newblock \bibinfo{title}{{Fixed-point feedforward deep neural network design
  using weights +1, 0, and -1}}, in: \bibinfo{booktitle}{2014 IEEE Workshop on
  Signal Processing Systems (SiPS)}, \bibinfo{publisher}{IEEE}. pp.
  \bibinfo{pages}{1--6}.
\newblock \URLprefix
  \url{https://ieeexplore.ieee.org/abstract/document/6986082/},
  \DOIprefix\doi{10.1109/SiPS.2014.6986082}.
\bibitem[{Iandola et~al.(2016)Iandola, Han, Moskewicz, Ashraf, Dally and
  Keutzer}]{Iandola2016}
\bibinfo{author}{Iandola, F.N.}, \bibinfo{author}{Han, S.},
  \bibinfo{author}{Moskewicz, M.W.}, \bibinfo{author}{Ashraf, K.},
  \bibinfo{author}{Dally, W.J.}, \bibinfo{author}{Keutzer, K.},
  \bibinfo{year}{2016}.
\newblock \bibinfo{title}{{SqueezeNet: AlexNet-level accuracy with 50x fewer
  parameters and <0.5MB model size}}, in: \bibinfo{booktitle}{ArXiv e-prints}.
\newblock \URLprefix \url{https://arxiv.org/abs/1602.07360
  http://arxiv.org/abs/1602.07360}, \DOIprefix\doi{10.1007/978-3-319-24553-9}.
\bibitem[{Ignatov et~al.(2019)Ignatov, Timofte, Kulik, Yang, Wang, Baum, Wu, Xu
  and Van~Gool}]{Ignatov2019}
\bibinfo{author}{Ignatov, A.}, \bibinfo{author}{Timofte, R.},
  \bibinfo{author}{Kulik, A.}, \bibinfo{author}{Yang, S.},
  \bibinfo{author}{Wang, K.}, \bibinfo{author}{Baum, F.}, \bibinfo{author}{Wu,
  M.}, \bibinfo{author}{Xu, L.}, \bibinfo{author}{Van~Gool, L.},
  \bibinfo{year}{2019}.
\newblock \bibinfo{title}{{AI benchmark: All about deep learning on smartphones
  in 2019}}.
\newblock \bibinfo{journal}{Proceedings - 2019 International Conference on
  Computer Vision Workshop, ICCVW 2019} , \bibinfo{pages}{3617--3635}\URLprefix
  \url{https://developer.arm.com/documentation/ddi0487/latest},
  \DOIprefix\doi{10.1109/ICCVW.2019.00447}.
\bibitem[{{Imagination}()}]{Imagination}
\bibinfo{author}{{Imagination}}, .
\newblock \bibinfo{title}{{PowerVR - embedded graphics processors powering
  iconic products}}.
\newblock \URLprefix \url{https://www.imgtec.com/graphics-processors/}.
\bibitem[{{Intel}()}]{Intel}
\bibinfo{author}{{Intel}}, .
\newblock \bibinfo{title}{{OpenVINO™ Toolkit}}.
\newblock \URLprefix \url{https://docs.openvinotoolkit.org/latest/index.html}.
\bibitem[{Ioffe and Szegedy(2015)}]{Ioffe2015}
\bibinfo{author}{Ioffe, S.}, \bibinfo{author}{Szegedy, C.},
  \bibinfo{year}{2015}.
\newblock \bibinfo{title}{{Batch normalization: Accelerating deep network
  training by reducing internal covariate shift}}, in:
  \bibinfo{booktitle}{International Conference on Machine Learning (ICML)}, pp.
  \bibinfo{pages}{448--456}.
\newblock \URLprefix \url{http://arxiv.org/abs/1502.03167}.
\bibitem[{Jacob et~al.(2018)Jacob, Kligys, Chen, Zhu, Tang, Howard, Adam and
  Kalenichenko}]{Jacob2017}
\bibinfo{author}{Jacob, B.}, \bibinfo{author}{Kligys, S.},
  \bibinfo{author}{Chen, B.}, \bibinfo{author}{Zhu, M.}, \bibinfo{author}{Tang,
  M.}, \bibinfo{author}{Howard, A.}, \bibinfo{author}{Adam, H.},
  \bibinfo{author}{Kalenichenko, D.}, \bibinfo{year}{2018}.
\newblock \bibinfo{title}{{Quantization and Training of Neural Networks for
  Efficient Integer-Arithmetic-Only Inference}}, in:
  \bibinfo{booktitle}{IEEE/CVF Conference on Computer Vision and Pattern
  Recognition (CVPR)}, \bibinfo{publisher}{IEEE}. pp.
  \bibinfo{pages}{2704--2713}.
\newblock \URLprefix \url{https://ieeexplore.ieee.org/document/8578384/},
  \DOIprefix\doi{10.1109/CVPR.2018.00286}.
\bibitem[{Jia et~al.(2019)Jia, Tillman, Maggioni and Scarpazza}]{Jia2019}
\bibinfo{author}{Jia, Z.}, \bibinfo{author}{Tillman, B.},
  \bibinfo{author}{Maggioni, M.}, \bibinfo{author}{Scarpazza, D.P.},
  \bibinfo{year}{2019}.
\newblock \bibinfo{title}{{Dissecting the graphcore IPU architecture via
  microbenchmarking}}.
\newblock \bibinfo{journal}{ArXiv preprint} .
\bibitem[{{Jia Deng} et~al.(2009){Jia Deng}, {Wei Dong}, Socher, {Li-Jia Li},
  {Kai Li} and {Li Fei-Fei}}]{JiaDeng2009}
\bibinfo{author}{{Jia Deng}}, \bibinfo{author}{{Wei Dong}},
  \bibinfo{author}{Socher, R.}, \bibinfo{author}{{Li-Jia Li}},
  \bibinfo{author}{{Kai Li}}, \bibinfo{author}{{Li Fei-Fei}},
  \bibinfo{year}{2009}.
\newblock \bibinfo{title}{{ImageNet: A large-scale hierarchical image
  database}}.
\newblock \bibinfo{journal}{IEEE/CVF Conference on Computer Vision and Pattern
  Recognition (CVPR)} ,
  \bibinfo{pages}{248--255}\DOIprefix\doi{10.1109/cvprw.2009.5206848}.
\bibitem[{{Jianchang Mao} et~al.(1994){Jianchang Mao}, Mohiuddin and
  Jain}]{JianchangMao1994}
\bibinfo{author}{{Jianchang Mao}}, \bibinfo{author}{Mohiuddin, K.},
  \bibinfo{author}{Jain, A.}, \bibinfo{year}{1994}.
\newblock \bibinfo{title}{{Parsimonious network design and feature selection
  through node pruning}}, in: \bibinfo{booktitle}{Proceedings of the 12th IAPR
  International Conference on Pattern Recognition, Vol. 3 - Conference C:
  Signal Processing (Cat. No.94CH3440-5)}, \bibinfo{publisher}{IEEE Comput.
  Soc. Press}. pp. \bibinfo{pages}{622--624}.
\newblock \URLprefix \url{http://ieeexplore.ieee.org/document/577060/},
  \DOIprefix\doi{10.1109/icpr.1994.577060}.
\bibitem[{Jiao et~al.(2020)Jiao, Han and Long}]{Jiao2020}
\bibinfo{author}{Jiao, Y.}, \bibinfo{author}{Han, L.}, \bibinfo{author}{Long,
  X.}, \bibinfo{year}{2020}.
\newblock \bibinfo{title}{{Hanguang 800 NPU – The Ultimate AI Inference
  Solution for Data Centers}}, in: \bibinfo{booktitle}{2020 IEEE Hot Chips 32
  Symposium (HCS)}, \bibinfo{publisher}{IEEE}. pp. \bibinfo{pages}{1--29}.
\newblock \URLprefix \url{https://ieeexplore.ieee.org/document/9220619/},
  \DOIprefix\doi{10.1109/HCS49909.2020.9220619}.
\bibitem[{Jouppi et~al.(2017)Jouppi, Borchers, Boyle, Cantin, Chao, Clark,
  Coriell, Daley, Dau, Dean, Gelb, Young, Ghaemmaghami, Gottipati, Gulland,
  Hagmann, Ho, Hogberg, Hu, Hundt, Hurt, Ibarz, Patil, Jaffey, Jaworski,
  Kaplan, Khaitan, Killebrew, Koch, Kumar, Lacy, Laudon, Law, Patterson, Le,
  Leary, Liu, Lucke, Lundin, MacKean, Maggiore, Mahony, Miller, Nagarajan,
  Agrawal, Narayanaswami, Ni, Nix, Norrie, Omernick, Penukonda, Phelps, Ross,
  Ross, Salek, Bajwa, Samadiani, Severn, Sizikov, Snelham, Souter, Steinberg,
  Swing, Tan, Thorson, Tian, Bates, Toma, Tuttle, Vasudevan, Walter, Wang,
  Wilcox, Yoon, Bhatia and Boden}]{Jouppi2017}
\bibinfo{author}{Jouppi, N.P.}, \bibinfo{author}{Borchers, A.},
  \bibinfo{author}{Boyle, R.}, \bibinfo{author}{Cantin, P.l.},
  \bibinfo{author}{Chao, C.}, \bibinfo{author}{Clark, C.},
  \bibinfo{author}{Coriell, J.}, \bibinfo{author}{Daley, M.},
  \bibinfo{author}{Dau, M.}, \bibinfo{author}{Dean, J.}, \bibinfo{author}{Gelb,
  B.}, \bibinfo{author}{Young, C.}, \bibinfo{author}{Ghaemmaghami, T.V.},
  \bibinfo{author}{Gottipati, R.}, \bibinfo{author}{Gulland, W.},
  \bibinfo{author}{Hagmann, R.}, \bibinfo{author}{Ho, C.R.},
  \bibinfo{author}{Hogberg, D.}, \bibinfo{author}{Hu, J.},
  \bibinfo{author}{Hundt, R.}, \bibinfo{author}{Hurt, D.},
  \bibinfo{author}{Ibarz, J.}, \bibinfo{author}{Patil, N.},
  \bibinfo{author}{Jaffey, A.}, \bibinfo{author}{Jaworski, A.},
  \bibinfo{author}{Kaplan, A.}, \bibinfo{author}{Khaitan, H.},
  \bibinfo{author}{Killebrew, D.}, \bibinfo{author}{Koch, A.},
  \bibinfo{author}{Kumar, N.}, \bibinfo{author}{Lacy, S.},
  \bibinfo{author}{Laudon, J.}, \bibinfo{author}{Law, J.},
  \bibinfo{author}{Patterson, D.}, \bibinfo{author}{Le, D.},
  \bibinfo{author}{Leary, C.}, \bibinfo{author}{Liu, Z.},
  \bibinfo{author}{Lucke, K.}, \bibinfo{author}{Lundin, A.},
  \bibinfo{author}{MacKean, G.}, \bibinfo{author}{Maggiore, A.},
  \bibinfo{author}{Mahony, M.}, \bibinfo{author}{Miller, K.},
  \bibinfo{author}{Nagarajan, R.}, \bibinfo{author}{Agrawal, G.},
  \bibinfo{author}{Narayanaswami, R.}, \bibinfo{author}{Ni, R.},
  \bibinfo{author}{Nix, K.}, \bibinfo{author}{Norrie, T.},
  \bibinfo{author}{Omernick, M.}, \bibinfo{author}{Penukonda, N.},
  \bibinfo{author}{Phelps, A.}, \bibinfo{author}{Ross, J.},
  \bibinfo{author}{Ross, M.}, \bibinfo{author}{Salek, A.},
  \bibinfo{author}{Bajwa, R.}, \bibinfo{author}{Samadiani, E.},
  \bibinfo{author}{Severn, C.}, \bibinfo{author}{Sizikov, G.},
  \bibinfo{author}{Snelham, M.}, \bibinfo{author}{Souter, J.},
  \bibinfo{author}{Steinberg, D.}, \bibinfo{author}{Swing, A.},
  \bibinfo{author}{Tan, M.}, \bibinfo{author}{Thorson, G.},
  \bibinfo{author}{Tian, B.}, \bibinfo{author}{Bates, S.},
  \bibinfo{author}{Toma, H.}, \bibinfo{author}{Tuttle, E.},
  \bibinfo{author}{Vasudevan, V.}, \bibinfo{author}{Walter, R.},
  \bibinfo{author}{Wang, W.}, \bibinfo{author}{Wilcox, E.},
  \bibinfo{author}{Yoon, D.H.}, \bibinfo{author}{Bhatia, S.},
  \bibinfo{author}{Boden, N.}, \bibinfo{year}{2017}.
\newblock \bibinfo{title}{{In-Datacenter Performance Analysis of a Tensor
  Processing Unit}}.
\newblock \bibinfo{journal}{ACM SIGARCH Computer Architecture News}
  \bibinfo{volume}{45}, \bibinfo{pages}{1--12}.
\newblock \URLprefix \url{http://dl.acm.org/citation.cfm?doid=3140659.3080246},
  \DOIprefix\doi{10.1145/3140659.3080246}.
\bibitem[{Judd et~al.(2017)Judd, Delmas, Sharify and Moshovos}]{Judd2017}
\bibinfo{author}{Judd, P.}, \bibinfo{author}{Delmas, A.},
  \bibinfo{author}{Sharify, S.}, \bibinfo{author}{Moshovos, A.},
  \bibinfo{year}{2017}.
\newblock \bibinfo{title}{{Cnvlutin2: Ineffectual-Activation-and-Weight-Free
  Deep Neural Network Computing}}.
\newblock \bibinfo{journal}{ArXiv preprint} , \bibinfo{pages}{1--6}\URLprefix
  \url{https://arxiv.org/abs/1705.00125}.
\bibitem[{Jung et~al.(2018)Jung, Son, Lee, Son, Kwak, Han, Hwang and
  Choi}]{Jung2018}
\bibinfo{author}{Jung, S.}, \bibinfo{author}{Son, C.}, \bibinfo{author}{Lee,
  S.}, \bibinfo{author}{Son, J.}, \bibinfo{author}{Kwak, Y.},
  \bibinfo{author}{Han, J.J.}, \bibinfo{author}{Hwang, S.J.},
  \bibinfo{author}{Choi, C.}, \bibinfo{year}{2018}.
\newblock \bibinfo{title}{{Learning to Quantize Deep Networks by Optimizing
  Quantization Intervals with Task Loss}}.
\newblock \bibinfo{journal}{Revue Internationale de la Croix-Rouge et Bulletin
  international des Soci{\'{e}}t{\'{e}}s de la Croix-Rouge} \URLprefix
  \url{http://arxiv.org/abs/1808.05779}, \DOIprefix\doi{arXiv:1808.05779v2}.
\bibitem[{Kathail(2020)}]{Kathail2020}
\bibinfo{author}{Kathail, V.}, \bibinfo{year}{2020}.
\newblock \bibinfo{title}{{Xilinx Vitis Unified Software Platform}}, in:
  \bibinfo{booktitle}{Proceedings of the 2020 ACM/SIGDA International Symposium
  on Field-Programmable Gate Arrays}, \bibinfo{publisher}{ACM},
  \bibinfo{address}{New York, NY, USA}. pp. \bibinfo{pages}{173--174}.
\newblock \URLprefix \url{https://dl.acm.org/doi/10.1145/3373087.3375887},
  \DOIprefix\doi{10.1145/3373087.3375887}.
\bibitem[{{Keil}(2018)}]{Keil2018}
\bibinfo{author}{{Keil}}, \bibinfo{year}{2018}.
\newblock \bibinfo{title}{{CMSIS NN Software Library}}.
\newblock \URLprefix
  \url{https://arm-software.github.io/CMSIS_5/NN/html/index.html}.
\bibitem[{K{\"{o}}ster et~al.(2017)K{\"{o}}ster, Webb, Wang, Nassar, Bansal,
  Constable, Elibol, Gray, Hall, Hornof, Khosrowshahi, Kloss, Pai and
  Rao}]{Koster2017}
\bibinfo{author}{K{\"{o}}ster, U.}, \bibinfo{author}{Webb, T.J.},
  \bibinfo{author}{Wang, X.}, \bibinfo{author}{Nassar, M.},
  \bibinfo{author}{Bansal, A.K.}, \bibinfo{author}{Constable, W.H.},
  \bibinfo{author}{Elibol, O.H.}, \bibinfo{author}{Gray, S.},
  \bibinfo{author}{Hall, S.}, \bibinfo{author}{Hornof, L.},
  \bibinfo{author}{Khosrowshahi, A.}, \bibinfo{author}{Kloss, C.},
  \bibinfo{author}{Pai, R.J.}, \bibinfo{author}{Rao, N.}, \bibinfo{year}{2017}.
\newblock \bibinfo{title}{{Flexpoint: An Adaptive Numerical Format for
  Efficient Training of Deep Neural Networks}}.
\newblock \bibinfo{journal}{ArXiv preprint} \URLprefix
  \url{http://arxiv.org/abs/1711.02213}.
\bibitem[{Krishnamoorthi(2018)}]{Krishnamoorthi2018}
\bibinfo{author}{Krishnamoorthi, R.}, \bibinfo{year}{2018}.
\newblock \bibinfo{title}{{Quantizing deep convolutional networks for efficient
  inference: A whitepaper}}.
\newblock \bibinfo{journal}{ArXiv preprint} \bibinfo{volume}{8},
  \bibinfo{pages}{667--668}.
\newblock \URLprefix \url{http://cn.arxiv.org/pdf/1806.08342.pdf
  http://arxiv.org/abs/1806.08342}, \DOIprefix\doi{arXiv:1806.08342v1}.
\bibitem[{Krizhevsky(2009)}]{Krizhevsky2009}
\bibinfo{author}{Krizhevsky, A.}, \bibinfo{year}{2009}.
\newblock \bibinfo{title}{{Learning Multiple Layers of Features from Tiny
  Images}}.
\newblock \bibinfo{journal}{Science Department, University of Toronto, Tech.}
  \DOIprefix\doi{10.1.1.222.9220}.
\bibitem[{Krizhevsky et~al.(2012)Krizhevsky, Sutskever and
  Hinton}]{Krizhevsky2012}
\bibinfo{author}{Krizhevsky, A.}, \bibinfo{author}{Sutskever, I.},
  \bibinfo{author}{Hinton, G.E.}, \bibinfo{year}{2012}.
\newblock \bibinfo{title}{{ImageNet Classification with Deep Convolutional
  Neural Networks}}, in: \bibinfo{booktitle}{Advances in Neural Information
  Processing Systems (NIPS)}, pp. \bibinfo{pages}{1--9}.
\newblock \URLprefix \url{http://code.google.com/p/cuda-convnet/},
  \DOIprefix\doi{http://dx.doi.org/10.1016/j.protcy.2014.09.007}.
\bibitem[{Lattner et~al.(2020)Lattner, Amini, Bondhugula, Cohen, Davis,
  Pienaar, Riddle, Shpeisman, Vasilache and Zinenko}]{Lattner2020}
\bibinfo{author}{Lattner, C.}, \bibinfo{author}{Amini, M.},
  \bibinfo{author}{Bondhugula, U.}, \bibinfo{author}{Cohen, A.},
  \bibinfo{author}{Davis, A.}, \bibinfo{author}{Pienaar, J.},
  \bibinfo{author}{Riddle, R.}, \bibinfo{author}{Shpeisman, T.},
  \bibinfo{author}{Vasilache, N.}, \bibinfo{author}{Zinenko, O.},
  \bibinfo{year}{2020}.
\newblock \bibinfo{title}{{MLIR: A Compiler Infrastructure for the End of
  Moore's Law}}.
\newblock \bibinfo{journal}{ArXiv preprint} \URLprefix
  \url{http://arxiv.org/abs/2002.11054}.
\bibitem[{Lavin and Gray(2016)}]{Lavin2016}
\bibinfo{author}{Lavin, A.}, \bibinfo{author}{Gray, S.}, \bibinfo{year}{2016}.
\newblock \bibinfo{title}{{Fast Algorithms for Convolutional Neural Networks}},
  in: \bibinfo{booktitle}{IEEE/CVF Conference on Computer Vision and Pattern
  Recognition (CVPR)}, \bibinfo{publisher}{IEEE}. pp.
  \bibinfo{pages}{4013--4021}.
\newblock \URLprefix \url{http://ieeexplore.ieee.org/document/7780804/
  http://arxiv.org/abs/1312.5851}, \DOIprefix\doi{10.1109/CVPR.2016.435}.
\bibitem[{Lebedev and Lempitsky(2016)}]{Lebedev2016}
\bibinfo{author}{Lebedev, V.}, \bibinfo{author}{Lempitsky, V.},
  \bibinfo{year}{2016}.
\newblock \bibinfo{title}{{Fast ConvNets Using Group-Wise Brain Damage}}, in:
  \bibinfo{booktitle}{IEEE/CVF Conference on Computer Vision and Pattern
  Recognition (CVPR)}, \bibinfo{publisher}{IEEE}. pp.
  \bibinfo{pages}{2554--2564}.
\newblock \URLprefix
  \url{http://openaccess.thecvf.com/content_cvpr_2016/html/Lebedev_Fast_ConvNets_Using_CVPR_2016_paper.html
  http://ieeexplore.ieee.org/document/7780649/},
  \DOIprefix\doi{10.1109/CVPR.2016.280}.
\bibitem[{Lebedev and Lempitsky(2018)}]{Lebedev2018}
\bibinfo{author}{Lebedev, V.}, \bibinfo{author}{Lempitsky, V.},
  \bibinfo{year}{2018}.
\newblock \bibinfo{title}{{Speeding-up convolutional neural networks: A
  survey}}.
\newblock \bibinfo{journal}{BULLETIN OF THE POLISH ACADEMY OF SCIENCES
  TECHNICAL SCIENCES} \bibinfo{volume}{66}, \bibinfo{pages}{2018}.
\newblock \URLprefix
  \url{http://www.czasopisma.pan.pl/Content/109869/PDF/05_799-810_00925_Bpast.No.66-6_31.12.18_K2.pdf?handler=pdf
  http://www.czasopisma.pan.pl/Content/109869/PDF/05_799-810_00925_Bpast.No.66-6_31.12.18_K2.pdf},
  \DOIprefix\doi{10.24425/bpas.2018.125927}.
\bibitem[{Lecun et~al.(2015)Lecun, Bengio and Hinton}]{Lecun2015}
\bibinfo{author}{Lecun, Y.}, \bibinfo{author}{Bengio, Y.},
  \bibinfo{author}{Hinton, G.}, \bibinfo{year}{2015}.
\newblock \bibinfo{title}{{Deep learning}}.
\newblock \bibinfo{journal}{Nature} \bibinfo{volume}{521},
  \bibinfo{pages}{436--444}.
\newblock \DOIprefix\doi{10.1038/nature14539}.
\bibitem[{LeCun et~al.(1998)LeCun, Bottou, Bengio and Haffner}]{LeCun1998}
\bibinfo{author}{LeCun, Y.}, \bibinfo{author}{Bottou, L.},
  \bibinfo{author}{Bengio, Y.}, \bibinfo{author}{Haffner, P.},
  \bibinfo{year}{1998}.
\newblock \bibinfo{title}{{Gradient-based learning applied to document
  recognition}}.
\newblock \bibinfo{journal}{Proceedings of the IEEE} \bibinfo{volume}{86},
  \bibinfo{pages}{2278--2323}.
\newblock \URLprefix \url{http://ieeexplore.ieee.org/document/726791/},
  \DOIprefix\doi{10.1109/5.726791}.
\bibitem[{LeCun et~al.(1990)LeCun, Denker and Solla}]{LeCun1990}
\bibinfo{author}{LeCun, Y.}, \bibinfo{author}{Denker, J.S.},
  \bibinfo{author}{Solla, S.A.}, \bibinfo{year}{1990}.
\newblock \bibinfo{title}{{Optimal Brain Damage}}, in:
  \bibinfo{booktitle}{Advances in Neural Information Processing Systems
  (NIPS)}, p. \bibinfo{pages}{598–605}.
\newblock \DOIprefix\doi{10.5555/109230.109298}.
\bibitem[{Lee et~al.(2019)Lee, Ajanthan and Torr}]{Lee2019a}
\bibinfo{author}{Lee, N.}, \bibinfo{author}{Ajanthan, T.},
  \bibinfo{author}{Torr, P.H.}, \bibinfo{year}{2019}.
\newblock \bibinfo{title}{{SnIP: Single-shot network pruning based on
  connection sensitivity}}, in: \bibinfo{booktitle}{International Conference on
  Learning Representations(ICLR)}.
\bibitem[{Lei et~al.(2018)Lei, Gao, Song, Wang and Song}]{Lei2018}
\bibinfo{author}{Lei, J.}, \bibinfo{author}{Gao, X.}, \bibinfo{author}{Song,
  J.}, \bibinfo{author}{Wang, X.L.}, \bibinfo{author}{Song, M.L.},
  \bibinfo{year}{2018}.
\newblock \bibinfo{title}{{Survey of Deep Neural Network Model Compression}}.
\newblock \bibinfo{journal}{Ruan Jian Xue Bao/Journal of Software}
  \bibinfo{volume}{29}, \bibinfo{pages}{251--266}.
\newblock \URLprefix
  \url{https://www.scopus.com/inward/record.uri?eid=2-s2.0-85049464636&doi=10.13328%2Fj.cnki.jos.005428&partnerID=40&md5=5a79dfdff4a05f188c5d553fb3b3123a},
  \DOIprefix\doi{10.13328/j.cnki.jos.005428}.
\bibitem[{Lei et~al.(2017)Lei, Chen and Wu}]{Lei2017}
\bibinfo{author}{Lei, W.}, \bibinfo{author}{Chen, H.}, \bibinfo{author}{Wu,
  Y.}, \bibinfo{year}{2017}.
\newblock \bibinfo{title}{{Compressing Deep Convolutional Networks Using
  K-means Based on Weights Distribution}}, in: \bibinfo{booktitle}{Proceedings
  of the 2nd International Conference on Intelligent Information Processing -
  IIP'17}, \bibinfo{publisher}{ACM Press}, \bibinfo{address}{New York, New
  York, USA}. pp. \bibinfo{pages}{1--6}.
\newblock \URLprefix \url{http://dl.acm.org/citation.cfm?doid=3144789.3144803},
  \DOIprefix\doi{10.1145/3144789.3144803}.
\bibitem[{Leng et~al.(2018)Leng, Li, Zhu and Jin}]{Leng2017}
\bibinfo{author}{Leng, C.}, \bibinfo{author}{Li, H.}, \bibinfo{author}{Zhu,
  S.}, \bibinfo{author}{Jin, R.}, \bibinfo{year}{2018}.
\newblock \bibinfo{title}{{Extremely Low Bit Neural Network: Squeeze the Last
  Bit Out with ADMM}}.
\newblock \bibinfo{journal}{The Thirty-Second AAAI Conference on Artificial
  Intelligence (AAAI-18)} \URLprefix \url{http://arxiv.org/abs/1707.09870}.
\bibitem[{Leroux et~al.(2017)Leroux, Bohez, De~Coninck, Verbelen,
  Vankeirsbilck, Simoens and Dhoedt}]{Leroux2017}
\bibinfo{author}{Leroux, S.}, \bibinfo{author}{Bohez, S.},
  \bibinfo{author}{De~Coninck, E.}, \bibinfo{author}{Verbelen, T.},
  \bibinfo{author}{Vankeirsbilck, B.}, \bibinfo{author}{Simoens, P.},
  \bibinfo{author}{Dhoedt, B.}, \bibinfo{year}{2017}.
\newblock \bibinfo{title}{{The cascading neural network: building the Internet
  of Smart Things}}.
\newblock \bibinfo{journal}{Knowledge and Information Systems}
  \bibinfo{volume}{52}, \bibinfo{pages}{791--814}.
\newblock \URLprefix \url{http://link.springer.com/10.1007/s10115-017-1029-1},
  \DOIprefix\doi{10.1007/s10115-017-1029-1}.
\bibitem[{Li et~al.(2016)Li, Zhang and Liu}]{Li2016}
\bibinfo{author}{Li, F.}, \bibinfo{author}{Zhang, B.}, \bibinfo{author}{Liu,
  B.}, \bibinfo{year}{2016}.
\newblock \bibinfo{title}{{Ternary Weight Networks}}, in:
  \bibinfo{booktitle}{Advances in Neural Information Processing Systems
  (NIPS)}.
\newblock \URLprefix \url{http://arxiv.org/abs/1605.04711}.
\bibitem[{Li et~al.(2017a)Li, Kadav, Durdanovic, Samet and Graf}]{Li2017}
\bibinfo{author}{Li, H.}, \bibinfo{author}{Kadav, A.},
  \bibinfo{author}{Durdanovic, I.}, \bibinfo{author}{Samet, H.},
  \bibinfo{author}{Graf, H.P.}, \bibinfo{year}{2017}a.
\newblock \bibinfo{title}{{Pruning Filters for Efficient ConvNets}}, in:
  \bibinfo{booktitle}{International Conference on Learning Representations
  (ICLR)}.
\newblock \URLprefix \url{http://arxiv.org/abs/1608.08710},
  \DOIprefix\doi{10.1029/2009GL038531}.
\bibitem[{Li et~al.(2019)Li, Zhang, Qi, Ruigang and Huang}]{Li2019}
\bibinfo{author}{Li, H.}, \bibinfo{author}{Zhang, H.}, \bibinfo{author}{Qi,
  X.}, \bibinfo{author}{Ruigang, Y.}, \bibinfo{author}{Huang, G.},
  \bibinfo{year}{2019}.
\newblock \bibinfo{title}{{Improved Techniques for Training Adaptive Deep
  Networks}}, in: \bibinfo{booktitle}{2019 IEEE/CVF International Conference on
  Computer Vision (ICCV)}, \bibinfo{publisher}{IEEE}. pp.
  \bibinfo{pages}{1891--1900}.
\newblock \URLprefix \url{https://ieeexplore.ieee.org/document/9010043/},
  \DOIprefix\doi{10.1109/ICCV.2019.00198}.
\bibitem[{Li et~al.(2020a)Li, Liu, Liu, Sun, You, Yang, Luan, Gan, Yang and
  Qian}]{Li2020}
\bibinfo{author}{Li, M.}, \bibinfo{author}{Liu, Y.I.}, \bibinfo{author}{Liu,
  X.}, \bibinfo{author}{Sun, Q.}, \bibinfo{author}{You, X.I.N.},
  \bibinfo{author}{Yang, H.}, \bibinfo{author}{Luan, Z.}, \bibinfo{author}{Gan,
  L.}, \bibinfo{author}{Yang, G.}, \bibinfo{author}{Qian, D.},
  \bibinfo{year}{2020}a.
\newblock \bibinfo{title}{{The Deep Learning Compiler: A Comprehensive
  Survey}}.
\newblock \bibinfo{journal}{ArXiv preprint} \bibinfo{volume}{1},
  \bibinfo{pages}{1--36}.
\newblock \URLprefix \url{http://arxiv.org/abs/2002.03794}.
\bibitem[{Li et~al.(2020b)Li, Gu, Mayer, Van~Gool and Timofte}]{Li2020b}
\bibinfo{author}{Li, Y.}, \bibinfo{author}{Gu, S.}, \bibinfo{author}{Mayer,
  C.}, \bibinfo{author}{Van~Gool, L.}, \bibinfo{author}{Timofte, R.},
  \bibinfo{year}{2020}b.
\newblock \bibinfo{title}{{Group Sparsity: The Hinge Between Filter Pruning and
  Decomposition for Network Compression}}, in: \bibinfo{booktitle}{2020
  IEEE/CVF Conference on Computer Vision and Pattern Recognition (CVPR)},
  \bibinfo{publisher}{IEEE}. pp. \bibinfo{pages}{8015--8024}.
\newblock \URLprefix \url{https://ieeexplore.ieee.org/document/9157445/},
  \DOIprefix\doi{10.1109/CVPR42600.2020.00804}.
\bibitem[{Li et~al.(2017b)Li, Wang, Zhi and Chen}]{Li2017a}
\bibinfo{author}{Li, Z.}, \bibinfo{author}{Wang, Y.}, \bibinfo{author}{Zhi,
  T.}, \bibinfo{author}{Chen, T.}, \bibinfo{year}{2017}b.
\newblock \bibinfo{title}{{A survey of neural network accelerators}}.
\newblock \bibinfo{journal}{Frontiers of Computer Science}
  \bibinfo{volume}{11}, \bibinfo{pages}{746--761}.
\newblock \URLprefix \url{http://link.springer.com/10.1007/s11704-016-6159-1},
  \DOIprefix\doi{10.1007/s11704-016-6159-1}.
\bibitem[{Li et~al.(2020c)Li, Zhang, Wang and Lai}]{Li2020a}
\bibinfo{author}{Li, Z.}, \bibinfo{author}{Zhang, Y.}, \bibinfo{author}{Wang,
  J.}, \bibinfo{author}{Lai, J.}, \bibinfo{year}{2020}c.
\newblock \bibinfo{title}{{A survey of FPGA design for AI era}}.
\newblock \bibinfo{journal}{Journal of Semiconductors} \bibinfo{volume}{41}.
\newblock \DOIprefix\doi{10.1088/1674-4926/41/2/021402}.
\bibitem[{Lin et~al.(2017a)Lin, Rao, Lu and Zhou}]{Lin2017}
\bibinfo{author}{Lin, J.}, \bibinfo{author}{Rao, Y.}, \bibinfo{author}{Lu, J.},
  \bibinfo{author}{Zhou, J.}, \bibinfo{year}{2017}a.
\newblock \bibinfo{title}{{Runtime Neural Pruning}}, in:
  \bibinfo{booktitle}{Advances in Neural Information Processing Systems
  (NIPS)}, pp. \bibinfo{pages}{2178--2188}.
\newblock \URLprefix
  \url{https://papers.nips.cc/paper/6813-runtime-neural-pruning.pdf}.
\bibitem[{Lin et~al.(2014)Lin, Chen and Yan}]{Lin2014}
\bibinfo{author}{Lin, M.}, \bibinfo{author}{Chen, Q.}, \bibinfo{author}{Yan,
  S.}, \bibinfo{year}{2014}.
\newblock \bibinfo{title}{{Network in network}}, in:
  \bibinfo{booktitle}{International Conference on Learning
  Representations(ICLR)}, pp. \bibinfo{pages}{1--10}.
\bibitem[{Lin et~al.(2017b)Lin, Zhao and Pan}]{Lin2017a}
\bibinfo{author}{Lin, X.}, \bibinfo{author}{Zhao, C.}, \bibinfo{author}{Pan,
  W.}, \bibinfo{year}{2017}b.
\newblock \bibinfo{title}{{Towards accurate binary convolutional neural
  network}}, in: \bibinfo{booktitle}{Advances in Neural Information Processing
  Systems (NIPS)}, pp. \bibinfo{pages}{345--353}.
\bibitem[{Lin et~al.(2016)Lin, Courbariaux, Memisevic and Bengio}]{Lin2015}
\bibinfo{author}{Lin, Z.}, \bibinfo{author}{Courbariaux, M.},
  \bibinfo{author}{Memisevic, R.}, \bibinfo{author}{Bengio, Y.},
  \bibinfo{year}{2016}.
\newblock \bibinfo{title}{{Neural Networks with Few Multiplications}}, in:
  \bibinfo{booktitle}{International Conference on Learning
  Representations(ICLR)}.
\newblock \URLprefix \url{https://github.com/hantek/
  http://arxiv.org/abs/1510.03009 https://arxiv.org/abs/1510.03009}.
\bibitem[{Liu et~al.(2013)Liu, Musialski, Wonka and Ye}]{Liu2013}
\bibinfo{author}{Liu, J.}, \bibinfo{author}{Musialski, P.},
  \bibinfo{author}{Wonka, P.}, \bibinfo{author}{Ye, J.}, \bibinfo{year}{2013}.
\newblock \bibinfo{title}{{Tensor Completion for Estimating Missing Values in
  Visual Data}}.
\newblock \bibinfo{journal}{IEEE Transactions on Pattern Analysis and Machine
  Intelligence} \bibinfo{volume}{35}, \bibinfo{pages}{208--220}.
\newblock \URLprefix \url{http://ieeexplore.ieee.org/document/6138863/},
  \DOIprefix\doi{10.1109/TPAMI.2012.39}.
\bibitem[{Liu et~al.(2017)Liu, Li, Shen, Huang, Yan and Zhang}]{Liu2017a}
\bibinfo{author}{Liu, Z.}, \bibinfo{author}{Li, J.}, \bibinfo{author}{Shen,
  Z.}, \bibinfo{author}{Huang, G.}, \bibinfo{author}{Yan, S.},
  \bibinfo{author}{Zhang, C.}, \bibinfo{year}{2017}.
\newblock \bibinfo{title}{{Learning Efficient Convolutional Networks through
  Network Slimming}}, in: \bibinfo{booktitle}{IEEE International Conference on
  Computer Vision (ICCV)}, \bibinfo{publisher}{IEEE}. pp.
  \bibinfo{pages}{2755--2763}.
\newblock \URLprefix \url{http://ieeexplore.ieee.org/document/8237560/},
  \DOIprefix\doi{10.1109/ICCV.2017.298}.
\bibitem[{Liu et~al.(2019a)Liu, Mu, Zhang, Guo, Yang, Cheng and Sun}]{Liu2019}
\bibinfo{author}{Liu, Z.}, \bibinfo{author}{Mu, H.}, \bibinfo{author}{Zhang,
  X.}, \bibinfo{author}{Guo, Z.}, \bibinfo{author}{Yang, X.},
  \bibinfo{author}{Cheng, T.K.T.}, \bibinfo{author}{Sun, J.},
  \bibinfo{year}{2019}a.
\newblock \bibinfo{title}{{MetaPruning: Meta Learning for Automatic Neural
  Network Channel Pruning}}, in: \bibinfo{booktitle}{IEEE International
  Conference on Computer Vision}.
\newblock \URLprefix \url{http://arxiv.org/abs/1903.10258}.
\bibitem[{Liu et~al.(2019b)Liu, Sun, Zhou, Huang and
  Darrell}]{Liu2019Rethingking}
\bibinfo{author}{Liu, Z.}, \bibinfo{author}{Sun, M.}, \bibinfo{author}{Zhou,
  T.}, \bibinfo{author}{Huang, G.}, \bibinfo{author}{Darrell, T.},
  \bibinfo{year}{2019}b.
\newblock \bibinfo{title}{{Rethinking the Value of Network Pruning}}, in:
  \bibinfo{booktitle}{International Conference on Learning Representations
  (ICLR)}, pp. \bibinfo{pages}{1--11}.
\newblock \URLprefix \url{http://arxiv.org/abs/1810.05270}.
\bibitem[{Liu et~al.(2018)Liu, Wu, Luo, Yang, Liu and Cheng}]{Liu2018c}
\bibinfo{author}{Liu, Z.}, \bibinfo{author}{Wu, B.}, \bibinfo{author}{Luo, W.},
  \bibinfo{author}{Yang, X.}, \bibinfo{author}{Liu, W.},
  \bibinfo{author}{Cheng, K.T.}, \bibinfo{year}{2018}.
\newblock \bibinfo{title}{{Bi-Real Net: Enhancing the performance of 1-bit CNNs
  with improved representational capability and advanced training algorithm}}.
\newblock \bibinfo{journal}{Lecture Notes in Computer Science (including
  subseries Lecture Notes in Artificial Intelligence and Lecture Notes in
  Bioinformatics)} \bibinfo{volume}{11219 LNCS}, \bibinfo{pages}{747--763}.
\newblock \DOIprefix\doi{10.1007/978-3-030-01267-0{\_}44}.
\bibitem[{Liu and Mattina(2019)}]{Liu2019c}
\bibinfo{author}{Liu, Z.G.}, \bibinfo{author}{Mattina, M.},
  \bibinfo{year}{2019}.
\newblock \bibinfo{title}{{Learning low-precision neural networks without
  Straight-Through Estimator (STE)}}, in: \bibinfo{booktitle}{IJCAI
  International Joint Conference on Artificial Intelligence},
  \bibinfo{publisher}{International Joint Conferences on Artificial
  Intelligence Organization}, \bibinfo{address}{California}. pp.
  \bibinfo{pages}{3066--3072}.
\newblock \URLprefix \url{https://www.ijcai.org/proceedings/2019/425},
  \DOIprefix\doi{10.24963/ijcai.2019/425}.
\bibitem[{Luo and Wu(2020)}]{Luo2020}
\bibinfo{author}{Luo, J.H.}, \bibinfo{author}{Wu, J.}, \bibinfo{year}{2020}.
\newblock \bibinfo{title}{{AutoPruner: An end-to-end trainable filter pruning
  method for efficient deep model inference}}.
\newblock \bibinfo{journal}{Pattern Recognition} \bibinfo{volume}{107},
  \bibinfo{pages}{107461}.
\newblock \URLprefix
  \url{https://linkinghub.elsevier.com/retrieve/pii/S0031320320302648},
  \DOIprefix\doi{10.1016/j.patcog.2020.107461}.
\bibitem[{Luo et~al.(2017)Luo, Wu and Lin}]{Luo2017}
\bibinfo{author}{Luo, J.H.H.}, \bibinfo{author}{Wu, J.}, \bibinfo{author}{Lin,
  W.}, \bibinfo{year}{2017}.
\newblock \bibinfo{title}{{ThiNet: A Filter Level Pruning Method for Deep
  Neural Network Compression}}.
\newblock \bibinfo{journal}{Proceedings of the IEEE International Conference on
  Computer Vision (ICCV)} \bibinfo{volume}{2017-Octob},
  \bibinfo{pages}{5068--5076}.
\newblock \URLprefix \url{http://ieeexplore.ieee.org/document/8237803/},
  \DOIprefix\doi{10.1109/ICCV.2017.541}.
\bibitem[{Ma et~al.(2016)Ma, Suda, Cao, Seo and Vrudhula}]{Ma2016}
\bibinfo{author}{Ma, Y.}, \bibinfo{author}{Suda, N.}, \bibinfo{author}{Cao,
  Y.}, \bibinfo{author}{Seo, J.S.}, \bibinfo{author}{Vrudhula, S.},
  \bibinfo{year}{2016}.
\newblock \bibinfo{title}{{Scalable and modularized RTL compilation of
  Convolutional Neural Networks onto FPGA}}.
\newblock \bibinfo{journal}{FPL 2016 - 26th International Conference on
  Field-Programmable Logic and Applications}
  \DOIprefix\doi{10.1109/FPL.2016.7577356}.
\bibitem[{Macchi(1975)}]{Macchi1975}
\bibinfo{author}{Macchi, O.}, \bibinfo{year}{1975}.
\newblock \bibinfo{title}{{Coincidence Approach To Stochastic Point Process.}}
\newblock \bibinfo{journal}{Advances in Applied Probability}
  \bibinfo{volume}{7}, \bibinfo{pages}{83--122}.
\newblock \DOIprefix\doi{10.1017/s0001867800040313}.
\bibitem[{Mariet and Sra(2016)}]{Mariet2016}
\bibinfo{author}{Mariet, Z.}, \bibinfo{author}{Sra, S.}, \bibinfo{year}{2016}.
\newblock \bibinfo{title}{{Diversity Networks: Neural Network Compression Using
  Determinantal Point Processes}}, in: \bibinfo{booktitle}{International
  Conference on Learning Representations(ICLR)}, pp. \bibinfo{pages}{1--13}.
\newblock \URLprefix \url{http://arxiv.org/abs/1511.05077}.
\bibitem[{Mathieu et~al.(2013)Mathieu, Henaff and LeCun}]{Mathieu2013}
\bibinfo{author}{Mathieu, M.}, \bibinfo{author}{Henaff, M.},
  \bibinfo{author}{LeCun, Y.}, \bibinfo{year}{2013}.
\newblock \bibinfo{title}{{Fast Training of Convolutional Networks through
  FFTs}}.
\newblock \bibinfo{journal}{ArXiv preprint} \URLprefix
  \url{http://arxiv.org/abs/1312.5851}.
\bibitem[{Medina(2019)}]{Medina2019}
\bibinfo{author}{Medina, E.}, \bibinfo{year}{2019}.
\newblock \bibinfo{title}{{Habana Labs presentation}}.
\newblock \bibinfo{journal}{2019 IEEE Hot Chips 31 Symposium, HCS 2019}
  \DOIprefix\doi{10.1109/HOTCHIPS.2019.8875670}.
\bibitem[{Mellempudi et~al.(2017)Mellempudi, Kundu, Mudigere, Das, Kaul and
  Dubey}]{Mellempudi2017}
\bibinfo{author}{Mellempudi, N.}, \bibinfo{author}{Kundu, A.},
  \bibinfo{author}{Mudigere, D.}, \bibinfo{author}{Das, D.},
  \bibinfo{author}{Kaul, B.}, \bibinfo{author}{Dubey, P.},
  \bibinfo{year}{2017}.
\newblock \bibinfo{title}{{Ternary Neural Networks with Fine-Grained
  Quantization}}.
\newblock \bibinfo{journal}{ArXiv preprint} \URLprefix
  \url{http://arxiv.org/abs/1705.01462}.
\bibitem[{Merolla et~al.(2016)Merolla, Appuswamy, Arthur, Esser and
  Modha}]{Merolla2016}
\bibinfo{author}{Merolla, P.}, \bibinfo{author}{Appuswamy, R.},
  \bibinfo{author}{Arthur, J.}, \bibinfo{author}{Esser, S.K.},
  \bibinfo{author}{Modha, D.}, \bibinfo{year}{2016}.
\newblock \bibinfo{title}{{Deep neural networks are robust to weight
  binarization and other non-linear distortions}}.
\newblock \bibinfo{journal}{ArXiv preprint} \URLprefix
  \url{https://arxiv.org/abs/1606.01981 http://arxiv.org/abs/1606.01981}.
\bibitem[{Micikevicius et~al.(2017)Micikevicius, Narang, Alben, Diamos, Elsen,
  Garcia, Ginsburg, Houston, Kuchaiev, Venkatesh and Wu}]{Micikevicius2017}
\bibinfo{author}{Micikevicius, P.}, \bibinfo{author}{Narang, S.},
  \bibinfo{author}{Alben, J.}, \bibinfo{author}{Diamos, G.},
  \bibinfo{author}{Elsen, E.}, \bibinfo{author}{Garcia, D.},
  \bibinfo{author}{Ginsburg, B.}, \bibinfo{author}{Houston, M.},
  \bibinfo{author}{Kuchaiev, O.}, \bibinfo{author}{Venkatesh, G.},
  \bibinfo{author}{Wu, H.}, \bibinfo{year}{2017}.
\newblock \bibinfo{title}{{Mixed Precision Training}}, in:
  \bibinfo{booktitle}{International Conference on Learning
  Representations(ICLR)}.
\newblock \URLprefix \url{http://arxiv.org/abs/1710.03740}.
\bibitem[{Migacz(2017)}]{Migacz2017}
\bibinfo{author}{Migacz, S.}, \bibinfo{year}{2017}.
\newblock \bibinfo{title}{{8-bit inference with TensorRT}}.
\newblock \bibinfo{journal}{GPU Technology Conference} \bibinfo{volume}{2},
  \bibinfo{pages}{7}.
\newblock \URLprefix
  \url{https://on-demand.gputechconf.com/gtc/2017/presentation/s7310-8-bit-inference-with-tensorrt.pdf}.
\bibitem[{Mishra et~al.(2018)Mishra, Nurvitadhi, Cook and Marr}]{Mishra2018}
\bibinfo{author}{Mishra, A.}, \bibinfo{author}{Nurvitadhi, E.},
  \bibinfo{author}{Cook, J.J.}, \bibinfo{author}{Marr, D.},
  \bibinfo{year}{2018}.
\newblock \bibinfo{title}{{WRPN: Wide reduced-precision networks}}, in:
  \bibinfo{booktitle}{International Conference on Learning
  Representations(ICLR)}, pp. \bibinfo{pages}{1--11}.
\bibitem[{Miyashita et~al.(2016)Miyashita, Lee and Murmann}]{Miyashita2016}
\bibinfo{author}{Miyashita, D.}, \bibinfo{author}{Lee, E.H.},
  \bibinfo{author}{Murmann, B.}, \bibinfo{year}{2016}.
\newblock \bibinfo{title}{{Convolutional Neural Networks using Logarithmic Data
  Representation}}.
\newblock \bibinfo{journal}{ArXiv preprint} \URLprefix
  \url{http://cn.arxiv.org/pdf/1603.01025.pdf http://arxiv.org/abs/1603.01025}.
\bibitem[{Molchanov et~al.(2017)Molchanov, Ashukha and Vetrov}]{Molchanov2017a}
\bibinfo{author}{Molchanov, D.}, \bibinfo{author}{Ashukha, A.},
  \bibinfo{author}{Vetrov, D.}, \bibinfo{year}{2017}.
\newblock \bibinfo{title}{{Variational dropout sparsifies deep neural
  networks}}, in: \bibinfo{booktitle}{International Conference on Machine
  Learning (ICML)}, pp. \bibinfo{pages}{3854--3863}.
\newblock \URLprefix \url{https://dl.acm.org/citation.cfm?id=3305939}.
\bibitem[{Molchanov et~al.(2016)Molchanov, Tyree, Karras, Aila and
  Kautz}]{Molchanov2017}
\bibinfo{author}{Molchanov, P.}, \bibinfo{author}{Tyree, S.},
  \bibinfo{author}{Karras, T.}, \bibinfo{author}{Aila, T.},
  \bibinfo{author}{Kautz, J.}, \bibinfo{year}{2016}.
\newblock \bibinfo{title}{{Pruning Convolutional Neural Networks for Resource
  Efficient Inference}}, in: \bibinfo{booktitle}{International Conference on
  Learning Representations (ICLR)}, pp. \bibinfo{pages}{1--17}.
\newblock \URLprefix \url{http://arxiv.org/abs/1611.06440}.
\bibitem[{Moss et~al.(2017)Moss, Nurvitadhi, Sim, Mishra, Marr, Subhaschandra
  and Leong}]{Moss2017}
\bibinfo{author}{Moss, D.J.M.}, \bibinfo{author}{Nurvitadhi, E.},
  \bibinfo{author}{Sim, J.}, \bibinfo{author}{Mishra, A.},
  \bibinfo{author}{Marr, D.}, \bibinfo{author}{Subhaschandra, S.},
  \bibinfo{author}{Leong, P.H.W.}, \bibinfo{year}{2017}.
\newblock \bibinfo{title}{{High performance binary neural networks on the
  Xeon+FPGA™ platform}}, in: \bibinfo{booktitle}{2017 27th International
  Conference on Field Programmable Logic and Applications (FPL)},
  \bibinfo{publisher}{IEEE}. pp. \bibinfo{pages}{1--4}.
\newblock \URLprefix
  \url{https://ieeexplore.ieee.org/abstract/document/8056823/},
  \DOIprefix\doi{10.23919/FPL.2017.8056823}.
\bibitem[{Moudgill et~al.(2020)Moudgill, Glossner, Huang, Tian, Xu, Yang, Wang,
  Liang, Shi, Zhang, Iancu, Nacer and Li}]{Moudgill2020}
\bibinfo{author}{Moudgill, M.}, \bibinfo{author}{Glossner, J.},
  \bibinfo{author}{Huang, W.}, \bibinfo{author}{Tian, C.}, \bibinfo{author}{Xu,
  C.}, \bibinfo{author}{Yang, N.}, \bibinfo{author}{Wang, L.},
  \bibinfo{author}{Liang, T.}, \bibinfo{author}{Shi, S.},
  \bibinfo{author}{Zhang, X.}, \bibinfo{author}{Iancu, D.},
  \bibinfo{author}{Nacer, G.}, \bibinfo{author}{Li, K.}, \bibinfo{year}{2020}.
\newblock \bibinfo{title}{{Heterogeneous Edge CNN Hardware Accelerator}}, in:
  \bibinfo{booktitle}{The 12th International Conference on Wireless
  Communications and Signal Processing}, pp. \bibinfo{pages}{6--11}.
\bibitem[{Muller and Indiveri(2015)}]{Muller2015}
\bibinfo{author}{Muller, L.K.}, \bibinfo{author}{Indiveri, G.},
  \bibinfo{year}{2015}.
\newblock \bibinfo{title}{{Rounding Methods for Neural Networks with Low
  Resolution Synaptic Weights}}.
\newblock \bibinfo{journal}{ArXiv preprint} \URLprefix
  \url{http://arxiv.org/abs/1504.05767}.
\bibitem[{Muthukrishnan and Rohini(2016)}]{Muthukrishnan2016}
\bibinfo{author}{Muthukrishnan, R.}, \bibinfo{author}{Rohini, R.},
  \bibinfo{year}{2016}.
\newblock \bibinfo{title}{{LASSO: A feature selection technique in predictive
  modeling for machine learning}}, in: \bibinfo{booktitle}{2016 IEEE
  International Conference on Advances in Computer Applications (ICACA)},
  \bibinfo{publisher}{IEEE}. pp. \bibinfo{pages}{18--20}.
\newblock \URLprefix \url{http://ieeexplore.ieee.org/document/7887916/},
  \DOIprefix\doi{10.1109/ICACA.2016.7887916}.
\bibitem[{Neill(2020)}]{Neill2020}
\bibinfo{author}{Neill, J.O.}, \bibinfo{year}{2020}.
\newblock \bibinfo{title}{{An Overview of Neural Network Compression}}.
\newblock \bibinfo{journal}{ArXiv preprint} , \bibinfo{pages}{1--73}\URLprefix
  \url{http://arxiv.org/abs/2006.03669}.
\bibitem[{{NVIDIA Corporation}(2014)}]{NVIDIACorporation2014}
\bibinfo{author}{{NVIDIA Corporation}}, \bibinfo{year}{2014}.
\newblock \bibinfo{title}{{NVIDIA GeForce GTX 980 Featuring Maxwell, The Most
  Advanced GPU Ever Made.}}
\newblock \bibinfo{journal}{White Paper} , \bibinfo{pages}{1--32}\URLprefix
  \url{http://international.download.nvidia.com/geforce-com/international/pdfs/GeForce_GTX_980_Whitepaper_FINAL.PDF}.
\bibitem[{{NVIDIA Corporation}(2015)}]{NVIDIACorporation2015}
\bibinfo{author}{{NVIDIA Corporation}}, \bibinfo{year}{2015}.
\newblock \bibinfo{title}{{NVIDIA Tesla P100}}.
\newblock \bibinfo{journal}{White Paper} \URLprefix
  \url{https://www.nvidia.com/en-us/data-center/tesla-p100/}.
\bibitem[{{NVIDIA Corporation}(2017a)}]{NVIDIACorporation2017}
\bibinfo{author}{{NVIDIA Corporation}}, \bibinfo{year}{2017}a.
\newblock \bibinfo{title}{{NVIDIA DGX-1 With Tesla V100 System Architecture}}.
\newblock \bibinfo{journal}{White Paper} \URLprefix
  \url{http://images.nvidia.com/content/pdf/dgx1-v100-system-architecture-whitepaper.pdf}.
\bibitem[{{NVIDIA Corporation}(2017b)}]{NVIDIACorporation2017a}
\bibinfo{author}{{NVIDIA Corporation}}, \bibinfo{year}{2017}b.
\newblock \bibinfo{title}{{NVIDIA Tesla V100 GPU Volta Architecture}}.
\newblock \bibinfo{journal}{White Paper} , \bibinfo{pages}{53}\URLprefix
  \url{http://images.nvidia.com/content/volta-architecture/pdf/volta-architecture-whitepaper.pdf%0Ahttp://www.nvidia.com/content/gated-pdfs/Volta-Architecture-Whitepaper-v1.1.pdf}.
\bibitem[{{NVIDIA Corporation}(2018a)}]{NVIDIACorporation2018a}
\bibinfo{author}{{NVIDIA Corporation}}, \bibinfo{year}{2018}a.
\newblock \bibinfo{title}{{NVIDIA A100 Tensor Core GPU}}.
\newblock \bibinfo{journal}{White Paper} , \bibinfo{pages}{20--21}.
\bibitem[{{NVIDIA Corporation}(2018b)}]{NVIDIACorporation2018}
\bibinfo{author}{{NVIDIA Corporation}}, \bibinfo{year}{2018}b.
\newblock \bibinfo{title}{{NVIDIA Turing GPU Architecture}}.
\newblock \bibinfo{journal}{White Paper} \URLprefix
  \url{https://gpltech.com/wp-content/uploads/2018/11/NVIDIA-Turing-Architecture-Whitepaper.pdf}.
\bibitem[{Odena et~al.(2017)Odena, Lawson and Olah}]{Odena2017}
\bibinfo{author}{Odena, A.}, \bibinfo{author}{Lawson, D.},
  \bibinfo{author}{Olah, C.}, \bibinfo{year}{2017}.
\newblock \bibinfo{title}{{Changing Model Behavior at Test-Time Using
  Reinforcement Learning}}, in: \bibinfo{booktitle}{International Conference on
  Learning Representations Workshops (ICLRW)},
  \bibinfo{publisher}{International Conference on Learning Representations,
  ICLR}.
\newblock \URLprefix \url{http://arxiv.org/abs/1702.07780}.
\bibitem[{{ONNX}()}]{ONNX}
\bibinfo{author}{{ONNX}}, .
\newblock \bibinfo{title}{{onnx/onnx: Open standard for machine learning
  interoperability}}.
\newblock \URLprefix \url{https://github.com/onnx/onnx}.
\bibitem[{Ouyang et~al.(2020)Ouyang, Noh, Wang, Qi, Ma, Gu, Kim, Hong, Bae,
  Zhao, Wang, Wu, Gong, Shi, Zhu and Du}]{Ouyang2020}
\bibinfo{author}{Ouyang, J.}, \bibinfo{author}{Noh, M.}, \bibinfo{author}{Wang,
  Y.}, \bibinfo{author}{Qi, W.}, \bibinfo{author}{Ma, Y.}, \bibinfo{author}{Gu,
  C.}, \bibinfo{author}{Kim, S.}, \bibinfo{author}{Hong, K.i.},
  \bibinfo{author}{Bae, W.K.}, \bibinfo{author}{Zhao, Z.},
  \bibinfo{author}{Wang, J.}, \bibinfo{author}{Wu, P.}, \bibinfo{author}{Gong,
  X.}, \bibinfo{author}{Shi, J.}, \bibinfo{author}{Zhu, H.},
  \bibinfo{author}{Du, X.}, \bibinfo{year}{2020}.
\newblock \bibinfo{title}{{Baidu Kunlun An AI processor for diversified
  workloads}}, in: \bibinfo{booktitle}{2020 IEEE Hot Chips 32 Symposium (HCS)},
  \bibinfo{publisher}{IEEE}. pp. \bibinfo{pages}{1--18}.
\newblock \URLprefix \url{https://ieeexplore.ieee.org/document/9220641/},
  \DOIprefix\doi{10.1109/HCS49909.2020.9220641}.
\bibitem[{Park et~al.(2017)Park, Ahn and Yoo}]{Park2017}
\bibinfo{author}{Park, E.}, \bibinfo{author}{Ahn, J.}, \bibinfo{author}{Yoo,
  S.}, \bibinfo{year}{2017}.
\newblock \bibinfo{title}{{Weighted-Entropy-Based Quantization for Deep Neural
  Networks}}, in: \bibinfo{booktitle}{IEEE/CVF Conference on Computer Vision
  and Pattern Recognition (CVPR)}, \bibinfo{publisher}{IEEE}. pp.
  \bibinfo{pages}{7197--7205}.
\newblock \URLprefix \url{http://ieeexplore.ieee.org/document/8100244/},
  \DOIprefix\doi{10.1109/CVPR.2017.761}.
\bibitem[{Paszke et~al.(2019)Paszke, Gross, Bradbury, Lin, Devito, Massa,
  Steiner, Killeen and Yang}]{Paszke2019}
\bibinfo{author}{Paszke, A.}, \bibinfo{author}{Gross, S.},
  \bibinfo{author}{Bradbury, J.}, \bibinfo{author}{Lin, Z.},
  \bibinfo{author}{Devito, Z.}, \bibinfo{author}{Massa, F.},
  \bibinfo{author}{Steiner, B.}, \bibinfo{author}{Killeen, T.},
  \bibinfo{author}{Yang, E.}, \bibinfo{year}{2019}.
\newblock \bibinfo{title}{{PyTorch : An Imperative Style , High-Performance
  Deep Learning Library}}.
\newblock \bibinfo{journal}{ArXiv preprint} .
\bibitem[{Pilipovi{\'{c}} et~al.(2018)Pilipovi{\'{c}}, Buli{\'{c}} and
  Risojevi{\'{c}}}]{Pilipovic2018}
\bibinfo{author}{Pilipovi{\'{c}}, R.}, \bibinfo{author}{Buli{\'{c}}, P.},
  \bibinfo{author}{Risojevi{\'{c}}, V.}, \bibinfo{year}{2018}.
\newblock \bibinfo{title}{{Compression of convolutional neural networks: A
  short survey}}, in: \bibinfo{booktitle}{2018 17th International Symposium on
  INFOTEH-JAHORINA, INFOTEH 2018 - Proceedings}, \bibinfo{publisher}{IEEE}. pp.
  \bibinfo{pages}{1--6}.
\newblock \URLprefix \url{https://ieeexplore.ieee.org/document/8345545/},
  \DOIprefix\doi{10.1109/INFOTEH.2018.8345545}.
\bibitem[{Polyak and Wolf(2015)}]{Polyak2015}
\bibinfo{author}{Polyak, A.}, \bibinfo{author}{Wolf, L.}, \bibinfo{year}{2015}.
\newblock \bibinfo{title}{{Channel-level acceleration of deep face
  representations}}.
\newblock \bibinfo{journal}{IEEE Access} \bibinfo{volume}{3},
  \bibinfo{pages}{2163--2175}.
\newblock \URLprefix \url{http://ieeexplore.ieee.org/document/7303876/},
  \DOIprefix\doi{10.1109/ACCESS.2015.2494536}.
\bibitem[{Preuser et~al.(2018)Preuser, Gambardella, Fraser and
  Blott}]{Preuser2018}
\bibinfo{author}{Preuser, T.B.}, \bibinfo{author}{Gambardella, G.},
  \bibinfo{author}{Fraser, N.}, \bibinfo{author}{Blott, M.},
  \bibinfo{year}{2018}.
\newblock \bibinfo{title}{{Inference of quantized neural networks on
  heterogeneous all-programmable devices}}, in: \bibinfo{booktitle}{2018
  Design, Automation {\&} Test in Europe Conference {\&} Exhibition (DATE)},
  \bibinfo{publisher}{IEEE}. pp. \bibinfo{pages}{833--838}.
\newblock \URLprefix \url{http://ieeexplore.ieee.org/document/8342121/},
  \DOIprefix\doi{10.23919/DATE.2018.8342121}.
\bibitem[{Prost-Boucle et~al.(2017)Prost-Boucle, Bourge, Petrot, Alemdar,
  Caldwell and Leroy}]{Prost-Boucle2017}
\bibinfo{author}{Prost-Boucle, A.}, \bibinfo{author}{Bourge, A.},
  \bibinfo{author}{Petrot, F.}, \bibinfo{author}{Alemdar, H.},
  \bibinfo{author}{Caldwell, N.}, \bibinfo{author}{Leroy, V.},
  \bibinfo{year}{2017}.
\newblock \bibinfo{title}{{Scalable high-performance architecture for
  convolutional ternary neural networks on FPGA}}, in: \bibinfo{booktitle}{2017
  27th International Conference on Field Programmable Logic and Applications
  (FPL)}, \bibinfo{publisher}{IEEE}. pp. \bibinfo{pages}{1--7}.
\newblock \URLprefix \url{https://hal.archives-ouvertes.fr/hal-01563763
  http://ieeexplore.ieee.org/document/8056850/},
  \DOIprefix\doi{10.23919/FPL.2017.8056850}.
\bibitem[{Qin et~al.(2020a)Qin, Gong, Liu, Bai, Song and Sebe}]{Qin2020}
\bibinfo{author}{Qin, H.}, \bibinfo{author}{Gong, R.}, \bibinfo{author}{Liu,
  X.}, \bibinfo{author}{Bai, X.}, \bibinfo{author}{Song, J.},
  \bibinfo{author}{Sebe, N.}, \bibinfo{year}{2020}a.
\newblock \bibinfo{title}{{Binary neural networks: A survey}}.
\newblock \bibinfo{journal}{Pattern Recognition} \bibinfo{volume}{105},
  \bibinfo{pages}{107281}.
\newblock \URLprefix
  \url{https://linkinghub.elsevier.com/retrieve/pii/S0031320320300856},
  \DOIprefix\doi{10.1016/j.patcog.2020.107281}.
\bibitem[{Qin et~al.(2020b)Qin, Gong, Liu, Shen, Wei, Yu and Song}]{Qin2020a}
\bibinfo{author}{Qin, H.}, \bibinfo{author}{Gong, R.}, \bibinfo{author}{Liu,
  X.}, \bibinfo{author}{Shen, M.}, \bibinfo{author}{Wei, Z.},
  \bibinfo{author}{Yu, F.}, \bibinfo{author}{Song, J.}, \bibinfo{year}{2020}b.
\newblock \bibinfo{title}{{Forward and Backward Information Retention for
  Accurate Binary Neural Networks}}, in: \bibinfo{booktitle}{IEEE/CVF
  Conference on Computer Vision and Pattern Recognition (CVPR)},
  \bibinfo{publisher}{IEEE}. pp. \bibinfo{pages}{2247--2256}.
\newblock \URLprefix \url{https://ieeexplore.ieee.org/document/9157443/},
  \DOIprefix\doi{10.1109/CVPR42600.2020.00232}.
\bibitem[{Rastegari et~al.(2016)Rastegari, Ordonez, Redmon and
  Farhadi}]{Rastegari2016}
\bibinfo{author}{Rastegari, M.}, \bibinfo{author}{Ordonez, V.},
  \bibinfo{author}{Redmon, J.}, \bibinfo{author}{Farhadi, A.},
  \bibinfo{year}{2016}.
\newblock \bibinfo{title}{{XNOR-Net: ImageNet Classification Using Binary
  Convolutional Neural Networks}}, in: \bibinfo{booktitle}{European Conference
  on Computer Vision}, \bibinfo{publisher}{Springer}. pp.
  \bibinfo{pages}{525--542}.
\newblock \URLprefix \url{http://arxiv.org/abs/1603.05279
  http://link.springer.com/10.1007/978-3-319-46493-0_32},
  \DOIprefix\doi{10.1007/978-3-319-46493-0{\_}32}.
\bibitem[{Reed(1993)}]{Reed1993}
\bibinfo{author}{Reed, R.}, \bibinfo{year}{1993}.
\newblock \bibinfo{title}{{Pruning Algorithms - A Survey}}.
\newblock \bibinfo{journal}{IEEE Transactions on Neural Networks}
  \bibinfo{volume}{4}, \bibinfo{pages}{740--747}.
\newblock \URLprefix \url{http://ieeexplore.ieee.org/document/248452/},
  \DOIprefix\doi{10.1109/72.248452}.
\bibitem[{Reuther et~al.(2019)Reuther, Michaleas, Jones, Gadepally, Samsi and
  Kepner}]{Reuther2019}
\bibinfo{author}{Reuther, A.}, \bibinfo{author}{Michaleas, P.},
  \bibinfo{author}{Jones, M.}, \bibinfo{author}{Gadepally, V.},
  \bibinfo{author}{Samsi, S.}, \bibinfo{author}{Kepner, J.},
  \bibinfo{year}{2019}.
\newblock \bibinfo{title}{{Survey and Benchmarking of Machine Learning
  Accelerators}}, in: \bibinfo{booktitle}{2019 IEEE High Performance Extreme
  Computing Conference (HPEC)}, \bibinfo{publisher}{IEEE}. pp.
  \bibinfo{pages}{1--9}.
\newblock \URLprefix \url{https://ieeexplore.ieee.org/document/8916327/},
  \DOIprefix\doi{10.1109/HPEC.2019.8916327}.
\bibitem[{{Richard Chuang} et~al.(2020){Richard Chuang}, Oliyide and
  Garrett}]{RichardChuang2020}
\bibinfo{author}{{Richard Chuang}}, \bibinfo{author}{Oliyide, O.},
  \bibinfo{author}{Garrett, B.}, \bibinfo{year}{2020}.
\newblock \bibinfo{title}{{Introducing the Intel{\textregistered} Vision
  Accelerator Design with Intel{\textregistered} Arria{\textregistered} 10
  FPGA}}.
\newblock \bibinfo{journal}{White Paper} .
\bibitem[{Rodriguez et~al.(2018)Rodriguez, Segal, Meiri, Fomenko, Kim and
  Shen}]{Rodriguez2018}
\bibinfo{author}{Rodriguez, A.}, \bibinfo{author}{Segal, E.},
  \bibinfo{author}{Meiri, E.}, \bibinfo{author}{Fomenko, E.},
  \bibinfo{author}{Kim, Y.J.}, \bibinfo{author}{Shen, H.},
  \bibinfo{year}{2018}.
\newblock \bibinfo{title}{{Lower Numerical Precision Deep Learning Inference
  and Training}}.
\newblock \bibinfo{journal}{Intel White Paper} ,
  \bibinfo{pages}{1--19}\URLprefix
  \url{https://software.intel.com/sites/default/files/managed/db/92/Lower-Numerical-Precision-Deep-Learning-Jan2018.pdf}.
\bibitem[{Rotem et~al.(2018)Rotem, Fix, Abdulrasool, Catron, Deng, Dzhabarov,
  Gibson, Hegeman, Lele, Levenstein, Montgomery, Maher, Nadathur, Olesen, Park,
  Rakhov, Smelyanskiy and Wang}]{Rotem2018}
\bibinfo{author}{Rotem, N.}, \bibinfo{author}{Fix, J.},
  \bibinfo{author}{Abdulrasool, S.}, \bibinfo{author}{Catron, G.},
  \bibinfo{author}{Deng, S.}, \bibinfo{author}{Dzhabarov, R.},
  \bibinfo{author}{Gibson, N.}, \bibinfo{author}{Hegeman, J.},
  \bibinfo{author}{Lele, M.}, \bibinfo{author}{Levenstein, R.},
  \bibinfo{author}{Montgomery, J.}, \bibinfo{author}{Maher, B.},
  \bibinfo{author}{Nadathur, S.}, \bibinfo{author}{Olesen, J.},
  \bibinfo{author}{Park, J.}, \bibinfo{author}{Rakhov, A.},
  \bibinfo{author}{Smelyanskiy, M.}, \bibinfo{author}{Wang, M.},
  \bibinfo{year}{2018}.
\newblock \bibinfo{title}{{Glow: Graph lowering compiler techniques for neural
  networks}}.
\newblock \bibinfo{journal}{ArXiv preprint} .
\bibitem[{Ruffy and Chahal(2019)}]{Ruffy2019}
\bibinfo{author}{Ruffy, F.}, \bibinfo{author}{Chahal, K.},
  \bibinfo{year}{2019}.
\newblock \bibinfo{title}{{The State of Knowledge Distillation for
  Classification}}.
\newblock \bibinfo{journal}{ArXiv preprint} \URLprefix
  \url{http://arxiv.org/abs/1912.10850}.
\bibitem[{Russakovsky et~al.(2015)Russakovsky, Deng, Su, Krause, Satheesh, Ma,
  Huang, Karpathy, Khosla, Bernstein, Berg and Fei-Fei}]{Russakovsky2015}
\bibinfo{author}{Russakovsky, O.}, \bibinfo{author}{Deng, J.},
  \bibinfo{author}{Su, H.}, \bibinfo{author}{Krause, J.},
  \bibinfo{author}{Satheesh, S.}, \bibinfo{author}{Ma, S.},
  \bibinfo{author}{Huang, Z.}, \bibinfo{author}{Karpathy, A.},
  \bibinfo{author}{Khosla, A.}, \bibinfo{author}{Bernstein, M.},
  \bibinfo{author}{Berg, A.C.}, \bibinfo{author}{Fei-Fei, L.},
  \bibinfo{year}{2015}.
\newblock \bibinfo{title}{{ImageNet Large Scale Visual Recognition Challenge}}.
\newblock \bibinfo{journal}{International Journal of Computer Vision}
  \bibinfo{volume}{115}, \bibinfo{pages}{211--252}.
\newblock \URLprefix \url{http://link.springer.com/10.1007/s11263-015-0816-y},
  \DOIprefix\doi{10.1007/s11263-015-0816-y}.
\bibitem[{Saad and Marom(1990)}]{Saad1990}
\bibinfo{author}{Saad, D.}, \bibinfo{author}{Marom, E.}, \bibinfo{year}{1990}.
\newblock \bibinfo{title}{{Training Feed Forward Nets with Binary Weights Via a
  Modified CHIR Algorithm}}.
\newblock \bibinfo{journal}{Complex Systems} \bibinfo{volume}{4},
  \bibinfo{pages}{573--586}.
\newblock \URLprefix \url{https://www.complex-systems.com/pdf/04-5-5.pdf}.
\bibitem[{Sabour et~al.(2017)Sabour, Frosst and Hinton}]{Sabour2017}
\bibinfo{author}{Sabour, S.}, \bibinfo{author}{Frosst, N.},
  \bibinfo{author}{Hinton, G.E.}, \bibinfo{year}{2017}.
\newblock \bibinfo{title}{{Dynamic routing between capsules}}, in:
  \bibinfo{booktitle}{Advances in Neural Information Processing Systems
  (NIPS)}, pp. \bibinfo{pages}{3857--3867}.
\bibitem[{Santurkar et~al.(2018)Santurkar, Tsipras, Ilyas and
  Madry}]{Santurkar2018}
\bibinfo{author}{Santurkar, S.}, \bibinfo{author}{Tsipras, D.},
  \bibinfo{author}{Ilyas, A.}, \bibinfo{author}{Madry, A.},
  \bibinfo{year}{2018}.
\newblock \bibinfo{title}{{How does batch normalization help optimization?}},
  in: \bibinfo{booktitle}{Advances in Neural Information Processing Systems
  (NIPS)}, pp. \bibinfo{pages}{2483--2493}.
\bibitem[{Sermanet et~al.(2013)Sermanet, Eigen, Zhang, Mathieu, Fergus and
  LeCun}]{Sermanet2013}
\bibinfo{author}{Sermanet, P.}, \bibinfo{author}{Eigen, D.},
  \bibinfo{author}{Zhang, X.}, \bibinfo{author}{Mathieu, M.},
  \bibinfo{author}{Fergus, R.}, \bibinfo{author}{LeCun, Y.},
  \bibinfo{year}{2013}.
\newblock \bibinfo{title}{{OverFeat: Integrated Recognition, Localization and
  Detection using Convolutional Networks}}, in:
  \bibinfo{booktitle}{International Conference on Learning
  Representations(ICLR)}.
\newblock \URLprefix \url{http://arxiv.org/abs/1312.6229}.
\bibitem[{Settle et~al.(2018)Settle, Bollavaram, D'Alberto, Delaye, Fernandez,
  Fraser, Ng, Sirasao and Wu}]{Settle2018}
\bibinfo{author}{Settle, S.O.}, \bibinfo{author}{Bollavaram, M.},
  \bibinfo{author}{D'Alberto, P.}, \bibinfo{author}{Delaye, E.},
  \bibinfo{author}{Fernandez, O.}, \bibinfo{author}{Fraser, N.},
  \bibinfo{author}{Ng, A.}, \bibinfo{author}{Sirasao, A.}, \bibinfo{author}{Wu,
  M.}, \bibinfo{year}{2018}.
\newblock \bibinfo{title}{{Quantizing Convolutional Neural Networks for
  Low-Power High-Throughput Inference Engines}}.
\newblock \bibinfo{journal}{ArXiv preprint} \URLprefix
  \url{http://arxiv.org/abs/1805.07941}.
\bibitem[{Shen et~al.(2019)Shen, Han, Xu and Wang}]{Shen2019}
\bibinfo{author}{Shen, M.}, \bibinfo{author}{Han, K.}, \bibinfo{author}{Xu,
  C.}, \bibinfo{author}{Wang, Y.}, \bibinfo{year}{2019}.
\newblock \bibinfo{title}{{Searching for accurate binary neural
  architectures}}.
\newblock \bibinfo{journal}{Proceedings - 2019 International Conference on
  Computer Vision Workshop, ICCVW 2019} ,
  \bibinfo{pages}{2041--2044}\DOIprefix\doi{10.1109/ICCVW.2019.00256}.
\bibitem[{Shen et~al.(2016)Shen, Yi, Zhang, Shu and Liu}]{Shen2016}
\bibinfo{author}{Shen, X.}, \bibinfo{author}{Yi, B.}, \bibinfo{author}{Zhang,
  Z.}, \bibinfo{author}{Shu, J.}, \bibinfo{author}{Liu, H.},
  \bibinfo{year}{2016}.
\newblock \bibinfo{title}{{Automatic Recommendation Technology for Learning
  Resources with Convolutional Neural Network}}, in:
  \bibinfo{booktitle}{Proceedings - 2016 International Symposium on Educational
  Technology, ISET 2016}, pp. \bibinfo{pages}{30--34}.
\newblock \DOIprefix\doi{10.1109/ISET.2016.12}.
\bibitem[{Sheng et~al.(2018)Sheng, Feng, Zhuo, Zhang, Shen and
  Aleksic}]{Sheng2018}
\bibinfo{author}{Sheng, T.}, \bibinfo{author}{Feng, C.}, \bibinfo{author}{Zhuo,
  S.}, \bibinfo{author}{Zhang, X.}, \bibinfo{author}{Shen, L.},
  \bibinfo{author}{Aleksic, M.}, \bibinfo{year}{2018}.
\newblock \bibinfo{title}{{A Quantization-Friendly Separable Convolution for
  MobileNets}}.
\newblock \bibinfo{journal}{2018 1st Workshop on Energy Efficient Machine
  Learning and Cognitive Computing for Embedded Applications (EMC2)} ,
  \bibinfo{pages}{14--18}\URLprefix
  \url{https://ieeexplore.ieee.org/document/8524017/},
  \DOIprefix\doi{10.1109/EMC2.2018.00011}.
\bibitem[{Simons and Lee(2019)}]{Simons2019}
\bibinfo{author}{Simons, T.}, \bibinfo{author}{Lee, D.J.},
  \bibinfo{year}{2019}.
\newblock \bibinfo{title}{{A review of binarized neural networks}}.
\newblock \bibinfo{journal}{Electronics (Switzerland)} \bibinfo{volume}{8}.
\newblock \DOIprefix\doi{10.3390/electronics8060661}.
\bibitem[{Simonyan and Zisserman(2014)}]{Simonyan2014}
\bibinfo{author}{Simonyan, K.}, \bibinfo{author}{Zisserman, A.},
  \bibinfo{year}{2014}.
\newblock \bibinfo{title}{{Very Deep Convolutional Networks for Large-Scale
  Image Recognition}}, in: \bibinfo{booktitle}{International Conference on
  Learning Representations(ICLR)}, pp. \bibinfo{pages}{1--14}.
\newblock \URLprefix \url{http://arxiv.org/abs/1409.1556}.
\bibitem[{Singh et~al.(2019)Singh, Kumar~Verma, Rai and Namboodiri}]{Singh2019}
\bibinfo{author}{Singh, P.}, \bibinfo{author}{Kumar~Verma, V.},
  \bibinfo{author}{Rai, P.}, \bibinfo{author}{Namboodiri, V.P.},
  \bibinfo{year}{2019}.
\newblock \bibinfo{title}{{Play and Prune: Adaptive Filter Pruning for Deep
  Model Compression}}, in: \bibinfo{booktitle}{Proceedings of the Twenty-Eighth
  International Joint Conference on Artificial Intelligence},
  \bibinfo{publisher}{International Joint Conferences on Artificial
  Intelligence Organization}, \bibinfo{address}{California}. pp.
  \bibinfo{pages}{3460--3466}.
\newblock \URLprefix \url{https://www.ijcai.org/proceedings/2019/480},
  \DOIprefix\doi{10.24963/ijcai.2019/480}.
\bibitem[{Society and Committee(2008)}]{Society2008}
\bibinfo{author}{Society, I.C.}, \bibinfo{author}{Committee, M.S.},
  \bibinfo{year}{2008}.
\newblock \bibinfo{title}{{IEEE Standard for Floating-Point Arithmetic}}.
\newblock \bibinfo{journal}{IEEE Std 754-2008} \bibinfo{volume}{2008},
  \bibinfo{pages}{1--70}.
\newblock \DOIprefix\doi{10.1109/IEEESTD.2008.4610935}.
\bibitem[{Soudry et~al.(2014)Soudry, Hubara and Meir}]{Soudry2014}
\bibinfo{author}{Soudry, D.}, \bibinfo{author}{Hubara, I.},
  \bibinfo{author}{Meir, R.}, \bibinfo{year}{2014}.
\newblock \bibinfo{title}{{Expectation backpropagation: Parameter-free training
  of multilayer neural networks with continuous or discrete weights}}, in:
  \bibinfo{booktitle}{Advances in Neural Information Processing Systems
  (NIPS)}, pp. \bibinfo{pages}{963--971}.
\newblock \URLprefix \url{https://dl.acm.org/doi/abs/10.5555/2968826.2968934}.
\bibitem[{Srinivas and Babu(2015)}]{Srinivas2015}
\bibinfo{author}{Srinivas, S.}, \bibinfo{author}{Babu, R.V.},
  \bibinfo{year}{2015}.
\newblock \bibinfo{title}{{Data-free parameter pruning for Deep Neural
  Networks}}, in: \bibinfo{booktitle}{Procedings of the British Machine Vision
  Conference 2015}, \bibinfo{publisher}{British Machine Vision Association}.
  pp. \bibinfo{pages}{1--31}.
\newblock \URLprefix
  \url{http://www.bmva.org/bmvc/2015/papers/paper031/index.html
  http://arxiv.org/abs/1507.06149}, \DOIprefix\doi{10.5244/C.29.31}.
\bibitem[{Srivastava et~al.(2014)Srivastava, Hinton, {\ldots} and
  2014}]{Srivastava2014}
\bibinfo{author}{Srivastava, N.}, \bibinfo{author}{Hinton, G.},
  \bibinfo{author}{{\ldots}, A.K.T.j.o.m.}, \bibinfo{author}{2014, U.},
  \bibinfo{year}{2014}.
\newblock \bibinfo{title}{{Dropout: a simple way to prevent neural networks
  from overfitting}}.
\newblock \bibinfo{journal}{The journal of machine learning research}
  \bibinfo{volume}{15}, \bibinfo{pages}{1929--1958}.
\newblock \URLprefix
  \url{http://www.jmlr.org/papers/volume15/srivastava14a/srivastava14a.pdf?utm_content=buffer79b43&utm_medium=social&utm_source=twitter.com&utm_campaign=buffer},
  \DOIprefix\doi{10.5555/2627435.2670313}.
\bibitem[{Sun et~al.(2020)Sun, Luo, Gao, Wang, Gao and Yang}]{Sun2020}
\bibinfo{author}{Sun, J.}, \bibinfo{author}{Luo, X.}, \bibinfo{author}{Gao,
  H.}, \bibinfo{author}{Wang, W.}, \bibinfo{author}{Gao, Y.},
  \bibinfo{author}{Yang, X.}, \bibinfo{year}{2020}.
\newblock \bibinfo{title}{{Categorizing Malware via A Word2Vec-based Temporal
  Convolutional Network Scheme}}.
\newblock \bibinfo{journal}{Journal of Cloud Computing} \bibinfo{volume}{9}.
\newblock \DOIprefix\doi{10.1186/s13677-020-00200-y}.
\bibitem[{Sun et~al.(2017)Sun, Song, Jiang, Pan and Pang}]{Sun2017}
\bibinfo{author}{Sun, M.}, \bibinfo{author}{Song, Z.}, \bibinfo{author}{Jiang,
  X.}, \bibinfo{author}{Pan, J.}, \bibinfo{author}{Pang, Y.},
  \bibinfo{year}{2017}.
\newblock \bibinfo{title}{{Learning Pooling for Convolutional Neural Network}}.
\newblock \bibinfo{journal}{Neurocomputing} \bibinfo{volume}{224},
  \bibinfo{pages}{96--104}.
\newblock \URLprefix \url{http://dx.doi.org/10.1016/j.neucom.2016.10.049},
  \DOIprefix\doi{10.1016/j.neucom.2016.10.049}.
\bibitem[{Sze et~al.(2017)Sze, Chen, Yang and Emer}]{Sze2017}
\bibinfo{author}{Sze, V.}, \bibinfo{author}{Chen, Y.H.H.},
  \bibinfo{author}{Yang, T.J.J.}, \bibinfo{author}{Emer, J.S.},
  \bibinfo{year}{2017}.
\newblock \bibinfo{title}{{Efficient Processing of Deep Neural Networks: A
  Tutorial and Survey}}.
\newblock \bibinfo{journal}{Proceedings of the IEEE} \bibinfo{volume}{105},
  \bibinfo{pages}{2295--2329}.
\newblock \URLprefix \url{http://ieeexplore.ieee.org/document/8114708/},
  \DOIprefix\doi{10.1109/JPROC.2017.2761740}.
\bibitem[{Szegedy et~al.(2015)Szegedy, Liu, Jia, Sermanet, Reed, Anguelov,
  Erhan, Vanhoucke and Rabinovich}]{Szegedy2015}
\bibinfo{author}{Szegedy, C.}, \bibinfo{author}{Liu, W.}, \bibinfo{author}{Jia,
  Y.}, \bibinfo{author}{Sermanet, P.}, \bibinfo{author}{Reed, S.},
  \bibinfo{author}{Anguelov, D.}, \bibinfo{author}{Erhan, D.},
  \bibinfo{author}{Vanhoucke, V.}, \bibinfo{author}{Rabinovich, A.},
  \bibinfo{year}{2015}.
\newblock \bibinfo{title}{{Going deeper with convolutions}}, in:
  \bibinfo{booktitle}{Proceedings of the IEEE Computer Society Conference on
  Computer Vision and Pattern Recognition}, \bibinfo{publisher}{IEEE}. pp.
  \bibinfo{pages}{1--9}.
\newblock \URLprefix \url{http://ieeexplore.ieee.org/document/7298594/},
  \DOIprefix\doi{10.1109/CVPR.2015.7298594}.
\bibitem[{{TansorFlow}()}]{tflite}
\bibinfo{author}{{TansorFlow}}, .
\newblock \bibinfo{title}{{Fixed Point Quantization}}.
\newblock \URLprefix \url{https://www.tensorflow.org/lite/guide}.
\bibitem[{Technologies(2019)}]{SNPE}
\bibinfo{author}{Technologies, Q.}, \bibinfo{year}{2019}.
\newblock \bibinfo{title}{{Snapdragon Neural Processing Engine SDK}}.
\newblock \URLprefix \url{https://developer.qualcomm.com/docs/snpe/index.html}.
\bibitem[{{Tencent}(2019)}]{Tencent2019}
\bibinfo{author}{{Tencent}}, \bibinfo{year}{2019}.
\newblock \bibinfo{title}{{NCNN is a high-performance neural network inference
  framework optimized for the mobile platform}}.
\newblock \URLprefix \url{https://github.com/Tencent/ncnn}.
\bibitem[{Tishbirani(1996)}]{Tishbirani1996}
\bibinfo{author}{Tishbirani, R.}, \bibinfo{year}{1996}.
\newblock \bibinfo{title}{{Regression shrinkage and selection via the Lasso}}.
\newblock \URLprefix \url{https://statweb.stanford.edu/~tibs/lasso/lasso.pdf}.
\bibitem[{Umuroglu et~al.(2016)Umuroglu, Fraser, Gambardella, Blott, Leong,
  Jahre and Vissers}]{Umuroglu2016a}
\bibinfo{author}{Umuroglu, Y.}, \bibinfo{author}{Fraser, N.J.},
  \bibinfo{author}{Gambardella, G.}, \bibinfo{author}{Blott, M.},
  \bibinfo{author}{Leong, P.}, \bibinfo{author}{Jahre, M.},
  \bibinfo{author}{Vissers, K.}, \bibinfo{year}{2016}.
\newblock \bibinfo{title}{{FINN: A Framework for Fast, Scalable Binarized
  Neural Network Inference}}.
\newblock \bibinfo{journal}{Proceedings of the 2017 ACM/SIGDA International
  Symposium on Field-Programmable Gate Arrays - FPGA '17} ,
  \bibinfo{pages}{65--74}\URLprefix
  \url{http://dl.acm.org/citation.cfm?doid=3020078.3021744},
  \DOIprefix\doi{10.1145/3020078.3021744}.
\bibitem[{Vanholder(2016)}]{Vanholder2016}
\bibinfo{author}{Vanholder, H.}, \bibinfo{year}{2016}.
\newblock \bibinfo{title}{{Efficient Inference with TensorRT}}.
\newblock \bibinfo{type}{Technical Report}.
\bibitem[{Vanhoucke et~al.(2011)Vanhoucke, Senior and Mao}]{Vanhoucke2011}
\bibinfo{author}{Vanhoucke, V.}, \bibinfo{author}{Senior, A.},
  \bibinfo{author}{Mao, M.Z.}, \bibinfo{year}{2011}.
\newblock \bibinfo{title}{{Improving the speed of neural networks on CPUs}}
  \URLprefix \url{https://research.google/pubs/pub37631/}.
\bibitem[{Venieris et~al.(2018)Venieris, Kouris and Bouganis}]{Venieris2018}
\bibinfo{author}{Venieris, S.I.}, \bibinfo{author}{Kouris, A.},
  \bibinfo{author}{Bouganis, C.S.}, \bibinfo{year}{2018}.
\newblock \bibinfo{title}{{Toolflows for Mapping Convolutional Neural Networks
  on FPGAs}}.
\newblock \bibinfo{journal}{ACM Computing Surveys} \bibinfo{volume}{51},
  \bibinfo{pages}{1--39}.
\newblock \URLprefix \url{http://dl.acm.org/citation.cfm?doid=3212709.3186332},
  \DOIprefix\doi{10.1145/3186332}.
\bibitem[{Venkatesh et~al.(2017)Venkatesh, Nurvitadhi and
  Marr}]{Venkatesh2017b}
\bibinfo{author}{Venkatesh, G.}, \bibinfo{author}{Nurvitadhi, E.},
  \bibinfo{author}{Marr, D.}, \bibinfo{year}{2017}.
\newblock \bibinfo{title}{{Accelerating Deep Convolutional Networks using
  low-precision and sparsity}}, in: \bibinfo{booktitle}{2017 IEEE International
  Conference on Acoustics, Speech and Signal Processing (ICASSP)},
  \bibinfo{publisher}{IEEE}. pp. \bibinfo{pages}{2861--2865}.
\newblock \URLprefix \url{https://arxiv.org/pdf/1610.00324.pdf
  http://ieeexplore.ieee.org/document/7952679/},
  \DOIprefix\doi{10.1109/ICASSP.2017.7952679}.
\bibitem[{Wang et~al.(2019a)Wang, Liu, Lin, Lin and Han}]{Wang2018}
\bibinfo{author}{Wang, K.}, \bibinfo{author}{Liu, Z.}, \bibinfo{author}{Lin,
  Y.}, \bibinfo{author}{Lin, J.}, \bibinfo{author}{Han, S.},
  \bibinfo{year}{2019}a.
\newblock \bibinfo{title}{{HAQ: Hardware-Aware Automated Quantization With
  Mixed Precision}}, in: \bibinfo{booktitle}{2019 IEEE/CVF Conference on
  Computer Vision and Pattern Recognition (CVPR)}, \bibinfo{publisher}{IEEE}.
  pp. \bibinfo{pages}{8604--8612}.
\newblock \URLprefix \url{http://arxiv.org/abs/1811.08886
  https://ieeexplore.ieee.org/document/8954415/},
  \DOIprefix\doi{10.1109/CVPR.2019.00881}.
\bibitem[{Wang et~al.(2018a)Wang, Choi, Brand, Chen and
  Gopalakrishnan}]{Wang2018b}
\bibinfo{author}{Wang, N.}, \bibinfo{author}{Choi, J.}, \bibinfo{author}{Brand,
  D.}, \bibinfo{author}{Chen, C.Y.}, \bibinfo{author}{Gopalakrishnan, K.},
  \bibinfo{year}{2018}a.
\newblock \bibinfo{title}{{Training deep neural networks with 8-bit floating
  point numbers}}, in: \bibinfo{booktitle}{Advances in Neural Information
  Processing Systems (NIPS)}, pp. \bibinfo{pages}{7675--7684}.
\bibitem[{Wang and Cheng(2017)}]{Wang2017}
\bibinfo{author}{Wang, P.}, \bibinfo{author}{Cheng, J.}, \bibinfo{year}{2017}.
\newblock \bibinfo{title}{{Fixed-Point Factorized Networks}}, in:
  \bibinfo{booktitle}{IEEE/CVF Conference on Computer Vision and Pattern
  Recognition (CVPR)}, \bibinfo{publisher}{IEEE}. pp.
  \bibinfo{pages}{3966--3974}.
\newblock \URLprefix \url{http://ieeexplore.ieee.org/document/8099905/},
  \DOIprefix\doi{10.1109/CVPR.2017.422}.
\bibitem[{Wang et~al.(2018b)Wang, Hu, Zhang, Zhang, Liu and Cheng}]{Wang2018a}
\bibinfo{author}{Wang, P.}, \bibinfo{author}{Hu, Q.}, \bibinfo{author}{Zhang,
  Y.}, \bibinfo{author}{Zhang, C.}, \bibinfo{author}{Liu, Y.},
  \bibinfo{author}{Cheng, J.}, \bibinfo{year}{2018}b.
\newblock \bibinfo{title}{{Two-Step Quantization for Low-bit Neural Networks}},
  in: \bibinfo{booktitle}{Proceedings of the IEEE/CVF Conference on Computer
  Vision and Pattern Recognition (CVPR)}, pp. \bibinfo{pages}{4376--4384}.
\newblock \DOIprefix\doi{10.1109/CVPR.2018.00460}.
\bibitem[{Wang et~al.(2019b)Wang, Lu, Tao, Zhou and Tian}]{Wang2019}
\bibinfo{author}{Wang, Z.}, \bibinfo{author}{Lu, J.}, \bibinfo{author}{Tao,
  C.}, \bibinfo{author}{Zhou, J.}, \bibinfo{author}{Tian, Q.},
  \bibinfo{year}{2019}b.
\newblock \bibinfo{title}{{Learning channel-wise interactions for binary
  convolutional neural networks}}, in: \bibinfo{booktitle}{Proceedings of the
  IEEE/CVF Conference on Computer Vision and Pattern Recognition (CVPR)}, pp.
  \bibinfo{pages}{568--577}.
\newblock \DOIprefix\doi{10.1109/CVPR.2019.00066}.
\bibitem[{Wen et~al.(2016)Wen, Wu, Wang, Chen and Li}]{Wen2016a}
\bibinfo{author}{Wen, W.}, \bibinfo{author}{Wu, C.}, \bibinfo{author}{Wang,
  Y.}, \bibinfo{author}{Chen, Y.}, \bibinfo{author}{Li, H.},
  \bibinfo{year}{2016}.
\newblock \bibinfo{title}{{Learning Structured Sparsity in Deep Neural
  Networks}}, in: \bibinfo{booktitle}{Advances in Neural Information Processing
  Systems (NIPS)}, \bibinfo{publisher}{IEEE}. pp. \bibinfo{pages}{2074--2082}.
\newblock \URLprefix \url{https://dl.acm.org/doi/abs/10.5555/3157096.3157329},
  \DOIprefix\doi{10.1016/j.ccr.2008.06.009}.
\bibitem[{Wu et~al.(2020)Wu, Judd, Zhang, Isaev and Micikevicius}]{Wu2020}
\bibinfo{author}{Wu, H.}, \bibinfo{author}{Judd, P.}, \bibinfo{author}{Zhang,
  X.}, \bibinfo{author}{Isaev, M.}, \bibinfo{author}{Micikevicius, P.},
  \bibinfo{year}{2020}.
\newblock \bibinfo{title}{{Integer quantization for deep learning inference:
  Principles and empirical evaluation}}.
\newblock \bibinfo{journal}{ArXiv preprint} , \bibinfo{pages}{1--20}.
\bibitem[{Wu et~al.(2016)Wu, Leng, Wang, Hu and Cheng}]{Wu2016}
\bibinfo{author}{Wu, J.}, \bibinfo{author}{Leng, C.}, \bibinfo{author}{Wang,
  Y.}, \bibinfo{author}{Hu, Q.}, \bibinfo{author}{Cheng, J.},
  \bibinfo{year}{2016}.
\newblock \bibinfo{title}{{Quantized Convolutional Neural Networks for Mobile
  Devices}}, in: \bibinfo{booktitle}{IEEE/CVF Conference on Computer Vision and
  Pattern Recognition (CVPR)}, \bibinfo{publisher}{IEEE}. pp.
  \bibinfo{pages}{4820--4828}.
\newblock \URLprefix \url{http://arxiv.org/abs/1512.06473
  http://ieeexplore.ieee.org/document/7780890/},
  \DOIprefix\doi{10.1109/CVPR.2016.521}.
\bibitem[{Wu et~al.(2018a)Wu, Li, Chen and Shi}]{Wu2018}
\bibinfo{author}{Wu, S.}, \bibinfo{author}{Li, G.}, \bibinfo{author}{Chen, F.},
  \bibinfo{author}{Shi, L.}, \bibinfo{year}{2018}a.
\newblock \bibinfo{title}{{Training and Inference with Integers in Deep Neural
  Networks}}, in: \bibinfo{booktitle}{International Conference on Learning
  Representations (ICLR)}.
\newblock \URLprefix \url{http://arxiv.org/abs/1802.04680}.
\bibitem[{Wu et~al.(2019)Wu, Li, Deng, Liu, Wu, Xie and Shi}]{Wu2019}
\bibinfo{author}{Wu, S.}, \bibinfo{author}{Li, G.}, \bibinfo{author}{Deng, L.},
  \bibinfo{author}{Liu, L.}, \bibinfo{author}{Wu, D.}, \bibinfo{author}{Xie,
  Y.}, \bibinfo{author}{Shi, L.}, \bibinfo{year}{2019}.
\newblock \bibinfo{title}{{L1-Norm Batch Normalization for Efficient Training
  of Deep Neural Networks}}.
\newblock \bibinfo{journal}{IEEE Transactions on Neural Networks and Learning
  Systems} \bibinfo{volume}{30}, \bibinfo{pages}{2043--2051}.
\newblock \URLprefix
  \url{https://ieeexplore.ieee.org/abstract/document/8528524/
  https://ieeexplore.ieee.org/document/8528524/},
  \DOIprefix\doi{10.1109/TNNLS.2018.2876179}.
\bibitem[{Wu et~al.(2018b)Wu, Nagarajan, Kumar, Rennie, Davis, Grauman and
  Feris}]{Wu2018a}
\bibinfo{author}{Wu, Z.}, \bibinfo{author}{Nagarajan, T.},
  \bibinfo{author}{Kumar, A.}, \bibinfo{author}{Rennie, S.},
  \bibinfo{author}{Davis, L.S.}, \bibinfo{author}{Grauman, K.},
  \bibinfo{author}{Feris, R.}, \bibinfo{year}{2018}b.
\newblock \bibinfo{title}{{BlockDrop: Dynamic Inference Paths in Residual
  Networks}}, in: \bibinfo{booktitle}{IEEE/CVF Conference on Computer Vision
  and Pattern Recognition (CVPR)}, \bibinfo{publisher}{IEEE}. pp.
  \bibinfo{pages}{8817--8826}.
\newblock \URLprefix \url{https://ieeexplore.ieee.org/document/8579017/},
  \DOIprefix\doi{10.1109/CVPR.2018.00919}.
\bibitem[{{Xiaomi}(2019)}]{Xiaomi2019}
\bibinfo{author}{{Xiaomi}}, \bibinfo{year}{2019}.
\newblock \bibinfo{title}{{MACE is a deep learning inference framework
  optimized for mobile heterogeneous computing platforms}}.
\newblock \URLprefix \url{https://github.com/XiaoMi/mace/}.
\bibitem[{{Xilinx} and {Inc}(2018)}]{Xilinx2018}
\bibinfo{author}{{Xilinx}}, \bibinfo{author}{{Inc}}, \bibinfo{year}{2018}.
\newblock \bibinfo{title}{{Accelerating DNNs with Xilinx Alveo Accelerator
  Cards (WP504)}}.
\newblock \bibinfo{journal}{White Paper} \bibinfo{volume}{504},
  \bibinfo{pages}{1--11}.
\newblock \URLprefix \url{www.xilinx.com1}.
\bibitem[{Xu et~al.(2019)Xu, Huan, Zheng and Zou}]{Xu2019}
\bibinfo{author}{Xu, J.}, \bibinfo{author}{Huan, Y.}, \bibinfo{author}{Zheng,
  L.R.}, \bibinfo{author}{Zou, Z.}, \bibinfo{year}{2019}.
\newblock \bibinfo{title}{{A Low-Power Arithmetic Element for Multi-Base
  Logarithmic Computation on Deep Neural Networks}}, in:
  \bibinfo{booktitle}{International System on Chip Conference},
  \bibinfo{publisher}{IEEE}. pp. \bibinfo{pages}{260--265}.
\newblock \URLprefix \url{https://ieeexplore.ieee.org/document/8618560/},
  \DOIprefix\doi{10.1109/SOCC.2018.8618560}.
\bibitem[{Xu et~al.(2020)Xu, Huang, Chen and Zhang}]{Xu2020a}
\bibinfo{author}{Xu, S.}, \bibinfo{author}{Huang, A.}, \bibinfo{author}{Chen,
  L.}, \bibinfo{author}{Zhang, B.}, \bibinfo{year}{2020}.
\newblock \bibinfo{title}{{Convolutional Neural Network Pruning: A Survey}},
  in: \bibinfo{booktitle}{2020 39th Chinese Control Conference (CCC)},
  \bibinfo{publisher}{IEEE}. pp. \bibinfo{pages}{7458--7463}.
\newblock \URLprefix \url{https://ieeexplore.ieee.org/document/9189610/},
  \DOIprefix\doi{10.23919/CCC50068.2020.9189610}.
\bibitem[{Xu et~al.(2018a)Xu, Lu, Yang, Hu, Chen, Hu and Shi}]{Xu2018a}
\bibinfo{author}{Xu, X.}, \bibinfo{author}{Lu, Q.}, \bibinfo{author}{Yang, L.},
  \bibinfo{author}{Hu, S.}, \bibinfo{author}{Chen, D.}, \bibinfo{author}{Hu,
  Y.}, \bibinfo{author}{Shi, Y.}, \bibinfo{year}{2018}a.
\newblock \bibinfo{title}{{Quantization of Fully Convolutional Networks for
  Accurate Biomedical Image Segmentation}}, in: \bibinfo{booktitle}{Proceedings
  of the IEEE/CVF Conference on Computer Vision and Pattern Recognition
  (CVPR)}, pp. \bibinfo{pages}{8300--8308}.
\newblock \DOIprefix\doi{10.1109/CVPR.2018.00866}.
\bibitem[{Xu et~al.(2018b)Xu, Hsu and Huang}]{Xu2018b}
\bibinfo{author}{Xu, Z.}, \bibinfo{author}{Hsu, Y.C.}, \bibinfo{author}{Huang,
  J.}, \bibinfo{year}{2018}b.
\newblock \bibinfo{title}{{Training shallow and thin networks for acceleration
  via knowledge distillation with conditional adversarial networks}}, in:
  \bibinfo{booktitle}{International Conference on Learning Representations
  (ICLR) - Workshop}.
\bibitem[{Yang et~al.(2019)Yang, Shen, Xing, Tian, Li, Deng, Huang and
  Hua}]{Yang2019}
\bibinfo{author}{Yang, J.}, \bibinfo{author}{Shen, X.}, \bibinfo{author}{Xing,
  J.}, \bibinfo{author}{Tian, X.}, \bibinfo{author}{Li, H.},
  \bibinfo{author}{Deng, B.}, \bibinfo{author}{Huang, J.},
  \bibinfo{author}{Hua, X.s.}, \bibinfo{year}{2019}.
\newblock \bibinfo{title}{{Quantization Networks}}, in:
  \bibinfo{booktitle}{2019 IEEE/CVF Conference on Computer Vision and Pattern
  Recognition (CVPR)}, \bibinfo{publisher}{IEEE}. pp.
  \bibinfo{pages}{7300--7308}.
\newblock \URLprefix \url{https://ieeexplore.ieee.org/document/8953531/},
  \DOIprefix\doi{10.1109/CVPR.2019.00748}.
\bibitem[{Yang et~al.(2020)Yang, Deng, Wu, Yan, Xie and Li}]{Yang2020}
\bibinfo{author}{Yang, Y.}, \bibinfo{author}{Deng, L.}, \bibinfo{author}{Wu,
  S.}, \bibinfo{author}{Yan, T.}, \bibinfo{author}{Xie, Y.},
  \bibinfo{author}{Li, G.}, \bibinfo{year}{2020}.
\newblock \bibinfo{title}{{Training high-performance and large-scale deep
  neural networks with full 8-bit integers}}.
\newblock \bibinfo{journal}{Neural Networks} \bibinfo{volume}{125},
  \bibinfo{pages}{70--82}.
\newblock \DOIprefix\doi{10.1016/j.neunet.2019.12.027}.
\bibitem[{Ye et~al.(2018)Ye, Lu, Lin and Wang}]{Ye2018}
\bibinfo{author}{Ye, J.}, \bibinfo{author}{Lu, X.}, \bibinfo{author}{Lin, Z.},
  \bibinfo{author}{Wang, J.Z.}, \bibinfo{year}{2018}.
\newblock \bibinfo{title}{{Rethinking the Smaller-Norm-Less-Informative
  Assumption in Channel Pruning of Convolution Layers}}.
\newblock \bibinfo{journal}{ArXiv preprint} \URLprefix
  \url{http://arxiv.org/abs/1802.00124}.
\bibitem[{Yin et~al.(2019)Yin, Zhang, Lyu, Osher, Qi and Xin}]{Yin2019}
\bibinfo{author}{Yin, P.}, \bibinfo{author}{Zhang, S.}, \bibinfo{author}{Lyu,
  J.}, \bibinfo{author}{Osher, S.}, \bibinfo{author}{Qi, Y.},
  \bibinfo{author}{Xin, J.}, \bibinfo{year}{2019}.
\newblock \bibinfo{title}{{Blended coarse gradient descent for full
  quantization of deep neural networks}}.
\newblock \bibinfo{journal}{Research in Mathematical Sciences}
  \bibinfo{volume}{6}.
\newblock \DOIprefix\doi{10.1007/s40687-018-0177-6}.
\bibitem[{Yogatama and Mann(2014)}]{Yogatama2014}
\bibinfo{author}{Yogatama, D.}, \bibinfo{author}{Mann, G.},
  \bibinfo{year}{2014}.
\newblock \bibinfo{title}{{Efficient Transfer Learning Method for Automatic
  Hyperparameter Tuning}}, in: \bibinfo{editor}{Kaski, S.},
  \bibinfo{editor}{Corander, J.} (Eds.), \bibinfo{booktitle}{Proceedings of the
  Seventeenth International Conference on Artificial Intelligence and
  Statistics}, \bibinfo{publisher}{PMLR}, \bibinfo{address}{Reykjavik,
  Iceland}. pp. \bibinfo{pages}{1077--1085}.
\newblock \URLprefix \url{http://proceedings.mlr.press/v33/yogatama14.html}.
\bibitem[{Yu et~al.(2017)Yu, Lukefahr, Palframan, Dasika, Das and
  Mahlke}]{Yu2017a}
\bibinfo{author}{Yu, J.}, \bibinfo{author}{Lukefahr, A.},
  \bibinfo{author}{Palframan, D.}, \bibinfo{author}{Dasika, G.},
  \bibinfo{author}{Das, R.}, \bibinfo{author}{Mahlke, S.},
  \bibinfo{year}{2017}.
\newblock \bibinfo{title}{{Scalpel: Customizing DNN pruning to the underlying
  hardware parallelism}}.
\newblock \bibinfo{journal}{ACM SIGARCH Computer Architecture News}
  \bibinfo{volume}{45}, \bibinfo{pages}{548--560}.
\newblock \URLprefix \url{http://dl.acm.org/citation.cfm?doid=3140659.3080215},
  \DOIprefix\doi{10.1145/3140659.3080215}.
\bibitem[{Yu et~al.(2018)Yu, Yang, Xu, Yang and Huang}]{Yu2018a}
\bibinfo{author}{Yu, J.}, \bibinfo{author}{Yang, L.}, \bibinfo{author}{Xu, N.},
  \bibinfo{author}{Yang, J.}, \bibinfo{author}{Huang, T.},
  \bibinfo{year}{2018}.
\newblock \bibinfo{title}{{Slimmable Neural Networks}}, in:
  \bibinfo{booktitle}{International Conference on Learning
  Representations(ICLR)}, \bibinfo{publisher}{International Conference on
  Learning Representations, ICLR}. pp. \bibinfo{pages}{1--12}.
\newblock \URLprefix \url{http://arxiv.org/abs/1812.08928}.
\bibitem[{Yuan and Lin(2006)}]{Yuan2006}
\bibinfo{author}{Yuan, M.}, \bibinfo{author}{Lin, Y.}, \bibinfo{year}{2006}.
\newblock \bibinfo{title}{{Model selection and estimation in regression with
  grouped variables}}.
\newblock \bibinfo{journal}{Journal of the Royal Statistical Society: Series B
  (Statistical Methodology)} \bibinfo{volume}{68}, \bibinfo{pages}{49--67}.
\newblock \URLprefix
  \url{http://doi.wiley.com/10.1111/j.1467-9868.2005.00532.x},
  \DOIprefix\doi{10.1111/j.1467-9868.2005.00532.x}.
\bibitem[{Yuan et~al.(2020)Yuan, Hu, Wu and Ban}]{Yuan2020}
\bibinfo{author}{Yuan, Z.}, \bibinfo{author}{Hu, J.}, \bibinfo{author}{Wu, D.},
  \bibinfo{author}{Ban, X.}, \bibinfo{year}{2020}.
\newblock \bibinfo{title}{{A dual-attention recurrent neural network method for
  deep cone thickener underflow concentration prediction}}.
\newblock \bibinfo{journal}{Sensors (Switzerland)} \bibinfo{volume}{20},
  \bibinfo{pages}{1--18}.
\newblock \DOIprefix\doi{10.3390/s20051260}.
\bibitem[{Zhang et~al.(2018)Zhang, Yang, Ye and Hua}]{Zhang2018d}
\bibinfo{author}{Zhang, D.}, \bibinfo{author}{Yang, J.}, \bibinfo{author}{Ye,
  D.}, \bibinfo{author}{Hua, G.}, \bibinfo{year}{2018}.
\newblock \bibinfo{title}{{LQ-Nets: Learned quantization for highly accurate
  and compact deep neural networks}}, in: \bibinfo{booktitle}{Lecture Notes in
  Computer Science (including subseries Lecture Notes in Artificial
  Intelligence and Lecture Notes in Bioinformatics)}, pp.
  \bibinfo{pages}{373--390}.
\newblock \DOIprefix\doi{10.1007/978-3-030-01237-3{\_}23}.
\bibitem[{Zhang et~al.(2019a)Zhang, Zhang, Chen, Sun, Ma and Yu}]{Zhang2019a}
\bibinfo{author}{Zhang, Q.}, \bibinfo{author}{Zhang, M.},
  \bibinfo{author}{Chen, T.}, \bibinfo{author}{Sun, Z.}, \bibinfo{author}{Ma,
  Y.}, \bibinfo{author}{Yu, B.}, \bibinfo{year}{2019}a.
\newblock \bibinfo{title}{{Recent Advances in Convolutional Neural Network
  Acceleration}}.
\newblock \bibinfo{journal}{Neurocomputing} \bibinfo{volume}{323},
  \bibinfo{pages}{37--51}.
\newblock \URLprefix
  \url{https://linkinghub.elsevier.com/retrieve/pii/S0925231218311007},
  \DOIprefix\doi{10.1016/j.neucom.2018.09.038}.
\bibitem[{Zhang et~al.(2016a)Zhang, Du, Zhang, Lan, Liu, Li, Guo, Chen and
  Chen}]{Zhang2016}
\bibinfo{author}{Zhang, S.}, \bibinfo{author}{Du, Z.}, \bibinfo{author}{Zhang,
  L.}, \bibinfo{author}{Lan, H.}, \bibinfo{author}{Liu, S.},
  \bibinfo{author}{Li, L.}, \bibinfo{author}{Guo, Q.}, \bibinfo{author}{Chen,
  T.}, \bibinfo{author}{Chen, Y.}, \bibinfo{year}{2016}a.
\newblock \bibinfo{title}{{Cambricon-X: An accelerator for sparse neural
  networks}}, in: \bibinfo{booktitle}{2016 49th Annual IEEE/ACM International
  Symposium on Microarchitecture (MICRO)}, \bibinfo{publisher}{IEEE}. pp.
  \bibinfo{pages}{1--12}.
\newblock \URLprefix \url{http://ieeexplore.ieee.org/document/7783723/},
  \DOIprefix\doi{10.1109/MICRO.2016.7783723}.
\bibitem[{Zhang et~al.(2016b)Zhang, Wu, Che, Lin, Memisevic, Salakhutdinov and
  Bengio}]{Zhang2016b}
\bibinfo{author}{Zhang, S.}, \bibinfo{author}{Wu, Y.}, \bibinfo{author}{Che,
  T.}, \bibinfo{author}{Lin, Z.}, \bibinfo{author}{Memisevic, R.},
  \bibinfo{author}{Salakhutdinov, R.}, \bibinfo{author}{Bengio, Y.},
  \bibinfo{year}{2016}b.
\newblock \bibinfo{title}{{Architectural complexity measures of recurrent
  neural networks}}, in: \bibinfo{booktitle}{Advances in Neural Information
  Processing Systems (NIPS)}, pp. \bibinfo{pages}{1830--1838}.
\bibitem[{Zhang et~al.(2019b)Zhang, Zhao, Ni, Zhang and Deng}]{Zhang2019}
\bibinfo{author}{Zhang, Y.}, \bibinfo{author}{Zhao, C.}, \bibinfo{author}{Ni,
  B.}, \bibinfo{author}{Zhang, J.}, \bibinfo{author}{Deng, H.},
  \bibinfo{year}{2019}b.
\newblock \bibinfo{title}{{Exploiting Channel Similarity for Accelerating Deep
  Convolutional Neural Networks}}.
\newblock \bibinfo{journal}{ArXiv preprint} , \bibinfo{pages}{1--14}\URLprefix
  \url{http://arxiv.org/abs/1908.02620}.
\bibitem[{Zhao et~al.(2017)Zhao, Song, Zhang, Xing, Lin, Srivastava, Gupta and
  Zhang}]{Zhao2017}
\bibinfo{author}{Zhao, R.}, \bibinfo{author}{Song, W.}, \bibinfo{author}{Zhang,
  W.}, \bibinfo{author}{Xing, T.}, \bibinfo{author}{Lin, J.H.},
  \bibinfo{author}{Srivastava, M.}, \bibinfo{author}{Gupta, R.},
  \bibinfo{author}{Zhang, Z.}, \bibinfo{year}{2017}.
\newblock \bibinfo{title}{{Accelerating Binarized Convolutional Neural Networks
  with Software-Programmable FPGAs}}, in: \bibinfo{booktitle}{Proceedings of
  the 2017 ACM/SIGDA International Symposium on Field-Programmable Gate Arrays
  - FPGA '17}, \bibinfo{publisher}{ACM Press}, \bibinfo{address}{New York, New
  York, USA}. pp. \bibinfo{pages}{15--24}.
\newblock \URLprefix \url{http://dl.acm.org/citation.cfm?doid=3020078.3021741},
  \DOIprefix\doi{10.1145/3020078.3021741}.
\bibitem[{Zhong et~al.(2020)Zhong, Zhao, Ning, Zeng, Guo, Wang and
  Yang}]{Zhong2020}
\bibinfo{author}{Zhong, K.}, \bibinfo{author}{Zhao, T.}, \bibinfo{author}{Ning,
  X.}, \bibinfo{author}{Zeng, S.}, \bibinfo{author}{Guo, K.},
  \bibinfo{author}{Wang, Y.}, \bibinfo{author}{Yang, H.}, \bibinfo{year}{2020}.
\newblock \bibinfo{title}{{Towards Lower Bit Multiplication for Convolutional
  Neural Network Training}}.
\newblock \bibinfo{journal}{ArXiv preprint} \URLprefix
  \url{http://arxiv.org/abs/2006.02804}.
\bibitem[{Zhou et~al.(2017a)Zhou, Yao, Guo, Xu and Chen}]{Zhou2017a}
\bibinfo{author}{Zhou, A.}, \bibinfo{author}{Yao, A.}, \bibinfo{author}{Guo,
  Y.}, \bibinfo{author}{Xu, L.}, \bibinfo{author}{Chen, Y.},
  \bibinfo{year}{2017}a.
\newblock \bibinfo{title}{{Incremental Network Quantization: Towards Lossless
  CNNs with Low-Precision Weights}}, in: \bibinfo{booktitle}{International
  Conference on Learning Representations(ICLR)}.
\newblock \URLprefix \url{https://github.com/Zhouaojun/Incremental-
  http://arxiv.org/abs/1702.03044 http://cn.arxiv.org/pdf/1702.03044.pdf}.
\bibitem[{Zhou et~al.(2016a)Zhou, Alvarez and Porikli}]{Zhou2016}
\bibinfo{author}{Zhou, H.}, \bibinfo{author}{Alvarez, J.M.},
  \bibinfo{author}{Porikli, F.}, \bibinfo{year}{2016}a.
\newblock \bibinfo{title}{{Less Is More: Towards Compact CNNs}}, in:
  \bibinfo{booktitle}{European Conference on Computer Vision}, pp.
  \bibinfo{pages}{662--677}.
\newblock \URLprefix
  \url{https://link.springer.com/chapter/10.1007/978-3-319-46493-0_40
  http://link.springer.com/10.1007/978-3-319-46493-0_40},
  \DOIprefix\doi{10.1007/978-3-319-46493-0{\_}40}.
\bibitem[{Zhou et~al.(2018)Zhou, Kannan and Prasanna}]{Zhou2018}
\bibinfo{author}{Zhou, S.}, \bibinfo{author}{Kannan, R.},
  \bibinfo{author}{Prasanna, V.K.}, \bibinfo{year}{2018}.
\newblock \bibinfo{title}{{Accelerating low rank matrix completion on FPGA}},
  in: \bibinfo{booktitle}{2017 International Conference on Reconfigurable
  Computing and FPGAs, ReConFig 2017}, \bibinfo{publisher}{IEEE}. pp.
  \bibinfo{pages}{1--7}.
\newblock \URLprefix \url{http://ieeexplore.ieee.org/document/8279771/},
  \DOIprefix\doi{10.1109/RECONFIG.2017.8279771}.
\bibitem[{Zhou et~al.(2016b)Zhou, Wu, Ni, Zhou, Wen and Zou}]{Zhou2016a}
\bibinfo{author}{Zhou, S.}, \bibinfo{author}{Wu, Y.}, \bibinfo{author}{Ni, Z.},
  \bibinfo{author}{Zhou, X.}, \bibinfo{author}{Wen, H.}, \bibinfo{author}{Zou,
  Y.}, \bibinfo{year}{2016}b.
\newblock \bibinfo{title}{{DoReFa-Net: Training Low Bitwidth Convolutional
  Neural Networks with Low Bitwidth Gradients}}.
\newblock \bibinfo{journal}{ArXiv preprint} \bibinfo{volume}{abs/1606.0},
  \bibinfo{pages}{1--13}.
\newblock \URLprefix \url{https://arxiv.org/abs/1606.06160}.
\bibitem[{Zhou et~al.(2017b)Zhou, Wang, Wen, He and Zou}]{Zhou2017c}
\bibinfo{author}{Zhou, S.C.}, \bibinfo{author}{Wang, Y.Z.},
  \bibinfo{author}{Wen, H.}, \bibinfo{author}{He, Q.Y.}, \bibinfo{author}{Zou,
  Y.H.}, \bibinfo{year}{2017}b.
\newblock \bibinfo{title}{{Balanced Quantization: An Effective and Efficient
  Approach to Quantized Neural Networks}}.
\newblock \bibinfo{journal}{Journal of Computer Science and Technology}
  \bibinfo{volume}{32}, \bibinfo{pages}{667--682}.
\newblock \DOIprefix\doi{10.1007/s11390-017-1750-y}.
\bibitem[{Zhu et~al.(2017)Zhu, Han, Mao and Dally}]{Zhu2016}
\bibinfo{author}{Zhu, C.}, \bibinfo{author}{Han, S.}, \bibinfo{author}{Mao,
  H.}, \bibinfo{author}{Dally, W.J.}, \bibinfo{year}{2017}.
\newblock \bibinfo{title}{{Trained Ternary Quantization}}, in:
  \bibinfo{booktitle}{International Conference on Learning Representations
  (ICLR)}, pp. \bibinfo{pages}{1--10}.
\newblock \URLprefix \url{http://arxiv.org/abs/1612.01064}.
\bibitem[{Zhu et~al.()Zhu, Gong, Yu, Liu, Wang, Li, Yang and Yan}]{Zhu2019}
\bibinfo{author}{Zhu, F.}, \bibinfo{author}{Gong, R.}, \bibinfo{author}{Yu,
  F.}, \bibinfo{author}{Liu, X.}, \bibinfo{author}{Wang, Y.},
  \bibinfo{author}{Li, Z.}, \bibinfo{author}{Yang, X.}, \bibinfo{author}{Yan,
  J.}, .
\newblock \bibinfo{title}{{Towards Unified INT8 Training for Convolutional
  Neural Network}}, in: \bibinfo{booktitle}{Proceedings of the IEEE/CVF
  Conference on Computer Vision and Pattern Recognition (CVPR)}.
\newblock \URLprefix \url{http://arxiv.org/abs/1912.12607}.
\bibitem[{Zhuang et~al.(2019)Zhuang, Shen, Tan, Liu and Reid}]{Zhuang2019}
\bibinfo{author}{Zhuang, B.}, \bibinfo{author}{Shen, C.}, \bibinfo{author}{Tan,
  M.}, \bibinfo{author}{Liu, L.}, \bibinfo{author}{Reid, I.},
  \bibinfo{year}{2019}.
\newblock \bibinfo{title}{{Structured binary neural networks for accurate image
  classification and semantic segmentation}}.
\newblock \bibinfo{journal}{Proceedings of the IEEE Computer Society Conference
  on Computer Vision and Pattern Recognition} \bibinfo{volume}{2019-June},
  \bibinfo{pages}{413--422}.
\newblock \DOIprefix\doi{10.1109/CVPR.2019.00050}.
\bibitem[{Zoph et~al.(2017)Zoph, Vasudevan, Shlens and Le}]{Zoph2017}
\bibinfo{author}{Zoph, B.}, \bibinfo{author}{Vasudevan, V.},
  \bibinfo{author}{Shlens, J.}, \bibinfo{author}{Le, Q.V.},
  \bibinfo{year}{2017}.
\newblock \bibinfo{title}{{Learning Transferable Architectures for Scalable
  Image Recognition}}.
\newblock \bibinfo{journal}{Proceedings of the IEEE/CVF Conference on Computer
  Vision and Pattern Recognition (CVPR)} ,
  \bibinfo{pages}{8697--8710}\URLprefix
  \url{https://ieeexplore.ieee.org/abstract/document/8579005/}.

\end{thebibliography}
